\documentclass[twoside,11pt]{article}

\usepackage{blindtext}
\usepackage[normalem]{ulem}
\usepackage{jmlr2e}

\usepackage[utf8]{inputenc} % allow utf-8 input
\usepackage[T1]{fontenc}    % use 8-bit T1 fonts
\usepackage{hyperref}       % hyperlinks
\usepackage{url}            % simple URL typesetting
\usepackage{booktabs}       % professional-quality tables
\usepackage{amsfonts}       % blackboard math symbols
\usepackage{nicefrac}       % compact symbols for 1/2, etc.
\usepackage{microtype}      % microtypography
\usepackage{xcolor}         % colors

\usepackage{mathtools}
\usepackage{fancyhdr,graphicx,amsmath,amssymb}
\usepackage[ruled,vlined]{algorithm2e}
\usepackage[e]{esvect}

\usepackage{algorithm2e}

\usepackage{varwidth}
\newtheorem{condition}{Condition}

\usepackage{multirow}
\usepackage{comment}
\newcommand{\norm}[1]{\left\lVert#1\right\rVert}

\DeclareMathOperator*{\argmax}{arg\,max}
\DeclareMathOperator*{\argmin}{arg\,min}

\DeclarePairedDelimiter{\ceil}{\lceil}{\rceil}

\usepackage{lastpage}

\ShortHeadings{Optimizing oblique splits}{}
\firstpageno{1}

\begin{document}

\title{Optimizing High-Dimensional Oblique Splits}

\author{\name Chien-Ming Chi\thanks{
This work was supported by Grant 113-2118-M-001-008-MY2 from the National Science and Technology Council, Taiwan.} 
 \email xbbchi@stat.sinica.edu.tw \\
       \addr Institute of Statistical Science\\
       Academia Sinica\\
       Taipei, Taiwan}

\editor{}

\maketitle

\begin{abstract}%   <- trailing '%' for backward compatibility of .sty file

Evidence suggests that oblique splits can significantly enhance the performance of decision trees. This paper explores the optimization of high-dimensional oblique splits for decision tree construction, establishing the Sufficient Impurity Decrease (SID) convergence that takes into account $s_0$-sparse oblique splits. We demonstrate that the SID function class expands as  sparsity parameter $s_0$ increases, enabling the model to capture complex data-generating processes such as the $s_0$-dimensional XOR function. Thus, \(s_0\) represents the unknown potential complexity of the underlying data-generating function. Furthermore, we establish that learning these complex functions necessitates greater computational resources. This highlights a fundamental trade-off between statistical accuracy, which is governed by the $s_0$-dependent size of the SID function class, and computational cost. Particularly, for challenging problems, the required candidate oblique split set can become prohibitively large, rendering standard ensemble approaches computationally impractical. To address this, we propose progressive trees that optimize oblique splits through an iterative refinement process rather than a single-step optimization. These splits are integrated alongside traditional orthogonal splits into ensemble models like Random Forests to enhance finite-sample performance. The effectiveness of our approach is validated through simulations and real-data experiments, where it consistently outperforms various existing oblique tree models.

\end{abstract}


\section{Introduction}

Trees are recognized as reliable predictors and classifiers. Various tree-based models, including ensembles~\citep{breiman2001random}, boosting methods~\citep{chen2016xgboost, friedman2001greedy}, and additive models~\citep{chipman2010bart}, have become well-established and are widely used for prediction tasks. Despite the impressive predictive power of deep learning, these tree-based approaches continue to remain competitive with neural networks in certain tasks, such as tabular regression problems~\citep{grinsztajn2022tree, mcelfresh2023neural}. The tree-based models mentioned earlier rely on axis-aligned (or orthogonal) decision rules, recursively partitioning the feature space by using a single feature at a time. However, the machine learning and statistics communities have increasingly recognized the advantages of oblique trees, which allow decision rules to incorporate multiple features simultaneously. Several notable oblique tree models have been proposed, including Forest-RC~\citep{breiman2001random}, SPORF~\citep{tomita2020sparse}, TAO~\citep{carreira2018alternating}, and oblique BART~\citep{nguyen2025oblique}.

From a theoretical perspective, oblique trees are also widely recognized as effective approximators. Recent advancements have established them as universal approximators~\citep{cattaneo2024convergence, zhan2022consistency}. However, existing theoretical works often overlook scenarios involving high-dimensional feature spaces. This paper focuses on the theory and practice of optimizing high-dimensional sparse oblique splits within the set 
\begin{equation*}    
    \widehat{W}_{p, s} = \{(\vv{w}, \vv{w}^{\top} \boldsymbol{X}_i) : i \in \{1, \dots, n\}, \vv{w} \in \mathbb{R}^p, \| \vv{w} \|_2 = 1, \| \vv{w} \|_0 \leq s \},
\end{equation*}
where \( \{\boldsymbol{X}_i\}_{i=1}^{n} \) represents a sample of $p$-dimensional feature vectors and $s$ is a user-defined sparsity parameter. The resulting tree model is constructed using these $s$-sparse oblique splits. The dimension $p$ is permitted to grow at an arbitrary polynomial rate relative to $n$, thereby accommodating a wide range of both high-dimensional and low-dimensional settings. Each split \( (\vv{w}, c) \) defines a decision hyperplane that partitions the feature space \( [0, 1]^p \) into two subsets: \( \{\vv{x} \in [0, 1]^p : \vv{w}^{\top} \vv{x} > c\} \) and \( \{\vv{x} \in [0, 1]^p : \vv{w}^{\top} \vv{x} \leq c\} \). These splits are recursively applied to partition the feature space, forming the foundation for subsequent steps in approximating the data-generating functions.

In this paper, we establish Sufficient Impurity Decrease (SID) convergence rates~\citep{chi2022asymptotic, mazumder2024convergence} for oblique trees. Specifically, in the context of sparse oblique splits, we demonstrate that SID depends not only on the convergence rate parameter but also on the sparsity parameter $s_0$ pertaining to the underlying data-generating function. As $s_0$ increases, the SID function class expands, capturing more complex data-generating functions, such as the $s_0$-dimensional XOR function (see Example~\ref{example.xor}). However, we also show that the computational resources required to learn functions in this expanded SID class grow with $s_0$. To achieve the SID convergence rate for a data-generating function with $s_0$-sparse splits, we prove that the computational runtime must scale at least proportionally to $\binom{p}{s_0}$, rather than $\binom{p}{s}$, when using our iterative $s$-sparse oblique trees with $s \geq s_0$ (see \eqref{lowers_0}).

These results are empirically validated through simulation experiments. Our findings suggest that each data-generating function is associated with a specific sparsity level $s_0$ of SID, which reflects its intrinsic approximation difficulty and is often unknown in practice. Additionally, our results reveal a fundamental trade-off between computational cost and statistical accuracy, contributing to the broader literature on this topic~\citep{chandrasekaran2013computational}. These findings, coupled with the uncertainty surrounding intrinsic approximation difficulty, expose a significant challenge in tuning the candidate split set for oblique split optimization. Specifically, an insufficient search space restricts approximation accuracy, whereas an excessively large space results in redundant computational overhead.

To address this tuning difficulty, our second contribution introduces the progressive tree, which optimizes high-dimensional oblique splits within a single tree structure through iterative refinement. Drawing inspiration from transfer learning in neural networks \citep{pratt1992discriminability}, this approach enables efficient optimization without exhaustive parameter searching. Specifically, we initialize the selected split set $\mathcal{S}^{(0)} = \emptyset$ and update it at each iteration $l \in \{1, \dots, b\}$ according to
\begin{equation}\label{oho}
\mathcal{S}^{(l)} = \{ \text{Splits constituting the tree optimized over the set } W_{l} \cup \mathcal{S}^{(l-1)} \},
\end{equation}
where $W_{l} \subseteq \widehat{W}_{p, s}$ is a sampled subset of predetermined size. In this framework, the $l$-th tree is constructed using the union of a newly sampled split set $W_{l}$ and the splits retained from the previous iteration $\mathcal{S}^{(l-1)}$. The underlying heuristic is that as the number of iterations $b$ increases, $\mathcal{S}^{(b)}$ progressively approximates the set of optimized splits for an ideal decision tree of depth $H$ grown over the complete set $\widehat{W}_{p, s}$. Such an ideal construction is typically unfeasible in practice due to prohibitive computational costs. Our iterative refinement approach eliminates the necessity of pre-tuning a static candidate split set size by utilizing $b$ as the primary control parameter. While higher approximation precision scales with $b$, this iterative framework permits early stopping once the splits reach a sufficient quality threshold. Crucially, halting the process at any arbitrary iteration $l$ yields a single decision tree (rather than an ensemble); consequently, the cardinality of the selected split set remains bounded by $\texttt{\#}\mathcal{S}^{(l)} \le 2^{H}$ for a tree of depth $H$, ensuring efficiency.

Our optimization approach \eqref{oho} is more efficient than existing methods for optimizing oblique trees because it does not require advanced tree-building techniques such as bagging, column subsampling~\citep{breiman2001random}, boosting~\citep{friedman2001greedy}, or pruning. However, to maximize predictive accuracy on finite samples, incorporating advanced tree-building techniques is essential.  To achieve this, we propose  an efficient strategy for utilizing the optimized oblique splits in \( \mathcal{S}^{(b)} \). Specifically, after \( b \) iterations, the selected oblique splits \( \mathcal{S}^{(b)} \) can be transferred and integrated into well-established tree-based models, such as Random Forests~\citep{breiman2001random}. These selected splits can be combined with orthogonal splits to enhance stability in prediction performance. This approach ensures that the method fully exploits the information in \( \widehat{W}_{p, s} \) when computational resources are sufficient while maintaining reliable performance improvements even under resource constraints. By adopting this strategy, practitioners can effectively balance computational cost and model performance based on their available resources.

The structure of this paper is as follows. Sections~\ref{Sec2.1.population} to~\ref{sec2.2.1} present our progressive tree algorithms, while Section~\ref{Sec4} introduces our main predictive tree model, Random Forests (RF) + $\mathcal{S}^{(b)}$. Section~\ref{Sec2.4} presents the memory transfer property, which enables step-by-step optimization based on available computational resources. In Section~\ref{Sec3}, we analyze our progressive tree model by introducing the SID criterion for oblique splits. Finally, the empirical study is presented in Sections~\ref{Sec5.0}--\ref{Sec6}. The Python source code for our experiments is publicly available at \url{https://github.com/xbb66kw/ohos}. All technical proofs are provided in the Supplementary Material.

\subsection{Related work}\label{Sec1.1}

This work provides a formal analysis of optimizing high-dimensional oblique trees, addressing the efficiency challenges inherent in sparse oblique split optimization. We establish standard SID convergence rates for oblique trees in both high-dimensional regimes ($p \gg n$) and lower-dimensional cases, expanding upon previous orthogonal split analyses that assumed $s_0=1$~\citep{chi2022asymptotic, mazumder2024convergence}. We demonstrate that the SID function class expands with oblique splits, enabling the model to capture complex structures like multidimensional XOR problems. This is a critical advancement, as orthogonal trees are inconsistent for 2-dimensional XOR problems~\citep{tan2024statistical, syrgkanis2020estimation}. Furthermore, whereas existing research~\citep{cattaneo2024convergence, zhan2022consistency} establishes consistency or slow logarithmic rates for ridge functions when $p \leq n$, our results provide a faster convergence framework tailored to the sparsity levels of high-dimensional settings.

Existing implementations of oblique trees primarily differ in how they optimize oblique splits. OC1~\citep{murthy1994system} and the original TAO~\citep{carreira2018alternating} refine the split hyperplane at each node based on the optimal orthogonal split. Optimal greedy oblique tree~\citep{bertsimas2017optimal, dunn2018optimal} directly optimizes splits using the full available oblique split set. Other methods leverage linear discriminants~\citep{lopez2013fisher, loh1988tree}, logistic regression~\citep{carreira2018alternating, truong2009fast}, or the more general projection pursuit~\citep{da2021projection} for split optimization. To mitigate local minima, randomization is often incorporated. For instance, OC1, RR-RF~\citep{blaser2016random}, SPORF~\citep{tomita2020sparse}, and Forest-RC~\citep{breiman2001random} introduce random normal vectors into their oblique split optimization to varying degrees. Among these, SPORF and Forest-RC are particularly comparable to our approach. We provide a brief overview of these methods below.

SPORF and Forest-RC both rely on independent bagging trees for making predictions.
At each node of a bagging tree, both methods sample a set of random oblique splits to optimize them. This set of splits is similar to our \( W_l \), but with their own choices for the number of splits  and sparsity \( s \). In contrast, we introduce an iterative training approach where each tree passes its selected splits to the next, progressively refining split quality. This `memory transfer` mechanism provides practical advantages—it allows early stopping when sufficient accuracy is reached or continued training if additional computational resources are available. This is one of our key practical contributions. Unlike our approach, existing oblique tree methods focus solely on optimizing individual trees without leveraging memory transfer, making it difficult to balance computational efficiency and predictive accuracy.

\subsection{Notation}
Let $(\Omega, \mathcal{F}, \mathbb{P})$ be a probability space. For any $a, b \in \mathbb{R}$, we define $a \wedge b = \min\{a, b\}$ and $a \vee b = \max\{a, b\}$. The ceiling function $\lceil x \rceil$ and floor function $\lfloor x \rfloor$ denote the smallest integer $n \geq x$ and the largest integer $n \leq x$, respectively, while $\log{x}$ refers to the natural logarithm. A set is understood as a collection of distinct elements, and each vector $\vv{x} \in \mathbb{R}^p$ is written as $\vv{x} = (x_1, \dots, x_p)^\top$. We define $\|\vv{x}\|_{0} = \texttt{\#} \{|x_{j}| > 0\}$ as the number of non-zero components and $\|\vv{x}\|_{l}^l = \sum_{j=1}^{p} |x_{j}|^{l}$ for any integer $l \geq 1$. Unless otherwise specified, the volume of a $p$-dimensional set refers to its Lebesgue measure. The indicator function is denoted by $\boldsymbol{1}_{A}$ or $\boldsymbol{1}\{A\}$, taking the value 1 if event $A$ occurs and 0 otherwise. Furthermore, following established conventions, we use numeric subscripts to distinguish technical parameters such as $K_{0}$ and $s_0$; these subscripts serve as identifiers only and do not imply a specific ordering or hierarchy.

\section{Optimization of oblique splits via iterative refinement}\label{Sec2}

At the population level, let the response $Y$, with $\mathbb{E}(Y^2) < \infty$, and the $p$-dimensional feature vector $\boldsymbol{X} \in [0, 1]^p$ represent the target variable and its predictors, respectively. At the sample level, we consider $n$ i.i.d. (independent and identically distributed) observations $\{Y_i, \boldsymbol{X}_i\}_{i=1}^n$ drawn from the same distribution as $(Y, \boldsymbol{X})$.

\subsection{Population algorithm of oblique decision trees}\label{Sec2.1.population}

In this section, we examine the regression problem of predicting the response \( Y \) using the \( p \)-dimensional feature vector \( \boldsymbol{X} \) at the population level. Let \( N_{1} = \dots = N_{H} = \emptyset \) and \( N_{0} = \{(\emptyset, (\vv{0}, 0), [0, 1]^p)\} \) represent the structure of the tree at depth \( h \). For each \( h \in \{1, \dots, H\} \), we perform the following operation for each  \( (\boldsymbol{t}^{'}, (\vv{v}, a), \boldsymbol{t}) \in N_{h-1} \), 
\begin{equation}
    \begin{split}\label{tspursuit.1}
        V & =  \inf_{(\vv{w}^{'}, c^{'})\in \mathcal{W}}  \mathbb{E} [ \textnormal{Var} ( Y \mid \boldsymbol{1}_{\boldsymbol{X} \in \boldsymbol{t}} , \boldsymbol{1}_{\vv{w}^{'\top}\boldsymbol{X}>c^{'}} ) \boldsymbol{1}_{\boldsymbol{X} \in \boldsymbol{t}} ] ,\\
    (\vv{w}, c)  \in  E(\rho,  \boldsymbol{t},\mathcal{W}) & \coloneqq \argmin_{ (\vv{w}^{'}, c^{'})\in \mathcal{W}} \max \Big\{ V  + 2\rho, \ \ \mathbb{E} [ \textnormal{Var} ( Y \mid \boldsymbol{1}_{\boldsymbol{X} \in \boldsymbol{t}} , \boldsymbol{1}_{\vv{w}^{'\top}\boldsymbol{X}>c^{'}} ) \boldsymbol{1}_{\boldsymbol{X} \in \boldsymbol{t}} ]  \Big\}.
    \end{split}
\end{equation}
The node set is then updated by replacing $\boldsymbol{t}$ with two children,
$\boldsymbol{t}_1=\{\vv{x}\in\boldsymbol{t}:\vv{w}^{\top}\vv{x}>c\}$ and
$\boldsymbol{t}_2=\boldsymbol{t}\setminus\boldsymbol{t}_1$, that is,
$N_h\leftarrow N_h\cup\{(\boldsymbol{t},(\vv{w},c),\boldsymbol{t}_1),
(\boldsymbol{t},(\vv{w},c),\boldsymbol{t}_2)\}$. Note that $\boldsymbol{t}$ is a child node resulting from the split of its parent node $\boldsymbol{t}'$.

The set $\mathcal{W}$ in \eqref{tspursuit.1} denotes a generic candidate set of oblique splits. This set satisfies $\mathcal{W}\subseteq\widehat{W}_{p,s}$, where $\widehat{W}_{p,s}$ is given in the Introduction. As a clarifying remark, $V$ in \eqref{tspursuit.1} represents the minimum achievable residual variance within the node $\boldsymbol{t}$ after an optimal split from $\mathcal{W}$ is applied. In comparison, $\mathbb{E}[\text{Var}(Y \mid \boldsymbol{1}_{\boldsymbol{X} \in \boldsymbol{t}}) \boldsymbol{1}_{\boldsymbol{X} \in \boldsymbol{t}}]$ is the node variance of $Y$ prior to any splitting within the region defined by $\boldsymbol{t}$. The difference between these two quantities signifies the maximum variance reduction (or improvement in purity) attainable by a single split selected from $\mathcal{W}$ at the population level.

However, in practice, there would be estimation variance due to finite samples, making it difficult to pinpoint the exact population minimizer. The parameter $\rho\ge0$ accounts for this estimation uncertainty. When $\rho=0$, $E(\rho,\boldsymbol{t},\mathcal{W})$ consists of strictly optimal splits within $\mathcal{W}$. Conversely, for an appropriately chosen $\rho>0$ (which typically depends on the sample size), the set expands to include approximately optimal splits. This reflects the reality that multiple candidate splits may be statistically indistinguishable from the true population optimum when working with finite data.

\subsection{Sample algorithm of oblique decision trees}\label{Sec2.2.boosting}

Let $\widehat{N}_{1} = \dots = \widehat{N}_{H} =\emptyset$, and $\widehat{N}_{0} = \{(\emptyset, (\vv{0}, 0), [0, 1]^p)\} $ initially. For each $h\in \{1, \dots, H\}$, we perform the following operation in each \( (\boldsymbol{t}^{'}, (\vv{v}, a), \boldsymbol{t}) \in \widehat{N}_{h-1} \), 
\begin{equation}
    \begin{split}
\label{tspursuit.4}
           L(\vv{w}^{'}, c^{'}, \vv{\beta}^{'}, \boldsymbol{t}) & =   \sum_{i=1}^{n} \boldsymbol{1}_{ \boldsymbol{X}_{i}\in \boldsymbol{t} }\times  \big[   Y_{i}  -  \beta_{1}^{'} \times \boldsymbol{1}_{ \vv{w}^{'\top} \boldsymbol{X}_{i} > c^{'}  } - \beta_{2}^{'} \times \boldsymbol{1}_{ \vv{w}^{'\top} \boldsymbol{X}_{i} \le c^{'}  }     \big]^2,\\
           (\vv{w}, c, \vv{\beta}) & \in  \argmin_{ (\vv{w}^{'}, c^{'})\in\mathcal{W}, \vv{\beta}^{'} \in \mathbb{R}^{2} } L (\vv{w}^{'}, c^{'}, \vv{\beta}^{'}, \boldsymbol{t}) ,
    \end{split}
\end{equation}
with $\widehat{N}_{h} \leftarrow  \widehat{N}_{h}\cup \{ (\boldsymbol{t}, (\vv{w}, c) , \boldsymbol{t}_{1}), (\boldsymbol{t}, (\vv{w}, c) ,\boldsymbol{t}_{2})\}$, $\boldsymbol{t}_{1} = \{\vv{x} \in \boldsymbol{t}: \vv{w}^{\top}\vv{x}>c\}$, and $\boldsymbol{t}_{2} = \boldsymbol{t} \backslash\{\boldsymbol{t}_{1}\}$.

In \eqref{tspursuit.4}, we define $\widehat{\beta}(\boldsymbol{t}_{1})$ and $\widehat{\beta}(\boldsymbol{t}_{2})$ such that $(\widehat{\beta}(\boldsymbol{t}_{1}), \widehat{\beta}(\boldsymbol{t}_{2}))^{\top} = \vv{\beta}$, where $\vv{\beta}$ denotes the least squares coefficients. Additionally, we ensure that ties are broken randomly. After the tree training, \(\texttt{\#}\widehat{N}_{h} = 2^{h}\) for each $h\in \{1, \dots, H\}$, but the number of distinct splits in $\{(\vv{w}, c) : (\boldsymbol{t}^{'}, (\vv{w}, c), \boldsymbol{t}) \in \widehat{N}_{h} \}$ is $2^{h-1}$. For the edge cases related to \eqref{tspursuit.4}, we define the following: if a daughter node contains no samples, the corresponding least squares coefficient is set to zero. Additionally, if \(\boldsymbol{t}\) is an empty set, \(L(\vv{w}^{'}, c^{'}, \vv{\beta}^{'}, \boldsymbol{t})\) is defined to be zero.

\subsection{Iterative refinement algorithm}\label{sec2.2.1}

The \emph{ideal tree} is defined by the solution to~\eqref{tspursuit.4} using the candidate set $\mathcal{W} = \widehat{W}_{p,s}$. Directly optimizing an ideal tree incurs substantial computational overhead and may also lead to memory storage issues. To establish a more practical alternative, consider a sequence of oblique split sets $\{\Lambda_q\}_{q \ge 1}$ defined by
\begin{equation}
    \label{lambda.sampling}
\Lambda_q
=
\left\{
\bigl(\vv{w}_q,\; \vv{w}_q^{\top}\boldsymbol{X}_i\bigr)
:\;
i \in \{1,\dots,n\}
\right\},
\end{equation}
where each weight vector $\vv{w}_q$ is sampled independently from $\{\vv{w}\in\mathbb{R}^p:\|\vv{w}\|_2=1,\ \|\vv{w}\|_0\le s\}$.
The \emph{one-shot tree} is obtained by running~\eqref{tspursuit.4} with
$\mathcal{W} = \bigcup_{q=1}^{B} \Lambda_q$ for $B < \infty$, yielding node estimates $\widehat{N}_{1}^{(0)}, \dots, \widehat{N}_{H}^{(0)}$. When $B = \infty$, the one-shot tree is defined to be the ideal tree.

The parameter $B$ is analogous to the \textsf{max\_features} parameter in oblique tree software such as F-RC, as both define the number of weight vectors sampled for candidate split optimization. As noted in the Introduction, selecting an appropriate value for $B$ involves a difficult trade-off: an insufficient $B$ restricts the search space and limits accuracy, whereas an excessively large $B$ results in prohibitive computational overhead. To formally analyze this trade-off and provide a practical resolution, we introduce the \emph{progressive tree} framework.

A progressive tree represents a single oblique tree of depth $H$ whose splits are updated iteratively. After a sufficiently large number of update rounds $b$, the resulting progressive tree is comparable to a one-shot tree constructed with $\mathcal{W} = \bigcup_{q=1}^{B} \Lambda_q$. The iterative refinement construction proceeds as follows: we initialize $\mathcal{S}^{(0)} = \emptyset$ and, for $l \in \{1,\dots,b\}$, repeat the following steps.
\begin{enumerate}
\item[(i)] Let \(W_l = \bigcup_{q \in S} \Lambda_q\), where \(S \subseteq \{1,\ldots,B\}\) is constructed as follows. If \(B < \infty\), \(S\) is sampled uniformly at random from \(\{1,\ldots,B\}\). If $B = \infty$, we define $S$ as the $l$th set of $\texttt{\#}S$ consecutive integers for each $l \ge 1$.

\item[(ii)] Run \eqref{tspursuit.4} with $\mathcal{W} = \mathcal{S}^{(l-1)} \cup W_l$, and obtain $\widehat{N}_h^{(l)} = \widehat{N}_h$ for each $h \in \{1,\ldots,H\}$.

\item[(iii)] Let $\mathcal{S}^{(l)}$ denote the set of oblique splits that constitute the tree obtained in step (ii).

\end{enumerate}

After $b$ iterations, the procedure yields a single oblique tree model of depth $H$, parameterized by the set of splits $\mathcal{S}^{(b)}$. For clarity, $\mathcal{S}^{(l)}$ is the collection of all splits defining the progressive tree structure $\widehat{N}_1^{(l)}, \dots, \widehat{N}_H^{(l)}$ at each iteration $l \in \{1, \dots, b\}$; this indexing distinguishes the progressive tree from the one-shot structure $\widehat{N}_1^{(0)}, \dots, \widehat{N}_H^{(0)}$. Notably, the iterative procedure \eqref{oho} corresponds to the limiting regime where $B = \infty$. Furthermore, in our numerical experiments, we fix the random subset size at $\texttt{\#}S = 100$.

In the subsequent sections, we demonstrate how increasing $B$ and $b$ improves prediction accuracy from both theoretical and empirical perspectives within the $B < \infty$ setting. Furthermore, we empirically evaluate the impact of $b$ when $B = \infty$ to provide a comprehensive analysis of these tuning parameters. As our results illustrate, the $B = \infty$ configuration effectively eliminates the necessity of pre-tuning both $B$ and $b$ simultaneously.

The iterative execution of our algorithm facilitates the transfer of knowledge from selected splits to subsequent iterations of the learning task. We adopt the term ``knowledge'' following established conventions in neural network literature, where model coefficients such as weights and biases are frequently characterized as ``memory'' or ``knowledge''~\citep{frosst2017distilling, pratt1992discriminability}. Given the documented structural relationship between oblique splits and neural network parameters \citep{wan2021nbdt, frosst2017distilling}, this terminology reflects how our refinement process retains and builds upon previously learned split information.

While iterative update strategies are common in existing oblique tree frameworks, our approach differs fundamentally in its optimization objective and theoretical foundation. For instance, Oblique BART \citep{nguyen2025oblique, chipman2010bart} utilizes Markov Chain Monte Carlo iterations to generate samples from a fixed posterior distribution, gradually refining its additive trees to better approximate the posterior distribution. Meanwhile, TAO \citep{carreira2018alternating} employs an alternating optimization strategy that iteratively updates the parameters of a single node while keeping all other nodes fixed, effectively performing a local search via coordinate descent. In contrast, our iterative refinement is specifically designed to achieve asymptotic equivalence with a one-shot tree constructed over a massive candidate set. This strategy is distinct because it is supported by formal theoretical guarantees established through the \emph{memory transfer property}, which we investigate rigorously in Proposition~\ref{prop1} in the next subsection.

\subsection{Memory transfer property}\label{Sec2.4}

Now, we formally introduce how the memory transfer property allows us to progressively shape the progressive tree into the one-shot tree. Proposition~\ref{prop1} below ensures that our progressive tree is asymptotically equivalent to the one-shot tree as $b \rightarrow \infty$, requiring only Condition~\ref{regular.tree} to avoid minor edge-case considerations.

% A different quality of equivalence in their ability in controlling estimation variance is derived in Corollary~\ref{corollary1} later, with explicit lower bound that is required for ensuring such results.

 \begin{condition}[Regular splits] \label{regular.tree}
  %\textnormal{1)} 
  The $\mathbb{L}^2$ loss in \eqref{tspursuit.4}, when splitting $\boldsymbol{t}$, exhibits ties for two splits, $(\vv{w}, c)$ and $(\vv{u}, a)$, if and only if 
  $\sum_{i=1}^n \boldsymbol{1}_{\boldsymbol{X}_{i}\in \boldsymbol{t}} |\boldsymbol{1}_{\vv{w}^{\top} \boldsymbol{X}_{i} > c} - \boldsymbol{1}_{\vv{u}^{\top} \boldsymbol{X}_{i} > a}| \in \{ 0, \sum_{i=1}^n \boldsymbol{1}_{\boldsymbol{X}_{i}\in \boldsymbol{t}}\}$.

 \end{condition}

 In Proposition~\ref{prop1}, the sample is given, and all hyperparameters, except for the number of search iterations, are fixed. In addition, if the two trees at the \( h \)th layer, \( \widehat{N}_{h}^{(0)} \) and \( \widehat{N}_{h}^{(b)} \) for some $b\ge 1$, are \emph{sample-equivalent}, then for each \( (\boldsymbol{t}^{'}, (\vv{w}, c), \boldsymbol{t}) \in \widehat{N}_{h}^{(0)} \), there exists some \( (\boldsymbol{s}^{'}, (\vv{u}, a), \boldsymbol{s}) \in \widehat{N}_{h}^{(b)} \) such that  $\sum_{i=1}^n |\boldsymbol{1}_{\boldsymbol{X}_{i} \in \boldsymbol{t}^{'}} - \boldsymbol{1}_{\boldsymbol{X}_{i} \in \boldsymbol{s}^{'}}| = \sum_{i=1}^n |\boldsymbol{1}_{\boldsymbol{X}_{i} \in \boldsymbol{t}} - \boldsymbol{1}_{\boldsymbol{X}_{i} \in \boldsymbol{s}}| = 0.$
This implies that while the feature space partitions, \( \{ \boldsymbol{t}: (\boldsymbol{t}^{'}, (\vv{u}, a), \boldsymbol{t}) \in \widehat{N}_{h}^{(l)}\} \) for \( l \in \{0, b\} \), may differ, the corresponding sample index partitions, \( \{\{i: \boldsymbol{X}_{i} \in \boldsymbol{t} \}: (\boldsymbol{t}^{'}, (\vv{u}, a), \boldsymbol{t}) \in \widehat{N}_{h}^{(l)}\} \) for \( l \in \{0, b\} \), remain identical. The proof of Proposition~\ref{prop1} is given in Section~\ref{proof.prop1}.

\begin{proposition}
    \label{prop1}
    Assume Condition~\ref{regular.tree}. Then,
    \textnormal{1)} for each $b_1\ge 1$, each $b$ with $b\ge b_{1}$, and each $h\in \{1, \dots, H\}$, the event  $\Theta_{b_{1}, h} = \{ \widehat{N}_{q}^{(0)} \textnormal{ and } \widehat{N}_{q}^{(b_{1})} \textnormal{ are sample-equivalent  for } q\in \{1,\dots, h\} \}$ is a subset of $\Theta_{b, h}$.
    \textnormal{2)} With probability approaching one as $b\rightarrow\infty$, $\widehat{N}_{h}^{(0)}$ and $\widehat{N}_{h}^{(b)}$ are sample-equivalent for each $h\in \{1, \dots, H\}$.
\end{proposition}

The first part of Proposition~\ref{prop1} establishes a memory transfer property, ensuring that information learned prior to the $b_1$th iteration is preserved across all subsequent steps. Building on this, the second part guarantees that the progressive tree, represented by $\widehat{N}_{1}^{(b)}, \dots, \widehat{N}_{H}^{(b)}$, becomes equivalent to the one-shot tree, represented by $\widehat{N}_{1}^{(0)}, \dots, \widehat{N}_{H}^{(0)}$, in probability as $b \rightarrow \infty$. This foundational equivalence requires only Condition~\ref{regular.tree}. To establish statistical accuracy beyond this baseline equivalence, further regularity conditions are necessary. With these conditions additionally assumed, Corollary~\ref{corollary1} in Section~\ref{Sec3.1} demonstrates that the progressive tree controls estimation variance as effectively as its one-shot counterpart, and it derives the theoretical lower bounds on $b$ required for these results. Ultimately, these variance control findings provide the formal foundation for establishing the SID convergence rates detailed in Theorem~\ref{theorem2} in Section~\ref{Sec3.2}.

Among existing models, Forest-RC (F-RC)~\citep{breiman2001random} and greedy oblique trees~\citep{dunn2018optimal} can approximate the ideal tree given sufficient computational resources. An F-RC model is an ensemble of independently trained trees using bagging; in this paper, an individual tree within this ensemble is termed a Breiman oblique tree.  We provide a detailed review and comparison of this approach in Table~\ref{tab:comparison} of Section~\ref{Sec2.5}.

Other existing oblique trees, such as oblique BART~\citep{nguyen2025oblique} and TAO~\citep{carreira2018alternating}, sequentially update splits using their respective optimization strategies. In contrast, optimal oblique trees~\citep{bertsimas2017optimal, zhu2020scalable} aim to jointly optimize all splits throughout the tree structure. While joint optimization provides a more global solution, it is generally more computationally intensive than greedy methods like F-RC, BART, or our proposed approach.

\subsection{Comparison with Breiman oblique trees}\label{Sec2.5}

\begin{table}[ht]
\centering
\small
\begin{tabular}{p{2cm} |  p{4.5cm}p{4cm}p{3cm}}
\hline
Tree Type &  Optimization strategy &  Transfer strategy & Usage in this paper\\ [0.5ex] 
\hline
Breiman  Oblique Tree &  Sample a set of oblique splits for node-wise optimization at \textbf{each node}, with a size equivalent to that of $\bigcup_{q=1}^{B} \Lambda_q$ &  Not used. & Base learners of F-RC \\
\hline
One-Shot Tree   &  Running~\eqref{tspursuit.4} with candidate set \(\mathcal{W} = \bigcup_{q=1}^{B} \Lambda_q\). Optimize splits over $\mathcal{W}$ for \textbf{all nodes}. &  Not used.& For theoretical clarity only.\\
\hline
Progressive Tree   & At each iteration, sample a set of oblique splits as described in Section~\ref{sec2.2.1}, and optimize splits for \textbf{all nodes} based this split set. & Iterative refinement for $b$ iterations.  & $B < \infty$ is analyzed theoretically, while $B = \infty$ is recommended in practice.  \\
\hline
Ideal Tree   & Running~\eqref{tspursuit.4} with full candidate set \(\mathcal{W} = \widehat{W}_{p,s}\). Optimize splits over $\mathcal{W}$ for \textbf{all nodes}. & Not used.  & For theoretical clarity only. \\
\hline
\end{tabular}
\caption{Comparison of single tree models. The parameter $B$ corresponds to the \textsf{max\_features} setting in standard implementations of Breiman oblique tree. In this tree model, each node is optimized over the candidate split set $\{(\vec{w}_{k}, \vec{w}_{k}^{\top}\boldsymbol{X}_i) \mid 1\le k\le B, 1\le i\le n\}$, where $\vec{w}_{1}, \dots, \vec{w}_{B}$ are sampled independently from \eqref{J.s} at each node.}
% All methods share identical hyperparameters for tree depth and node constraints. 
\label{tab:comparison}
\end{table}

The Breiman oblique tree, the base learner of F-RC~\citep{breiman2001random}, is closely related to our one-shot tree given $\textsf{max\_features} = B$ and $\textsf{feature\_combinations} = s$. Here, $B$ defines the number of candidate weight vectors for optimization, while $s$ determines the sparsity of each vector. These two models differ only in their optimization strategies, as shown in Table~\ref{tab:comparison}.

Proposition~\ref{prop3} establishes the probabilistic equivalence between the progressive tree, the one-shot tree, the Breiman oblique tree, and the ideal tree as $B$ approaches infinity. For the progressive tree, this equivalence further requires the number of iterations $b$ to approach infinity, a parameter specific to its iterative search strategy. Under Condition~\ref{regular.tree}, equivalence implies that all splits in the model match those of the ideal tree. If this condition is not met, ideal tree splits may not be unique. In such cases, equivalence signifies that the splits of the tree match one of the possible ideal tree. For this proposition, the training sample and hyperparameters such as depth $H$ and sparsity $s$ are fixed. The detailed proof is provided in Section~\ref{proof.prop3}.

\begin{proposition}\label{prop3}
Fix a sample $\{ \boldsymbol{X}_i, Y_i \}_{i=1}^{n}$ where $\boldsymbol{X}_i \in [0, 1]^p$ for all $1 \le i \le n$. A Breiman oblique tree and a one-shot tree are each equivalent to their corresponding ideal trees with probability approaching one as $B \to \infty$. Furthermore, under Condition~\ref{regular.tree}, for a sufficiently large finite $B$ or the limiting case $B = \infty$ (see Section~\ref{sec2.2.1} for details), the progressive tree is equivalent to its corresponding ideal tree with probability approaching one as $b \to \infty$.
\end{proposition}

This proposition implies that the performance gap between the progressive tree, the one-shot tree, and the Breiman oblique tree vanishes as $b$ and $B$ increase. Because the training sample is fixed, this convergence in probability pertains exclusively to the randomness induced by the algorithmic split sampling, rather than the data generating process. As we will establish later in Theorem~\ref{theorem2}, increasing $b$ and $B$ improves theoretical convergence rates for the progressive tree; however, while our theoretical analysis establishes that exceedingly large parameters are sufficient to guarantee the resolution of complex structures, such as multidimensional XOR problems, our empirical evidence indicates that large values are indeed required in practice. Given that the underlying data-generating process is typically unknown, a fixed selection of $B$ or $b$ creates a practical dilemma: a large value wastes computational time on simple tasks, whereas a small value fails to approximate complex ones. In Sections~\ref{Sec5.0} to \ref{Sec6}, we demonstrate that our iterative refinement approach effectively addresses this trade-off. By removing the need for predetermined parameters, this method enables the search process to dynamically adapt to the underlying complexity of the problem.

\section{Analysis of the iterative refinement algorithm}\label{Sec3} 

\subsection{Estimation variance of high-dimensional oblique splits}\label{Sec3.1}

We require the following regularity conditions for our theoretical analysis. The notation in Condition~\ref{regularity.1} follows that of Section~\ref{Sec2}; specifically, $s$-sparse oblique splits are used for growing the proposed oblique tree models. This parameter should not be confused with $s_0$, which represents the underlying approximation difficulty (see Condition~\ref{sid}).

\begin{condition}
\label{regularity.1}
Assume either \textnormal{(a)} or \textnormal{(b)}:  
\textnormal{(a)} \( \mathbb{P}(\boldsymbol{X} \in \{0, 1\}^p) = 1 \), and for every \( \vv{v} \in \{0, 1\}^s \) and $J\subseteq \{1, \dots, p\}$ with \( \texttt{\#}J = s \), we have \( \mathbb{P}((X_j, j \in J)^{\top} = \vv{v}) \ge D_{\min} 2^{-s} \) for some constant \( D_{\min} > 0 \), \textnormal{(b)} \( \mathbb{P}(\boldsymbol{X} \in [0, 1]^p) = 1 \), and for any measurable set \( A \subseteq [0, 1]^p \), \( \mathbb{P}(\boldsymbol{X} \in A) \leq (\textnormal{Volume of } A) \times D_{\max} \) for some constant \( D_{\max} \geq 1 \).  Assume that $Y = m(\boldsymbol{X}) + \varepsilon$, where $\varepsilon$ is an independent model error with $\mathbb{E} (\varepsilon) = 0$ and $|\varepsilon| \le M_{\epsilon}$ almost surely for some $M_{\epsilon} \ge 1$, and $\sup_{\vv{x} \in [0, 1]^p} |m(\vv{x})|$ is bounded by some constant. Assume that $(\boldsymbol{X}_{1}, Y_{1}), \dots, (\boldsymbol{X}_{n}, Y_{n}), (\boldsymbol{X}, Y)$ are i.i.d.
\end{condition}

Condition \ref{regularity.1} imposes mild regularity assumptions to ensure the statistical stability of the tree-partitioning procedure. In the discrete setting, the constant $D_{\min}$ enforces a uniform lower bound on the probability of all $s$-dimensional feature vectors, which excludes perfectly redundant interactions within any sparse subspace. In the continuous setting, the constant $D_{\max}$ prevents probability mass from concentrating on lower-dimensional subsets, ensuring that node probabilities scale proportionally with their geometric volume. 

These lower and upper probability bounds are standard in the theoretical analysis of tree-based methods and are sufficient to control empirical variance estimates along the partitioning path. For example, $D_{\min} = 1$ for the discrete uniform $\{0, 1\}^p$, while $D_{\max} = 1$ for the continuous uniform case on $[0, 1]^p$. For simplicity, we assume the model error is almost surely bounded by $M_{\epsilon} \ge 1$. Similar conclusions hold under sub-Gaussian noise or finite high-order moment assumptions through standard truncation arguments. In Theorem~\ref{theorem1} below, recall the definition of $E(\cdot, \cdot, \cdot)$ from \eqref{tspursuit.1}. For each $q \le p$, we define 
$$W_{p,q} = \{ (\vv{w}, c) : \vv{w} \in \mathbb{R}^p, \|\vv{w}\|_2 = 1, \|\vv{w}\|_0 \le q, c \in \mathbb{R} \}.$$ 
The proof of Theorem~\ref{theorem1} is provided in Section~\ref{proof.theorem1}.

\begin{theorem}
    \label{theorem1}
    Assume Condition~\ref{regularity.1}. Run \eqref{tspursuit.4} with an arbitrary split set $\mathcal{W}\subseteq W_{p, s}$. For all large $n$,  each $1 \le p \le n^{K_{0}}$ for an arbitrary fixed constant $K_{0} > 1$, each $M_{\epsilon} \ge 1$, each $s\ge 1$, each $H\ge 1$ with $\sqrt{2}n^{-\frac{1}{2}} s \sqrt{H(1\vee\log{s})} \log{n}\le 1$, the following holds. For  each $h \in \{1, \dots, H\}$ and every $(\boldsymbol{t}^{'}, (\vv{w}, c), \boldsymbol{t})\in \widehat{N}_{h}$, it holds that $(\vv{w}, c) \in  E(\rho_{n}, \boldsymbol{t}^{'}, \mathcal{W})$, in which $\rho_{n} = C_{0}  \times n^{-\frac{1}{2}} s \sqrt{H(1\vee\log{s})} \log{n} \times \left(M_{\epsilon} + \sup_{\vv{x} \in [0, 1]^p} |m(\vv{x})|\right)^2$ for sufficiently large $C_{0} >0$  with probability at least $1 - 10\exp{ (- (\log{n})^2 / 2)}$.
    
\end{theorem}

Theorem \ref{theorem1} establishes that the splits selected by the sample-based algorithm \eqref{tspursuit.4} are nearly as effective as the optimal ones. Specifically, when implementing \eqref{tspursuit.4} with a candidate split set $\mathcal{W}$, the resulting sample tree satisfies \eqref{tspursuit.1}, where the theoretical $N_h$ and $\rho$ are respectively replaced by the sample-based $\widehat{N}_h$ constructed via \eqref{tspursuit.4} and the margin $\rho_n$. This result demonstrates that each split in the sample tree reduces variance as effectively as the best possible split in $\mathcal{W}$, subject to an estimation variance margin of $2\rho_n$, according to the definition of $E(\cdot, \cdot, \cdot)$. To apply this theorem, one may set $\mathcal{W} = \widehat{W}_{p, s}$ for the ideal tree or $\mathcal{W} = \cup_{q=1}^{B} \Lambda_{q}$ for the one-shot tree, as detailed in Table \ref{tab:comparison}. Notably, Theorem~\ref{theorem1} does not pertain to the progressive tree.

% In Corollary~\ref{corollary1}, we consider a progressive tree constructed through $b$ rounds of iterative refinement.

 Theorem~\ref{theorem1} provides non-asymptotic results, accounting for various parameters, including the model error upper bound \( M_{\epsilon} \), tree depth \( H \), sparsity \( s \), and the number of features \( p \). These results are further used in Corollary~\ref{corollary1} below, where $\widehat{N}_{h}^{(b)}$ represents the progressive tree structure after $b$ rounds of iterative refinement, with $\mathcal{W} = \cup_{q=1}^{B} \Lambda_{q}$. In Corollary~\ref{corollary1}, we define an event $\mathcal{B}$ such that, on $\mathcal{B}$, every split $(\vv{w}, c)$ satisfies $(\vv{w}, c) \in E(\rho_{n}, \boldsymbol{t}^{'}, \cup_{q=1}^{B} \Lambda_{q})$ for each tree depth $h \in \{1, \dots, H\}$ and every node transition $(\boldsymbol{t}^{'}, (\vv{w}, c), \boldsymbol{t})\in \widehat{N}_{h}^{(b)}$. Under event $\mathcal{B}$, the progressive tree is equivalent to its corresponding one-shot tree. Each split in these sample trees reduces variance as effectively as the best possible split in $\mathcal{W} = \cup_{q=1}^{B} \Lambda_{q}$, subject to an estimation margin of $2\rho_n$.
 
 Event $\mathcal{B}$ depends on the sample size $n$, iteration count $b$, and other configuration parameters. It also depends on the training data $\mathcal{X}_n = \{\boldsymbol{X}_i, Y_{i}\}_{i=1}^{n}$ and the set of randomly sampled candidate splits $\bigcup_{q=1}^B \Lambda_{q}$. We omit these dependencies for brevity. Following Section~\ref{sec2.2.1}, $\texttt{\#}S$ denotes the size of the random subset drawn from $\{1, \dots, B\}$ at each step. This parameter is held as a fixed constant throughout our data experiments. Corollary~\ref{corollary1} establishes a probability upper bound for the complement $\mathcal{B}^c$, given sufficiently many iterations. The full proof is in Section~\ref{proof.corollary1}.

\begin{corollary}
    \label{corollary1}

Assume Condition~\ref{regular.tree}, and all conditions for Theorem~\ref{theorem1} with $\rho_{n}$ given in the same theorem. Then, for all large $n$, each $1 \le p \le n^{K_{0}}$ for an arbitrary fixed constant $K_{0} > 1$, each $M_{\epsilon} \ge 0$, each $s\ge 1$, each $H\ge 1$ with $\sqrt{2}n^{-\frac{1}{2}} s \sqrt{H(1\vee\log{s})} \log{n}\le 1$, and either \textnormal{(i)} or \textnormal{(ii)}:
\begin{equation*}
	\begin{split}		
 &\textnormal{(i) Condition}~\ref{regularity.1}(b) \textnormal{ holds with } \texttt{\#}S \times b  \ge n2^{H} B\log{\left(2^H B\right)} ,\\
& \textnormal{(ii) Condition}~\ref{regularity.1}(a) \textnormal{ holds with } \texttt{\#}S \times b  \ge D_{\min}^{-1}2^{H + s + 1} B\log{\left(2^H B\right)} \textnormal{ and  } s < \frac{\log{n}}{2},
\end{split}
\end{equation*}
it holds that $\mathbb{P}(\mathcal{B}^c ) \le n^{-1} + B^{-1}$. 
\end{corollary}

Corollary~\ref{corollary1} implies that, given sufficient samples, the probabilistic difference between these tree types, $\mathbb{P}(\mathcal{B}^c)$, is controlled by ensuring adequate computational resources. This cost is characterized by the lower bounds on $\texttt{\#}S \times b$ derived in the corollary, which both include a factor of $2^H$ to represent the maximum number of splits in a tree of depth $H$. These bounds also account for the number of distinct candidate biases for each weight vector $\vv{w} \in \mathcal{J}(s)$ based on the sample, quantified as $\texttt{\#} \{ (\vv{w}, \vv{w}^{\top} \boldsymbol{X}_{i}): i \in \{1, \dots, n\} \}$, where
\begin{equation}
    \label{J.s}
    \mathcal{J}(s)  =\{\vv{w}\in\mathbb{R}^p:\|\vv{w}\|_2=1,\ \|\vv{w}\|_0\le s\}.
\end{equation}
In the continuous case (i), this cardinality can reach $n$, whereas in the discrete case (ii), it is bounded by $2^s$ because the set contains only unique elements. Remaining factors in the lower bounds result from technical calculations. These bounds are utilized in \eqref{sb1} and \eqref{sb2} of Theorem~\ref{theorem2}, and are further elaborated in the discussion following Remark~\ref{s_s_0} and in \eqref{lowers_0}, respectively. Finally, $B$ must be sufficiently large to ensure a satisfactory approximation as detailed in Theorem~\ref{theorem2}.

Both Proposition~\ref{prop1} and Corollary~\ref{corollary1} establish equivalences between the progressive tree and its corresponding one-shot tree, although they address slightly different contexts. Specifically, Corollary~\ref{corollary1} establishes that both models share the same upper bound on estimation variance for each oblique split. These results provide the theoretical foundation for the consistency analysis in Theorem~\ref{theorem2}. Furthermore, the corollary explicitly derives the lower bounds on the number of iterations $b$ required to ensure the results of Corollary~\ref{corollary1}.

\subsection{Trade-offs between statistical accuracy and computation cost}\label{Sec3.2}

 The convergence rates of oblique trees depend on the Sufficient Impurity Decrease (SID) condition~\citep{chi2022asymptotic, mazumder2024convergence}, which we state below alongside several illustrative examples. Recall that $m(\boldsymbol{X})$ denotes the regression function.

\begin{condition}\label{sid}
There exist $\alpha_{0}\in [0, 1)$ and an integer $s_{0} > 0$ such that for every $l>0$ and every $\boldsymbol{t} = \cap_{i=1}^{l} A_{i}$ where $A_{i} \in \{\{ \vv{x}\in [0, 1]^p : \vv{w}_{i}^{\top}\vv{x} > c_{i}\}, \{ \vv{x}\in [0, 1]^p : \vv{w}_{i}^{\top}\vv{x} \le c_{i}\}\}$ and $(\vv{w}_{i}, c_{i}) \in W_{p, p}$, 
$\inf_{(\vv{w}, c) \in W_{p, s_{0}}}  \mathbb{E} [ \textnormal{Var} ( m(\boldsymbol{X}) \mid \boldsymbol{1}_{\boldsymbol{X} \in \boldsymbol{t}} , \boldsymbol{1}_{\vv{w}^{\top}\boldsymbol{X}>c} ) \boldsymbol{1}_{\boldsymbol{X} \in \boldsymbol{t}} ] \le \alpha_{0} \times \mathbb{E} [ \textnormal{Var} ( m(\boldsymbol{X}) | \boldsymbol{1}_{\boldsymbol{X} \in \boldsymbol{t}} ) \boldsymbol{1}_{\boldsymbol{X} \in \boldsymbol{t}} ].$
\end{condition}

Let the SID function class be denoted by
$$ \text{SID}(\alpha, s) = \{ m(\boldsymbol{X}) : m(\boldsymbol{X}) \text{ satisfies Condition~\ref{sid} with } \alpha_0 = \alpha \text{ and } s_0 = s \}. $$ 
By definition, $\text{SID}(\alpha_1, s_1) \subseteq\text{SID}(\alpha_2, s_2)$ if $\alpha_1 \leq \alpha_2$ and $s_1 \leq s_2$. We note that the SID function class also depends on the distribution of $\boldsymbol{X}$. Below, we provide examples that satisfy the SID condition, with detailed proofs included in Sections~\ref{proof.example.linear}–\ref{proof.example.convex}.

\begin{example}
    \label{example.linear}
    Let $m(\boldsymbol{X}) = \beta\boldsymbol{1}_{\vv{\gamma}^{\top}\boldsymbol{X} > \gamma_{0}}$ for some $(\beta, \gamma_{0}) \in\mathbb{R}^2$ and $\vv{\gamma} \in \mathbb{R}^p$ with $\norm{\vv{\gamma}}_{2} = 1$ and $\norm{\vv{\gamma}}_{0} \le s_0$, in which $\mathbb{P}(\boldsymbol{X}\in [0, 1]^p)=1$. Then, $m(\boldsymbol{X})\in\textnormal{SID}(0, s_0)$.% with $\alpha_{0} = 0$ and every $s$ with  $s\ge 1$.
\end{example}

\begin{example}
    \label{example.linear2}
    Let \( m(\boldsymbol{X}) = \vv{\gamma}^{\top} \boldsymbol{X} \), where \( \vv{\gamma} \in \mathbb{R}^p \) satisfies \( \| \vv{\gamma} \|_2 = 1 \) and \( \| \vv{\gamma} \|_0 \le s_0 \). Assume that the probability density function of the distribution of \( \boldsymbol{X} \) exists, is bounded both above and below over \( [0, 1]^p \), and that \( \mathbb{P}(\boldsymbol{X} \in [0, 1]^p ) = 1 \). Then, \( m(\boldsymbol{X}) \in \textnormal{SID}(\alpha_0, s_0)\) for some $\alpha_0 \in [0, 1)$.
\end{example}

\begin{example}
    \label{example.xor}
    
    Let $m(\boldsymbol{X}) = \beta \big[\boldsymbol{1}_{(X_{1}, \dots, X_{s_{0}}) \in \mathcal{V}_{1}} -  \boldsymbol{1}_{(X_{1}, \dots, X_{s_{0}}) \in \mathcal{V}_{2}} \big] $, where $\mathcal{V}_{1} = \{\vv{v} \in \{0, 1\}^{s_{0}}: \sum_{j=1}^{s_{0}} v_{j} \textnormal{ is odd}\}$, $\mathcal{V}_{2} = \{\vv{v} \in \{0, 1\}^{s_{0}}: \sum_{j=1}^{s_{0}} v_{j} \textnormal{ is even}\}$, \(\beta \in \mathbb{R}\), and \(\boldsymbol{X}\) is uniformly distributed over \(\{0, 1\}^p\). Then, \(m(\boldsymbol{X}) \in \textnormal{SID}(1 - 2^{-s_{0}}, s_0)\).
\end{example}

\begin{example}\label{example.mono}
Assume that for each measurable \( A \subseteq [0, 1]^p \), \((\text{Volume of } A) \times D_{\min} \leq \mathbb{P}(\boldsymbol{X} \in A) \leq (\text{Volume of } A) \times D_{\max}\) for some \( D_{\max} \geq D_{\min} > 0 \). Let \( m(\vv{x}) \) be constant over \( G_{l+1} \setminus G_{l} \) for each \( l \in \{0, \dots, L-1\} \), and assume \( m(\vv{x}_{l+1}) - m(\vv{x}_{l}) \) strictly increases in \( l \in \{1, \dots, L-1\} \), where \( \vv{x}_{l} \in G_{l+1} \setminus G_{l} \). Here, \( G_{l} = \{\vv{x} \in [0, 1]^p : \vv{v}_{l}^\top \vv{x} > a_{l}\} \), with \(\emptyset = G_{0}\subseteq G_{1} \subseteq \dots \subseteq G_{L} = [0, 1]^p \). Assume that $\{\vv{x} \in [0, 1]^p : \vv{v}_{l}^\top \vv{x} = a_{l}\}\cap \{\vv{x} \in [0, 1]^p : \vv{v}_{k}^\top \vv{x} = a_{k}\}=\emptyset$ for $l\not=k$, that \(\texttt{\#}\{j : |v_{l, j}| > 0, l \in \{1, \dots, L\}\} \leq s_{0}\) for some integer \(s_{0} > 0\), and that all weight vectors here have unit length. Then, \(m(\boldsymbol{X})\in \textnormal{SID}(\alpha_0, s_0)\) for some $\alpha_0 \in [0, 1)$.
\end{example}

\begin{example}\label{example.convex}
Assume that for each measurable \( A \subseteq [0, 1]^p \), \((\text{Volume of } A) \times D_{\min} \leq \mathbb{P}(\boldsymbol{X} \in A) \leq (\text{Volume of } A) \times D_{\max}\) for some \( D_{\max} \geq D_{\min} > 0 \). There exists some \(\vv{x}_{0} \in [0, 1]^p\) and integer $K_{1} > 0$ such that for each $k\in \{1, \dots, K_{1}\}$, it holds that $\{ \vv{x}\in [0, 1]^p: \vv{v}_{j}^{(k)\top} \vv{x} < a_{j}^{(k)}\} \subseteq \{ \vv{x}\in [0, 1]^p: \vv{v}_{j+1}^{(k)\top} \vv{x} < a_{j+1}^{(k)}\}$ for $j\in \{1, \dots, n_{k}-1\}$, and that \(\vv{v}_{1}^{(k)\top} \vv{x}_{0} < a_{1}^{(k)}\). Additionally, for each \(l \neq k\), \(\{\vv{x} \in [0, 1]^p : \vv{v}_{1}^{(l)\top} \vv{x} - a_{1}^{(l)} = 0\}\subseteq\{\vv{x} \in [0, 1]^p : \vv{v}_{1}^{(k)\top} \vv{x} - a_{1}^{(k)} < 0\}\). Furthermore, assume that no two hyperplanes intersect within \([0, 1]^p\), that the total number of nonzero coordinates across all weight vectors satisfies \(\texttt{\#}\{j : |v_{l, j}^{(k)}| > 0, l \in \{1, \dots, n_{k}\}, k \in \{1, \dots, K_{1}\}\} \leq s_{0}\), for some integer \(s_{0} > 0\), where \(\vv{v}_{l}^{(k)} = (v_{l, 1}^{(k)}, \dots, v_{l, p}^{(k)})^{\top}\), and that all weight vectors in this example have unit length. Let \(m(\vv{x})\) be constant over \(G_{j}^{(k)}\) for each \(j \in \{0, \dots, n_{k}\}\) and \(k \in \{1, \dots, K_{1}\}\), and assume that \(m(\vv{x}_{j}^{(k)})\) strictly increases with \(j\) for each \(k\), where \(\vv{x}_{j}^{(k)} \in G_{j}^{(k)}\). The regions \(G_{j}^{(k)}\)'s are defined as follows: 
$G_{j}^{(k)} = \{\vv{x} \in [0, 1]^p : \vv{v}_{j+1}^{(k)\top} \vv{x} < a_{j+1}^{(k)}, \vv{v}_{j}^{(k)\top} \vv{x} \geq a_{j}^{(k)}\}$ for  $1 \leq j < n_{k}$, and $G_{j}^{(k)} =\{\vv{x} \in [0, 1]^p : \vv{v}_{n_{k}}^{(k)\top} \vv{x} \geq a_{n_{k}}^{(k)}\}$  for $j = n_{k}$. 
Additionally, let \(G_{0}^{(1)} = \dots = G_{0}^{(K_{1})} = \bigcap_{k=1}^{K_{1}} \{\vv{x} \in [0, 1]^p : \vv{v}_{1}^{(k)\top} \vv{x} < a_{1}^{(k)}\}\).  Under these conditions, \(m(\boldsymbol{X})\in \textnormal{SID}(\alpha_0, s_0)\) for some $\alpha_0 \in [0, 1)$. 
\end{example}

In Example~\ref{example.convex}, the collection $\{G_{0}^{(1)}\} \cup \{G_{j}^{(k)} : j \in \{1, \dots, n_{k}\}, k \in \{1, \dots, K_{1}\}\}$ constitutes a partition of $[0, 1]^p$. The regression function of Example~\ref{example.convex} forms a valley centered at \(\vv{x}_0\), with \(K_1\) uphill slopes rising around the valley's low point \(\vv{x}_0\). When ``strictly increases'' is replaced with ``strictly decreases,'' Example~\ref{example.convex} describes a mountain-shaped regression function with a peak at \(\vv{x}_0\). 
Example~\ref{example.mono} represents a specific case of Example~\ref{example.convex}. Similarly, the linear model in Example~\ref{example.linear} is a special instance of Example~\ref{example.convex} that features only a single uphill side.

Example~\ref{example.xor} presents the $s_0$-dimensional XOR function, which serves as our illustrative case for examining the trade-off between computational runtime and statistical accuracy. In Section~\ref{Sec5.3}, we demonstrate that as $s_0$ increases, the time required to learn the XOR function also grows. Notably, previous work~\citep{tan2024statistical} has shown that an orthogonal tree is not consistent for a 2-dimensional XOR function. Intuitively, an \( s_0 \)-dimensional XOR function likely does not belong to SID$(\alpha_0, k)$ for any \( k < s_0 \) and $\alpha_0 \in[0, 1)$, though a formal proof is not provided here.

Example~\ref{example.linear2} illustrates that continuous functions can satisfy SID when using oblique splits. Moreover, it also satisfies SID under orthogonal splits, as shown in~\citet{mazumder2024convergence}. Empirical studies~\citep{breiman2001random} have shown that oblique trees achieve a faster convergence rate for additive models than orthogonal trees. 

Apart from Example~\ref{example.xor}, the distributional requirements for SID remain quite flexible. For further discussions on verifying SID for additive and interaction functions under orthogonal splits, see~\citet{mazumder2024convergence, chi2022asymptotic}. Additionally, results from~\citet{mazumder2024convergence} may extend to verifying SID for linear combinations of ridge functions, as studied in~\citet{cattaneo2024convergence}.

In addition to SID, Theorem~\ref{theorem2} requires that our sampling procedure satisfies Condition~\ref{sampling.1} below. When sampling a sparse unit-length split weight vector from \( \mathcal{J}(s) \), we first determine the number of nonzero coordinates as specified in Condition~\ref{sampling.1}, then sample the specific set of nonzero coordinates, and finally assign their directions.
\begin{condition}\label{sampling.1}
	A weight vector \( \vv{w} \in \mathcal{J}(s) \) is sampled by first determining the number of nonzero elements in $\vv{w}$, which is drawn uniformly from \( \{1, \dots, s\} \).  
\end{condition}

 Theorem~\ref{theorem2} establishes the convergence rates for the progressive tree and the ideal tree, which are defined as:
\begin{equation}
    \label{pt.1}
    \begin{split}        
    \widehat{R}(\boldsymbol{X}) &= \sum_{(\boldsymbol{t}^{'}, (\vv{w}, c), \boldsymbol{t}) \in \widehat{N}_{H}^{(b)}} \boldsymbol{1}_{\boldsymbol{X} \in \boldsymbol{t}} \times \widehat{\beta}(\boldsymbol{t}),\\
    \widehat{I}(\boldsymbol{X}) &= \sum_{(\boldsymbol{t}^{'}, (\vv{w}, c), \boldsymbol{t}) \in \widehat{N}_H} \boldsymbol{1}_{\boldsymbol{X} \in \boldsymbol{t}} \times \widehat{\beta}_{\text{ideal}}(\boldsymbol{t}),
    \end{split}
\end{equation}
using the notation from \eqref{tspursuit.4} and Section~\ref{sec2.2.1}. Here, $\widehat{N}_H$ represents the tree constructed via \eqref{tspursuit.4} using the exhaustive split space $\mathcal{W} = \widehat{W}_{p, s}$. To clearly distinguish the two models, we let $\widehat{\beta}_{\text{ideal}}(\cdot)$ denote the corresponding least squares coefficients derived from \eqref{tspursuit.4} for the ideal tree.

In Theorem~\ref{theorem2}, $B$ is the number of candidate oblique splits. The parameters $b$ and $S$ represent the optimization iterations and splits per iteration, respectively, both of which are specific to our progressive tree. Additionally, $s$ is the maximum sparsity of the oblique splits, $s_0$ is the SID complexity level, and $H$ is the tree depth. The proof of Theorem~\ref{theorem2} is provided in Section~\ref{proof.theorem2}.

\begin{theorem}\label{theorem2}

Assume that $m(\boldsymbol{X})\in \textnormal{SID} (\alpha_0, s_0)$ for some $\alpha_{0}\in (0, 1]$ and $s_{0}\ge 1$, that all conditions for Corollary~\ref{corollary1} hold with $s\ge s_0$ for some constant $s$, and that Condition~\ref{sampling.1} is satisfied. Let an arbitrary constant $\nu\in (0, 1)$ be given. Then, under Condition~\ref{regularity.1}(b), there is some constant $C >0$ depending on $s$ such that for all large $n$, each $1 \le p \le n^{K_{0}}$ for an arbitrary fixed constant $K_{0} > 1$, each $M_{\epsilon} \ge 0$, each $H\ge 1$ with $\sqrt{2}n^{-\frac{1}{2}} s \sqrt{H(1\vee\log{s})} \log{n}\le 1$, each $B\ge 1$ and $\iota > 0$ with $B \ge  s 2^{2s_0}\binom{p}{s_0} s_0^{\frac{s_0}{2}} \iota^{-s_0} \times \log{ \left(n 2^{2s_0}  \binom{p}{s_0}  s_0^{\frac{s_0}{2}} \iota^{-s_0}\right)}$, and for any $S$ and $b$ satisfying
\begin{equation}
     \label{sb1}
     \texttt{\#}S \times b \ge n2^{H} B\log{\left(2^H B\right)},
\end{equation}     
it holds that
\begin{align}    
        \mathbb{E}\{[ \widehat{R}(\boldsymbol{X}) - m(\boldsymbol{X})]^2\} & \le \left( \sup_{\vv{x} \in [0, 1]^p} |m(\vv{x})|\right)^2  \left(\alpha_{0}^{H} + \frac{8}{n} + \frac{4}{B}\right) +  2^H C (\rho_{n} + D_{\max}A_{n}), \nonumber \\
        \mathbb{E}\{[ \widehat{I}(\boldsymbol{X}) - m(\boldsymbol{X})]^2\} & \le 2\left[\left( \sup_{\vv{x} \in [0, 1]^p} |m(\vv{x})|\right)^2  \left(\alpha_{0}^{H} + \frac{8}{n}\right) +  2^H C \rho_{n}\right],     \label{ideal.tree.consistency}
\end{align}
 where $A_{n} = \left( \sup_{\vv{x}\in [0, 1]^p}|m(\vv{x})|\right)^2 \times (\iota + n^{-\nu})$, and $\rho_{n}$ is given in Theorem~\ref{theorem1}. Additionally, under Condition~\ref{regularity.1}(a),  there is some constant $C>0$ such that for all large $n$, each $1\le p  \le n^{K_{0}}$, each $M_{\epsilon} \ge 0$, each $H\ge 1$  with $\sqrt{2}n^{-\frac{1}{2}} s \sqrt{H(1\vee\log{s})} \log{n}\le 1$, each $B\ge (\log{n})^3 \binom{p}{s_0}$, and for any $S$ and $b$ satisfying
 \begin{equation}
     \label{sb2}
     \texttt{\#}S \times b \ge D_{\min}^{-1}2^{H + s + 1} B\log{(2^H B)},
 \end{equation}
 the bound in \eqref{ideal.tree.consistency} remains valid, and furthermore:
\begin{equation}
    \begin{split}\label{condib}
        & \mathbb{E}\{[\widehat{R}(\boldsymbol{X}) - m(\boldsymbol{X})]^2  \}  \le \left(\sup_{\vv{x} \in [0, 1]^p} |m(\vv{x})|\right)^2  \left(\alpha_{0}^{H} + \frac{8}{n} + \frac{4}{B}\right)  + C\rho_{n}  2^{H}.
    \end{split}
\end{equation}
 
\end{theorem}

We now discuss the theoretical contributions of Theorem~\ref{theorem2}. The convergence rates presented align with the common SID rates for tree consistency \citep{chi2022asymptotic}. Our rates highlight a trade-off between $2^H \rho_n$ and $\alpha_0^H$ in terms of tree depth $H$, where $2^H$, $\alpha_0^H$, and $\rho_n$ represent model complexity, bias convergence, and estimation variance, respectively. The sum of these two terms reaches its minimum at
\begin{equation}
\label{convergence.1}
\inf_{H > 0} \left( 2^H \rho_n + \alpha_0^H \right) = 2 n^{\left(\frac{1}{2} - \epsilon\right) \frac{\log \alpha_0}{\log(2/\alpha_0)}}
\end{equation}
where we denote $\rho_n = n^{-\frac{1}{2} + \epsilon}$ for some $\epsilon > 0$ for brevity. A similar minimum value can be obtained when $H$ is restricted to integer values. Our rate approaches $2 n^{-(\frac{1}{2} - \epsilon)}$ as $\alpha_0 \to 0$, given a sufficiently large $B$; here, a smaller $\alpha_0 \ge 0$ indicates that oblique splits are more suitable for the learning task (see Examples~\ref{example.linear} to \ref{example.convex}).

Additionally, our rates formally characterize the trade-off between the complexity of the function class $\text{SID}(\alpha_0, s_0)$ and the computational cost of oblique trees. An important implication of Theorem~\ref{theorem2} is that the required number of candidate oblique splits $B$ must be surprisingly large when approximating difficult learning problems within the SID class. For example, when $p$ is large relative to $s_0$ and $\log n$, and the lower bound \eqref{sb2} on $\texttt{\#}S \times b$ is satisfied, $B$ is required to be approximately proportional to $\binom{p}{s_0}$ to establish \eqref{condib}. This necessity for a large $B$ on challenging problems complicates the tuning process, further motivating the iterative refinement approach discussed in Sections~\ref{sec2.2.1} and~\ref{Sec5.0}.

Theorem~\ref{theorem2} also establishes an approximate log-linear relationship between the required number of iterations and $s_0$. For instance, when $p$ is large relative to $s_0$ and $\log n$ with $B =  (\lceil\log n \rceil)^3 \binom{p}{s_0}$, the lower bound on $\log b$ derived in \eqref{sb2} is approximately proportional to:
\begin{equation}\label{lowers_0}
s_0 \log p.
\end{equation}
This relationship stems from the dominant term in the lower bound $\log{B} \ge \log \binom{p}{s_0} \approx s_0 \log p$. Consequently, our results suggest that the logarithm of the required runtime, measured by the number of iterations, scales linearly with the intrinsic sparsity $s_0$ rather than the user-defined $s$ as a requirement for learning functions in the $\text{SID}(\alpha_0, s_0)$ class.

We acknowledge that the lower bounds on $B$ and $b$ in Theorem~\ref{theorem2} may be conservative. This is primarily due to the technical challenges of analyzing progressive trees as opposed to one-shot trees, as well as the need to account for the broad range of functions within the $\text{SID}(\alpha_0, s_0)$ class. Nevertheless, the necessity for a large $B$ and the identified log-linear relationship represent fundamental requirements, both of which are empirically validated in Section~\ref{Sec5.3}.

\begin{remark}\label{s_s_0}
While the analysis in Theorem~\ref{theorem1} allows for a flexible range of $s$, we treat $s$ as a fixed constant in Theorem~\ref{theorem2} to maintain analytical clarity. This choice aligns with conventions in~\citet{breiman2001random} and~\citet{tomita2020sparse}, who suggest $s=3$ and $s=5$, respectively. Consequently, we assume a fixed $s$ for Theorem~\ref{theorem2} and use $s=5$ for all numerical experiments in Sections~\ref{Sec5.0}--\ref{Sec6}. Furthermore, although Theorem~\ref{theorem2} is formulated for a specific $s_0$, the results hold for any $s_0 \in \{1, \dots, s\}$ given that $s$ is constant. In such cases, \eqref{lowers_0} remains valid, though specific constants like $K_{0}$ may vary.
\end{remark}

On the other hand, for continuous features, Theorem~\ref{theorem2} investigates the trade-off between the desired approximation accuracy $\iota$ and the total computational cost, defined by $\texttt{\#}S \times b$. Specifically, the computational requirement in \eqref{sb1}, after omitting logarithmic factors, is proportional to $n  s 2^{2s_0 + H} \binom{p}{s_0} s_0^{s_0/2} \iota^{-s_0}$. This expression represents the cardinality of the oblique split candidate set necessary to achieve an approximation error bounded by $\iota$ for a given sample size $n$. The computational burden is significantly higher for continuous features because the search space for potential splits is far more vast than in the discrete case~\citep{murthy1994system}. For example, if $p=1$ and $\mathbb{P}(X_{1} \in \{0, 1\}) = 1$, a single threshold separating the sample into $\{i: X_{i1} > c\}$ and $\{i: X_{i1} \le c\}$ for $c\in [0, 1)$ is exhaustive; any further split candidates are redundant. In contrast, continuous features necessitate a much denser sampling of the split space to ensure the optimal boundary is approximated within the specified tolerance $\iota$.

Our convergence rates are slower than the rates achieved in \citet{mazumder2024convergence}. For a fixed feature dimension, the lower bound for the $L^2$ consistency rate of any partition based estimator \citep{gyorfi2002distribution}, including trees, is $n^{-2/(2+p)}$ up to a logarithmic factor of $n$. This optimal rate is reached by the rates in \citet{mazumder2024convergence} for the univariate case where $p=1$ with linear data generating functions. In contrast, our Theorem~\ref{theorem2} aligns with the less tight bounds established by \citet{chi2022asymptotic}, which are generally no faster than $n^{-1/2}$ as discussed in \eqref{convergence.1}.

The analysis in \citet{mazumder2024convergence} employs a sharper balance between the high estimation noise in small leaf nodes and the negligible fraction of the population those nodes represent. By weighting these factors more precisely, they achieve tighter convergence rates than those in our current analysis. We do not adopt this more granular approach to circumvent technical challenges specific to analyzing rates for oblique trees. Notably, our results address the approximation error between a one-shot tree and the ideal tree, a factor not considered in existing studies on tree convergence rates \citep{mazumder2024convergence, cattaneo2024convergence}. This added complexity makes it unclear whether the results from \citet{mazumder2024convergence} are directly applicable to our setting, and we leave this as an open problem for future investigation.

In a related context, \citet{cattaneo2024convergence} demonstrate that oblique trees provide effective approximations for a wide range of data generating functions. They establish this by proving a slow logarithmic rate of convergence for finite linear combinations of ridge functions in settings where $p \leq n$.

%%%
%%%

\section{Integrating selective oblique splits into Random Forests}\label{Sec4}

In this section, we leverage Theorem~\ref{theorem2} by transferring optimized oblique splits to advanced ensemble models. While Theorem~\ref{theorem2} implies that optimizing oblique splits does not strictly require techniques like bagging or boosting for asymptotic convergence, incorporating these methods is essential for maximizing predictive accuracy on finite samples. We propose a two-step construction for the RF+$\mathcal{S}^{(b)}$ model:
\begin{enumerate}
\item Iterative Refinement Phase: We execute the iterative refinement algorithm for $b$ search iterations to obtain the set of optimized oblique splits $\mathcal{S}^{(b)}$, as detailed in Section~\ref{sec2.2.1}. 

\item Integration Phase: We integrate the splits from $\mathcal{S}^{(b)}$ into a Random Forest model alongside standard orthogonal splits. This strategy allows the ensemble to benefit from high-quality oblique partitions while avoiding the prohibitive computational cost of re-calculating oblique directions for every individual tree in the bagging process.
\end{enumerate}

The integration treats the optimized weight vectors as feature extractors. Specifically, for an oblique tree of depth $H$, we define $\mathcal{S}^{(b)} = \{(\vv{w}_{k}, c_{k})\}_{k=1}^{2^H - 1}$. We then perform the following regression:
\begin{equation}\label{augment.1}
\textnormal{Regress } Y_{i} \textnormal{ on }  \big(\boldsymbol{X}_{i}, \vv{w}_{1}^{\top}\boldsymbol{X}_{i}, \dots, \vv{w}_{2^{h}-1}^{\top}\boldsymbol{X}_{i}  \big) \textnormal{ using Random Forests,}
\end{equation}
where $h \in \{0, \dots, H\}$ is a tuning hyperparameter. The thresholds $c_k$ from $\mathcal{S}^{(b)}$ are omitted in \eqref{augment.1} because the Random Forest naturally determines optimal split points along the projected features $\vv{w}_{k}^{\top}\boldsymbol{X}_{i}$. We apply standard hyperparameter tuning \citep{grinsztajn2022tree} to optimize $h$ and other Random Forest parameters.

This framework offers two distinct advantages. First, when $h = 0$, the model reverts to a standard Random Forest, ensuring that RF+$\mathcal{S}^{(b)}$ can achieve at least the accuracy of its orthogonal counterpart. Second, it provides greater stability than a single progressive tree by benefiting from bagging and column subsampling.

State-of-the-art tree models typically rely on multiple tuning parameters that control tree depth, node sizes, splitting criteria, and overall tree structure. These hyperparameters are crucial for achieving accurate predictions in finite-sample settings. However, theoretical analyses of tree models often concentrate on their fundamental structural properties, while overlooking finer details such as the effects of node sizes and the use of bagging~\citep{mazumder2024convergence, chi2022asymptotic, klusowski2024large, cattaneo2024convergence}. In our case, Theorem~\ref{theorem2} provides a theoretical foundation for the progressive tree for optimizing oblique splits, which forms the basis of the RF+\(\mathcal{S}^{(b)}\) model. We defer a detailed analysis of the effects of bagging and column subsampling in RF+\(\mathcal{S}^{(b)}\) to future work.

\subsection{Warm-starting via transfer learning}

Our RF+$\mathcal{S}^{(b)}$ framework is closely related to warm-start strategies in transfer learning, which reuse previously learned relationships among inputs, latent representations, and outputs to improve training~\citep{humbird2018deep}. Related ideas appear in the TAO framework~\citep{carreira2018alternating}, where split coefficients are initialized from pilot tree models. Beyond accelerating convergence, such knowledge transfer can substantially improve performance when training data are limited~\citep{oquab2014learning}.

\subsection{Hyperparameter tuning procedure}\label{Sec4.1}

To optimize model performance, we partition the dataset into training (80\%) and validation (20\%) sets. Using the training sample, we generate \(\mathcal{S}^{(b)}\) after \(b\) search iterations, following the procedure described in Section~\ref{sec2.2.1}, with parameter settings \( (s, H, \texttt{\#}S, B) = (5, 3, 100, \infty) \). The value $s=5$ balances the need to capture complex variable relationships against the computational constraints implied by the complexity bounds in Theorem~\ref{theorem2}. Furthermore, we suggest maintaining a shallow progressive tree of depth $H=3$ for feature extraction, as oblique splits typically reduce bias more efficiently than their orthogonal counterparts \citep{tomita2020sparse}.

We subsequently sample a set of Random Forest hyperparameters, including the maximum tree depth of the forest, the column subsampling rate, and the minimum node size. It is important to note that the forest depth is tuned independently of the fixed depth $H=3$ used for the initial progressive tree. For each configuration, a Random Forest model is trained on the training set and evaluated against the validation set. This procedure is repeated for $R \in \{30, 100\}$ iterations, utilizing the \texttt{hyperopt} Python package to minimize the validation loss. Once the optimal configuration is identified, we retrain the final model on the pooled training and validation data. The performance of this final model is then reported on the held-out test set. A comprehensive list of the hyperparameter search space for each forest model is provided in Section~\ref{SecA.1} of the Supplementary Material.

\section{Simulated experiments}\label{Sec5.0}

We validate the asymptotic theory of Theorem~\ref{theorem2} by comparing the progressive tree (see Table~\ref{tab:comparison}) with both the well-established Forest-RC (F-RC~\citep{breiman2001random}) and its base learners. These base learners are Breiman oblique trees, as described in Section~\ref{Sec2.5}. For our experiments, F-RC is implemented using the Python package \texttt{treeple}.

In Sections~\ref{Sec5.3}--\ref{Sec5.2}, we simulate an i.i.d. sample \(\{(\boldsymbol{X}_i, Y_i, \varepsilon_i)\}_{i=1}^{n}\) of size \( n \), where the feature vector \(\boldsymbol{X}_i = (X_{i1}, X_{i2}, \dots, X_{ip})\) follows a uniform distribution over \(\{0,1\}^p\). The response variable is defined as  
\begin{equation}
    \label{data-generating.model}
    Y_i = \boldsymbol{1}\left\{ \left(\sum_{j\in \mathcal{U}}X_{ij}\right) \text{ is odd} \right\} - \boldsymbol{1}\left\{ \left(\sum_{j\in \mathcal{U}}X_{ij}\right) \text{ is even}\right\} + \varepsilon_i,
\end{equation}
where \(\mathcal{U} \subseteq \{1, \dots, p\}\) is a randomly selected subset of indices with \(\texttt{\#}\mathcal{U} = s_0\), and \(\varepsilon_i\) are i.i.d. zero-mean Gaussian errors with standard deviation $\sigma\in \{0, 1\}$. The available training sample consists of \(\{(\boldsymbol{X}_i, Y_i)\}_{i=1}^{n}\), with \(\varepsilon_i\) being unobserved. The data-generating function \eqref{data-generating.model} is an $s_0$-dimensional XOR problem, as in Example~\ref{example.xor}.

We use the XOR function as our illustrative example because it is one of the simplest yet complex functions that effectively demonstrates the trade-off between computational runtime and model accuracy. The approximation difficulty is directly influenced by the value of \( s_0 \). However, real-world prediction tasks do not necessarily follow an \( s_0 \)-dimensional XOR structure. Therefore, in Section~\ref{Sec6}, we assess the predictive performance of RF+$\mathcal{S}^{(b)}$ using real datasets.

For all experiments, an 80/20 train-validation split is used whenever hyperparameter tuning is required, as detailed in Section~\ref{Sec4}. The R\(^2\) scores, defined as 
\begin{equation}
    \label{r_square.1}
    \textnormal{R}^2 \textnormal{ score } = 1 - \frac{\textnormal{Mean squared error on the test set}}{\textnormal{Empirical variance of the response in the test set}},  
\end{equation}
are computed. In all experiments in this section, the test sample is an independent errorless sample of size 5,000, which is generated from \eqref{data-generating.model} with the corresponding $s_0$ and $\sigma = 0$.

\subsection{Comparison between progressive trees and Breiman oblique trees}\label{Sec5.3}

Let us introduce the simulation set-up in this section. We consider the  \( s_0 \)-dimensional XOR problem from \eqref{data-generating.model} with $\sigma = 0$ (noiseless), \( p = 20 \), and \( n = 250 \), with \( s_0 \in \{1, 2, 3, 4\}\). Tables \ref{tab:combined_results} and \ref{tab:3} report the average R$^{2}$ scores over 50 independent trials for the progressive tree and Breiman oblique tree, respectively. Results are rounded to two decimal places, with standard deviations in parentheses and runtimes (in seconds) following the slash.

For both the progressive tree (Table~\ref{tab:combined_results}) and Breiman oblique tree (Table~\ref{tab:3}), we set the minimum number of samples for a leaf node and an internal node split to $10$. The tree depth is fixed at $H = 3$, and the number of variables for each oblique split is $s = 5$ (i.e., \texttt{feature\_combinations}). For the progressive tree specifically, we maintain the configuration $(s, H, \texttt{\#}S) = (5, 3, 100)$ while varying $B$ (i.e.,  \textsf{max\_features}) and $b$ as detailed in Table~\ref{tab:combined_results}. In Setting I, $B$ is defined as $(\lceil\log n \rceil)^3 \binom{p}{s_0} \in \{3366, 31982, 191896, 815559\}$ to satisfy the discrete case requirements of \eqref{condib} in Theorem~\ref{theorem2}.

\begin{table}[h]
\centering
\footnotesize 
\setlength{\tabcolsep}{3pt} 
\begin{tabular}{c c c | lllll} 
& $s_0$ & $B$ & \multicolumn{5}{c}{Update Iterations ($b$)} \\
 \hline
 \hline
  & & & 2 & 20 & 200 & 2000 & 20000 \\ [0.5ex] 
 \hline
 % Paper Settings
\multirow{4}{*}{I}  & $1$ & {\tiny $3.4 \times 10^3$} & 1.00(0.00)/0.0 & 1.00(0.00)/0.2 & 1.00(0.00)/1.1 & 1.00(0.00)/10.3 & 1.00(0.00)/102.4 \\
 & $2$ & {\tiny $3.2 \times 10^4$} & 0.84(0.22)/0.1 & 1.00(0.00)/0.3 & 1.00(0.00)/1.4 & 1.00(0.00)/10.8 & 1.00(0.00)/104.1 \\
 & $3$ & {\tiny $1.9 \times 10^5$} & 0.13(0.33)/0.6 & 0.69(0.28)/1.4 & 1.00(0.00)/2.9 & 1.00(0.00)/12.9 & 1.00(0.00)/107.6 \\
 & $4$ & {\tiny $8.2 \times 10^5$} & -0.17(0.20)/2.6 & 0.05(0.43)/5.2 & 0.49(0.43)/8.8 & 0.93(0.12)/20.8 & 0.99(0.03)/117.7 \\
 \hline
 % B = 5 X 10^3 Settings
\multirow{4}{*}{II}   & $1$ & {\tiny $5 \times 10^{3}$} & 1.00(0.00)/0.0 & 1.00(0.00)/0.2 & 1.00(0.00)/1.1 & 1.00(0.00)/10.4 & 1.00(0.00)/102.7 \\
 & $2$ & {\tiny $5 \times 10^{3}$} & 0.77(0.25)/0.0 & 1.00(0.02)/0.2 & 1.00(0.00)/1.1 & 1.00(0.00)/10.4 & 1.00(0.00)/103.0 \\
 & $3$ & {\tiny $5 \times 10^{3}$} & 0.27(0.39)/0.0 & 0.72(0.30)/0.2 & 0.88(0.17)/1.1 & 0.87(0.19)/10.4 & 0.89(0.17)/103.4 \\
 & $4$ & {\tiny $5 \times 10^{3}$} & -0.20(0.11)/0.0 & -0.14(0.20)/0.2 & 0.20(0.47)/1.1 & 0.23(0.47)/10.6 & 0.07(0.46)/104.3 \\
 \hline
 % B = 10^6 Settings
 \multirow{4}{*}{III}   & $1$ & {\tiny $10^{6}$} & 1.00(0.00)/3.2 & 1.00(0.00)/6.4 & 1.00(0.00)/10.5 & 1.00(0.00)/23.0 & 1.00(0.00)/119.1 \\
 & $2$ & {\tiny $10^{6}$} & 0.89(0.19)/3.2 & 1.00(0.00)/6.4 & 1.00(0.00)/10.5 & 1.00(0.00)/22.9 & 1.00(0.00)/119.5 \\
 & $3$ & {\tiny $10^{6}$} & 0.24(0.39)/3.2 & 0.70(0.25)/6.4 & 0.99(0.05)/10.5 & 1.00(0.00)/23.1 & 1.00(0.00)/120.6 \\
 & $4$ & {\tiny $10^{6}$} & -0.18(0.10)/3.2 & 0.03(0.40)/6.4 & 0.58(0.40)/10.5 & 0.94(0.10)/23.1 & 1.00(0.02)/120.8 \\
 \hline
 \multirow{4}{*}{IV}   & $1$ & $\infty$ & 1.00(0.00)/0.0 & 1.00(0.00)/0.1 & 1.00(0.00)/1.3 & 1.00(0.00)/12.5 & 1.00(0.00)/125.0 \\
 & $2$ & $\infty$ & 0.83(0.22)/0.0 & 1.00(0.00)/0.1 & 1.00(0.00)/1.3 & 1.00(0.00)/12.6 & 1.00(0.00)/125.4 \\
 & $3$ & $\infty$ & 0.18(0.38)/0.0 & 0.66(0.29)/0.1 & 0.99(0.07)/1.3 & 1.00(0.00)/12.7 & 1.00(0.00)/126.3 \\
 & $4$ & $\infty$ & -0.20(0.12)/0.0 & -0.04(0.32)/0.1 & 0.53(0.39)/1.3 & 0.96(0.07)/12.7 & 1.00(0.00)/126.5 \\
 \hline
\end{tabular} 
\caption{Progressive tree results across sparsity $s_0$ and iterations $b$. Cells report mean R$^2$ (standard deviation) / average runtime in seconds over 50 trials. The setups for $B < \infty$ (Settings I, II, and III) and $B = \infty$ (Setting IV) are defined in Section~\ref{sec2.2.1}.}
\label{tab:combined_results}
\end{table}

\begin{table}[h]
\centering
\footnotesize % Matches the first table's font size
\setlength{\tabcolsep}{5pt} % Reduces horizontal padding for a tighter fit
\begin{tabular}{c| lllll} 
 \hline
 & \multicolumn{5}{c}{\textsf{max\_features} ($B$)} \\
 $s_0$  & $10^3$ & $10^4$ & $10^5$ & $10^6$ & $10^7$\\ 
 \hline
 1 & 1.00(0.01)/0.0 & 1.00(0.00)/0.03 & 1.00(0.00)/0.35 & 1.00(0.00)/3.74 & 1.00(0.00)/40.57 \\
2 & 0.60(0.23)/0.01 & 1.00(0.03)/0.06 & 1.00(0.00)/0.59 & 1.00(0.00)/6.36 & 1.00(0.00)/66.71 \\
3 & 0.11(0.23)/0.01 & 0.75(0.25)/0.09 & 1.00(0.03)/0.7 & 1.00(0.00)/7.48 & 1.00(0.00)/78.63 \\
4 & -0.14(0.12)/0.01 & 0.16(0.27)/0.09 & 0.69(0.18)/0.89 & 0.95(0.09)/8.93 & 0.96(0.08)/91.9 \\
 \hline
\end{tabular}
\caption{Results for Breiman oblique trees across sparsity $s_0$ and candidate set size $B$. Cells report mean R$^2$ (standard deviation) / average runtime in seconds over 50 trials.}
\label{tab:3}
\end{table}

The empirical results in Setting I of Table \ref{tab:combined_results} confirm that a minimum number of iterations $b$ is essential for convergence. However, the theoretical lower bound for \eqref{condib} provided by Theorem \ref{theorem2} is highly conservative in practice. Under the current setup in Setting I, \eqref{sb2} dictate that $b$ is required to be at least as large as $B$. In stark contrast to this strict requirement, empirical R$^2$ scores reach 1.00 for $s_0 = 3$ at only $b = 200$, and a high accuracy of 0.93 is achieved for $s_0 = 4$ at $b = 2000$. Despite this numerical gap, Setting I reveals a clear log-linear relationship between the iterations $b$ and the learning difficulty $s_0$. Specifically, the logarithm of $b$ scales linearly with $s_0$, an observation that aligns perfectly with the scaling behavior predicted in \eqref{lowers_0}. This trend generally persists across Settings I, III, and IV.

Although Theorem~\ref{theorem2} also implies a log-linear relationship between $B$ and $s_0$, the requirements for the candidate split size $B$ cannot be isolated from the number of iterations $b$ for empirical validation within the progressive tree framework, as both parameters jointly govern the iterative refinement process. Nevertheless, owing to their structural parallels, we find partial support for this scaling through XOR experiments using Breiman oblique trees. As shown in Table \ref{tab:3}, these results reveal a similar approximate log-linear relationship between $B$ and $s_0$; specifically, $B = 10^4$ suffices for R$^2 = 1.00$ at $s_0 = 2$, whereas $s_0 = 4$ requires $B = 10^6$ to achieve R$^2 = 0.95$. Breiman oblique tree is slightly faster for these specific benchmarks, averaging 8.93s to reach R$^2 > 0.90$ compared to 12.7s for the progressive tree ($b = 2000$) under our recommended $B = \infty$ setup (Setting IV). Additionally, the runtime in Table \ref{tab:3} increases with $s_0$ despite $B$ being fixed. This suggests that the computational efficiency of the \texttt{treeple} implementation is sensitive to the complexity of the interaction, even when the search space and tree depth remain constant.

Settings I through III in Table~\ref{tab:combined_results} support Theorem~\ref{theorem2}, confirming that accurate approximation requires a sufficiently large $B$. Specifically, Setting II fails on higher $s_0$ interactions due to an insufficient $B$, while Setting III demonstrates that excessively large $B$ values introduce significant computational overhead. Since $s_0$ is typically unknown, tuning $B$ to balance computational efficiency with predictive accuracy is a nontrivial challenge, as further evidenced by the results for Breiman oblique tree in Table~\ref{tab:3}. This difficulty motivates our iterative refinement strategy, which circumvents the need to pre-determine $B$ and $b$. Setting IV ($B = \infty$) offers a practical alternative by sampling splits independently at each node. This approach avoids the costs of maintaining a large global set and is efficient in early-stopping scenarios where the iteration count $b$ is gradually increased. Consequently, for applications where pre-tuning a global split set is impractical, we recommend the $B = \infty$ configuration with $\texttt{\#}S = 100$.

\subsection{Two-dimensional XOR problems in a high-dimensional feature space}\label{Sec5.2}

In this section, we evaluate the predictive performance of tree models on XOR problems defined by \eqref{data-generating.model}, with parameters $\sigma = 1$, \( s_0 = 2 \), \( n \in\{ 200, 400\} \), and \( p = 2n \), as detailed in Table~\ref{tab:1}. We compare seven models as detailed in Table~\ref{tab:method_comparison}. Specifically, models (i) progressive tree (PT) through (iii) PT, which represent the progressive tree with finite $B$ and increasing $b$, provide evidence for the asymptotic theory derived in \eqref{condib}. Meanwhile, (v) Breiman oblique trees (OT) utilize a much higher $B$, where $B$ is equivalent to the \textsf{max\_features} parameter, than (viii) F-RC. Model (vi) RF+$\mathcal{S}^{(8000)}$ inherits its parameter configuration from method (iv) because $\mathcal{S}^{(8000)}$ is generated through that process (see Section~\ref{Sec4.1}). Hyperparameter tuning for (vi) RF+$\mathcal{S}^{(8000)}$, (vii) RF, and (viii) F-RC follows the procedure outlined in Section~\ref{Sec4.1}, with search spaces for the latter three defined in Section~\ref{SecA.1}.

\begin{table}[ht]
\centering
\footnotesize 
\caption{Configuration settings for all evaluated models. For models (i)--(v), the minimum number of samples for a leaf node and an internal node split is fixed at 10.}
\label{tab:method_comparison}
\begin{tabular}{cllccc}
\toprule
\textbf{ID} & \textbf{Method} & \textbf{Parameters} $(B, b, \texttt{\#}S, H, s)$ & \textbf{Tuning} & \texttt{\#}\textbf{Bagging Trees} \\ 
\midrule
(i) & Progressive Tree (PT) & $(10^6, 8000, 100, 3, 5)$ & No & -- \\
(ii) & Progressive Tree (PT) & $(10^6, 16000, 100, 3, 5)$ & No & -- \\
(iii) & Progressive Tree (PT) & $(10^6, 24000, 100, 3, 5)$ & No & -- \\
(iv) & Progressive Tree (PT) & $(\infty, 8000, 100, 3, 5)$ & No & -- \\
(v) & Breiman Oblique Tree (OT) & $(10^6, \text{N/A}, \text{N/A}, 3, 5)$ & No & -- \\
\midrule
(vi) & RF+$\mathcal{S}^{(8000)}$ & $\mathcal{S}^{(8000)}$ depends on (iv) & Yes & 100 \\
(vii) & Random Forest (RF) & \text{N/A} & Yes & 100 \\
(viii) & Forest-RC (F-RC) & $B = \min\{1000, p^2\}$ & Yes & 100 \\
\bottomrule
\end{tabular}
\end{table}

%  \begin{table}[ht]
% \centering
% \footnotesize % Matches the first table's font size
% \setlength{\tabcolsep}{5pt}
% \caption{Comparison of methods and configuration settings. The minimum number of samples for a leaf node and an internal node split to $10$ for (i)--(v).  }
% \label{tab:method_comparison}
% \begin{tabular}{lp{6cm}l}
% \hline
% Method & Model Parameter Setup & Hyperparameter Tuning \\ \hline
% (i) progressive tree (PT) & $(B, b, \texttt{\#}S, H, s) = (10^6, 8000, 100, 3, 5)$ & No \\
% (ii) progressive tree (PT) & $(B, b, \texttt{\#}S, H, s) = (10^6, 16000, 100, 3, 5)$ & No \\
% (iii) progressive tree (PT) & $(B, b, \texttt{\#}S, H, s) = (10^6, 24000, 100, 3, 5)$ & No \\
% (iv) progressive tree (PT) & $(B, b, \texttt{\#}S, H, s) = (\infty, 8000, 100, 3, 5)$ & No \\
% (v) Breiman oblique tree (OT) & $(B, H, s) = (10^6, 3, 5)$ & No  \\
% (vi) RF+$\mathcal{S}^{(8000)}$ & Inherits parameters from (iv) and (vii); \texttt{\#}bagging trees = 100 & Yes \\
% (vii) RF & \texttt{\#}bagging trees = 100 & Yes \\
% (viii) F-RC & $B = \min\{1000, p^2\}$; \texttt{\#}bagging trees = 100 & Yes \\
% \hline
%  \end{tabular}
% \end{table}

In Table~\ref{tab:1}, each entry reports the average R$^2$ score, computed according to \eqref{r_square.1}, alongside the average runtime in seconds. These metrics are calculated over 50 independent trials, with standard deviations for the R$^2$ scores provided in parentheses. The reported runtimes are inclusive of the total duration required for hyperparameter optimization to ensure a fair comparison of computational efficiency across all models.

The results in Table~\ref{tab:1} demonstrate that the progressive tree becomes both more accurate and more reliable as the search progresses, as evidenced by the rising R$^2$ scores and the significant reduction in standard deviation from model (i) to (ii). The stabilization observed between models (ii) and (iii) PT indicates that the progressive tree achieves convergence to the corresponding one-shot tree without overfitting, even as iterations extend to 24,000. Furthermore, the upward trend in R$^2$ scores for models (ii) and (iii) as sample sizes increase suggests that the model moves toward consistency when given a sufficiently large number of iterations and a large split set $B$. These findings provide empirical support for the theoretical foundation in Theorem \ref{theorem2}, implying that estimation error vanishes as the sample size grows provided that both the number of iterations and $B$ are sufficiently large.

It is worth noting that the theoretical lower bounds on $B$ and $b$ for deriving \eqref{condib} are not strictly met in this study. Specifically, when $(n, p, s_0) = (400, 800, 2)$, the lower bound on $B$ exceeds 68 million. Such a high value is impractical for empirical study given the runtimes for both the progressive tree and the Breiman oblique tree shown in Table~\ref{tab:1}. For this same reason, we limit the experiment to $n \le 400$; increasing the sample size while maintaining the $p = 2n$ setup would require $p$, $B$, and $b$ to grow simultaneously, leading to prohibitive computational costs. Despite this, our results show that the model achieves respectable performance with $B = 10^6$. This suggests that while the theoretical bounds are sufficient to guarantee results, they are likely not strictly necessary for achieving convergence in practical applications.

Model (v) OT outperforms (viii) F-RC in Table~\ref{tab:1} because the latter uses an insufficiently small split set size. If (viii) F-RC were to use $B = 10^6$, the estimated total runtime would be the product of the runtime for (v) OT, the number of tuning rounds, the number of bagging trees, and the total trials. For the $(n, p) = (400, 800)$ case, this would require approximately $58.01 \times 30 \times 100 \times 50$ seconds, totaling roughly 2,300 hours. Such a requirement is computationally impractical and justifies its omission from our experiments. Instead, we recommend (vi) RF+$\mathcal{S}^{(8000)}$, which is based on the progressive tree with the $B = \infty$ setup (similar to (iv) PT), as a robust solution for exploiting optimized oblique splits in noisy applications. Table~\ref{tab:1} illustrates the empirical advantages of model (vi), including superior finite-sample performance ($\ge 90\% $ R$^2$ scores) and the elimination of the need to pre-specify $B$. This approach balances the reduced approximation error of (iv) PT with the variance control of the Random Forest framework. It is important to note that the runtime of (vi) RF+$\mathcal{S}^{(8000)}$ is comparable to (iv) PT rather than the sum of (iv) and (vii). This efficiency occurs because (vi) utilizes only 80\% of the training data for construction, reserving the remaining 20\% for validation during the tuning process.

Furthermore, model (v) OT underperforms compared to models (i) through (iii) in Table~\ref{tab:1}, even though all utilize a candidate split set of size $B=10^6$. While comparing a progressive tree directly to a Breiman oblique tree is structurally complex, comparing a one-shot tree to a Breiman oblique tree is straightforward. As discussed previously, (iii) PT with $b = 24,000$ closely approximates its corresponding one-shot tree with $B=10^6$. The primary architectural difference between the Breiman oblique tree and a one-shot tree lies in their optimization strategies: the former employs node-wise optimization, whereas the latter optimizes splits using a global split set $\mathcal{W}$ across all nodes (as summarized in Table~\ref{tab:comparison}). Consequently, the performance gap between (iii) PT and (v) OT strongly suggests that global split optimization yields significantly higher prediction accuracy in this specific setting.

\begin{table}[ht]
\centering
\scriptsize
\setlength{\tabcolsep}{2.3pt}
\begin{tabular}{c c | llllllll} 
 \hline
 $p / n$ &  & (i)PT & (ii)PT & (iii)PT & (iv)PT & (v)OT &  (vi)RF+$\mathcal{S}^{(8000)}$ & (vii)RF & (viii)F-RC  \\ 
 \hline
 \multirow{2}{*}{400/200} & R$^2$ & 0.51 (0.22) & 0.55 (0.09) & 0.55 (0.10) & 0.56 (0.07) & 0.17 (0.55) & 0.93 (0.07) & -0.03 (0.04) & -0.03 (0.04) \\
 & t & 69.52  & 126.80  & 183.73 & 67.40  & 28.04 & 67.57  & 8.68 &  154.00  \\
 \hline
\multirow{2}{*}{800/400} & R$^2$ & 0.49 (0.39) & 0.63 (0.23) & 0.67 (0.16) & 0.64 (0.27) & 0.23 (0.32) & 0.90 (0.17) & -0.01 (0.05) & -0.01 (0.04) \\
 & t & 210.86  & 396.09 & 578.15  & 206.57  & 58.01  & 213.31 & 50.63  & 410.96  \\
 \hline
\end{tabular}
\caption{Benchmark comparison based on data-generating model \eqref{data-generating.model} with $(\sigma, s_0)= (1, 2)$. R$^2$ scores and runtime (in seconds) are averaged over 50 trials. Standard deviations of R$^2$ scores are shown in parentheses.}
\label{tab:1}
\end{table}
%  \multirow{2}{*}{400/200} & R$^2$ & 0.51 (0.22) & 0.55 (0.09) & 0.55 (0.10) & 0.56 (0.07) & 0.17 (0.55) & 0.93 (0.07) & -0.03 (0.04) & -0.03 (0.04) \\
%  & t & 69.52 (0.46) & 126.80 (1.02) & 183.73 (1.20) & 67.40 (0.62) & 28.04 (3.04) & 67.57 (4.09) & 8.68 (2.71) &  154.00 (26.56) \\
%  \hline
% \multirow{2}{*}{800/400} & R$^2$ & 0.49 (0.39) & 0.63 (0.23) & 0.67 (0.16) & 0.64 (0.27) & 0.23 (0.32) & 0.90 (0.17) & -0.01 (0.05) & -0.01 (0.04) \\
%  & t & 210.86 (23.42) & 396.09 (44.89) & 578.15 (62.50) & 206.57 (26.11) & 58.01 (5.53) & 213.31 (27.63) & 50.63 (17.96) & 410.96 (128.61) \\
 % & runtime & 212.05 (24.79) & 398.61 (47.10) & 580.44 (65.76) & 208.04 (27.42) & 56.98 (6.03) & 212.58 (26.15) & 51.51 (17.41) & 435.92 (140.68) \\

\section{Evaluation of prediction accuracy using real data}\label{Sec6}

\begin{table}[ht!]
\centering
\scriptsize
\setlength{\tabcolsep}{4pt}
\caption{Dataset summary. High-dimensional (HD) cases are those where the total feature count exceeds the training size $n = 400$. Total features include expanded sets such as interactions used during optimization.}
\label{tab:dataset_features_10rows}
\begin{tabular}{|l|p{1.5cm}p{1.5cm}||l|p{1.5cm}p{1.5cm}|}
\hline
Dataset Name & \texttt{\#}Original Features & \texttt{\#}Total Features & Dataset Name &  \texttt{\#}Original Features & \texttt{\#}Total Features \\ \hline
Ailerons (HD) & 33 & 594 & medical\_charges & 3 & 9\\
superconduct (HD) & 79 & 3239 & MiamiHousing2016 & 13 & 104\\
yprop\_4\_1 (HD) & 42 & 945 & sulfur & 6 & 27\\
cpu\_act & 21 & 252 & houses & 8 & 44 \\
pol & 26 & 377 & abalone & 7 & 35 \\
house\_16H & 16 & 152 & wine\_quality & 11 & 77 \\
Brazilian\_houses & 8 & 44 & diamonds & 6 & 27 \\
Bike\_Sharing\_Demand & 6 & 27 & house\_sales & 15 & 135 \\
nyc-taxi-green-dec-2016 & 9 & 54 & delay\_zurich\_transport & 8 & 44 \\
elevators & 16 & 152 & &&   \\
\hline
\end{tabular}
\end{table}

By using the real benchmark datasets, we assess the predictive performance of RF+\(\mathcal{S}^{(b)}\) and compare it against Random Forests, F-RC from \citet{breiman2001random}, and Manifold Oblique Random Forests (MORF~\citep{li2023manifold}). MORF, which is implemented using the Python package \texttt{treeple}, represent a more recent approach that integrates both oblique splits and feature distances, specifically designed for image regression tasks. Since SPORF~\citep{tomita2020sparse} is highly similar to F-RC, we exclude it from our comparison. Our evaluation approach follows the methodology established in \citet{grinsztajn2022tree}, which provides a state-of-the-art benchmark procedure and simulation framework. We evaluate performance using R$^2$ scores across 19 selected OpenML datasets, where 400 training and 2,000 test samples are randomly drawn from each dataset without overlap. See Section~\ref{Sec8} for details on obtaining the datasets. The names of these datasets are given in Table~\ref{tab:dataset_features_10rows}, with other details available in \citet{grinsztajn2022tree}.

To evaluate performance in high-dimensional settings, we augment each dataset with pairwise interactions and squared terms ($X_{i}X_{j}$ for $1 \le i \le j \le p$). As shown in the ``Total Features'' column of Table~\ref{tab:dataset_features_10rows}, this augmentation results in three specific datasets, namely Ailerons, superconduct, and yprop\_4\_1, where the feature dimension $p$ exceeds the training size $n=400$. We categorize these as our High-Dimensional (HD) cases to assess each model's capability when $n \le p$. Each R\(^2\) score \eqref{r_square.1} is computed based on an independent test sample of size 2000. % The training sample is further divided into training (80\%) and validation (20\%) sets. 

We train a progressive tree for $b$ iterations ($\textnormal{PT}^{(b)}$) using the full training set without hyperparameter tuning, with the configuration $(\texttt{\#}S, H, B, s) = (100, 3, \infty, 5)$. Predictive performance and runtimes are recorded at $b \in \{1000, 2000, 4000, 8000\}$. Simultaneously, we train the $\textnormal{RF}+\mathcal{S}^{(b)}$ model with hyperparameter tuning as described in Section~\ref{Sec4} at the same $b$ intervals. Note that to maintain consistency with the hyperparameter tuning scheme, the progressive tree utilized to generate the candidate splits $\mathcal{S}^{(b)}$ for the $\textnormal{RF}+\mathcal{S}^{(b)}$ model is trained on a 80\% subsample of the training data rather than the full set. The hyperparameters for $\textnormal{RF}+\mathcal{S}^{(b)}$, F-RC, RF, and MORF are optimized over 100 iterations as described in Section~\ref{Sec4}. To manage the computational demand, we set the number of bagging trees to 30 during the tuning phase for these ensembles, whereas 100 trees are used for calculating the final prediction scores.

To ensure reliable results, we repeat each evaluation 20 times across 19 datasets. For each trial, we compute a relative R$^2$ score (RRS) to normalize performance across different data distributions. For a method $m$ on dataset $d$, the score is:
$$RRS_{m,d} = \frac{\text{R}^2_{m,d} - \min(\text{R}^2_{d})}{\max(\text{R}^2_{d}) - \min(\text{R}^2_{d})}$$
Here, $\min(\text{R}^2_{d})$ and $\max(\text{R}^2_{d})$ are the lowest and highest scores achieved by any compared method on that dataset $d$. This interpolation assigns a 1 to the best performer and a 0 to the worst, scaling all other results proportionally in between. We compute RRS for the seven methods in Tables~\ref{tab:overall.b} and \ref{tab:overall.a}, excluding standalone progressive trees. As shown in Table~\ref{tab:r_square_updated_high}, their significantly lower R$^2$ scores would compress the ensemble results toward 1. This exclusion prevents a loss of resolution and ensures that performance differences among the top models remain visible. Next, for each independent trial, we calculate the average RRS across the datasets for each method. We term the average across all 19 datasets as ALL-RRS and the average across the three high-dimensional datasets (Table~\ref{tab:dataset_features_10rows}) as HD-RRS. We then report the maximum, average, and minimum of these HD-RRS and ALL-RRS values across 20 independent trials in Tables~\ref{tab:overall.b}--\ref{tab:overall.a}. This metric facilitates direct comparisons across different datasets and models. Similar benchmarking approaches are commonly employed in current literature \citep{grinsztajn2022tree}.

As shown in Table~\ref{tab:overall.b}, RF+$\mathcal{S}^{(b)}$ consistently outperforms F-RC, RF, and MORF. Specifically, it maintains a superior average HD-RRS with a margin exceeding $0.1$ across all values of $b$ for the three high-dimensional datasets. Furthermore, RF+$\mathcal{S}^{(8000)}$ requires less than 35\% of the runtime of F-RC on average (Table~\ref{tab:runtime_comparison}). These results validate our strategy of optimizing high-dimensional oblique splits through the progressive tree framework.

In Table~\ref{tab:overall.a}, RF+$\mathcal{S}^{(b)}$ demonstrates stable performance across 19 datasets, consistently outperforming all competitors by a margin of approximately $0.03$. While this improvement is more modest than in high-dimensional settings, the fact that the minimum ALL-RRS for RF+$\mathcal{S}^{(b)}$ is comparable to the average ALL-RRS of standard RF underscores its robustness. The superior performance of both RF+$\mathcal{S}^{(b)}$ and F-RC relative to axis-aligned tree methods suggests that real-world data benefit significantly from the bias reduction of oblique splits. Note that minor fluctuations across $b$ values persist due to the inherent randomness of split-weight sampling.

Table~\ref{tab:r_square_updated_high} presents the maximum, average, and minimum R$^2$ scores across the evaluated datasets, complementing the findings in Tables~\ref{tab:overall.b}--\ref{tab:overall.a}. In general, RF+$\mathcal{S}^{(b)}$, F-RC, and RF exhibit comparable R$^2$ performance, with the winning margins between them typically remaining below 0.09. A notable exception is the elevators dataset, where RF+$\mathcal{S}^{(b)}$ significantly outperforms the standard RF and F-RC; for instance, at $b=4000$, RF+$\mathcal{S}^{(b)}$ achieves an average R$^2$ of 0.601, whereas RF and F-RC trail at 0.509 and 0.526, respectively.

As expected, a single progressive tree (PT) generally does not match the predictive accuracy of ensemble models. This gap is partly due to the inherent variance-reduction benefits of bagging and Random Forests \citep{breiman2001random}. Furthermore, our progressive trees are intentionally evaluated as ``off-the-shelf'' estimators without the hyperparameter tuning typically applied to ensemble methods, as their primary design goal is the efficient optimization of oblique splits. Among the evaluated ensemble methods, MORF tends to underperform compared to the others, suggesting that feature distances may not be a dominant factor in the underlying structure of these specific datasets.

Table~\ref{tab:runtime_comparison} summarizes the computational efficiency of the evaluated methods. F-RC consistently exhibits the highest average runtimes, significantly exceeding both $PT^{(8000)}$ and RF+$\mathcal{S}^{(8000)}$ in all cases except for the low-dimensional medical\_charge dataset (9 features). Excluding this outlier, RF+$\mathcal{S}^{(8000)}$ typically requires less than 50\% of the training time of F-RC, with even greater efficiency gains observed on several high-dimensional tasks. Among ensemble models, the standard axis-aligned RF remains the most computationally efficient.

The training time for progressive trees scales linearly with $b$ and remains remarkably stable. This consistency stems from the omission of hyperparameter tuning, which avoids the unpredictable runtime overhead typical of exhaustive parameter searches.  For most datasets, $PT^{(8000)}$ requires between 70 and 100 seconds on average, with increased runtimes only on the three highest-dimensional datasets: Ailerons, superconduct, and yprop\_4\_1. Notably, the training time for the RF+$\mathcal{S}^{(b)}$ ensemble is often lower than the sum of $PT^{(b)}$ and standard RF runtimes. This mirrors the behavior observed in Section~\ref{Sec5.2}, where the same efficiency occurs because the progressive tree within our ensemble utilizes only 80\% of the training data. Finally, it is important to note that tree-based runtimes depend not only on sample size and dimensionality but also on the recursive node construction process. For example, the pol dataset (377 features) exhibits significantly shorter runtimes for the progressive tree compared to datasets with similar dimensions, reflecting the complexity of its underlying split structure.

In summary, RF+$\mathcal{S}^{(b)}$ significantly improves performance on high-dimensional datasets and remains a top-tier competitor across all 19 benchmarks when compared to standard oblique models like F-RC and axis-aligned RF. Notably, its training runtime is often less than 50\% of that required by F-RC. While the progressive tree exhibits lower predictive accuracy as a standalone model, this is consistent with the known instability of single decision trees compared to ensemble methods \citep{breiman2001random, james2013introduction}. Consequently, we recommend utilizing these optimized oblique trees within a Random Forest framework to maximize performance.

\begin{table}[t]
\centering
\scriptsize
    \begin{center}
        \begin{tabular}[t]{ |lccc| } \hline
            & maximum ${\textnormal{HD-RRS}}$ & average ${\textnormal{HD-RRS}}$ & minimum ${\textnormal{HD-RRS}}$ \\
            {\tiny RF+$\mathcal{S}^{(1000)}$} & 0.933 & 0.724 & 0.552 \\
            {\tiny RF+$\mathcal{S}^{(2000)}$} & 0.952 & 0.680 & 0.348 \\
            {\tiny RF+$\mathcal{S}^{(4000)}$} & 0.904 & 0.701 & 0.253 \\
            {\tiny RF+$\mathcal{S}^{(8000)}$} & 0.917 & 0.674 & 0.217 \\
            {\tiny F-RC} & 0.933 & 0.562 & 0.144 \\
            {\tiny RF} & 0.796 & 0.483 & 0.090 \\
            {\tiny MORF} & 0.664 & 0.270 & 0.000 \\
            \hline
        \end{tabular}
        \caption{The table reports, from left to right, the maximum, average, and minimum of HD-RRS scores across 20 independent evaluations for each method.  HD-RRS scores are calculated based on three high-dimensional datasets listed in Table~\ref{tab:dataset_features_10rows}.  } \label{tab:overall.b}
    \end{center}
\end{table}

\begin{table}[t]
\centering
\scriptsize
    \begin{center}
        \begin{tabular}[t]{ |lccc| } \hline
            & maximum ${\textnormal{ALL-RRS}}$ & average ${\textnormal{ALL-RRS}}$ & minimum ${\textnormal{ALL-RRS}}$ \\
            {\tiny RF+$\mathcal{S}^{(1000)}$} & 0.773 & 0.655 & 0.510 \\
            {\tiny RF+$\mathcal{S}^{(2000)}$} & 0.814 & 0.664 & 0.553 \\
            {\tiny RF+$\mathcal{S}^{(4000)}$} & 0.815 & 0.697 & 0.562 \\
            {\tiny RF+$\mathcal{S}^{(8000)}$} & 0.803 & 0.670 & 0.508 \\
            {\tiny F-RC} & 0.792 & 0.627 & 0.466 \\
            {\tiny RF} & 0.654 & 0.500 & 0.381 \\
            {\tiny MORF} & 0.424 & 0.305 & 0.191 \\
            \hline
        \end{tabular}
        \caption{The table reports, from left to right, the maximum, average, and minimum of ALL-RRS scores across 20 independent evaluations for each method. } \label{tab:overall.a}
    \end{center}
\end{table}

\begin{table}[ht!]\centering\scriptsize
\setlength{\tabcolsep}{2pt}
\resizebox{0.98\textwidth}{!}{
\begin{tabular}{|l|cccc|}\hline
   & Ailerons (594) & superconduct (3239) & yprop\_4\_1 (945) & cpu\_act (252) \\ \hline
            $PT^{(1000)}$ & (0.724, 0.690, 0.642) & (0.719, 0.672, 0.587) & (-0.093, -0.223, -0.325) & (0.957, 0.929, 0.856) \\
            $PT^{(2000)}$ & (0.744, 0.683, 0.625) & (0.722, 0.664, 0.609) & (-0.119, -0.250, -0.433) & (0.957, 0.927, 0.853) \\
            $PT^{(4000)}$ & (0.741, 0.694, 0.634) & (0.715, 0.672, 0.583) & (-0.169, -0.291, -0.504) & (0.958, 0.922, 0.798) \\
            $PT^{(8000)}$ & (0.737, 0.694, 0.614) & (0.725, 0.673, 0.607) & (-0.116, -0.285, -0.501) & (0.955, 0.917, 0.795) \\
            RF+$\mathcal{S}^{(1000)}$ & (0.820, 0.784, 0.727) & (0.807, 0.781, 0.752) & (0.044, 0.011, -0.010) & (0.977, 0.971, 0.955) \\
            RF+$\mathcal{S}^{(2000)}$ & (0.818, 0.784, 0.739) & (0.807, 0.781, 0.736) & (0.035, 0.001, -0.079) & (0.976, 0.970, 0.949) \\
            RF+$\mathcal{S}^{(4000)}$ & (0.827, 0.790, 0.752) & (0.802, 0.775, 0.698) & (0.042, 0.006, -0.063) & (0.976, 0.970, 0.958) \\
            RF+$\mathcal{S}^{(8000)}$ & (0.825, 0.790, 0.756) & (0.804, 0.776, 0.726) & (0.046, 0.003, -0.076) & (0.977, 0.971, 0.957) \\
            F-RC & (0.802, 0.776, 0.749) & (0.816, 0.781, 0.751) & (0.026, -0.002, -0.077) & (0.980, 0.974, 0.970) \\
            RF & (0.790, 0.758, 0.726) & (0.799, 0.775, 0.708) & (0.043, 0.009, -0.043) & (0.977, 0.971, 0.955) \\
            MORF & (0.789, 0.747, 0.673) & (0.822, 0.765, 0.668) & (0.038, -0.014, -0.077) & (0.977, 0.970, 0.961) \\
\hline
   & pol (377) & elevators (152) & wine\_quality (77) & houses (44) \\ \hline
            $PT^{(1000)}$ & (0.797, 0.730, 0.569) & (0.573, 0.441, 0.284) & (0.217, 0.143, 0.055) & (0.667, 0.607, 0.533) \\
            $PT^{(2000)}$ & (0.791, 0.726, 0.581) & (0.548, 0.442, 0.232) & (0.208, 0.130, 0.037) & (0.640, 0.600, 0.549) \\
            $PT^{(4000)}$ & (0.830, 0.738, 0.614) & (0.570, 0.435, 0.210) & (0.227, 0.128, 0.062) & (0.644, 0.604, 0.548) \\
            $PT^{(8000)}$ & (0.823, 0.743, 0.622) & (0.567, 0.439, 0.268) & (0.209, 0.125, 0.039) & (0.635, 0.599, 0.505) \\
            RF+$\mathcal{S}^{(1000)}$ & (0.925, 0.896, 0.828) & (0.674, 0.585, 0.433) & (0.337, 0.298, 0.169) & (0.763, 0.713, 0.624) \\
            RF+$\mathcal{S}^{(2000)}$ & (0.924, 0.898, 0.867) & (0.719, 0.588, 0.491) & (0.351, 0.296, 0.127) & (0.773, 0.718, 0.650) \\
            RF+$\mathcal{S}^{(4000)}$ & (0.929, 0.901, 0.859) & (0.694, 0.601, 0.509) & (0.346, 0.298, 0.240) & (0.773, 0.725, 0.685) \\
            RF+$\mathcal{S}^{(8000)}$ & (0.928, 0.895, 0.810) & (0.692, 0.601, 0.455) & (0.357, 0.299, 0.222) & (0.770, 0.723, 0.686) \\
            F-RC & (0.939, 0.907, 0.883) & (0.597, 0.526, 0.268) & (0.338, 0.301, 0.243) & (0.743, 0.706, 0.646) \\
            RF & (0.933, 0.901, 0.862) & (0.584, 0.509, 0.385) & (0.326, 0.297, 0.238) & (0.689, 0.641, 0.575) \\
            MORF & (0.925, 0.886, 0.848) & (0.529, 0.458, 0.308) & (0.340, 0.265, 0.086) & (0.736, 0.688, 0.602) \\
\hline
   & house\_16H (152) & diamonds (27) & Brazilian\_houses (44) & Bike\_Sharing\_Demand (27) \\ \hline
            $PT^{(1000)}$ & (0.433, 0.014, -1.005) & (0.925, 0.916, 0.907) & (0.970, 0.954, 0.929) & (0.504, 0.395, 0.210) \\
            $PT^{(2000)}$ & (0.430, 0.088, -0.757) & (0.923, 0.916, 0.906) & (0.970, 0.955, 0.928) & (0.523, 0.390, 0.218) \\
            $PT^{(4000)}$ & (0.465, 0.082, -0.743) & (0.922, 0.915, 0.903) & (0.972, 0.958, 0.943) & (0.544, 0.418, 0.306) \\
            $PT^{(8000)}$ & (0.451, 0.105, -0.929) & (0.926, 0.916, 0.902) & (0.971, 0.957, 0.930) & (0.510, 0.415, 0.200) \\
            RF+$\mathcal{S}^{(1000)}$ & (0.600, 0.433, 0.309) & (0.940, 0.935, 0.923) & (0.998, 0.985, 0.971) & (0.651, 0.562, 0.503) \\
            RF+$\mathcal{S}^{(2000)}$ & (0.585, 0.428, 0.333) & (0.940, 0.935, 0.921) & (0.998, 0.985, 0.970) & (0.654, 0.563, 0.523) \\
            RF+$\mathcal{S}^{(4000)}$ & (0.602, 0.432, 0.327) & (0.941, 0.935, 0.918) & (0.998, 0.986, 0.973) & (0.662, 0.563, 0.513) \\
            RF+$\mathcal{S}^{(8000)}$ & (0.568, 0.412, 0.288) & (0.940, 0.935, 0.926) & (0.998, 0.986, 0.972) & (0.660, 0.569, 0.516) \\
            F-RC & (0.607, 0.431, 0.294) & (0.942, 0.935, 0.923) & (0.997, 0.983, 0.964) & (0.622, 0.548, 0.455) \\
            RF & (0.601, 0.418, 0.247) & (0.939, 0.934, 0.918) & (0.992, 0.969, 0.945) & (0.645, 0.565, 0.523) \\
            MORF & (0.571, 0.392, 0.237) & (0.938, 0.930, 0.919) & (0.999, 0.989, 0.973) & (0.600, 0.517, 0.329) \\
\hline
   & nyc-taxi-green-dec-2016 (54) & house\_sales (135) & sulfur (27) & medical\_charges (9) \\ \hline
            $PT^{(1000)}$ & (0.299, 0.152, -0.084) & (0.663, 0.632, 0.550) & (0.732, 0.400, -0.985) & (0.953, 0.943, 0.927) \\
            $PT^{(2000)}$ & (0.261, 0.178, -0.035) & (0.669, 0.630, 0.581) & (0.665, 0.502, 0.088) & (0.953, 0.943, 0.927) \\
            $PT^{(4000)}$ & (0.264, 0.179, 0.018) & (0.658, 0.632, 0.601) & (0.669, 0.438, -0.500) & (0.953, 0.942, 0.924) \\
            $PT^{(8000)}$ & (0.255, 0.182, 0.124) & (0.681, 0.625, 0.529) & (0.704, 0.477, -0.261) & (0.953, 0.942, 0.914) \\
            RF+$\mathcal{S}^{(1000)}$ & (0.407, 0.356, 0.297) & (0.813, 0.788, 0.765) & (0.803, 0.648, 0.342) & (0.979, 0.973, 0.961) \\
            RF+$\mathcal{S}^{(2000)}$ & (0.441, 0.357, 0.293) & (0.807, 0.791, 0.762) & (0.792, 0.648, 0.364) & (0.979, 0.972, 0.960) \\
            RF+$\mathcal{S}^{(4000)}$ & (0.444, 0.361, 0.304) & (0.808, 0.787, 0.762) & (0.790, 0.664, 0.508) & (0.980, 0.973, 0.959) \\
            RF+$\mathcal{S}^{(8000)}$ & (0.444, 0.362, 0.304) & (0.804, 0.787, 0.770) & (0.791, 0.646, 0.353) & (0.979, 0.973, 0.960) \\
            F-RC & (0.427, 0.354, 0.247) & (0.794, 0.772, 0.732) & (0.782, 0.624, 0.470) & (0.979, 0.972, 0.959) \\
            RF & (0.418, 0.345, 0.267) & (0.804, 0.793, 0.769) & (0.784, 0.648, 0.353) & (0.979, 0.973, 0.960) \\
            MORF & (0.405, 0.337, 0.243) & (0.805, 0.767, 0.721) & (0.783, 0.605, 0.459) & (0.979, 0.969, 0.953) \\
\hline
   & MiamiHousing2016 (104) & abalone (35) & delay\_zurich\_transport (44)  & \\ \hline
            $PT^{(1000)}$ & (0.684, 0.631, 0.590) & (0.484, 0.408, 0.280) & (-0.074, -0.175, -0.271)  & \\
            $PT^{(2000)}$ & (0.689, 0.630, 0.525) & (0.493, 0.413, 0.295) & (-0.085, -0.176, -0.305)  & \\
            $PT^{(4000)}$ & (0.695, 0.638, 0.553) & (0.505, 0.409, 0.305) & (-0.070, -0.186, -0.321)  & \\
            $PT^{(8000)}$ & (0.705, 0.632, 0.587) & (0.506, 0.427, 0.319) & (-0.123, -0.204, -0.323)  & \\
            RF+$\mathcal{S}^{(1000)}$ & (0.815, 0.789, 0.739) & (0.534, 0.504, 0.468) & (0.015, -0.042, -0.167)  & \\
            RF+$\mathcal{S}^{(2000)}$ & (0.815, 0.794, 0.758) & (0.574, 0.515, 0.441) & (0.005, -0.030, -0.083)  & \\
            RF+$\mathcal{S}^{(4000)}$ & (0.822, 0.790, 0.750) & (0.554, 0.519, 0.430) & (0.017, -0.035, -0.177)  & \\
            RF+$\mathcal{S}^{(8000)}$ & (0.817, 0.792, 0.730) & (0.560, 0.517, 0.459) & (0.006, -0.039, -0.147)  & \\
            F-RC & (0.839, 0.816, 0.776) & (0.577, 0.545, 0.482) & (-0.008, -0.064, -0.164)  & \\
            RF & (0.815, 0.795, 0.742) & (0.530, 0.479, 0.419) & (0.015, -0.043, -0.179)  & \\
            MORF & (0.821, 0.760, 0.659) & (0.508, 0.461, 0.284) & (0.002, -0.052, -0.148)  & \\
\hline
\end{tabular}}
\caption{The values of (maximum, average, and minimum R$^2$ scores over 20 independent evaluations) are reported for each dataset and model. The numbers in parentheses following the dataset names represent the total number of features (see Table~\ref{tab:dataset_features_10rows})}\label{tab:r_square_updated_high}
\end{table}

\clearpage
\begin{table}[ht!]\centering\scriptsize
\setlength{\tabcolsep}{2pt}
\resizebox{0.98\textwidth}{!}{
\begin{tabular}{|l|cccc|}\hline
  & Ailerons (594) & superconduct (3239) & yprop\_4\_1 (945) & cpu\_act (252) \\ \hline
            $PT^{(1000)}$ & (15.1, 14.9, 14.8) & (24.7, 24.1, 23.3) & (17.0, 16.8, 15.4) & (12.6, 12.4, 12.3) \\
            $PT^{(2000)}$ & (30.1, 29.8, 29.7) & (48.8, 47.4, 45.1) & (34.1, 33.6, 32.2) & (25.1, 24.9, 24.5) \\
            $PT^{(4000)}$ & (59.7, 59.4, 59.2) & (96.5, 93.6, 87.1) & (68.0, 67.0, 65.7) & (50.1, 49.6, 49.0) \\
            $PT^{(8000)}$ & (119.2, 118.5, 117.9) & (191.3, 186.6, 178.0) & (135.2, 133.7, 132.2) & (99.8, 98.9, 98.0) \\
            RF+$\mathcal{S}^{(1000)}$ & (99.5, 52.2, 33.7) & (622.3, 417.3, 302.3) & (179.1, 80.5, 36.8) & (138.8, 63.1, 37.7) \\
            RF+$\mathcal{S}^{(2000)}$ & (146.5, 74.6, 43.8) & (835.4, 458.7, 319.8) & (202.1, 104.7, 48.3) & (106.6, 62.2, 42.8) \\
            RF+$\mathcal{S}^{(4000)}$ & (171.7, 101.9, 71.8) & (979.6, 551.7, 363.3) & (191.5, 115.1, 74.4) & (136.4, 82.9, 65.2) \\
            RF+$\mathcal{S}^{(8000)}$ & (203.4, 142.9, 117.2) & (1150.4, 595.3, 398.2) & (270.4, 173.5, 132.3) & (183.2, 135.6, 108.6) \\
            F-RC & (896.7, 480.3, 243.4) & (2756.0, 1906.9, 1241.3) & (1237.3, 510.5, 163.3) & (1551.1, 872.2, 452.2) \\
            RF & (106.7, 40.9, 22.5) & (1208.5, 449.9, 279.6) & (143.2, 63.8, 24.8) & (64.4, 38.7, 30.2) \\
            MORF & (737.6, 251.1, 151.7) & (1299.8, 844.7, 323.1) & (663.1, 374.3, 110.0) & (1522.4, 644.8, 364.0) \\
\hline
  & pol (377) & elevators (152) & wine\_quality (77) & houses (44) \\ \hline
            $PT^{(1000)}$ & (11.1, 10.9, 10.7) & (12.1, 11.9, 11.7) & (11.4, 11.2, 11.0) & (10.8, 10.6, 10.4) \\
            $PT^{(2000)}$ & (22.1, 21.9, 21.6) & (24.1, 23.8, 23.5) & (22.6, 22.4, 22.0) & (21.4, 21.1, 20.8) \\
            $PT^{(4000)}$ & (44.6, 43.6, 43.3) & (48.2, 47.5, 46.7) & (45.2, 44.7, 44.0) & (42.8, 42.0, 41.6) \\
            $PT^{(8000)}$ & (87.8, 87.1, 86.2) & (96.0, 94.8, 93.6) & (90.2, 89.1, 88.0) & (85.4, 84.0, 83.1) \\
            RF+$\mathcal{S}^{(1000)}$ & (27.7, 22.1, 18.5) & (58.1, 30.8, 19.7) & (25.7, 17.3, 15.0) & (27.3, 18.5, 14.9) \\
            RF+$\mathcal{S}^{(2000)}$ & (36.5, 31.5, 27.6) & (66.7, 40.1, 30.2) & (35.5, 26.2, 24.4) & (38.5, 26.7, 22.8) \\
            RF+$\mathcal{S}^{(4000)}$ & (53.4, 49.3, 46.1) & (84.3, 59.5, 51.7) & (46.0, 43.9, 41.1) & (53.0, 42.2, 39.7) \\
            RF+$\mathcal{S}^{(8000)}$ & (92.8, 85.4, 80.8) & (126.9, 100.7, 86.3) & (90.6, 80.9, 78.0) & (85.9, 76.2, 72.5) \\
            F-RC & (415.1, 334.2, 264.8) & (963.8, 412.0, 234.4) & (1435.5, 624.7, 249.1) & (1041.9, 570.6, 269.6) \\
            RF & (18.7, 13.1, 9.3) & (37.2, 19.1, 9.0) & (13.9, 9.1, 6.4) & (13.1, 8.4, 5.7) \\
            MORF & (410.3, 227.4, 137.0) & (879.2, 372.9, 217.5) & (1068.5, 335.8, 204.2) & (755.2, 335.9, 223.2) \\
\hline
  & house\_16H (152) & diamonds (27) & Brazilian\_houses (44) & Bike\_Sharing\_Demand (27) \\ \hline
            $PT^{(1000)}$ & (11.8, 11.6, 11.3) & (10.6, 10.5, 10.4) & (10.6, 10.5, 10.4) & (10.8, 10.7, 10.5) \\
            $PT^{(2000)}$ & (23.7, 23.2, 22.7) & (21.1, 20.9, 20.7) & (21.1, 21.0, 20.8) & (21.6, 21.3, 21.1) \\
            $PT^{(4000)}$ & (47.1, 46.3, 45.3) & (41.9, 41.6, 41.4) & (42.2, 41.8, 41.4) & (43.1, 42.6, 42.3) \\
            $PT^{(8000)}$ & (97.1, 92.6, 90.4) & (83.4, 83.0, 82.6) & (84.4, 83.6, 82.8) & (86.2, 85.1, 84.4) \\
            RF+$\mathcal{S}^{(1000)}$ & (36.8, 28.6, 23.1) & (19.4, 14.3, 11.8) & (27.1, 18.8, 14.3) & (20.8, 15.4, 13.2) \\
            RF+$\mathcal{S}^{(2000)}$ & (71.8, 43.4, 32.5) & (26.5, 22.6, 19.9) & (31.9, 26.5, 22.4) & (29.4, 24.9, 21.2) \\
            RF+$\mathcal{S}^{(4000)}$ & (73.5, 57.7, 50.6) & (43.5, 39.5, 37.1) & (51.1, 43.8, 39.3) & (46.6, 40.8, 38.2) \\
            RF+$\mathcal{S}^{(8000)}$ & (134.1, 101.0, 87.9) & (75.9, 72.8, 70.0) & (83.3, 76.9, 73.6) & (82.8, 75.3, 73.3) \\
            F-RC & (1090.9, 567.4, 278.5) & (722.2, 297.4, 164.8) & (1052.4, 555.8, 233.7) & (862.7, 508.7, 193.7) \\
            RF & (59.7, 28.4, 13.5) & (9.3, 5.6, 3.3) & (14.4, 7.6, 4.2) & (13.0, 6.9, 4.2) \\
            MORF & (1111.5, 492.5, 267.5) & (671.6, 297.8, 161.1) & (927.3, 473.3, 252.4) & (940.8, 303.0, 181.0) \\
\hline
  & nyc-taxi-green-dec-2016 (54) & house\_sales (135) & sulfur (27) & medical\_charges (9) \\ \hline
            $PT^{(1000)}$ & (11.1, 10.9, 10.7) & (11.5, 11.3, 11.2) & (11.3, 11.0, 10.7) & (9.7, 9.6, 9.5) \\
            $PT^{(2000)}$ & (22.2, 21.8, 21.4) & (22.9, 22.6, 22.4) & (22.4, 21.9, 21.5) & (19.3, 19.2, 19.0) \\
            $PT^{(4000)}$ & (44.1, 43.6, 42.8) & (45.7, 45.2, 44.7) & (44.7, 43.7, 42.9) & (38.7, 38.3, 37.9) \\
            $PT^{(8000)}$ & (88.0, 86.9, 85.8) & (91.0, 90.3, 89.5) & (89.1, 87.3, 85.4) & (77.3, 76.6, 75.9) \\
            RF+$\mathcal{S}^{(1000)}$ & (24.9, 16.4, 12.5) & (44.5, 31.1, 19.9) & (20.8, 16.0, 12.9) & (13.3, 11.8, 10.2) \\
            RF+$\mathcal{S}^{(2000)}$ & (32.7, 26.6, 22.2) & (64.1, 41.3, 30.5) & (32.7, 25.4, 21.9) & (21.0, 19.5, 17.9) \\
            RF+$\mathcal{S}^{(4000)}$ & (50.5, 43.3, 38.9) & (85.0, 59.1, 47.3) & (48.1, 43.2, 40.0) & (37.2, 35.3, 33.2) \\
            RF+$\mathcal{S}^{(8000)}$ & (86.8, 78.6, 73.6) & (127.4, 98.2, 84.6) & (87.7, 79.5, 75.4) & (67.1, 65.4, 63.7) \\
            F-RC & (1019.8, 505.8, 228.8) & (940.7, 457.3, 245.3) & (952.3, 394.4, 160.5) & (79.1, 40.7, 23.0) \\
            RF & (23.2, 8.4, 3.8) & (55.3, 27.8, 16.4) & (14.6, 7.9, 4.0) & (5.0, 3.4, 2.5) \\
            MORF & (860.9, 320.6, 222.1) & (1042.6, 472.0, 244.9) & (764.8, 371.0, 153.7) & (77.1, 48.5, 19.2) \\
\hline
  & MiamiHousing2016 (104) & abalone (35) & delay\_zurich\_transport (44)  & \\ \hline
            $PT^{(1000)}$ & (11.5, 11.3, 11.1) & (10.8, 10.7, 10.6) & (11.5, 11.1, 10.9)  & \\
            $PT^{(2000)}$ & (22.9, 22.6, 22.3) & (21.6, 21.3, 21.1) & (22.8, 22.1, 20.9)  & \\
            $PT^{(4000)}$ & (45.6, 45.0, 44.4) & (43.1, 42.6, 42.2) & (45.6, 44.2, 42.8)  & \\
            $PT^{(8000)}$ & (91.2, 89.8, 88.7) & (86.1, 84.9, 84.5) & (91.3, 88.4, 85.9)  & \\
            RF+$\mathcal{S}^{(1000)}$ & (30.2, 23.1, 19.9) & (21.1, 15.1, 12.8) & (26.3, 16.7, 13.4)  & \\
            RF+$\mathcal{S}^{(2000)}$ & (47.7, 33.8, 29.7) & (28.9, 23.2, 21.6) & (38.2, 26.1, 22.6)  & \\
            RF+$\mathcal{S}^{(4000)}$ & (75.7, 53.4, 46.9) & (42.1, 39.8, 38.4) & (59.5, 43.9, 39.3)  & \\
            RF+$\mathcal{S}^{(8000)}$ & (96.4, 86.7, 82.8) & (77.2, 73.8, 71.6) & (91.6, 79.7, 73.8)  & \\
            F-RC & (1124.2, 566.7, 320.8) & (1114.0, 496.9, 249.5) & (1125.8, 502.2, 234.7)  & \\
            RF & (37.4, 16.4, 9.2) & (7.7, 5.6, 4.1) & (21.1, 8.5, 4.1)  & \\
            MORF & (1109.4, 441.9, 245.5) & (809.8, 348.5, 227.6) & (1110.6, 465.7, 247.1)  & \\
\hline
\end{tabular}}
\caption{The maximum, average, and minimum computational runtimes (in seconds) over 20 independent evaluations are reported for each dataset and model. }\label{tab:runtime_comparison}
\end{table}

\section{Discussion}

In this paper, we establish SID convergence rates for high-dimensional oblique trees and introduce an iterative framework for optimizing oblique splits through a memory transfer strategy. This approach directly addresses the challenge of tuning candidate set sizes, as our iterative refinement effectively eliminates the need to pre-determine these parameters. To support these theoretical contributions, we have developed an open-source Python package that implements our method, thereby enhancing the accessibility and scalability of oblique trees for practical applications.

Several important research directions remain for future exploration. First, extending our theoretical results to fast SID convergence rates and incorporating bagging or column subsampling would provide a more comprehensive analysis of the RF+$\mathcal{S}^{(b)}$ framework. Second, it remains an open question whether faster algorithms can be developed to optimize high-dimensional oblique splits while maintaining the unique approximation power and SID convergence rates demonstrated by our progressive tree. Finally, it would be insightful to characterize the specific function classes targeted by alternative oblique methods, such as PPF, OC1, TAO, and LDA, to assess their efficiency relative to our iterative refinement approach.

% In this paper, we establish SID convergence rates for high-dimensional oblique trees and propose an iterative approach for optimizing oblique splits through transfer learning. Furthermore, we support our theoretical contributions by developing an open-source Python package that implements our method, improving the accessibility and scalability of oblique trees for practical, real-world applications.

% Several promising and important future research directions are worth exploring. First, an extension of our theoretical results to fast SID convergence rates~\citep{mazumder2024convergence} is a natural next step, along with incorporating bagging and column subsampling for a more comprehensive analysis of RF+$\mathcal{S}^{(b)}$. Second, it remains an open question whether a significantly faster algorithm can be developed for optimizing high-dimensional oblique splits—one that could leverage the unique approximation power of oblique trees, such as achieving SID convergence rates, while also reducing the observed computational cost of our progressive tree. Third, it would be insightful to examine the specific function classes targeted by other oblique tree methods, such as PPF~\citep{da2021projection}, OC1~\citep{murthy1994system}, TAO~\citep{carreira2018alternating}, and LDA~\citep{loh1988tree}, and to assess their efficiency in optimizing oblique splits compared to our progressive tree approach.

\section{Source Code and Data Availability}  \label{Sec8}
The Python source code for our experiments is publicly available at \url{https://github.com/xbb66kw/ohos}. The datasets used in this study are also available at the same repository. All real datasets were obtained from OpenML, though some have been subsampled to reduce file sizes for GitHub upload. For access to the original datasets, we refer to \citep{grinsztajn2022tree}.

% \bibliographystyle{unsrtnat}
% \bibliography{references}

%%%
%%%
%%%%
% Acknowledgements and Disclosure of Funding should go at the end, before appendices and references

% \acks{All acknowledgements go at the end of the paper before appendices and references. Moreover, you are required to declare funding (financial activities supporting the submitted work) and competing interests (related financial activities outside the submitted work). More information about this disclosure can be found on the JMLR website.}

% Manual newpage inserted to improve layout of sample file - not
% needed in general before appendices/bibliography.

\vskip 0.2in
\bibliography{sample}

\newpage
\appendix

\setcounter{page}{1}
\setcounter{section}{0}
\renewcommand{\theequation}{A.\arabic{equation}}
\renewcommand{\thesubsection}{A.\arabic{subsection}}
\setcounter{equation}{0}

\begin{center}{\bf \Large Supplementary Material to ``Optimizing High-Dimensional Oblique Splits"}
	
	\bigskip
	
	Chien-Ming Chi
\end{center}

\noindent This Supplementary Material includes the appendix of Section~\ref{Sec4} in Section~\ref{SecA.1}, the proofs of all main results in Sections~\ref{SecB}--\ref{SecC}, and additional technical details in Section~\ref{SecD}. All notation remains consistent with the main body of the paper.

\renewcommand{\thesubsection}{A.\arabic{subsection}}
\section{Hyperparameter spaces}\label{SecA.1}

To ensure reproducible results, we utilize the Python package \texttt{hyperopt} for hyperparameter optimization across all predictive models. The dataset is partitioned into training and validation sets, representing 80\% and 20\% of the available data, respectively. For each model, we perform $R$ search rounds. In each round, \texttt{hyperopt} samples a candidate configuration from the predefined hyperparameter space. The model is then trained on the training set and evaluated on the validation set. After $R$ iterations, the configuration yielding the highest validation score is selected as the optimal set. The final model is then trained using these optimal hyperparameters on the pooled training and validation data.

The hyperparameter spaces for Random Forests, Forest-RC, and MORF \citep{li2023manifold} are detailed in Tables~\ref{tab:validation_sapce_rf}--\ref{tab:validation_sapce_morf}. For additional technical details, we refer the reader to the official documentation of the respective software packages.

\begin{table}[ht]
		\begin{center}{
				\begin{tabular}[t]{ |ccc|}
    \hline
    Parameter name & Parameter type & Parameter space \\
                         \hline \textsf{n\_estimators} & integer & 100  \\
                        \textsf{gamma} &  non-negative real number & $\textnormal{Uniform}[0, 1]$ \\
                        \textsf{min\_samples\_split} & integer & $\{1, \dots, 20\}$\\ 
                        \textsf{min\_samples\_leaf} & integer& $\{2, \dots, 20\}$ \\
                        \textsf{min\_impurity\_decrease} & non-negative real number &  $ \textnormal{Uniform}\{0, 0.01, 0.02,0.05\}$ \\
                        \textsf{max\_depth} & non-negative integer &  Uniform$\{5, 10, 20, 50, \infty\}$\\
       \textsf{criterion} & string &  $[\textnormal{squared\_error}, \textnormal{absolute\_error}]$ \\
					 \hline 
			\end{tabular} }
			\caption{Hyperparameter space for tuning Random Forests. Details of these parameters can be found on the website \url{https://scikit-learn.org/stable/modules/generated/sklearn.ensemble.RandomForestClassifier.html}} \label{tab:validation_sapce_rf}
		\end{center}
	\end{table}

    \begin{table}[ht]
		\begin{center}{
        \resizebox{0.9\textwidth}{!}{
				\begin{tabular}[t]{ |ccc|}
    \hline
    Parameter name & Parameter type & Parameter space \\
                         \hline \textsf{n\_estimators} & integer & 100  \\
                        \textsf{min\_samples\_split} & integer & $\{1, \dots, 20\}$\\ 
                        \textsf{min\_samples\_leaf} & integer& $\{2, \dots, 20\}$ \\
                        \textsf{min\_impurity\_decrease} & non-negative real number &  $ \textnormal{Uniform}\{0, 0.01, 0.02,0.05\}$ \\
                        \textsf{max\_depth} & non-negative integer &  Uniform$\{5, 10, 20, 50, \infty\}$\\
       \textsf{criterion} & string &  $[\textnormal{squared\_error}, \textnormal{absolute\_error}]$ \\
       \textsf{bootstrap} & bool & [True, False]\\
       \textsf{feature\_combinations} & non-negative integer & [1, 2, 3, 4, 5, 6, 7, 8, 9 ,10, 20, None]\\
					 \hline 
			\end{tabular} } }
			\caption{Hyperparameter space for tuning Forest-RC. The values of the parameter \textsf{max\_features} is specified in the main text. Details of these hyperparameters can be found on the website \url{https://docs.neurodata.io/treeple/dev/generated/treeple.ObliqueRandomForestRegressor.html\#treeple.ObliqueRandomForestRegressor}.} \label{tab:validation_sapce_orf}
		\end{center}
	\end{table}

\begin{table}[ht]
		\begin{center}{
        \resizebox{0.9\textwidth}{!}{
				\begin{tabular}[t]{ |ccc|}
    \hline
    Parameter name & Parameter type & Parameter space \\
                         \hline \textsf{n\_estimators} & integer & 100  \\
                        \textsf{min\_samples\_split} & integer & $\{1, \dots, 20\}$\\ 
                        \textsf{min\_samples\_leaf} & integer& $\{2, \dots, 20\}$ \\
                        \textsf{min\_impurity\_decrease} & non-negative real number &  $ \textnormal{Uniform}\{0, 0.01, 0.02,0.05\}$ \\
                        \textsf{max\_depth} & non-negative integer &  Uniform$\{5, 10, 20, 50, \infty\}$\\
       \textsf{criterion} & string &  $[\textnormal{squared\_error}, \textnormal{absolute\_error}]$ \\
       \textsf{bootstrap} & bool & [True, False]\\
       \textsf{max\_patch\_dims} & [integer, integer] & [ 1, $\{1, \dots, 15\}$] \\
                \textsf{min\_patch\_dims} & [integer, integer] & [ 1, $\{1, \dots, 15\}$ (less than \textsf{max\_patch\_dims})]\\
					 \hline 
			\end{tabular} }}
			\caption{Hyperparameter space for tuning MORF. The values of the parameter \textsf{max\_features} is specified in the main text. Details of these hyperparameters can be found on the website \url{https://docs.neurodata.io/treeple/dev/generated/treeple.PatchObliqueRandomForestRegressor.html\#treeple.PatchObliqueRandomForestRegressor}.}\label{tab:validation_sapce_morf}
		\end{center}
	\end{table}

% TO get the appendix back, uncommented this
%\begin{comment}

% \renewcommand{\thesection}{\Alph{section}.\arabic{section}}
% \setcounter{section}{0}

\renewcommand{\thesubsection}{B.\arabic{subsection}} 
\setcounter{equation}{0}
\renewcommand\theequation{B.\arabic{equation}}

\section{Proofs of Theorems}\label{SecB}

\subsection{Proof of Theorem~\ref{theorem1}}\label{proof.theorem1}

%%%
%%%

Fix $h \in \{1, \dots, H\}$ and $(\boldsymbol{t}^{'}, (\vv{w}, c), \boldsymbol{t}) \in \widehat{N}_{h}$ as specified in Theorem~\ref{theorem1}. Given these parameters, we begin by establishing an upper bound for \eqref{theorem1.14} that holds for each $(\vv{w}^{'}, c^{'}) \in \mathcal{W}$.
\begin{equation}
    \begin{split}\label{theorem1.14}
        & \left|\mathbb{E} [ \textnormal{Var} ( Y  \mid \boldsymbol{1}_{\boldsymbol{X} \in \boldsymbol{t}^{'} }, \boldsymbol{1}_{\vv{w}^{'\top}\boldsymbol{X}>c^{'}} ) \boldsymbol{1}_{\boldsymbol{X} \in \boldsymbol{t}^{'} } ] - n^{-1} \inf_{ \vv{\gamma}^{'} \in \mathbb{R}^{2} } L (\vv{w}^{'}, c^{'}, \vv{\gamma}^{'}, \boldsymbol{t}^{'}) \right|,     
    \end{split}
\end{equation}
where the loss function $L (\cdot,\cdot,\cdot,\cdot)$ is defined by \eqref{tspursuit.4}.

As noted following \eqref{tspursuit.1}, the first term in \eqref{theorem1.14} represents the residual variance resulting from splitting node $\boldsymbol{X} \in \boldsymbol{t}^{'}$ at $(\vv{w}^{'}, c^{'})$. The second term then serves as the sample counterpart to this variance. After establishing the upper bound in \eqref{theorem1.36}, we prove that $(\vv{w}, c)$ achieves a nearly optimal variance reduction, within a margin of $2\rho_n$, as shown in \eqref{aim.2.b} and \eqref{aim.2} below, thereby completing the desired proof. Let us now establish a bound for \eqref{theorem1.14} as follows.

We deduce by the assumptions of $1\le p\le n^{K_{0}}$ and Condition~\ref{regularity.1} that on the event $E_{n}$, where  $E_{n}$ is defined by Lemma~\ref{lemma.3} in Section~\ref{proof.lemma3}, that for all large $n$, each $s\ge 1$, each $h\in \{1, \dots, H\}$, each $H\ge 1$ with $\sqrt{2}n^{-\frac{1}{2}} s \sqrt{H(1\vee\log{s})} \log{n}\le 1$, and each $(\vv{w}^{'}, c^{'}) \in \mathcal{W}$,
\begin{equation}
    \begin{split}\label{theorem1.36}
        \textnormal{\eqref{theorem1.14}} & \le  \rho_{n},
    \end{split}
\end{equation}
where  $\rho_{n}  = C_{0}  \times n^{-\frac{1}{2}} s \sqrt{H(1\vee\log{s})} \log{n} \times \left(M_{\epsilon} + \sup_{\vv{x} \in [0, 1]^p} |m(\vv{x})|\right)^2$, which is defined in Theorem~\ref{theorem1}.
The proof of \eqref{theorem1.36} is deferred to Section~\ref{proof.theorem1.36}. 

With \eqref{theorem1.36}, we derive that
\begin{equation}
    \begin{split}\label{aim.2.b}
         &  \mathbb{E} [ \textnormal{Var} ( Y  \mid \boldsymbol{1}_{\boldsymbol{X} \in \boldsymbol{t}^{'} }, \boldsymbol{1}_{\vv{w}^{\top}\boldsymbol{X}>c} ) \boldsymbol{1}_{\boldsymbol{X} \in \boldsymbol{t}^{'} } ] - \rho_{n} \\
         & \le \inf_{\vv{\gamma} \in \mathbb{R}^{2} } n^{-1} L (\vv{w}, c, \vv{\gamma}, \boldsymbol{t}^{'}) \\
         & = \inf_{(\vv{w}^{''}, c^{''}) \in \mathcal{W}, \vv{\gamma} \in \mathbb{R}^{2} } n^{-1} L (\vv{w}^{''}, c^{''}, \vv{\gamma}, \boldsymbol{t}^{'}) \\
        & \le \inf_{\vv{\gamma} \in \mathbb{R}^{2} } n^{-1} L (\vv{w}_{2}, c_{2}, \vv{\gamma}, \boldsymbol{t}^{'}) \\
        & \le  \mathbb{E} [ \textnormal{Var} ( Y  \mid \boldsymbol{1}_{\boldsymbol{X} \in \boldsymbol{t}^{'} }, \boldsymbol{1}_{\vv{w}_{2}^{\top}\boldsymbol{X}>c_{2}} ) \boldsymbol{1}_{\boldsymbol{X} \in \boldsymbol{t}^{'} } ] + \rho_{n}\\
        & = \inf_{(\vv{w}^{''}, c^{''})\in \mathcal{W}} \mathbb{E} [ \textnormal{Var} ( Y  \mid \boldsymbol{1}_{\boldsymbol{X} \in \boldsymbol{t}^{'} }, \boldsymbol{1}_{\vv{w}^{''\top}\boldsymbol{X}>c^{''}} ) \boldsymbol{1}_{\boldsymbol{X} \in \boldsymbol{t}^{'} } ]  +\rho_{n},
    \end{split}
\end{equation}
where 
\begin{equation*}
    \begin{split}
        (\vv{w}_{2}, c_{2}) &\in \argmin_{(\vv{w}^{''}, c^{''})\in \mathcal{W}} \mathbb{E} [ \textnormal{Var} ( Y  \mid \boldsymbol{1}_{\boldsymbol{X} \in \boldsymbol{t}^{'} }, \boldsymbol{1}_{\vv{w}^{''\top}\boldsymbol{X}>c^{''}} ) \boldsymbol{1}_{\boldsymbol{X} \in \boldsymbol{t}^{'} } ].
    \end{split}
\end{equation*}
Here, the first and third inequalities of \eqref{aim.2.b} holds because of \eqref{theorem1.36}, the second follows from the definition of the infimum, and the first equality holds because $ (\vv{w}, c)  \in \argmin_{ (\vv{w}, c) \in \mathcal{W}} \inf_{ \vv{\gamma} \in \mathbb{R}^{2} } L (\vv{w}, c, \vv{\gamma}, \boldsymbol{t}^{'})$, which follows from that $(\boldsymbol{t}^{'}, (\vv{w}, c), \boldsymbol{t})\in \widehat{N}_{h}$ and the definition of $\widehat{N}_{h}$ given by  \eqref{tspursuit.4}.

By \eqref{theorem1.36}--\eqref{aim.2.b}, we conclude that
\begin{equation}
    \begin{split}\label{aim.2}
         (\vv{w}, c) \in E(\rho_{n}, \boldsymbol{t}^{'}, \mathcal{W}),
    \end{split}
\end{equation}
where the notation $E(\cdot, \cdot, \cdot)$ is defined by \eqref{tspursuit.1}.

In addition, by Lemma~\ref{lemma.3} in Section~\ref{proof.lemma3}, it holds that 
\begin{equation}
    \begin{split}
        \label{theorem1.30}
        \mathbb{P}(E_{n}^c ) \le  4\exp{\left(-\frac{1}{2} s^2 (\log{n})^2 (1\vee\log{s})\right)}.
    \end{split}
\end{equation}

Additionally, the results of \eqref{aim.2} and \eqref{theorem1.30} hold for each $h \in \{1, \dots, H\}$ and every $(\boldsymbol{t}^{'}, (\vv{w}, c), \boldsymbol{t})\in \widehat{N}_{h}$. Therefore, we have completed the proof of Theorem~\ref{theorem1}.

\subsubsection{Proof of \texorpdfstring{\eqref{theorem1.36}}{LG}}\label{proof.theorem1.36}

Define that
\begin{equation*}
	\begin{split}
\widehat{R}(\boldsymbol{X}) &= \sum_{(\boldsymbol{t}^{'}, (\vv{w}, c), \boldsymbol{t}) \in \widehat{N}_{H}} \boldsymbol{1}_{\boldsymbol{X} \in \boldsymbol{t}} \times \widehat{\beta}(\boldsymbol{t}), \\
R(\boldsymbol{X}) & = \sum_{(\boldsymbol{t}^{'}, (\vv{w}, c), \boldsymbol{t}) \in \widehat{N}_{H}} \boldsymbol{1}_{\boldsymbol{X} \in \boldsymbol{t}} \times \mathbb{E}(Y| \boldsymbol{X}\in \boldsymbol{t}).
	\end{split}
\end{equation*}
 Consider some  $(\boldsymbol{t}^{'}, (\vv{w}, c), \boldsymbol{t})\in \widehat{N}_{h}$, and let  $(\vv{w}^{'}, c^{'}) \in \mathcal{W}$ be some generic split. Denote $\boldsymbol{t}_{1}^{'} = \{\vv{x}\in [0, 1]^p: \vv{w}^{'\top}\vv{x} > c^{'}, \vv{x} \in \boldsymbol{t}^{'}\}$ and $\boldsymbol{t}_{2}^{'} = \{\vv{x}\in [0, 1]^p: \vv{w}^{'\top}\vv{x} \le c^{'}, \vv{x} \in \boldsymbol{t}^{'}\}$, and that
\begin{equation}
    \begin{split}\label{theorem1.13}
         \widehat{\beta}(\boldsymbol{t}_{1}^{'}) & = \left[\sum_{i=1}^{n}  \boldsymbol{1}_{\boldsymbol{X}_{i} \in \boldsymbol{t}^{'}}\boldsymbol{1}_{\vv{w}^{'\top}\boldsymbol{X}_{i} > c^{'}} \right]^{-1}\times \sum_{i=1}^{n} \left[ Y_{i}   \boldsymbol{1}_{\boldsymbol{X}_{i} \in \boldsymbol{t}^{'}} \boldsymbol{1}_{\vv{w}^{'\top}\boldsymbol{X}_{i} > c^{'}} \right] \\
         & = \left[\sum_{i=1}^{n}  \boldsymbol{1}_{\boldsymbol{X}_{i} \in \boldsymbol{t}_{1}^{'}} \right]^{-1}\times \sum_{i=1}^{n} \left[ Y_{i}   \boldsymbol{1}_{\boldsymbol{X}_{i} \in \boldsymbol{t}_{1}^{'}} \right] ,\\
        \widehat{\beta}(\boldsymbol{t}_{2}^{'}) & = [\sum_{i=1}^{n}  \boldsymbol{1}_{\boldsymbol{X}_{i} \in \boldsymbol{t}^{'}}\boldsymbol{1}_{\vv{w}^{'\top}\boldsymbol{X}_{i} \le c^{'}} ]^{-1}\times \sum_{i=1}^{n} \left[ Y_{i}   \boldsymbol{1}_{\boldsymbol{X}_{i} \in \boldsymbol{t}^{'}} \boldsymbol{1}_{\vv{w}^{'\top}\boldsymbol{X}_{i} \le c^{'}} \right]\\
                 & = \left[\sum_{i=1}^{n}  \boldsymbol{1}_{\boldsymbol{X}_{i} \in \boldsymbol{t}_{2}^{'}} \right]^{-1}\times \sum_{i=1}^{n} \left[ Y_{i}   \boldsymbol{1}_{\boldsymbol{X}_{i} \in \boldsymbol{t}_{2}^{'}} \right] .\\
    \end{split}
\end{equation}

Let us derive that
\begin{equation}
    \begin{split}\label{theorem1.1.new}
        & \mathbb{E} [ \textnormal{Var} ( Y  \mid \boldsymbol{1}_{\boldsymbol{X} \in \boldsymbol{t}^{'} }, \boldsymbol{1}_{\vv{w}^{'\top}\boldsymbol{X}>c^{'}} ) \boldsymbol{1}_{\boldsymbol{X} \in \boldsymbol{t}^{'} } ] \\
        & = \mathbb{E} (  Y ^2 \boldsymbol{1}_{\boldsymbol{X} \in \boldsymbol{t}^{'} }) - [\mathbb{E}(Y| \boldsymbol{X}\in\boldsymbol{t}_{1}^{'})]^2 \times \mathbb{P}(\boldsymbol{X}\in\boldsymbol{t}_{1}^{'}) - [\mathbb{E}(Y| \boldsymbol{X}\in\boldsymbol{t}_{2}^{'})]^2 \times \mathbb{P}(\boldsymbol{X}\in\boldsymbol{t}_{2}^{'}).
    \end{split}
\end{equation}
Meanwhile, a simple calculation shows that
\begin{equation}
    \begin{split}\label{theorem1.2.new}
      \min_{ \vv{\gamma}^{'} \in \mathbb{R}^{2} } L (\vv{w}^{'}, c^{'}, \vv{\gamma}^{'}, \boldsymbol{t}^{'}) & =    \sum_{i=1}^{n} \boldsymbol{1}_{\boldsymbol{X}_{i}\in \boldsymbol{t}^{'}}\left[     Y_{i}  -  \boldsymbol{1}_{\vv{w}^{'\top}\boldsymbol{X} > c^{'}} \widehat{\beta}(\boldsymbol{t}_{1}^{'})  - \boldsymbol{1}_{\vv{w}^{'\top}\boldsymbol{X} \le c^{'}} \widehat{\beta}(\boldsymbol{t}_{2}^{'})  \right]^2 \\
& = \left[ \sum_{i=1}^{n} ( Y_{i} \boldsymbol{1}_{\boldsymbol{X}_{i}\in \boldsymbol{t}^{'}})^2 \right] -   \sum_{ l = 1}^{2 } [\widehat{\beta}(\boldsymbol{t}_{l}^{'})]^{2} \times \left[\sum_{i=1}^{n}  \boldsymbol{1}_{\boldsymbol{X}_{i} \in \boldsymbol{t}_{l}^{'}} \right].
    \end{split}
\end{equation}

Therefore, 
\begin{equation}
    \begin{split}\label{theorem1.3.new}
        & \textnormal{\eqref{theorem1.14}} \\
        & \le \left| \mathbb{E} (  Y ^2 \boldsymbol{1}_{\boldsymbol{X} \in \boldsymbol{t}^{'} })   - n^{-1} \left[ \sum_{i=1}^{n} \left( Y_{i} \boldsymbol{1}_{\boldsymbol{X}_{i}\in \boldsymbol{t}^{'}}\right)^2 \right] \right| \\
        &\quad + \left| \left[\sum_{l=1}^2 [\mathbb{E}(Y| \boldsymbol{X}\in\boldsymbol{t}_{l}^{'})]^2 \times \mathbb{P}(\boldsymbol{X}\in\boldsymbol{t}_{l}^{'}) \right]  - \sum_{ l = 1}^{2 } [\widehat{\beta}(\boldsymbol{t}_{l}^{'})]^{2} \times n^{-1} \left[\sum_{i=1}^{n}  \boldsymbol{1}_{\boldsymbol{X}_{i} \in \boldsymbol{t}_{l}^{'}}\right]  \right| .
    \end{split}
\end{equation}

To establish the the upper bound for the first term on the RHS of \eqref{theorem1.3.new}, we deduce that
\begin{equation}
    \begin{split}\label{theorem1.30.new}
        & \bigg| \mathbb{E} (  Y ^2 \boldsymbol{1}_{\boldsymbol{X} \in \boldsymbol{t}^{'} })   - n^{-1} \left[ \sum_{i=1}^{n} ( Y_{i} \boldsymbol{1}_{\boldsymbol{X}_{i}\in \boldsymbol{t}^{'}})^2 \right] \bigg|\\
        & = \bigg| \mathbb{E} (  [m(\boldsymbol{X})]^2 \boldsymbol{1}_{\boldsymbol{X} \in \boldsymbol{t}^{'} })   - n^{-1} \left[ \sum_{i=1}^{n} [m(\boldsymbol{X}_{i})]^2 \boldsymbol{1}_{\boldsymbol{X}_{i}\in \boldsymbol{t}^{'}} \right] \\
        & \quad + 2 \left[\mathbb{E} (  m(\boldsymbol{X}) \times \varepsilon\times \boldsymbol{1}_{\boldsymbol{X} \in \boldsymbol{t}^{'} }) -  n^{-1} \sum_{i=1}^{n}  m(\boldsymbol{X}_{i})\times \varepsilon_{i}\times \boldsymbol{1}_{\boldsymbol{X}_{i}\in \boldsymbol{t}^{'}} \right]\\
        & \quad + \mathbb{E} (  \varepsilon^2 \boldsymbol{1}_{\boldsymbol{X} \in \boldsymbol{t}^{'} })   - n^{-1} \left[ \sum_{i=1}^{n}  \varepsilon_{i}^2 \boldsymbol{1}_{\boldsymbol{X}_{i}\in \boldsymbol{t}^{'}} \right] \bigg|\\
        & \le 48 \left(M_{\epsilon} + \sup_{\vv{x} \in [0, 1]^p} |m(\vv{x})|\right) n^{-\frac{1}{2}} s \sqrt{H(1\vee\log{s})} \log{n},
    \end{split}
\end{equation}
where the last inequality is due to an application of Lemma~\ref{lemma.3} in Section~\ref{proof.lemma3} in light of Condition~\ref{regularity.1}.

Now, let us proceed to deal with establishing the upper bound for the second term on the RHS of \eqref{theorem1.3.new}. We derive that 
\begin{equation}
    \begin{split}\label{theorem1.33}
        & \left| \left[\sum_{l=1}^2 [\mathbb{E}(Y| \boldsymbol{X}\in\boldsymbol{t}_{l}^{'})]^2 \times \mathbb{P}(\boldsymbol{X}\in\boldsymbol{t}_{l}^{'}) \right]  - \sum_{ l = 1}^{2 } [\widehat{\beta}(\boldsymbol{t}_{l}^{'})]^{2} \times n^{-1} \left[\sum_{i=1}^{n}  \boldsymbol{1}_{\boldsymbol{X}_{i} \in \boldsymbol{t}_{l}^{'}}\right]    \right| \\
        & \le \bigg|  \left[\sum_{l=1}^2 [\mathbb{E}(Y| \boldsymbol{X}\in\boldsymbol{t}_{l}^{'})]^2 \times \mathbb{P}(\boldsymbol{X}\in\boldsymbol{t}_{l}^{'}) \right]  - \sum_{ l = 1}^{2 } [\mathbb{E}(Y| \boldsymbol{X}\in\boldsymbol{t}_{l}^{'})]^2 \times n^{-1} \left[\sum_{i=1}^{n}  \boldsymbol{1}_{\boldsymbol{X}_{i} \in \boldsymbol{t}_{l}^{'}}\right]    \\
        & \quad + \left[\sum_{l=1}^2 [\mathbb{E}(Y| \boldsymbol{X}\in\boldsymbol{t}_{l}^{'})]^2 \times n^{-1}\sum_{i=1}^{n}  \boldsymbol{1}_{\boldsymbol{X}_{i} \in \boldsymbol{t}_{l}^{'}} \right]  - \sum_{ l = 1}^{2 } [\widehat{\beta}(\boldsymbol{t}_{l}^{'})]^{2} \times n^{-1} \left[\sum_{i=1}^{n}  \boldsymbol{1}_{\boldsymbol{X}_{i} \in \boldsymbol{t}_{l}^{'}}\right]     \bigg|.
    \end{split}
\end{equation}

Next, there exists some constant $C_{0}>0$ such that
\begin{equation}
    \begin{split}\label{theorem1.31.new}
        & \textnormal{RHS of \eqref{theorem1.33}} \\ 
        & \le \sum_{l=1}^2 [\mathbb{E}(Y| \boldsymbol{X}\in\boldsymbol{t}_{l}^{'})]^2 \times \left| \mathbb{P}(\boldsymbol{X}\in\boldsymbol{t}_{l}^{'}) \  - n^{-1} \sum_{i=1}^{n}  \boldsymbol{1}_{\boldsymbol{X}_{i} \in \boldsymbol{t}_{l}^{'}}\right|  \\
        & \quad + \sum_{l=1}^2 \left| [\mathbb{E}(Y| \boldsymbol{X}\in\boldsymbol{t}_{l}^{'})]^2   - [\widehat{\beta}(\boldsymbol{t}_{l}^{'})]^{2} \right| \times n^{-1} \left[\sum_{i=1}^{n}  \boldsymbol{1}_{\boldsymbol{X}_{i} \in \boldsymbol{t}_{l}^{'}}\right]                   \\
        & \le \sum_{l=1}^2 [\mathbb{E}(Y| \boldsymbol{X}\in\boldsymbol{t}_{l}^{'})]^2 \times \left| \mathbb{P}(\boldsymbol{X}\in\boldsymbol{t}_{l}^{'}) \  - n^{-1} \sum_{i=1}^{n}  \boldsymbol{1}_{\boldsymbol{X}_{i} \in \boldsymbol{t}_{l}^{'}}\right|  \\
        & \quad + \sum_{l=1}^2 \bigg\{ 
        ( |\mathbb{E}(Y| \boldsymbol{X}\in\boldsymbol{t}_{l}^{'})| + | \widehat{\beta}(\boldsymbol{t}_{l}^{'})| )  \times n^{-1} \left[\sum_{i=1}^{n}  \boldsymbol{1}_{\boldsymbol{X}_{i} \in \boldsymbol{t}_{l}^{'}}\right] \\
        & \quad \quad \times \bigg( \frac{\mathbb{E}(Y\boldsymbol{1}_{ \boldsymbol{X}\in\boldsymbol{t}_{l}^{'} }) } { \mathbb{P}(\boldsymbol{X}\in \boldsymbol{t}_{l}^{'})}  - \frac{\mathbb{E}(Y\boldsymbol{1}_{ \boldsymbol{X}\in\boldsymbol{t}_{l}^{'} }) }{\sum_{i=1}^{n} \boldsymbol{1}_{ \boldsymbol{X}_{i}\in \boldsymbol{t}_{l}^{'}}  } + \frac{\mathbb{E}(Y\boldsymbol{1}_{ \boldsymbol{X}\in\boldsymbol{t}_{l}^{'}}) }{\sum_{i=1}^{n} \boldsymbol{1}_{ \boldsymbol{X}_{i}\in \boldsymbol{t}_{l}^{'}}  }  - \frac{\sum_{i=1}^{n} Y_{i}\boldsymbol{1}_{ \boldsymbol{X}_{i}\in \boldsymbol{t}_{l}^{'}}   }{\sum_{i=1}^{n} \boldsymbol{1}_{ \boldsymbol{X}_{i}\in \boldsymbol{t}_{l}^{'}}  } \bigg) \bigg\}.   
    \end{split}
\end{equation}
and
\begin{equation}
    \begin{split}\label{theorem1.31.new.2}
        & \textnormal{RHS of \eqref{theorem1.31.new}} \\        
        & \le \sum_{l=1}^2 [\mathbb{E}(Y| \boldsymbol{X}\in\boldsymbol{t}_{l}^{'})]^2 \times \left| \mathbb{P}(\boldsymbol{X}\in\boldsymbol{t}_{l}^{'}) \  - n^{-1} \sum_{i=1}^{n}  \boldsymbol{1}_{\boldsymbol{X}_{i} \in \boldsymbol{t}_{l}^{'}}\right|  \\
        & \quad + \sum_{l=1}^2 \bigg\{ \left( |\mathbb{E}(Y| \boldsymbol{X}\in\boldsymbol{t}_{l}^{'})| + | \widehat{\beta}(\boldsymbol{t}_{l}^{'})| \right) \\
        & \quad\quad \times \bigg[ \left|  \mathbb{P}(\boldsymbol{X}\in\boldsymbol{t}_{l}^{'} ) - n^{-1} \sum_{i=1}^{n}  \boldsymbol{1}_{\boldsymbol{X}_{i} \in \boldsymbol{t}_{l}^{'}}\right|  \times |\mathbb{E}(Y | \boldsymbol{X}\in\boldsymbol{t}_{l}^{'} ) | \\
        & \quad \quad\quad+ \sum_{l=1}^2 \left| \mathbb{E}(Y \boldsymbol{1}_{ \boldsymbol{X}\in\boldsymbol{t}_{l}^{'} }) - n^{-1}\sum_{i=1}^n Y_{i}\boldsymbol{1}_{\boldsymbol{X}_{i}\in \boldsymbol{t}_{l}^{'}} \right|  \bigg] \bigg\}\\
        & \le C_{0} \times n^{-\frac{1}{2}} s \sqrt{H(1\vee\log{s})} \log{n} \times \left(M_{\epsilon} + \sup_{\vv{x} \in [0, 1]^p} |m(\vv{x})|\right)^2,
    \end{split}
\end{equation}
where the last inequality follows from $Y = m(\boldsymbol{X}) + \varepsilon$, that $|Y| \le M_{\epsilon} + \sup_{\vv{x} \in [0, 1]^p} |m(\vv{x})|$ almost surely, and that $| \widehat{\beta}(\boldsymbol{t}_{l}^{'})| \le M_{\epsilon} + \sup_{\vv{x} \in [0, 1]^p} |m(\vv{x})|$ almost surely, and an application of Lemma~\ref{lemma.3} in Section~\ref{proof.lemma3}. The derivations of \eqref{theorem1.31.new} and \eqref{theorem1.31.new.2} are presented separately to enhance readability and ensure better text rendering.

By \eqref{theorem1.3.new}--\eqref{theorem1.31.new.2}, there exists some constant $C>0$ such that 
\begin{equation}
    \begin{split}
        \textnormal{\eqref{theorem1.14}} & \le  C  \times n^{-\frac{1}{2}} s \sqrt{H(1\vee\log{s})} \log{n} \times \left(M_{\epsilon} + \sup_{\vv{x} \in [0, 1]^p} |m(\vv{x})|\right)^2 \\
        & \le \rho_{n},
    \end{split}
\end{equation}
since $\rho_{n} = C_{0}  \times n^{-\frac{1}{2}} s \sqrt{H(1\vee\log{s})} \log{n} \times \left(M_{\epsilon} + \sup_{\vv{x} \in [0, 1]^p} |m(\vv{x})|\right)^2$ for some large constant $C_{0} >0$. 

We have completed the proof of \eqref{theorem1.36}.

\subsection{Proof of Theorem~\ref{theorem2}}\label{proof.theorem2}
Let
\begin{equation*}
    \begin{split}
        \widehat{R}(\boldsymbol{X}) & = \sum_{(\boldsymbol{t}^{'}, (\vv{w}, c), \boldsymbol{t}) \in \widehat{N}_{H}^{(b)} } \boldsymbol{1}_{\boldsymbol{X} \in \boldsymbol{t}} \times \widehat{\beta}(\boldsymbol{t}),\\
        R(\boldsymbol{X}) & = \sum_{(\boldsymbol{t}^{'}, (\vv{w}, c), \boldsymbol{t}) \in \widehat{N}_{H}^{(b)}} \boldsymbol{1}_{\boldsymbol{X} \in \boldsymbol{t}} \times \mathbb{E}(Y| \boldsymbol{X}\in \boldsymbol{t}).
    \end{split}
\end{equation*}
By the law of total expectation,
\begin{equation}\label{prop2.6.b}
    \mathbb{E}\{[\widehat{R}(\boldsymbol{X}) - m(\boldsymbol{X})]^2  \} = \mathbb{E}\{\mathbb{E}\{[\widehat{R}(\boldsymbol{X}) - m(\boldsymbol{X})]^2 | \mathcal{X}_{n}\}\},
\end{equation}
which can be upper bounded by an application of Cauchy-Schwarz inequality as follows.
\begin{equation}
    \begin{split}\label{prop2.6}
        \mathbb{E}\{[\widehat{R}(\boldsymbol{X}) - m(\boldsymbol{X})]^2 | \mathcal{X}_{n}\} & \le 2 \mathbb{E}\{[\widehat{R}(\boldsymbol{X}) - R(\boldsymbol{X}) ]^2 | \mathcal{X}_{n} \} + 2 \mathbb{E} \{ [ R(\boldsymbol{X}) -  m(\boldsymbol{X})]^2 |\mathcal{X}_{n} \}.
    \end{split}
\end{equation}

The first term on the right-hand side of \eqref{prop2.6} represents the estimation variance, while the second term represents the approximation error or bias. We will establish upper bounds for each of these terms across various contexts.

In the following, we first address the case of the progressive tree with continuous features under Condition~\ref{regularity.1}(b). We begin by establishing an upper bound on the approximation error in \eqref{prop2.6}. The proof strategy involves showing that the sample splits from the progressive tree are nearly as effective as the optimal splits from $W_{p, s_0}$ (as defined prior to Theorem~\ref{theorem1}) given sufficiently large $b$ and $B$, as established in \eqref{prop2.2.b}. This step is essential for our subsequent analysis of the SID convergence rate in \eqref{prop2.4}, which is developed at the population level. Note that, unlike $\widehat{W}_{p, s_0}$, the set $W_{p, s_0}$ is independent of the training sample.

Let us define an event $\mathcal{D}$ so that on $\mathcal{D}$, 
\begin{equation}
    \label{iota.3}
    \sup_{\vv{b} \in \mathcal{J}(s_0)} \min_{\vv{u} \in \Lambda} \norm{\vv{b} - \vv{u}}_{2} \le \iota,
\end{equation}
in which $\Lambda = \{ \vv{u}: (\vv{u}, a)\in \cup_{q=1}^{B} \Lambda_{q} \}$, and $\Lambda_{q}$'s are given in Section~\ref{sec2.2.1}. Let the event $\mathcal{B}$ be defined as in Corollary~\ref{corollary1} with candidate split set $\mathcal{W} = \cup_{q=1}^{B} \Lambda_{q}$. For the case with continuous features as in (b) of Condition~\ref{regularity.1}, there exists some constant $C>0$ depending on $s^s$ such that for every $h \in \{1, \dots, H\}$ and every $(\boldsymbol{t}^{'}, (\vv{w}, c), \boldsymbol{t})\in \widehat{N}_{h}^{(b)}$, it holds for all large $n$ that  on $\mathcal{B} \cap \mathcal{D}$,
\begin{equation}
\begin{split}
     \label{prop2.2.b}
     & \mathbb{E}[\textnormal{Var}(m(\boldsymbol{X})| \boldsymbol{1}_{\boldsymbol{X} \in \boldsymbol{t}^{'}}, \boldsymbol{1}_{\vv{w}^{\top}\boldsymbol{X}  > c} ) \boldsymbol{1}_{\boldsymbol{X} \in \boldsymbol{t}^{'}}] \\
      & \le \inf_{(\vv{w}^{'}, c^{'}) \in W_{p, s_0} } \mathbb{E}[\textnormal{Var}( m(\boldsymbol{X})| \boldsymbol{1}_{\boldsymbol{X} \in \boldsymbol{t}^{'}}, \boldsymbol{1}_{\vv{w}^{'\top}\boldsymbol{X}  > c^{'}} ) \boldsymbol{1}_{\boldsymbol{X} \in \boldsymbol{t}^{'}}] + 2(\rho_{n} + C D_{\max} A_{n}),
      \end{split}
 \end{equation}
 where $A_{n} = \left(\sup_{\vv{x}\in [0, 1]^p}|m(\vv{x})|\right)^2 \times (\iota + n^{-\nu})$ and $D_{\max}$ is from (b) of Condition~\ref{regularity.1}. Recall that in Theorem~\ref{theorem2}, we have defined \( s \) as a constant. The proof of \eqref{prop2.2.b} is deferred to Section~\ref{proof.prop2.2.b}.

Next, we derive that
\begin{equation}
    \begin{split}\label{prop2.4.c}
         & \mathbb{E} \{ [ R(\boldsymbol{X}) -  m(\boldsymbol{X})]^2  \ | \ \mathcal{X}_{n} \}  \\
        & = \mathbb{E}\left\{ \left\{\sum_{(\boldsymbol{t}^{'}, (\vv{w}, c), \boldsymbol{t}) \in \widehat{N}_{H}^{(b)}} \boldsymbol{1}_{\boldsymbol{X} \in \boldsymbol{t}} \times [ \mathbb{E}(Y| \boldsymbol{X}\in \boldsymbol{t}) - m(\boldsymbol{X})] \right\}^2 \ \middle| \ \mathcal{X}_{n} \right\} \\
        & = \sum_{(\boldsymbol{t}^{'}, (\vv{w}, c), \boldsymbol{t}) \in \widehat{N}_{H}^{(b)}}  \mathbb{E}\{  \boldsymbol{1}_{\boldsymbol{X} \in \boldsymbol{t}}\times [ \mathbb{E}(Y| \boldsymbol{X}\in \boldsymbol{t}) - m(\boldsymbol{X})]^2  \ | \ \mathcal{X}_{n}\}\\
        & = \sum_{(\boldsymbol{t}^{'}, (\vv{w}, c), \boldsymbol{t}) \in \widehat{N}_{H}^{(b)}}  \mathbb{E}\{ \boldsymbol{1}_{\boldsymbol{X} \in \boldsymbol{t}}\times  [ \mathbb{E}(Y| \boldsymbol{X}\in \boldsymbol{t}) - m(\boldsymbol{X})]^2 \}\\
        & = \sum_{(\boldsymbol{t}^{'}, (\vv{w}, c), \boldsymbol{t}) \in \widehat{N}_{H}^{(b)}}  \mathbb{E}\{ \boldsymbol{1}_{\boldsymbol{X} \in \boldsymbol{t}}\times  [ \mathbb{E}( m(\boldsymbol{X})| \boldsymbol{X}\in \boldsymbol{t}) - m(\boldsymbol{X})]^2 \},
     \end{split}
\end{equation}
where the first and second equalities follow from the fact that \( \sum_{(\boldsymbol{t}^{'}, (\vv{w}, c), \boldsymbol{t}) \in \widehat{N}_{H}^{(b)}} \boldsymbol{1}_{\boldsymbol{X} \in \boldsymbol{t}} = 1 \) almost surely. The third equality holds because the sample $\mathcal{X}_n$ and its population counterpart $(\boldsymbol{X}, Y)$ are independent. In the fourth equality, we rely on that $\mathbb{E}(Y|\boldsymbol{X}\in \boldsymbol{t}) = \mathbb{E}(m(\boldsymbol{X}) |\boldsymbol{X}\in \boldsymbol{t})$.

By \eqref{prop2.2.b}--\eqref{prop2.4.c}, it follows that on \(\mathcal{B} \cap \mathcal{D}\),
\begin{equation}
\begin{split}\label{prop2.4}
    & \mathbb{E} \{ [ R(\boldsymbol{X}) -  m(\boldsymbol{X})]^2  |\mathcal{X}_{n} \} \\
    & = \sum_{(\boldsymbol{t}^{'}, (\vv{w}, c), \boldsymbol{t})\in \widehat{N}_{H}^{(b)}} \mathbb{E}[\textnormal{Var} (m(\boldsymbol{X})|\boldsymbol{1}_{ \boldsymbol{X} \in \boldsymbol{t}}) \boldsymbol{1}_{ \boldsymbol{X} \in \boldsymbol{t}}] \\
    & = \frac{1}{2}\times \sum_{ (\boldsymbol{t}^{'}, (\vv{w}, c), \boldsymbol{t})\in \widehat{N}_{H}^{(b)} } \mathbb{E}[\textnormal{Var} (m(\boldsymbol{X})|\boldsymbol{1}_{ \boldsymbol{X} \in \boldsymbol{t}^{'}}, \boldsymbol{1}_{ \vv{w}^{\top}\boldsymbol{X} >c } ) \boldsymbol{1}_{ \boldsymbol{X} \in \boldsymbol{t}^{'}}] \\
    & \le (\rho_{n} + CD_{\max} A_{n}) 2^{H} \\
    & \quad + \frac{1}{2}\times \sum_{ (\boldsymbol{t}^{'}, (\vv{w}, c), \boldsymbol{t})\in \widehat{N}_{H}^{(b)} } \inf_{(\vv{w}^{'}, c^{'}) \in W_{p, s_0}}\mathbb{E}[\textnormal{Var} (m(\boldsymbol{X})|\boldsymbol{1}_{ \boldsymbol{X} \in \boldsymbol{t}^{'}}, \boldsymbol{1}_{ \vv{w}^{'\top}\boldsymbol{X} >c^{'} } ) \boldsymbol{1}_{ \boldsymbol{X} \in \boldsymbol{t}^{'}}] \\
    &\le (\rho_{n} + CD_{\max} A_{n}) 2^{H} + \frac{\alpha_{0}}{2} \times  \sum_{ (\boldsymbol{t}^{'}, (\vv{w}, c), \boldsymbol{t})\in \widehat{N}_{H}^{(b)} } \mathbb{E}[\textnormal{Var} (m(\boldsymbol{X})|\boldsymbol{1}_{ \boldsymbol{X} \in \boldsymbol{t}^{'}}) \boldsymbol{1}_{ \boldsymbol{X} \in \boldsymbol{t}^{'}}]\\
    &= (\rho_{n} + CD_{\max} A_{n}) 2^{H} + \alpha_{0} \times  \sum_{ (\boldsymbol{t}^{'}, (\vv{w}, c), \boldsymbol{t})\in \widehat{N}_{H - 1}^{(b)} } \mathbb{E}[\textnormal{Var} (m(\boldsymbol{X})|\boldsymbol{1}_{ \boldsymbol{X} \in \boldsymbol{t}}) \boldsymbol{1}_{ \boldsymbol{X} \in \boldsymbol{t}}]\\
    & \le \cdots\\
    & \le \alpha_{0}^{H} \times \textnormal{Var}(m(\boldsymbol{X})) + \sum_{h=1}^H(\rho_{n} + CD_{\max} A_{n}) 2^{h} ,
    \end{split}
\end{equation}
where the second equality holds because $\widehat{N}_{h}^{(b)}$ consists of $2^{h-1}$ splits and $2^h$ elements, the first inequality follows from \eqref{prop2.2.b}, the second inequality is due to Condition~\ref{sid}. These arguments are recursively applied to obtain the last two inequalities. On the other hand, on $\mathcal{B}^c \cup \mathcal{D}^c$,
\begin{equation}
\begin{split}\label{prop2.4.b}
    & \mathbb{E} \{ [ R(\boldsymbol{X}) -  m(\boldsymbol{X})]^2  |\mathcal{X}_{n} \} \le 4\left(\sup_{\vv{x} \in [0, 1]^p} |m(\vv{x})|\right)^2,
    \end{split}
    \end{equation}
since $\mathbb{E}(Y| \boldsymbol{X} \in \boldsymbol{t}) = \mathbb{E}(m(\boldsymbol{X})| \boldsymbol{X} \in \boldsymbol{t})$.

To establish the desired error bound, the probabilities of the events $\mathcal{B}^c$ and $\mathcal{D}^c$ must vanish as both $n$ and $B$ grow sufficiently large. While the probability $\mathbb{P}(\mathcal{B}^c)$ was previously analyzed in Corollary~\ref{corollary1}, we now focus on providing a corresponding upper bound for $\mathbb{P}(\mathcal{D}^c)$. When $B \ge  s 2^{2s_0}\binom{p}{s_0} s_0^{\frac{s_0}{2}} \iota^{-s_0} \times \log{ (n 2^{2s_0}  \binom{p}{s_0}  s_0^{\frac{s_0}{2}} \iota^{-s_0})}$, it holds that for large $n$,
\begin{equation}
    \begin{split}\label{prob.upper.1}    
        \mathbb{P}(\mathcal{D}^{c}) & \le \frac{1}{n}.
    \end{split}
\end{equation}
The proof of \eqref{prob.upper.1} is deferred to Section~\ref{proof.prob.upper.1}.

By \eqref{prop2.4.c}--\eqref{prob.upper.1} and Corollary~\ref{corollary1} with $\texttt{\#}S \times b \ge n2^{H} B\log{(2^H B)}$ and continuous features, it holds that for all large $n$,
\begin{equation}
\begin{split}\label{prop2.10}
    & \mathbb{E} \{ [ R(\boldsymbol{X}) -  m(\boldsymbol{X})]^2  |\mathcal{X}_{n} \} \\
    & \le \left(\sup_{\vv{x} \in [0, 1]^p} |m(\vv{x})|\right)^2 \times (8n^{-1} + 4B^{-1}) + \alpha_{0}^{H} \times \left(\sup_{\vv{x} \in [0, 1]^p} |m(\vv{x})|\right)^2 \\
    &\quad + (\rho_{n} + CD_{\max} A_{n}) 2^{H+1}.
\end{split}
\end{equation}

In the case of the progressive tree with continuous features under Condition~\ref{regularity.1}(b), we have addressed the approximation error and now move on to analyze the estimation variance as established in \eqref{prop2.6}. Recall that $\rho_{n} = C_{0}  \times n^{-\frac{1}{2}} s \sqrt{H(1\vee\log{s})} \log{n} \times \left(M_{\epsilon} + \sup_{\vv{x} \in [0, 1]^p} |m(\vv{x})|\right)^2$ for some sufficiently large $C_{0} >0$. By Lemma~\ref{lemma.3} in Section~\ref{proof.lemma3}, there is some constant $C_{2}>0$ such that for all large $n$, each $1\le p \le n^{K_{0}}$, each $s\ge 1$, each $H\ge 1$ with $\sqrt{2}n^{-\frac{1}{2}} s \sqrt{H(1\vee\log{s})} \log{n}\le 1$, it holds that 
% \begin{equation}
%     \begin{split}\label{prop2.5}
%          & \mathbb{E}\{[\widehat{R}(\boldsymbol{X}) - R(\boldsymbol{X}) ]^2 | \mathcal{X}_{n} \} \\
%          & \le \sum_{(\boldsymbol{t}^{'}, (\vv{w}, c), \boldsymbol{t}) \in \widehat{N}_{H}^{(b)}}  \mathbb{P}(\boldsymbol{X}\in \boldsymbol{t})\times  |\widehat{\beta}(\boldsymbol{t}) - \mathbb{E}(Y| \boldsymbol{X}\in \boldsymbol{t})|^2\\
%          & \le \sum_{(\boldsymbol{t}^{'}, (\vv{w}, c), \boldsymbol{t}) \in \widehat{N}_{H}^{(b)}} \mathbb{P} (\boldsymbol{X} \in \boldsymbol{t}) \bigg|   \frac{\mathbb{E}(Y\boldsymbol{1}_{ \boldsymbol{X}\in\boldsymbol{t} } )}{ \mathbb{P} (\boldsymbol{X} \in \boldsymbol{t}) } - \frac{ n^{-1} \sum_{i=1}^{n}  Y_{i} \boldsymbol{1}_{\boldsymbol{X}_{i} \in \boldsymbol{t}}}{ \mathbb{P} (\boldsymbol{X} \in \boldsymbol{t}) } \\
%          & \quad +  \frac{ n^{-1} \sum_{i=1}^{n}  Y_{i}\boldsymbol{1}_{\boldsymbol{X}_{i} \in \boldsymbol{t}}}{ \mathbb{P} (\boldsymbol{X} \in \boldsymbol{t}) } -  \frac{ n^{-1} \sum_{i=1}^{n}  Y_{i}\boldsymbol{1}_{\boldsymbol{X}_{i} \in \boldsymbol{t}} }{n^{-1} \sum_{i=1}^{n}  \boldsymbol{1}_{\boldsymbol{X}_{i} \in \boldsymbol{t}} }  \bigg|^2          
%     \end{split}
% \end{equation}
\begin{equation}
    \begin{split}\label{prop2.5}
         & \mathbb{E}\{[\widehat{R}(\boldsymbol{X}) - R(\boldsymbol{X}) ]^2 | \mathcal{X}_{n} \} \\
         & \le \sum_{(\boldsymbol{t}^{'}, (\vv{w}, c), \boldsymbol{t}) \in \widehat{N}_{H}^{(b)}}  \mathbb{P}(\boldsymbol{X}\in \boldsymbol{t})\times  |\widehat{\beta}(\boldsymbol{t}) - \mathbb{E}(Y| \boldsymbol{X}\in \boldsymbol{t})|^2\\
         & \le \sum_{(\boldsymbol{t}^{'}, (\vv{w}, c), \boldsymbol{t}) \in \widehat{N}_{H}^{(b)}} \mathbb{P} (\boldsymbol{X} \in \boldsymbol{t}) \bigg|   \frac{\mathbb{E}(Y\boldsymbol{1}_{ \boldsymbol{X}\in\boldsymbol{t} } )}{ \mathbb{P} (\boldsymbol{X} \in \boldsymbol{t}) } - \frac{ n^{-1} \sum_{i=1}^{n}  Y_{i} \boldsymbol{1}_{\boldsymbol{X}_{i} \in \boldsymbol{t}}}{ \mathbb{P} (\boldsymbol{X} \in \boldsymbol{t}) } \\
         & \quad +  \frac{ n^{-1} \sum_{i=1}^{n}  Y_{i}\boldsymbol{1}_{\boldsymbol{X}_{i} \in \boldsymbol{t}}}{ \mathbb{P} (\boldsymbol{X} \in \boldsymbol{t}) } -  \frac{ n^{-1} \sum_{i=1}^{n}  Y_{i}\boldsymbol{1}_{\boldsymbol{X}_{i} \in \boldsymbol{t}} }{n^{-1} \sum_{i=1}^{n}  \boldsymbol{1}_{\boldsymbol{X}_{i} \in \boldsymbol{t}} }  \bigg| \times 2\left(M_{\epsilon} + \sup_{\vv{x} \in [0, 1]^p} |m(\vv{x})|\right) \\
         & \le 2\left(M_{\epsilon} + \sup_{\vv{x} \in [0, 1]^p} |m(\vv{x})|\right) \sum_{(\boldsymbol{t}^{'}, (\vv{w}, c), \boldsymbol{t}) \in \widehat{N}_{H}^{(b)}} |   \mathbb{E}(Y\boldsymbol{1}_{ \boldsymbol{X}\in\boldsymbol{t} } ) - n^{-1} \sum_{i=1}^{n}  Y_{i} \boldsymbol{1}_{\boldsymbol{X}_{i} \in \boldsymbol{t}} | \\
         & \quad +  2\left(M_{\epsilon} + \sup_{\vv{x} \in [0, 1]^p} |m(\vv{x})|\right)^2 \sum_{(\boldsymbol{t}^{'}, (\vv{w}, c), \boldsymbol{t}) \in \widehat{N}_{H}^{(b)}}  | \mathbb{P} (\boldsymbol{X} \in \boldsymbol{t})  -  n^{-1} \sum_{i=1}^{n}  \boldsymbol{1}_{\boldsymbol{X}_{i} \in \boldsymbol{t}} |\\
         & \le C_{2}  \times 2^H \rho_{n}.
    \end{split}
\end{equation}
The first inequality follows the same reasoning as that used for \eqref{prop2.4.c}. The second inequality follows from the identity $(a+b)^2 = |a - b| |a+b|$ and by bounding the sum of the terms by using $|Y| \le M_{\epsilon} + \sup_{\vv{x} \in [0, 1]^p} |m(\vv{x})|$. The third inequality is obtained by 
\begin{equation*}
    \begin{split}
        & \mathbb{P}(\boldsymbol{X} \in \boldsymbol{t}) \times \left|\frac{ n^{-1} \sum_{i=1}^{n}  Y_{i}\boldsymbol{1}_{\boldsymbol{X}_{i} \in \boldsymbol{t}}}{ \mathbb{P} (\boldsymbol{X} \in \boldsymbol{t}) } -  \frac{ n^{-1} \sum_{i=1}^{n}  Y_{i}\boldsymbol{1}_{\boldsymbol{X}_{i} \in \boldsymbol{t}} }{n^{-1} \sum_{i=1}^{n}  \boldsymbol{1}_{\boldsymbol{X}_{i} \in \boldsymbol{t}} } \right|\\
        & \le \mathbb{P}(\boldsymbol{X} \in \boldsymbol{t})\times \left|\frac{ n^{-1} \sum_{i=1}^{n}  Y_{i}\boldsymbol{1}_{\boldsymbol{X}_{i} \in \boldsymbol{t}}} { n^{-1} \sum_{i=1}^{n}  \boldsymbol{1}_{\boldsymbol{X}_{i} \in \boldsymbol{t}} } \times \frac{n^{-1} \sum_{i=1}^{n}  \boldsymbol{1}_{\boldsymbol{X}_{i} \in \boldsymbol{t}}}{ \mathbb{P}(\boldsymbol{X} \in \boldsymbol{t}) } -  \frac{ n^{-1} \sum_{i=1}^{n}  Y_{i}\boldsymbol{1}_{\boldsymbol{X}_{i} \in \boldsymbol{t}} }{n^{-1} \sum_{i=1}^{n}  \boldsymbol{1}_{\boldsymbol{X}_{i} \in \boldsymbol{t}} }  \right|,
    \end{split}
\end{equation*}
and that 
$\widehat{\beta}(\boldsymbol{t}) = \frac{ n^{-1} \sum_{i=1}^{n}  Y_{i}\boldsymbol{1}_{\boldsymbol{X}_{i} \in \boldsymbol{t}} }{n^{-1} \sum_{i=1}^{n}  \boldsymbol{1}_{\boldsymbol{X}_{i} \in \boldsymbol{t}} }$, whose absolute value is bounded by $M_{\epsilon} + \sup_{\vv{x} \in [0, 1]^p} |m(\vv{x})|$.  The fourth inequality results from an application of Lemma~\ref{lemma.3}. Note that each leaf node in \(\widehat{N}_{H}^{(b)}\) is derived from at most \(2^{H-1}\) splits, ensuring the applicability of Lemma~\ref{lemma.3}. Furthermore, \eqref{prop2.5} applies to both continuous and discrete feature cases.

By \eqref{prop2.6.b}-\eqref{prop2.6}, \eqref{prop2.10}--\eqref{prop2.5}, and the definition of $\rho_{n}$ given by Theorem~\ref{theorem1}, there is some constant $C>0$ such that for all large $n$, each $1\le p \le n^{K_{0}}$, each $H\ge 1$ with
$$\sqrt{2}n^{-\frac{1}{2}} s \sqrt{H(1\vee\log{s})} \log{n}\le 1,$$ it holds that 
\begin{equation}
    \begin{split}\label{continuous.1}
        & \mathbb{E}\{[\widehat{R}(\boldsymbol{X}) - m(\boldsymbol{X})]^2  \} \\
        & \le \left(\sup_{\vv{x} \in [0, 1]^p} |m(\vv{x})|\right)^2 \times (8n^{-1} + 4B^{-1} + \alpha_{0}^{H} )  + (\rho_{n} + D_{\max}A_{n}) 2^{H} \times C,
    \end{split}
\end{equation}
which completes the desired proof of Theorem~\ref{theorem2} for the case with continuous features and progressive tree. 

On the other hand, to prove \eqref{ideal.tree.consistency}, we first observe that the concept of a finite candidate set size $B$ is not applicable to an ideal tree, as its splits are determined over $\widehat{W}_{p, s}$. However, according to Proposition~\ref{prop3}, the ideal tree can be analyzed through the one-shot tree by assuming a sufficiently large $B$. This approach is valid because the two trees are equivalent with arbitrarily high probability as $B$ increases, as discussed in the remark preceding Proposition~\ref{prop3}. Specifically, there exists an event $E_{B}$, defined in Section~\ref{proof.prop3}, such that $\lim_{B\rightarrow\infty}\mathbb{P}(E_{B} \mid \mathcal{X}_n) = 1$. On this event, the one-shot tree and the corresponding ideal tree are equivalent. Since we are not physically constructing the ideal tree, we can assume $B$ is large enough to ensure equivalence without considering computational constraints.

Additionally, for fixed $s_0$ and $p$, the term $\iota$ in \eqref{iota.3} can be made arbitrarily small by choosing a sufficiently large $B$. By setting $\iota \ll n^{-1}$, $\mathbb{P}(E_{B}^c \mid \mathcal{X}_n) \ll n^{-1}$, and $B \gg n$, and applying a constant $\nu > \frac{1}{2}$ in $A_n$, direct calculation shows that \eqref{ideal.tree.consistency} holds for all large $n$ under continuous features. The factor of $2$ in \eqref{ideal.tree.consistency} is used to bound minor differences, such as $4B^{-1} + 8n^{-1} \le 2\times 8n^{-1}$.

 Now, we proceed to prove the case with discrete features with $B\ge C_{3}s(s_0\log{n})^2 \binom{p}{s_0}$
for some $C_{3}>0$ depending on $s$, there is an event $\mathcal{G}$ such that for all large $n$, every $h \in \{1, \dots, H\}$, and every $(\boldsymbol{t}^{'}, (\vv{w}, c), \boldsymbol{t})\in \widehat{N}_{h}^{(b)}$, on $\mathcal{B} \cap \mathcal{G}$,
\begin{equation}
\begin{split}
     \label{discrete.1}
      & \mathbb{E}[\textnormal{Var}(m(\boldsymbol{X})| \boldsymbol{1}_{\boldsymbol{X} \in \boldsymbol{t}^{'}}, \boldsymbol{1}_{\vv{w}^{\top}\boldsymbol{X}  > c} ) \boldsymbol{1}_{\boldsymbol{X} \in \boldsymbol{t}^{'}}] \\
      & \quad\le \inf_{(\vv{w}^{'}, c^{'}) \in W_{p, s_0} } \mathbb{E}[\textnormal{Var}( m(\boldsymbol{X})| \boldsymbol{1}_{\boldsymbol{X} \in \boldsymbol{t}^{'}}, \boldsymbol{1}_{\vv{w}^{'\top}\boldsymbol{X}  > c^{'}} ) \boldsymbol{1}_{\boldsymbol{X} \in \boldsymbol{t}^{'}}] + 2\rho_{n},      
      \end{split}
 \end{equation}
with $\mathbb{P}(\mathcal{G}^c) \le n^{-1}$. The proof of \eqref{discrete.1} is deferred to Section~\ref{proof.discrete.1}.

\sloppy By arguments similar to those for \eqref{prop2.10}, but using \eqref{discrete.1} instead of \eqref{prop2.2.b} and applying Corollary~\ref{corollary1} with \( \texttt{\#}S \times b \geq D_{\min}^{-1} 2^{H + s + 1} B \log{(2^H B)} \) and discrete features, it follows that for all sufficiently large \( n \), each \( s \geq s_0 \), and each \( B \geq C_{3} s (s_0 \log{n})^2 \binom{p}{s_0} \) for some \( C_{3} > 0 \),
\begin{equation}
\begin{split}\label{prop2.10.b}
    & \mathbb{E} \{ [ R(\boldsymbol{X}) -  m(\boldsymbol{X})]^2  |\mathcal{X}_{n} \} \le \left(\sup_{\vv{x} \in [0, 1]^p} |m(\vv{x})|\right)^2 \times (8n^{-1} + 4B^{-1} + \alpha_{0}^{H}) + \rho_{n} 2^{H+1}.
\end{split}
\end{equation}

By \eqref{prop2.10.b} and arguments similar for \eqref{continuous.1}, there is some constant $C>0$ such that for all large $n$, each $1\le p \le n^{K_{0}}$, each $H\ge 1$, each (constant) $s\ge s_0$ with $\sqrt{2}n^{-\frac{1}{2}} s \sqrt{H(1\vee\log{s})} \log{n}\le 1$, each $B\ge (\log{n})^3 \binom{p}{s_0}$, and each $\texttt{\#}S \times b \ge  D_{\min}^{-1}2^{H + s + 1} B\log{(2^H B)}$, it holds that 
\begin{equation}
    \begin{split}
        & \mathbb{E}\{[\widehat{R}(\boldsymbol{X}) - m(\boldsymbol{X})]^2  \}  \le \left(\sup_{\vv{x} \in [0, 1]^p} |m(\vv{x})|\right)^2 \times (8n^{-1} + 4B^{-1} + \alpha_{0}^{H} )  + \rho_{n}  2^{H} \times C,
    \end{split}
\end{equation}
which completes the desired proof of Theorem~\ref{theorem2} for the case with discrete features and progressive tree. Note that we use the assumption that $s$ and $s_0$ are constants with $s_0 \le s$ to bound the term $C_3 \times s \times s_0^2$ by $\log n$ for all sufficiently large $n$ when deriving a compact lower bound for $B$.
 
Lastly, to prove \eqref{ideal.tree.consistency} under discrete features, we apply the same reasoning used for proving \eqref{ideal.tree.consistency} under the continuous case to establish the desired result. This concludes the proof of Theorem~\ref{theorem2}. 

%%%
%%%
%%%

\subsubsection{Proof of \texorpdfstring{\eqref{prop2.2.b}}{LG} } \label{proof.prop2.2.b}

     Let an arbitrary $h \in \{1, \dots, H\}$, an arbitrary  $(\boldsymbol{t}^{'}, (\vv{w}, c), \boldsymbol{t})\in \widehat{N}_{h}^{(b)}$ be given. In light of the definition of $\mathcal{B}$ (see Corollary~\ref{corollary1}), it holds that on $\mathcal{B}$, $(\vv{w}, c) \in  E(\rho_{n}, \boldsymbol{t}^{'}, \cup_{q=1}^{B} \Lambda_{q})$. Our goal is to show \eqref{prop2.2.b} for $(\vv{w}, c)$ on $\mathcal{B}\cap \mathcal{D}$.

The core challenge in this proof is that the candidate splits $\bigcup_{q=1}^{B}\Lambda_{q}$ are random and discrete, whereas the full set $W_{p, s_0}$ is continuous. To bridge this gap, we first construct a finite set of weight vectors, $\mathcal{P}$, that is sufficiently dense in the space of all possible $s_0$-sparse splits (see \eqref{vvu.con.1}). Next, for each weight vector $\vv{v} \in \mathcal{P}$, we show that the population-level threshold can be approximated by a sample counterpart $\vv{v}^{\top}\boldsymbol{X}_i$ for some $i\in \{1, \dots, n\}$ (see \eqref{exponential.prob.1.e}) on the event $\mathcal{I}$ defined in \eqref{I.def}. This construction ensures that on event $\mathcal{I}$, the sample-best split is nearly as effective as the population-best split in terms of variance reduction.

 In the following, we construct the set $\mathcal{P}$ by first covering $\mathcal{J}(s_0)$ with hypercubes of edge length $\eta_n$, where $\eta_{n} = (\lfloor n^{\nu} \rfloor)^{-1}$ for some $\nu \in (0, 1)$. Specifically, for each coordinate subset \( J \subseteq \{1, \dots, p\} \) with \( \texttt{\#}J = s_0 \), we construct a cover of  
\[
Q(J) = \{\vv{v} \in [-1, 1]^p : \max_{j \notin J} |v_j| = 0\}
\]  
 using hypercubes, each within \( Q(J) \), with edge length \( \eta_n \) in the \( s_0 \) active coordinates and collapsing to a point in the remaining \( p - s_0 \) dimensions. Next, from this collection of hypercubes, we retain only those that intersect the sphere  
\[
\{\vv{v} \in [-1, 1]^p : \max_{j \notin J} |v_j| = 0, \|\vv{v}\|_2 = 1\},
\]  
and denote the retained set of hypercubes by \(T(J)\). We repeat this process for each \(J \subseteq \{1, \dots, p\}\) with \(\texttt{\#}J = s_0\) and note that \(\mathcal{J}(s_0) \subseteq \bigcup_{J\subseteq \{1, \dots, p\}, \texttt{\#}J = s_0} \bigcup_{I \in T(J)} I\). This implies that \(\mathcal{J}(s_0)\) is covered by \(\binom{p}{s_0}\) distinct sets of hypercubes \(T(J)\), where each \(T(J)\) contains at most \(2^{s_0} \eta_n^{-s_0}\) hypercubes.

Next, we consider a unit vector within each hypercube in \(\{I : I \in T(J), \texttt{\#}J = s_0\}\). In \eqref{p.def} below, the superscript \(q\) indicates the index of each coordinate subset \(J\) with \(\texttt{\#}J = s_0\), and the subscript denotes the index of the hypercube within each \(T(J)\). Let the set of these unit vectors be denoted by \(\mathcal{P}\), defined as follows:
 \begin{equation}
     \label{p.def}
     \mathcal{P}\coloneqq \{\vv{v}_{1}^{(q)}, \dots, \vv{v}_{L_{q}}^{(q)} : q\in \{1\ ,\dots, \binom{p}{s_0}\}\}
 \end{equation} 
 such that $\vv{v}_{l}^{(q)} \in \mathcal{J}(s_0)$ and that
\begin{equation}
    \label{exponential.prob.11}
    \begin{split}        
     \sup_{\vv{u}\in \mathcal{J}(s_0)} \min_{1\le l \le L_{q}, q\in \{1\ ,\dots, \binom{p}{s_0}\}}  \norm{\vv{u} - \vv{v}_{l}^{(q)}}_{2} & \le \sqrt{s_0} \times \eta_{n}, \\
     \max_{q\in \{1\ ,\dots, \binom{p}{s_0}\}}L_{q} & \le 2^{s_0} \eta_{n}^{-s_0}. %(\eta_{n}^{-1} + 1)^s.
    \end{split}
\end{equation} 

While the set $\mathcal{P}$ covers the possible directions of the weight vectors in the sense of \eqref{exponential.prob.11}, we must also account for the split threshold (the bias). Since our candidate splits are restricted to sample points, we need to ensure the training data is distributed well enough that for any direction $\vv{v}$ in our cover, there exists a sample point $\boldsymbol{X}_i$ such that $\vv{v}^{\top}\boldsymbol{X}_i$ provides a near-optimal threshold.

Since $\mathbb{P}(\boldsymbol{X} \in [0, 1]^p) = 1$, it follows that $\mathbb{P}(-\sqrt{s_0} \le \vv{r}^{\top}\boldsymbol{X} \le \sqrt{s_0}) = 1$ for any unit vector $\vv{r}$ with $s_0$ active coordinates. To ensure that our sample-based splits can approximate any population split, we partition the range of the projected data into $N_n = \lfloor n^{\nu} \rfloor$ intervals. 
For each direction $\vv{v}_{l}^{(q)}$ in the covering set, we construct intervals $I_{r, l}^{(q)} = [g_{r, l}^{(q)}, g_{r+1, l}^{(q)}]$ for $r\in \{0, \dots, N_{n}-2\}$ with $g_{0, l}^{(q)} = -\sqrt{s_0}$ and $I_{N_{n}  -1, l}^{(q)} = [g_{N_{n} -1} , \sqrt{s_0}]$ such that 
\begin{equation}
    \begin{split}\label{exponential.prob.1.b}
        \mathbb{P} ( \vv{v}_{l}^{(q)\top}\boldsymbol{X}\in I_{r, l}^{(q)}) & = \eta_{n} \textnormal{ for } r \in \{0, \dots, N_{n}-1\}.
    \end{split}
\end{equation}
This construction is feasible under Condition~\ref{regularity.1}(b).

Then, by \eqref{exponential.prob.1.b}, for any fixed interval $r\in \{0, \dots, N_{n}-1\}$, the probability of no  sample falling in $I_{r, l}^{(q)}$ is upper bounded by 
\begin{equation}
\begin{split}\label{exponential.prob.1}
        \mathbb{P}( \{i: \vv{v}_{l}^{(q)\top}\boldsymbol{X}_{i} \in I_{r, l}^{(q)} \} = \emptyset) & \le  [1 - \mathbb{P}(\vv{v}_{l}^{(q)\top}\boldsymbol{X} \in I_{r, l}^{(q)})]^{n} \\
        & \le (1 - \eta_{n})^{n}\\        
        & \le  \exp{(-  n \eta_{n}    )},
    \end{split}
\end{equation}
where the first inequality follows from the assumption that $\boldsymbol{X}_{i}$'s are i.i.d., and the third inequality follows from that $1+x\le e^{x}$ for $x\in \mathbb{R}$ (setting $x = -\eta_n$).

Based on the construction of the intervals $I_{r, l}^{(q)}$, we observe that if every interval contains at least one sample point, then any arbitrary threshold $a \in \mathbb{R}$ can be closely approximated by some sampled split point. Formally, on the event $\cap_{r=0}^{N_{n}-1} \{\{i: \vv{v}_{l}^{(q)\top}\boldsymbol{X}_{i} \in I_{r, l}^{(q)} \} \not= \emptyset \}$, the nearest sample neighbor $k = \argmin_{1\le i\le n} |\vv{v}_{l}^{(q)\top}\boldsymbol{X}_{i} - a|$ satisfies:
\begin{equation}
\label{exponential.prob.1.e}
\mathbb{P}(\min\{\vv{v}_{l}^{(q)\top}\boldsymbol{X}_{k} , a\}\le \vv{v}_{l}^{(q)\top}\boldsymbol{X} \le \max\{\vv{v}_{l}^{(q)\top}\boldsymbol{X}_{k} , a\} \mid \mathcal{X}_{n}) \le 2\eta_{n}.
\end{equation}
The result of \eqref{exponential.prob.1.e} holds because, given the sample $\mathcal{X}_n$, the region between the target threshold $a$ and its nearest neighbor $\vv{v}_{l}^{(q)\top}\boldsymbol{X}_{k}$ is necessarily contained within at most two adjacent intervals. Specifically, for some $r \in \{0, \dots, N_n - 2\}$, we have:
$$\{\min\{\vv{v}_{l}^{(q)\top}\boldsymbol{X}_{k} , a\}\le \vv{v}_{l}^{(q)\top}\boldsymbol{X} \le \max\{\vv{v}_{l}^{(q)\top}\boldsymbol{X}_{k} , a\}\} \subseteq \{ \vv{v}_{l}^{(q)\top}\boldsymbol{X} \in I_{r, l}^{(q)} \cup I_{r+1, l}^{(q)}\}.$$
Since each interval carries a probability mass of $\eta_n$ by \eqref{exponential.prob.1.b}, the total probability mass of the difference is bounded by $2\eta_n$.

To ensure this approximation holds globally across all directions and thresholds, we define the event $\mathcal{I}$ as the intersection of these conditions over the entire cover:
\begin{equation}
    \label{I.def}
    \mathcal{I} = \bigcap_{q\in \{1\ ,\dots, \binom{p}{s_0}\}} \bigcap_{ l\in \{1, \dots, L_{q}\}} \bigcap_{k=0}^{N_{n} -1} \{\{i: \vv{v}_{l}^{(q)\top}\boldsymbol{X}_{i} \in I_{k, l}^{(q)} \} \not= \emptyset\}.
\end{equation}

By \eqref{exponential.prob.11}, \eqref{exponential.prob.1}, the definition that $N_{n} = \lfloor n^{\nu} \rfloor= \eta_{n}^{-1}$ for some $\nu < 1$, and the assumption that $1\le p \le n^{K_{0}}$ and that $s_0$ is a constant, it holds for all large $n$ that
\begin{equation}
    \begin{split}
        \label{exponential.prob.1.c}
\mathbb{P}(\mathcal{I}^c) \le \binom{p}{s_0}\times  2^{s_0} \eta_{n}^{-s_0}\times n^{\nu}\times \exp{(-  n \eta_{n}  )} \le \eta_{n}.
     \end{split}
\end{equation}

Consequently, on the high-probability event $\mathcal{I}$, any weight vector in $\mathcal{J}(s_0)$ and  target bias in $\mathbb{R}$ can be accurately approximated by $\vv{u}^{\dagger\top}\boldsymbol{X}_i$ for some $\vv{u}^{\dagger} \in \mathcal{P}$ and $i \in \{1, \dots, n\}$. This result provides the foundation for the variance bound established in \eqref{exponential.prob.23}.

% Recall that (at the beginning of this proof) we are considering an arbitrary $h \in \{1, \dots, H\}$ and an arbitrary  $(\boldsymbol{t}^{'}, (\vv{w}, c), \boldsymbol{t})\in \widehat{N}_{h}^{(b)}$  with $(\vv{w}, c) \in  E(\rho_{n}, \boldsymbol{t}^{'}, \cup_{q=1}^{B} \Lambda_{q})$ on $\mathcal{B}$. Now, let  one of the optimal split given the full available split be denoted by 
% \begin{equation}
%     \label{wstar.1}
%     (\vv{w}^{\star}, c^{\star}) \in E(0, \boldsymbol{t}^{'}, W_{p, s_0}).
% \end{equation}

% Additionally, let $\vv{u}^{\dagger} \in \mathcal{P}$, in which $\mathcal{P}$ is defined in \eqref{p.def}, be given such that
% \begin{equation}
%     \label{vvu.con.1}
%     \norm{\vv{u}^{\dagger} - \vv{w}^{\star}}_{2}\le \sqrt{s_0} \times \eta_{n}
% \end{equation}
% and $\{j : |u_{j}^{\dagger}| > 0\}  = \{j : |w_{j}^{\star}| > 0\}$, which can be done due to \eqref{exponential.prob.11} and that $(\vv{w}^{\star}, c^{\star}) \in W_{p, s_0}$. 

To set the stage for our primary calculations, we first define the population optimal split pair $(\vv{w}^{\star}, c^{\star})$ against the sample split $(\vv{w}, c)$. Recall that we are considering an arbitrary tree depth $h \in \{1, \dots, H\}$ and a sample split $(\boldsymbol{t}^{'}, (\vv{w}, c), \boldsymbol{t}) \in \widehat{N}_{h}^{(b)}$ that, on the event $\mathcal{B}$, satisfies the empirical goodness condition  $(\vv{w}, c) \in E(\rho_{n}, \boldsymbol{t}^{'}, \cup_{q=1}^{B} \Lambda_{q})$. We now define a corresponding optimal population split $(\vv{w}^{\star}, c^{\star})$ from the set of all possible $s_0$-sparse splits:
\begin{equation}\label{wstar.1}
(\vv{w}^{\star}, c^{\star}) \in E(0, \boldsymbol{t}^{'}, W_{p, s_0}).
\end{equation}

By utilizing the cover $\mathcal{P}$ defined in \eqref{p.def}, we can find a weight vector $\vv{u}^{\dagger} \in \mathcal{P}$ that closely approximates this optimal weight vector. Specifically, we choose $\vv{u}^{\dagger}$ such that it shares the same active coordinate support as $\vv{w}^{\star}$ and satisfies the distance bound:
\begin{equation}
\label{vvu.con.1}
\norm{\vv{u}^{\dagger} - \vv{w}^{\star}}_{2} \le \sqrt{s_0} \times \eta_{n}.
\end{equation}
The existence of this $\vv{u}^{\dagger}$ is guaranteed by the density of our construction in \eqref{exponential.prob.11}. This approximation is a prerequisite for the main variance reduction calculations that follow.

     We have for all large $n$ that on $\mathcal{B}\cap \mathcal{D}$,
\begin{equation}
    \begin{split}\label{exponential.prob.23}
        & \mathbb{E}[\textnormal{Var}(m(\boldsymbol{X})| \boldsymbol{1}_{\boldsymbol{X} \in \boldsymbol{t}^{'}}, \boldsymbol{1}_{\vv{w}^{\top}\boldsymbol{X} > c}) \boldsymbol{1}_{\boldsymbol{X} \in \boldsymbol{t}^{'}}] \\        
        & \le  \mathbb{E}[\textnormal{Var}(m(\boldsymbol{X})| \boldsymbol{1}_{\boldsymbol{X} \in \boldsymbol{t}^{'}}, \boldsymbol{1}_{\vv{w}^{\star\top}\boldsymbol{X} > c^{\star}}) \boldsymbol{1}_{\boldsymbol{X} \in \boldsymbol{t}^{'}}] + 2 Q_{n} + 2\rho_{n},
    \end{split}
\end{equation}
where $Q_{n}$ depends on several technical parameters such that
\begin{equation}
    \label{prop2.3}
    Q_{n} \le 12D_{\max}\left(\sup_{\vv{x}\in [0, 1]^p}|m(\vv{x})|\right)^2 \times [4s^{s + \frac{1}{2}} \iota + (4s^{s+1} +3 ) \eta_{n}].
\end{equation}
The proof of \eqref{exponential.prob.23} is deferred to Section~\ref{proof.exponential.prob.23}. We note that while the inequality in \eqref{exponential.prob.23} could be stated without explicitly referencing $\mathcal{P}$, $\mathcal{I}$, or the bound in \eqref{vvu.con.1}, these constructions are fundamental to its derivation. We present them here, rather than postponing them to the technical appendix, to provide a clearer and more cohesive overview of the proof strategy.

By \eqref{exponential.prob.23}, the definition of $(\vv{w}^{\star}, c^{\star})$ as in \eqref{wstar.1}, and the definition of $E(\cdot,\cdot, \cdot)$ as in \eqref{tspursuit.1}, we conclude for all large $n$ that on $\mathcal{B}\cap \mathcal{D}$, 
\begin{equation}
\begin{split}
     \label{prop2.2}
     % (\vv{w}, c) \in E(\rho_{n} + Q_{n}, \boldsymbol{t}^{'}, W_{p, s}).
     & \mathbb{E}[\textnormal{Var}(m(\boldsymbol{X})| \boldsymbol{1}_{\boldsymbol{X} \in \boldsymbol{t}^{'}}, \boldsymbol{1}_{\vv{w}^{\top}\boldsymbol{X}  > c} ) \boldsymbol{1}_{\boldsymbol{X} \in \boldsymbol{t}^{'}}] \\
      & \le \inf_{(\vv{w}^{'}, c^{'}) \in W_{p, s_0} } \mathbb{E}[\textnormal{Var}( m(\boldsymbol{X})| \boldsymbol{1}_{\boldsymbol{X} \in \boldsymbol{t}^{'}}, \boldsymbol{1}_{\vv{w}^{'\top}\boldsymbol{X}  > c^{'}} ) \boldsymbol{1}_{\boldsymbol{X} \in \boldsymbol{t}^{'}}] + 2(\rho_{n} + Q_{n}).
      \end{split}
 \end{equation}

By \eqref{prop2.3}--\eqref{prop2.2} and some simple calculations, we conclude the desired results of \eqref{prop2.2.b}.

\subsubsection{Proof of \texorpdfstring{\eqref{exponential.prob.23}}{LG} }\label{proof.exponential.prob.23}

Let us begin with deriving that on $\mathcal{B}$,
\begin{equation}
    \begin{split}\label{exponential.prob.40}
        & \mathbb{E}[\textnormal{Var}(m(\boldsymbol{X})| \boldsymbol{1}_{\boldsymbol{X} \in \boldsymbol{t}^{'}}, \boldsymbol{1}_{\vv{w}^{\top}\boldsymbol{X} > c}) \boldsymbol{1}_{\boldsymbol{X} \in \boldsymbol{t}^{'}}] \\
        & \le \inf_{(\vv{w}^{'}, c^{'}) \in \cup_{q=1}^{B}\Lambda_{q}} \mathbb{E}[\textnormal{Var}(m(\boldsymbol{X})| \boldsymbol{1}_{\boldsymbol{X} \in \boldsymbol{t}^{'}}, \boldsymbol{1}_{\vv{w}^{'\top}\boldsymbol{X} > c^{'}}) \boldsymbol{1}_{\boldsymbol{X} \in \boldsymbol{t}^{'}}]  + 2\rho_{n},        
    \end{split}
\end{equation}
where the inequality follows from the definitions of $(\vv{w}, c)$ and the event $\mathcal{B}$.

To further bound the right-hand side of \eqref{exponential.prob.40}, we identify a sample index $k^{\star}$ that serves as a proxy for the optimal threshold:
    $$k^{\star} = \argmin_{1\le i\le n} |\vv{u}^{\dagger\top}\boldsymbol{X}_{i} - c^{\star}|.$$
    Since $\vv{u}^{\dagger}\in \mathcal{J}(s_0)$, there is some $(\vv{u}, \vv{u}^{\top}\boldsymbol{X}_{k^{\star}}) \in \cup_{q=1}^{B}\Lambda_{q}$ such that on $\mathcal{D}$,
    \begin{equation}
    	\label{exponential.prob.22}
    	\norm{\vv{u}^{\dagger} - \vv{u}}_{2} \le \iota,
    \end{equation}
    where we recall that on $\mathcal{D}$, $\sup_{\vv{b} \in \mathcal{J}(s_0)} \min_{\vv{v} \in \Lambda} \norm{\vv{b} - \vv{v}}_{2} \le \iota$, in which $\Lambda = \{ \vv{v}: (\vv{v}, c)\in \cup_{q=1}^{B} \Lambda_{q} \}$. Then,
    \begin{equation}
    	\begin{split}\label{exponential.prob.23.b.x}
    		& \inf_{(\vv{w}^{'}, c^{'}) \in \cup_{q=1}^{B}\Lambda_{q}} \mathbb{E}[\textnormal{Var}(m(\boldsymbol{X})| \boldsymbol{1}_{\boldsymbol{X} \in \boldsymbol{t}^{'}}, \boldsymbol{1}_{\vv{w}^{'\top}\boldsymbol{X} > c^{'}}) \boldsymbol{1}_{\boldsymbol{X} \in \boldsymbol{t}^{'}}]  + 2\rho_{n}\\
    		& \le  \mathbb{E}[\textnormal{Var}(m(\boldsymbol{X})| \boldsymbol{1}_{\boldsymbol{X} \in \boldsymbol{t}^{'}}, \boldsymbol{1}_{\vv{u}^{\top}\boldsymbol{X} > \vv{u}^{\top}\boldsymbol{X}_{k^{\star}}}) \boldsymbol{1}_{\boldsymbol{X} \in \boldsymbol{t}^{'}}]  + 2\rho_{n},
    	\end{split}
    \end{equation}
since $(\vv{u}, \vv{u}^{\top}\boldsymbol{X}_{k^{\star}}) \in \cup_{q=1}^{B}\Lambda_{q}$.

The proof is completed by establishing the following bound on the approximation error between our discrete candidate split and the optimal population split:
\begin{equation}
    \begin{split}\label{exponential.prob.23.b}
        & \big|\mathbb{E}[\textnormal{Var}(m(\boldsymbol{X}) \mid \boldsymbol{1}_{\boldsymbol{X}\in \boldsymbol{t}^{'}}, \boldsymbol{1}_{\vv{u}^{\top}\boldsymbol{X} > \vv{u}^{\top}\boldsymbol{X}_{k^{\star}}}) \boldsymbol{1}_{\boldsymbol{X}\in \boldsymbol{t}^{'}}] \\
        & \quad - \mathbb{E}[\textnormal{Var}(m(\boldsymbol{X}) \mid \boldsymbol{1}_{\boldsymbol{X}\in \boldsymbol{t}^{'}}, \boldsymbol{1}_{\vv{w}^{\star\top}\boldsymbol{X} > c^{\star}}) \boldsymbol{1}_{\boldsymbol{X}\in \boldsymbol{t}^{'}}]\big| \\
        & \le 24\left(\sup_{\vv{x}\in [0, 1]^p}|m(\vv{x})|\right)^2  \\
        & \quad \quad\times \left[D_{\max} s^s  4\norm{ \vv{u} - \vv{u}^{\dagger} }_{2} \sqrt{s} + 3\eta_{n} +D_{\max} s^s  4\norm{\vv{u}^{\dagger} - \vv{w}^{\star}}_{2} \sqrt{s}\right]\\
        & \le 24\left(\sup_{\vv{x}\in [0, 1]^p}|m(\vv{x})|\right)^2  \\
        & \quad \quad\times \left[D_{\max} s^s  4\iota\times \sqrt{s} + 3\eta_{n} +D_{\max} s^s  4\sqrt{s}\times\eta_n \times \sqrt{s}\right],
    \end{split}
\end{equation}
where $\vv{u}^{\dagger} \in \mathcal{P}$ and $(\vv{u}, \vv{u}^{\top}\boldsymbol{X}_{k^{\star}}) \in \cup_{q=1}^{B}\Lambda_{q}$ with \eqref{vvu.con.1} and \eqref{exponential.prob.22}, and $(\vv{w}^{\star}, c^{\star}) \in W_{p, s_0}$. The second inequality holds because of \eqref{vvu.con.1} and \eqref{exponential.prob.22}. Recall that \(\mathcal{P}\) is defined as in \eqref{p.def}.

In the following, we show the first inequality of \eqref{exponential.prob.23.b}, starting with establishing an upper bound for $\mathbb{P}(\{|\boldsymbol{1}_{\vv{u}^{\top}\boldsymbol{X} > \vv{u}^{\top}\boldsymbol{X}_{k^{\star}} } - \boldsymbol{1}_{\vv{w}^{\star\top}\boldsymbol{X} > c^{\star}}| = 1 \} )$. Let us derive that
\begin{equation}
	\label{exponential.prob.4.c}
	\begin{split}    
		& \{|\boldsymbol{1}_{\vv{u}^{\top}\boldsymbol{X} > \vv{u}^{\top}\boldsymbol{X}_{k^{\star}} } - \boldsymbol{1}_{\vv{w}^{\star\top}\boldsymbol{X} > c^{\star}}| = 1 \} \\
		& \subseteq \{|\boldsymbol{1}_{\vv{u}^{\top}\boldsymbol{X} > \vv{u}^{\top}\boldsymbol{X}_{k^{\star}} } - \boldsymbol{1}_{\vv{u}^{\dagger\top}\boldsymbol{X} > \vv{u}^{\dagger\top}\boldsymbol{X}_{k^{\star}} }| = 1\} \cup \{ | \boldsymbol{1}_{\vv{u}^{\dagger\top}\boldsymbol{X} > \vv{u}^{\dagger\top}\boldsymbol{X}_{k^{\star}} } - \boldsymbol{1}_{\vv{w}^{\star\top}\boldsymbol{X} > c^{\star}}| = 1 \}\\
		& \subseteq \Big(\{|\boldsymbol{1}_{\vv{u}^{\top}\boldsymbol{X} > \vv{u}^{\top}\boldsymbol{X}_{k^{\star}} } - \boldsymbol{1}_{\vv{u}^{\dagger\top}\boldsymbol{X} > \vv{u}^{\dagger\top}\boldsymbol{X}_{k^{\star}} }| = 1\} \\
		& \quad \cup \{ | \boldsymbol{1}_{\vv{u}^{\dagger\top}\boldsymbol{X} > \vv{u}^{\dagger\top}\boldsymbol{X}_{k^{\star}} } - \boldsymbol{1}_{\vv{u}^{\dagger\top}\boldsymbol{X} > c^{\star}}| = 1 \}\\
		& \quad \cup \{ | \boldsymbol{1}_{\vv{u}^{\dagger\top}\boldsymbol{X} > c^{\star} } - \boldsymbol{1}_{\vv{w}^{\star\top}\boldsymbol{X} > c^{\star}}| = 1 \}\Big).
	\end{split}
\end{equation}

To establish an upper bound for $\mathbb{P}(\{|\boldsymbol{1}_{\vv{u}^{\top}\boldsymbol{X} > \vv{u}^{\top}\boldsymbol{X}_{k^{\star}} } - \boldsymbol{1}_{\vv{w}^{\star\top}\boldsymbol{X} > c^{\star}}| = 1 \} )$, we have to bound the probabilities of the three events on the RHS of \eqref{exponential.prob.4.c}. In what follows, we begin with dealing with the second event. The definition of $\mathcal{I}$ in \eqref{I.def} and \eqref{exponential.prob.1.e} imply that on $\mathcal{I}$,
\begin{equation*}     
	\mathbb{P}(\min\{\vv{u}^{\dagger\top}\boldsymbol{X}_{k^{\star}} , c^{\star}\}\le \vv{u}^{\dagger\top}\boldsymbol{X} \le \max\{\vv{u}^{\dagger\top}\boldsymbol{X}_{k^{\star}} , c^{\star}\} \mid \mathcal{X}_{n}) \le 2\eta_{n},
\end{equation*}
leading to that on $\mathcal{I}$,
\begin{equation}
	\label{exponential.prob.13}
	\mathbb{P}(\{ | \boldsymbol{1}_{\vv{u}^{\dagger\top}\boldsymbol{X} > \vv{u}^{\dagger\top}\boldsymbol{X}_{k^{\star}} } - \boldsymbol{1}_{\vv{u}^{\dagger\top}\boldsymbol{X} > c^{\star}}| = 1 \} \mid \mathcal{X}_{n}) \le  2\eta_{n}.
\end{equation}

Next, we proceed to deal with the other two events. By the Cauchy-Schwarz inequality, the definitions of $\vv{u}$ and $\vv{u}^{\dagger}$, and that $\mathbb{P}(\boldsymbol{X}\in [0, 1]^p) = 1$, it holds almost surely that
\begin{equation}
	\label{exponential.prob.3}
	\begin{split}
		& |\vv{u}^{\top}\boldsymbol{X} - \vv{u}^{\top}\boldsymbol{X}_{k^{\star}} - (\vv{u}^{\dagger\top}\boldsymbol{X} - \vv{u}^{\dagger\top}\boldsymbol{X}_{k^{\star}})  | \\
		& \le |\sum_{j \in \{k : |u_{k}| + |u_{k}^{\dagger}|> 0 \}} (u_{j} - u_{j}^{\dagger}) X_{j}| + |\sum_{j \in \{k : |u_{k}| + |u_{k}^{\dagger}|> 0 \}} (u_{j} - u_{j}^{\dagger}) X_{k^{\star}j}|\\
		& \le 2\norm{\vv{u} - \vv{u}^{\dagger}}_{2} \sqrt{s + s_0}\\
		& \le 4\norm{\vv{u} - \vv{u}^{\dagger}}_{2} \sqrt{s}.
	\end{split}
\end{equation}
Additionally, we derive that
\begin{equation}
	\begin{split}
		\label{exponential.prob.6}
		& \{|\boldsymbol{1}_{\vv{u}^{\top}\boldsymbol{X} > \vv{u}^{\top}\boldsymbol{X}_{k^{\star}} } - \boldsymbol{1}_{\vv{u}^{\dagger\top}\boldsymbol{X} > \vv{u}^{\dagger\top}\boldsymbol{X}_{k^{\star}}}| = 1 \} \\
		& \subseteq (\{ \vv{u}^{\top}\boldsymbol{X} > \vv{u}^{\top}\boldsymbol{X}_{k^{\star}} \} \cap \{\vv{u}^{\dagger\top}\boldsymbol{X}  \le \vv{u}^{\dagger\top}\boldsymbol{X}_{k^{\star}} \}) \\
		& \quad \cup (\{ \vv{u}^{\top}\boldsymbol{X} \le \vv{u}^{\top}\boldsymbol{X}_{k^{\star}}\} \cap \{\vv{u}^{\dagger\top}\boldsymbol{X}  >  \vv{u}^{\dagger\top}\boldsymbol{X}_{k^{\star}} \}).
	\end{split}
\end{equation}
By \eqref{exponential.prob.3}--\eqref{exponential.prob.6}, the following two sets are empty.
\begin{equation*}
	\begin{split}
		& (\{ \vv{u}^{\top}\boldsymbol{X} > \vv{u}^{\top}\boldsymbol{X}_{k^{\star}} \} \cap \{\vv{u}^{\dagger\top}\boldsymbol{X}  \le   \vv{u}^{\dagger\top}\boldsymbol{X}_{k^{\star}}\}) \cap \{| \vv{u}^{\dagger\top}\boldsymbol{X} -  \vv{u}^{\dagger\top}\boldsymbol{X}_{k^{\star}} | > 4\norm{\vv{u} - \vv{u}^{\dagger}}_{2} \sqrt{s}  \} ,\\
		& (\{ \vv{u}^{\top}\boldsymbol{X} \le \vv{u}^{\top}\boldsymbol{X}_{k^{\star}}\} \cap \{\vv{u}^{\dagger\top}\boldsymbol{X}  >  \vv{u}^{\dagger\top}\boldsymbol{X}_{k^{\star}}\}) \cap \{|\vv{u}^{\dagger\top}\boldsymbol{X} - \vv{u}^{\dagger\top}\boldsymbol{X}_{k^{\star}}| > 4\norm{\vv{u} - \vv{u}^{\dagger}}_{2} \sqrt{s}  \},
	\end{split}
\end{equation*}
which concludes that
\begin{equation}
	\label{exponential.prob.4}
	\{|\boldsymbol{1}_{\vv{u}^{\top}\boldsymbol{X} > \vv{u}^{\top}\boldsymbol{X}_{k^{\star}} } - \boldsymbol{1}_{\vv{u}^{\dagger\top}\boldsymbol{X} > \vv{u}^{\dagger\top}\boldsymbol{X}_{k^{\star}} }| = 1 \} \subseteq \{|\vv{u}^{\dagger\top}\boldsymbol{X} - \vv{u}^{\dagger\top}\boldsymbol{X}_{k^{\star}} | \le 4\norm{\vv{u} - \vv{u}^{\dagger}}_{2} \sqrt{s}\} .
\end{equation}
Similarly, we have that
\begin{equation}
	\label{exponential.prob.4.b}
	\{|\boldsymbol{1}_{\vv{u}^{\dagger\top}\boldsymbol{X} > c^{\star} } - \boldsymbol{1}_{\vv{w}^{\star\top}\boldsymbol{X} > c^{\star} }| = 1 \} \subseteq \{|\vv{w}^{\star\top}\boldsymbol{X} - c^{\star} | \le 4\norm{\vv{u}^{\dagger} - \vv{w}^{\star}}_{2} \sqrt{s}\} .
\end{equation}

To bound the probabilities of the RHS of \eqref{exponential.prob.4}--\eqref{exponential.prob.4.b}, we have to establish an upper bound for the volume in between two hyperplanes within the $p$-dimensional unit hypercube, which is derived as follows. Since the maximum intersection area between an \(s\)-dimensional hypercube with unit-length edges and a hyperplane is bounded by $\frac{\pi^{\frac{s-1}{2}} \times \left(\frac{\sqrt{s}}{2}\right)^{s-1}} {\Gamma (s-1 + \frac{1}{2})} \le s^{ s}$, in which $\frac{\sqrt{s}}{2}$ represents the radius of the smallest $(s-1)$-dimensional ball that surrounds the hypercube, we derive that for any $\vv{v}\in\mathbb{R}^s$ with $\norm{\vv{v}}_{2} = 1$, each $q\ge 0$, and each $b\in \mathbb{R}$,
\begin{equation*}
	\textnormal{ Volume of } \{\vv{x} \in [0, 1]^{s}: |\vv{v}^{\top} \vv{x} - b| \le q \} \le s^{ s} \times q,
\end{equation*}
which, in combination with $\norm{\vv{w}^{\star}}_{0}\le s$ and $\norm{\vv{u}^{\dagger}}_{0}\le s$ due to the definitions of $\vv{w}^{\star}$ and $\vv{u}^{\dagger}$, concludes that for each $b\in \mathbb{R}$ and $q\ge 0$,
\begin{equation}\label{exponential.prob.7}
	\begin{split}
		& \textnormal{ Volume of } \{\vv{x} \in [0, 1]^{p}: |\vv{w}^{\star\top} \vv{x} - b| \le q \} \le s^{ s} \times q,\\
		& \textnormal{ Volume of } \{\vv{x} \in [0, 1]^{p}: |\vv{u}^{\dagger\top} \vv{x} - b| \le q \} \le s^{ s} \times q.
	\end{split}    
\end{equation}

By \eqref{exponential.prob.4}--\eqref{exponential.prob.7} and the assumption (b) of Condition~\ref{regularity.1},
\begin{equation}
	\begin{split}\label{exponential.prob.14}
		& \mathbb{P}(\{|\boldsymbol{1}_{\vv{u}^{\top}\boldsymbol{X} > \vv{u}^{\top}\boldsymbol{X}_{k^{\star}} } - \boldsymbol{1}_{\vv{u}^{\dagger\top}\boldsymbol{X} > \vv{u}^{\dagger\top}\boldsymbol{X}_{k^{\star}} }| = 1 \}) \\
		& \quad \le\mathbb{P}\left(\left\{|\vv{u}^{\dagger\top}\boldsymbol{X} - \vv{u}^{\dagger\top}\boldsymbol{X}_{k^{\star}} | \le 4\norm{\vv{u} - \vv{u}^{\dagger}}_{2} \sqrt{s}\right\}\right) \\
		& \quad \le \sup_{c \in \mathbb{R}}\mathbb{P}\left(\left\{|\vv{u}^{\dagger\top}\boldsymbol{X} - c | \le 4\norm{\vv{u} - \vv{u}^{\dagger}}_{2} \sqrt{s}\right\}\right) \\
		& \quad\le s^s \times 4\norm{\vv{u} - \vv{u}^{\dagger}}_{2} \sqrt{s} D_{\max}, \\
	\end{split}    
\end{equation}
where the second inequality holds because $\boldsymbol{X}_{i}$'s and $\boldsymbol{X}$ are independent. Additionally,
\begin{equation}
	\begin{split}\label{exponential.prob.14.b}
		& \mathbb{P}(\{|\boldsymbol{1}_{\vv{u}^{\dagger\top}\boldsymbol{X} > c^{\star} } - \boldsymbol{1}_{\vv{w}^{\star\top}\boldsymbol{X} > c^{\star} }| = 1 \} )  \le s^s \times 4\norm{\vv{u}^{\dagger} - \vv{w}^{\star}}_{2} \sqrt{s} D_{\max}.
	\end{split}    
\end{equation}

By \eqref{exponential.prob.4.c}--\eqref{exponential.prob.13}, \eqref{exponential.prob.14}--\eqref{exponential.prob.14.b}, and \eqref{exponential.prob.1.c}, we conclude that
\begin{equation}
	\label{exponential.prob.20}
	\begin{split}            
		& \mathbb{P}(\{|\boldsymbol{1}_{\vv{u}^{\top}\boldsymbol{X} > \vv{u}^{\top}\boldsymbol{X}_{k^{\star}} } - \boldsymbol{1}_{\vv{w}^{\star\top}\boldsymbol{X} > c^{\star}}| = 1 \} ) \\
		& \le \mathbb{P}(\{|\boldsymbol{1}_{\vv{u}^{\top}\boldsymbol{X} > \vv{u}^{\top}\boldsymbol{X}_{k^{\star}} } - \boldsymbol{1}_{\vv{u}^{\dagger\top}\boldsymbol{X} > \vv{u}^{\dagger\top}\boldsymbol{X}_{k^{\star}} }| = 1\}) \\
		& \quad + \mathbb{E}[\mathbb{P}(\{ | \boldsymbol{1}_{\vv{u}^{\dagger\top}\boldsymbol{X} > \vv{u}^{\dagger\top}\boldsymbol{X}_{k^{\star}} } - \boldsymbol{1}_{\vv{u}^{\dagger\top}\boldsymbol{X} > c^{\star}}| = 1 \} \mid \mathcal{X}_{n}) \boldsymbol{1}_{\mathcal{I}}]  + \mathbb{P}(\mathcal{I}^c)\\
		& \quad + \mathbb{P}(\{ | \boldsymbol{1}_{\vv{u}^{\dagger\top}\boldsymbol{X} > c^{\star} } - \boldsymbol{1}_{\vv{w}^{\star\top}\boldsymbol{X} > c^{\star}}| = 1 \})  \\
		& \le s^s \times 4\norm{ \vv{u} - \vv{u}^{\dagger} }_{2} \sqrt{s} D_{\max}+ 3\eta_{n} + s^s \times 4\norm{\vv{u}^{\dagger} - \vv{w}^{\star}}_{2} \sqrt{s} D_{\max}. 
	\end{split}
\end{equation}

With \eqref{exponential.prob.20}, the desired first inequality of \eqref{exponential.prob.23.b} is established as follows. To simplify the following derivation in  \eqref{exponential.prob.9}--\eqref{exponential.prob.10}, we let $\zeta = \vv{u}^{\top}\boldsymbol{X}_{k^{\star}}$. We have that 
\begin{equation}
	\begin{split}\label{exponential.prob.9}
		& \big|\mathbb{E}[\textnormal{Var}(m(\boldsymbol{X}) \mid \boldsymbol{1}_{\boldsymbol{X}\in \boldsymbol{t}^{'}}, \boldsymbol{1}_{\vv{u}^{\top}\boldsymbol{X} > \zeta}) \boldsymbol{1}_{\boldsymbol{X}\in \boldsymbol{t}^{'}}] - \mathbb{E}[\textnormal{Var}(m(\boldsymbol{X}) \mid \boldsymbol{1}_{\boldsymbol{X}\in \boldsymbol{t}^{'}}, \boldsymbol{1}_{\vv{w}^{\star\top}\boldsymbol{X} > c^{\star}}) \boldsymbol{1}_{\boldsymbol{X}\in \boldsymbol{t}^{'}}] \big|\\
		& = \Big| \mathbb{E}\Big\{ \big[  \mathbb{E}(m(\boldsymbol{X}) \mid \boldsymbol{1}_{\boldsymbol{X}\in \boldsymbol{t}^{'}}, \boldsymbol{1}_{\vv{w}^{\star\top}\boldsymbol{X} > c^{\star}}) - \mathbb{E}(m(\boldsymbol{X}) \mid \boldsymbol{1}_{\boldsymbol{X}\in \boldsymbol{t}^{'}}, \boldsymbol{1}_{\vv{u}^{\top}\boldsymbol{X} > \zeta}) \big]\\
		& \quad\times \big[2m(\boldsymbol{X}) - \mathbb{E}(m(\boldsymbol{X}) \mid \boldsymbol{1}_{\boldsymbol{X}\in \boldsymbol{t}^{'}}, \boldsymbol{1}_{\vv{w}^{\star\top}\boldsymbol{X} > c^{\star}}) - \mathbb{E}(m(\boldsymbol{X}) \mid \boldsymbol{1}_{\boldsymbol{X}\in \boldsymbol{t}^{'}}, \boldsymbol{1}_{\vv{u}^{\top}\boldsymbol{X} > \zeta}) \big] \boldsymbol{1}_{\boldsymbol{X}\in \boldsymbol{t}^{'}}\Big\} \Big| \\
		& \le 4\left(\sup_{\vv{x}\in [0, 1]^p}|m(\vv{x})|\right) \\
		& \quad \times \mathbb{E} \left[ \big|  \mathbb{E}(m(\boldsymbol{X}) \mid \boldsymbol{1}_{\boldsymbol{X}\in \boldsymbol{t}^{'}}, \boldsymbol{1}_{\vv{w}^{\star\top}\boldsymbol{X} > c^{\star}}) - \mathbb{E}(m(\boldsymbol{X}) \mid \boldsymbol{1}_{\boldsymbol{X}\in \boldsymbol{t}^{'}}, \boldsymbol{1}_{\vv{u}^{\top}\boldsymbol{X} > \zeta}) \big| \times \boldsymbol{1}_{\boldsymbol{X}\in \boldsymbol{t}^{'}} \right]\\
		& \le 24\left(\sup_{\vv{x}\in [0, 1]^p}|m(\vv{x})|\right)^2 \times \mathbb{P}(\{|\boldsymbol{1}_{\vv{u}^{\top}\boldsymbol{X} > \vv{u}^{\top}\boldsymbol{X}_{k^{\star}} } - \boldsymbol{1}_{\vv{w}^{\star\top}\boldsymbol{X} > c^{\star}}| = 1 \}),
	\end{split}
\end{equation}
To establish the first equality, we use the conditional variance formula:
$$\text{Var}(m(\boldsymbol{X}) \mid \mathcal{F}) = \mathbb{E}\big[ (m(\boldsymbol{X}) - \mathbb{E}[m(\boldsymbol{X}) \mid \mathcal{F}])^2 \mid \mathcal{F} \big],$$
and that  $b^2 - a^2 = (b - a)(b + a)$ for real $a$ and $b$. The second inequality holds because of \eqref{exponential.prob.8}--\eqref{exponential.prob.10} below.

A direct calculation shows that
{\small \begin{equation}
		\begin{split}\label{exponential.prob.8}
			& \mathbb{E}\left[ \big|  \mathbb{E}(m(\boldsymbol{X}) \mid \boldsymbol{1}_{\boldsymbol{X}\in \boldsymbol{t}^{'}}, \boldsymbol{1}_{\vv{w}^{\star\top}\boldsymbol{X} > c^{\star}}) - 
			\mathbb{E}(m(\boldsymbol{X}) \mid \boldsymbol{1}_{\boldsymbol{X}\in \boldsymbol{t}^{'}}, \boldsymbol{1}_{\vv{u}^{\top}\boldsymbol{X} > \zeta}) \big| \times \boldsymbol{1}_{\boldsymbol{X}\in \boldsymbol{t}^{'}} \right] \\
			& \le \mathbb{E}\left[\boldsymbol{1}_{\boldsymbol{X} \in \boldsymbol{t}^{'}} \boldsymbol{1}_{\vv{w}^{\star\top}\boldsymbol{X} > c^{\star}}\boldsymbol{1}_{\vv{u}^{\top}\boldsymbol{X} > \zeta} \times \left| \frac{\mathbb{E}(m(\boldsymbol{X})  \boldsymbol{1}_{\boldsymbol{X}\in \boldsymbol{t}^{'}} \boldsymbol{1}_{\vv{w}^{\star\top}\boldsymbol{X} > c^{\star}} )}{  \mathbb{P}(\boldsymbol{X}\in \boldsymbol{t}^{'}, \vv{w}^{\star\top}\boldsymbol{X} > c^{\star})} - 
			\frac{\mathbb{E}(m(\boldsymbol{X})  \boldsymbol{1}_{\boldsymbol{X}\in \boldsymbol{t}^{'}} \boldsymbol{1}_{\vv{u}^{\top}\boldsymbol{X} > \zeta} )}{  \mathbb{P}(\boldsymbol{X}\in \boldsymbol{t}^{'}, \vv{u}^{\top}\boldsymbol{X} > \zeta)} \right|   \right]\\
			& \ + \mathbb{E}\left[\boldsymbol{1}_{\boldsymbol{X} \in \boldsymbol{t}^{'}} \boldsymbol{1}_{\vv{w}^{\star\top}\boldsymbol{X} \le c^{\star}}\boldsymbol{1}_{\vv{u}^{\top}\boldsymbol{X} \le \zeta} \times \left| \frac{\mathbb{E}(m(\boldsymbol{X})  \boldsymbol{1}_{\boldsymbol{X}\in \boldsymbol{t}^{'}} \boldsymbol{1}_{\vv{w}^{\star\top}\boldsymbol{X} \le c^{\star}} )}{  \mathbb{P}(\boldsymbol{X}\in \boldsymbol{t}^{'}, \vv{w}^{\star\top}\boldsymbol{X} \le c^{\star})} - 
			\frac{\mathbb{E}(m(\boldsymbol{X})  \boldsymbol{1}_{\boldsymbol{X}\in \boldsymbol{t}^{'}} \boldsymbol{1}_{\vv{u}^{\top}\boldsymbol{X} \le \zeta} )}{  \mathbb{P}(\boldsymbol{X}\in \boldsymbol{t}^{'}, \vv{u}^{\top}\boldsymbol{X} \le \zeta)} \right|   \right]\\
			& \ + \mathbb{E}\left[\boldsymbol{1}_{\boldsymbol{X} \in \boldsymbol{t}^{'}} \boldsymbol{1}_{\vv{w}^{\star\top}\boldsymbol{X} > c^{\star}}\boldsymbol{1}_{\vv{u}^{\top}\boldsymbol{X} \le \zeta} \times \left| \frac{\mathbb{E}(m(\boldsymbol{X})  \boldsymbol{1}_{\boldsymbol{X}\in \boldsymbol{t}^{'}} \boldsymbol{1}_{\vv{w}^{\star\top}\boldsymbol{X} > c^{\star}} )}{  \mathbb{P}(\boldsymbol{X}\in \boldsymbol{t}^{'}, \vv{w}^{\star\top}\boldsymbol{X} > c^{\star})} - 
			\frac{\mathbb{E}(m(\boldsymbol{X})  \boldsymbol{1}_{\boldsymbol{X}\in \boldsymbol{t}^{'}} \boldsymbol{1}_{\vv{u}^{\top}\boldsymbol{X} \le \zeta} )}{  \mathbb{P}(\boldsymbol{X}\in \boldsymbol{t}^{'}, \vv{u}^{\top}\boldsymbol{X} \le \zeta)} \right|   \right]\\
			& \ + \mathbb{E}\left[\boldsymbol{1}_{\boldsymbol{X} \in \boldsymbol{t}^{'}} \boldsymbol{1}_{\vv{w}^{\star\top}\boldsymbol{X} \le c^{\star}}\boldsymbol{1}_{\vv{u}^{\top}\boldsymbol{X} > \zeta} \times \left| \frac{\mathbb{E}(m(\boldsymbol{X})  \boldsymbol{1}_{\boldsymbol{X}\in \boldsymbol{t}^{'}} \boldsymbol{1}_{\vv{w}^{\star\top}\boldsymbol{X} \le c^{\star}} )}{  \mathbb{P}(\boldsymbol{X}\in \boldsymbol{t}^{'}, \vv{w}^{\star\top}\boldsymbol{X} \le c^{\star})} - 
			\frac{\mathbb{E}(m(\boldsymbol{X})  \boldsymbol{1}_{\boldsymbol{X}\in \boldsymbol{t}^{'}} \boldsymbol{1}_{\vv{u}^{\top}\boldsymbol{X} > \zeta} )}{  \mathbb{P}(\boldsymbol{X}\in \boldsymbol{t}^{'}, \vv{u}^{\top}\boldsymbol{X} > \zeta)} \right|   \right]        \\
			& \eqqcolon (I) + (II) + (III) + (IV).
		\end{split}
\end{equation}}%

To bound (I), we analyze the absolute difference between the two conditional expectations within its integrand:
{\small \begin{equation}
		\begin{split}
			& \left| \frac{\mathbb{E}(m(\boldsymbol{X})  \boldsymbol{1}_{\boldsymbol{X}\in \boldsymbol{t}^{'}} \boldsymbol{1}_{\vv{w}^{\star\top}\boldsymbol{X} > c^{\star}} )}{  \mathbb{P}(\boldsymbol{X}\in \boldsymbol{t}^{'}, \vv{w}^{\star\top}\boldsymbol{X} > c^{\star})} - 
			\frac{\mathbb{E}(m(\boldsymbol{X})  \boldsymbol{1}_{\boldsymbol{X}\in \boldsymbol{t}^{'}} \boldsymbol{1}_{\vv{u}^{\top}\boldsymbol{X} > \zeta} )}{  \mathbb{P}(\boldsymbol{X}\in \boldsymbol{t}^{'}, \vv{u}^{\top}\boldsymbol{X} > \zeta)} \right| \\
			& = \bigg| \frac{\mathbb{E}(m(\boldsymbol{X})  \boldsymbol{1}_{\boldsymbol{X}\in \boldsymbol{t}^{'}} \boldsymbol{1}_{\vv{w}^{\star\top}\boldsymbol{X} > c^{\star}} )}{  \mathbb{P}(\boldsymbol{X}\in \boldsymbol{t}^{'}, \vv{w}^{\star\top}\boldsymbol{X} > c^{\star})} 
			- \frac{\mathbb{E}(m(\boldsymbol{X})  \boldsymbol{1}_{\boldsymbol{X}\in \boldsymbol{t}^{'}} \boldsymbol{1}_{\vv{w}^{\star\top}\boldsymbol{X} > c^{\star}} )}{  \mathbb{P}(\boldsymbol{X}\in \boldsymbol{t}^{'}, \vv{u}^{\top}\boldsymbol{X} > \zeta)}  +\\
			& \quad + \frac{\mathbb{E}(m(\boldsymbol{X})  \boldsymbol{1}_{\boldsymbol{X}\in \boldsymbol{t}^{'}} \boldsymbol{1}_{\vv{w}^{\star\top}\boldsymbol{X} > c^{\star}} )}{  \mathbb{P}(\boldsymbol{X}\in \boldsymbol{t}^{'}, \vv{u}^{\top}\boldsymbol{X} > \zeta)}   - \frac{\mathbb{E}(m(\boldsymbol{X})  \boldsymbol{1}_{\boldsymbol{X}\in \boldsymbol{t}^{'}} \boldsymbol{1}_{\vv{u}^{\top}\boldsymbol{X} > \zeta} )}{  \mathbb{P}(\boldsymbol{X}\in \boldsymbol{t}^{'}, \vv{u}^{\top}\boldsymbol{X} > \zeta)}\bigg|\\
			& \le \frac{\left|\mathbb{E}(m(\boldsymbol{X}) | \{\boldsymbol{X}\in \boldsymbol{t}^{'} \} \cap \{\vv{w}^{\star\top}\boldsymbol{X} > c^{\star}\} )\right|  \left| \mathbb{P}(\boldsymbol{X}\in \boldsymbol{t}^{'}, \vv{w}^{\star\top}\boldsymbol{X} > c^{\star}) - \mathbb{P}(\boldsymbol{X}\in \boldsymbol{t}^{'}, \vv{u}^{\top}\boldsymbol{X} > \zeta) \right| } {\mathbb{P}(\boldsymbol{X}\in \boldsymbol{t}^{'}, \vv{u}^{\top}\boldsymbol{X} > \zeta) }\\
			& \quad + \frac{\left(\sup_{\vv{x}\in [0, 1]^p}|m(\vv{x})|\right) \times \mathbb{P}( | \boldsymbol{1}_{\vv{w}^{\star\top}\boldsymbol{X} > c^{\star}} - \boldsymbol{1}_{\vv{u}^{\top}\boldsymbol{X} > \zeta}| = 1 ) } {\mathbb{P}(\boldsymbol{X}\in \boldsymbol{t}^{'}, \vv{u}^{\top}\boldsymbol{X} > \zeta) } \\
			& \le 2 \frac{ \left(\sup_{\vv{x}\in [0, 1]^p}|m(\vv{x})|\right) \times \mathbb{P}( | \boldsymbol{1}_{\vv{w}^{\star\top}\boldsymbol{X} > c^{\star}} - \boldsymbol{1}_{\vv{u}^{\top}\boldsymbol{X} > \zeta}| = 1 ) } {\mathbb{P}(\boldsymbol{X}\in \boldsymbol{t}^{'}, \vv{u}^{\top}\boldsymbol{X} > \zeta) }.
		\end{split}
\end{equation}}%
The second inequality follows because for any events $A$, $B$, $D$, $E$, and $F$:
$$\mathbb{P}(E\backslash F) = \mathbb{P}(E) - \mathbb{P}(E\cap F) \ge \mathbb{P}(E) - \mathbb{P}(E\cap F) - \mathbb{P} (E^c\cap F) = \mathbb{P}(E) - \mathbb{P}(F),$$and$$\mathbb{P}(A \backslash B)\ge \mathbb{P}( (A \cap D) \backslash B) = \mathbb{P}( (A \cap D) \backslash (B\cap D ) ),$$
along with the identity
\begin{equation}\label{difference.1}
\mathbb{P}(|\boldsymbol{1}_{A} - \boldsymbol{1}_{B}| = 1)=\mathbb{P}(A\backslash B) + \mathbb{P}(B\backslash A).
\end{equation}
By letting $D = \{\boldsymbol{X} \in \boldsymbol{t}'\}$, $A = \{\vv{w}^{\star\top}\boldsymbol{X} > c^{\star}\}$, $B = \{\vv{u}^{\top}\boldsymbol{X} > \zeta\}$, $E = A\cap D$, and $F = B\cap D$, we derive the second inequality. Furthermore, these inequalities are established by bounding the conditional mean by the supremum of $m(\boldsymbol{X})$.

Similarly, for the term (II), we apply the same analysis to the absolute difference between the two conditional expectations within its integrand:
\begin{equation}
	\begin{split}
		& \left| \frac{\mathbb{E}(m(\boldsymbol{X})  \boldsymbol{1}_{\boldsymbol{X}\in \boldsymbol{t}^{'}} \boldsymbol{1}_{\vv{w}^{\star\top}\boldsymbol{X} \le c^{\star}} )}{  \mathbb{P}(\boldsymbol{X}\in \boldsymbol{t}^{'}, \vv{w}^{\star\top}\boldsymbol{X} \le c^{\star})} - 
		\frac{\mathbb{E}(m(\boldsymbol{X})  \boldsymbol{1}_{\boldsymbol{X}\in \boldsymbol{t}^{'}} \boldsymbol{1}_{\vv{u}^{\top}\boldsymbol{X} \le \zeta} )}{  \mathbb{P}(\boldsymbol{X}\in \boldsymbol{t}^{'}, \vv{u}^{\top}\boldsymbol{X} \le \zeta)} \right| \\       
		& \le 2 \frac{ \left(\sup_{\vv{x}\in [0, 1]^p}|m(\vv{x})|\right) \times \mathbb{P}( | \boldsymbol{1}_{\vv{w}^{\star\top}\boldsymbol{X} \le c^{\star}} - \boldsymbol{1}_{\vv{u}^{\top}\boldsymbol{X} \le \zeta}| = 1 ) } {\mathbb{P}(\boldsymbol{X}\in \boldsymbol{t}^{'}, \vv{u}^{\top}\boldsymbol{X} \le \zeta) }\\
		& = 2 \frac{ \left(\sup_{\vv{x}\in [0, 1]^p}|m(\vv{x})|\right) \times \mathbb{P}( | \boldsymbol{1}_{\vv{w}^{\star\top}\boldsymbol{X} > c^{\star}} - \boldsymbol{1}_{\vv{u}^{\top}\boldsymbol{X} > \zeta}| = 1 ) } {\mathbb{P}(\boldsymbol{X}\in \boldsymbol{t}^{'}, \vv{u}^{\top}\boldsymbol{X} \le \zeta) },
	\end{split}
\end{equation}
where the equality holds because $ | \boldsymbol{1}_{\vv{w}^{\star\top}\boldsymbol{X} > c^{\star}} - \boldsymbol{1}_{\vv{u}^{\top}\boldsymbol{X} > \zeta}| =  | \boldsymbol{1}_{\vv{w}^{\star\top}\boldsymbol{X} \le c^{\star}} - \boldsymbol{1}_{\vv{u}^{\top}\boldsymbol{X} \le \zeta}|$.

By similar calculation, \eqref{difference.1}, and recall that $\zeta = \vv{u}^{\top}\boldsymbol{X}_{k^{\star}}$, we have that
\begin{equation}
	\begin{split}  
		\label{exponential.prob.8.b}
		& (III) + (IV)\\
		& \le \left(\mathbb{E}[\boldsymbol{1}_{\vv{w}^{\star\top}\boldsymbol{X} > c^{\star}}\boldsymbol{1}_{\vv{u}^{\top}\boldsymbol{X} \le \zeta}] + \mathbb{E}[\boldsymbol{1}_{\vv{w}^{\star\top}\boldsymbol{X} \le c^{\star}}\boldsymbol{1}_{\vv{u}^{\top}\boldsymbol{X} > \zeta}]\right) \times 2\left(\sup_{\vv{x}\in [0, 1]^p}|m(\vv{x})|\right)\\
		&= 2\left(\sup_{\vv{x}\in [0, 1]^p}|m(\vv{x})|\right) \times \mathbb{P}( |\boldsymbol{1}_{\vv{u}^{\top}\boldsymbol{X} > \vv{u}^{\top}\boldsymbol{X}_{k^{\star}} } - \boldsymbol{1}_{\vv{w}^{\star\top}\boldsymbol{X} > c^{\star}}| = 1 ) .
	\end{split}
\end{equation}

By \eqref{exponential.prob.8}--\eqref{exponential.prob.8.b}, we deduce that
\begin{equation}
	\begin{split}  \label{exponential.prob.10}
		& (III) + (IV) + (I) + (II)\\
		&\le 6\left( \sup_{\vv{x}\in [0, 1]^p}|m(\vv{x})|\right) \times \mathbb{P}( |\boldsymbol{1}_{\vv{u}^{\top}\boldsymbol{X} > \vv{u}^{\top}\boldsymbol{X}_{k^{\star}} } - \boldsymbol{1}_{\vv{w}^{\star\top}\boldsymbol{X} > c^{\star}}| = 1 ).
	\end{split}
\end{equation}

Based on \eqref{exponential.prob.8} through \eqref{exponential.prob.10}, we establish the second inequality of \eqref{exponential.prob.9}. Combining \eqref{exponential.prob.20} and \eqref{exponential.prob.9}, the first inequality of \eqref{exponential.prob.23.b} follows.

Lastly, define
$$2Q_{n} = 24\left(\sup_{\vv{x}\in [0, 1]^p}|m(\vv{x})|\right)^2 \times [s^s \times 4 \iota \sqrt{s}D_{\max} + 3\eta_{n} + s^{s+1} \times 4\eta_n D_{\max}],$$ 
which, by that $D_{\max}\ge 1$ as given in (b) of Condition~\ref{regularity.1}, simplifies to
\begin{equation}
    \label{prop2.3}
    Q_{n} \le 12D_{\max}\left(\sup_{\vv{x}\in [0, 1]^p}|m(\vv{x})|\right)^2 \times \left[4s^{s + \frac{1}{2}} \iota + (4s^{s+1} +3 ) \eta_{n}\right].
\end{equation}

By \eqref{exponential.prob.40}, \eqref{exponential.prob.23.b.x}--\eqref{exponential.prob.23.b}, and \eqref{prop2.3}, we have compteled the proof of \eqref{exponential.prob.23}.

% Then, it follows that
% \begin{equation}
%     \begin{split}\label{exponential.prob.41}
%         & \inf_{(\vv{w}^{'}, c^{'}) \in \cup_{q=1}^{B}\Lambda_{q}} \mathbb{E}[\textnormal{Var}(m(\boldsymbol{X})| \boldsymbol{1}_{\boldsymbol{X} \in \boldsymbol{t}^{'}}, \boldsymbol{1}_{\vv{w}^{'\top}\boldsymbol{X} > c^{'}}) \boldsymbol{1}_{\boldsymbol{X} \in \boldsymbol{t}^{'}}]  + 2\rho_{n}\\
%         & \le  \mathbb{E}[\textnormal{Var}(m(\boldsymbol{X})| \boldsymbol{1}_{\boldsymbol{X} \in \boldsymbol{t}^{'}}, \boldsymbol{1}_{\vv{w}^{\star\top}\boldsymbol{X} > c^{\star}}) \boldsymbol{1}_{\boldsymbol{X} \in \boldsymbol{t}^{'}}] + 2 Q_{n} + 2\rho_{n},
%     \end{split}
% \end{equation}
% which in combination with \eqref{exponential.prob.40} shows that 

\subsubsection{Proof of \texorpdfstring{\eqref{prob.upper.1}}{LG}}\label{proof.prob.upper.1}

To establish the bound for $\mathbb{P}(\mathcal{D}^c)$, we follow a geometric covering argument combined with a concentration inequality. We first define the probability of sampling a vector with the target sparsity. Based on Condition~\ref{sampling.1} and given $s \ge s_0$, a vector $\vv{w}$ randomly sampled from $\mathcal{J}(s)$ satisfies:
\begin{equation}
\label{D.2}
\mathbb{P}(\norm{\vv{w}}_0 = s_0) = s^{-1}.
\end{equation}
Next, we construct a covering set. Let $z_n = s_0^{-1/2} (\iota/2)$ be the edge length of an $s_0$-dimensional hypercube. For any two points $\vv{v}, \vv{w}$ within such a hypercube, the Euclidean distance is bounded by:
\begin{equation}
\label{Is1}
\norm{\vv{w} - \vv{v}}_{2} \le \sqrt{s_0} \times z_n = \frac{\iota}{2}.
\end{equation}

We construct a collection of hypercubes $\mathcal{M}(s_0)$ that covers the set of all $s_0$-sparse vectors in $[-1, 1]^p$. Because $\mathcal{J}(s_0)$ is a subset of these sparse vectors, this collection also serves as a cover for $\mathcal{J}(s_0)$. The number of such hypercubes is bounded by $\texttt{\#}\mathcal{M}(s_0) \le \binom{p}{s_0} \times 2^{s_0} z_n^{-s_0}$. From this, we derive a corresponding set of hyperballs $\mathcal{H}$, each with a radius of $\iota/2$ and centered at a point within the intersection of a hypercube and $\mathcal{J}(s_0)$.  By the construction in \eqref{Is1}, each hypercube is entirely contained within its respective hyperball of radius $\iota/2$. This ensures that $\mathcal{J}(s_0) \subseteq \cup_{W \in \mathcal{H}} W$, where the total number of hyperballs satisfies 
\begin{equation}
    \label{D.3}
    \texttt{\#}\mathcal{H} \le \binom{p}{s_0} 2^{s_0} z_n^{-s_0}.
\end{equation}

By \eqref{D.2} and \eqref{D.3}, the probability that a random vector sampled from $\mathcal{J}(s)$ falls into a specific hyperball $W \in \mathcal{H}$ is at least:
\begin{equation}\label{Is6}
P_0 \coloneqq \frac{1}{s} \times \left[\binom{p}{s_0} 2^{s_0} z_n^{-s_0}\right]^{-1}.
\end{equation}

Let $\mathcal{D}'$ be the event that every hyperball $W \in \mathcal{H}$ contains at least one direction from our set of $B$ sampled splits $\Lambda = \{ \vv{w}: (\vv{w}, c)\in \cup_{q=1}^{B} \Lambda_{q} \}$.  If $\mathcal{D}'$ occurs, then for any $\vv{v} \in \mathcal{J}(s_0)$, there exists a $\vv{u} \in \Lambda$ such that both $\vv{v}$ and $\vv{u}$ lie within the same hyperball. Since each hyperball has a radius of $\iota/2$, the distance between any two points within it cannot exceed the diameter $\iota$. Thus, $\norm{\vv{v} - \vv{u}}_2 \le \iota$, which implies $\mathcal{D}' \subseteq \mathcal{D}$. Using the assumption of i.i.d. samples and the inequality $1+x \le e^x$, we bound the probability of the complement:
\begin{equation}
\begin{split}\label{d.upper.2}
\mathbb{P}(\mathcal{D}^c) & \le \mathbb{P}(\mathcal{D}'^c) \\
& \le \texttt{\#}\mathcal{H} \times \left(1 - P_0\right)^B \\
& \le \left[\binom{p}{s_0} 2^{s_0} z_n^{-s_0}\right] \exp\left( -B \times \frac{1}{s \binom{p}{s_0} 2^{s_0} z_n^{-s_0}} \right).
\end{split}
\end{equation}
By substituting $z_n = s_0^{-1/2} \frac{\iota}{2}$ and the condition $B \ge s 2^{2s_0}\binom{p}{s_0} s_0^{s_0/2} \iota^{-s_0} \log(n 2^{2s_0} \binom{p}{s_0} s_0^{s_0/2} \iota^{-s_0})$ in \eqref{d.upper.2}, we have $\mathbb{P}(\mathcal{D}^c) \le n^{-1}$, which concludes the desired result of \eqref{prob.upper.1}.

%%%
%%%
%%%

\subsubsection{Proof of \texorpdfstring{\eqref{discrete.1}}{LG}}\label{proof.discrete.1}

The proof strategy for \eqref{discrete.1} follows a sequence of approximations that bridge the gap between global theory and finite samples. By moving step-by-step from \eqref{lambda.5.b} to \eqref{lambda.20}, we establish the desired upper bound. Specifically, we show that, with high probability, the infimum over the global split set $W_{p, s_0}$ is identical to the infimum over the sample-based set $\widehat{W}_{p, s_0}$. We then demonstrate that the randomly sampled candidate splits $\cup_{q=1}^B \Lambda_{q}$ cover this space with high probability. This allows us to control the variance of the chosen sample split relative to the theoretical optimum via event $\mathcal{B}$, whose complementary probability is bounded by Corollary~\ref{corollary1}.

In the case of discrete features, it is possible to infer that sample equivalent splits are also almost surely equivalent splits at the population level when conditional on the event \(\mathcal{E}_{2}\) as defined below. 
Define 
$$\mathcal{E}_{2} = \bigcap_{J \subseteq \{1, \dots, p\}, \texttt{\#}J = s_0} \big\{ \{a \in \{0, 1\}^{s_0}: (X_{ij}, j\in J) = a \textnormal{ for some } i\in \{1, \dots, n\}\} = \{0, 1\}^{s_0} \big\}.$$

On the event $\mathcal{E}_{2}$, for each \(\vv{r} \in \mathcal{J}(s_0)\), splits  other than those in \(\{(\vv{r}, \vv{r}^\top \boldsymbol{X}_i): i\in \{1,\dots, n\}\}\) are equivalent splits to these splits. Specifically, for each \(\vv{r} \in \mathcal{J}(s_0)\) and each $g\in\mathbb{R}$, on $\mathcal{E}_2$, there is some integer variable $1\le \widehat{i} \le n$  depending on sample $\mathcal{X}_n$ and some $c_0\in \{0 , 1\}$  such that
\begin{equation}
    \label{equivalencet.1}
    \mathbb{P}( \boldsymbol{1}_{\vv{r}^{\top}{\boldsymbol{X}} > g} - \boldsymbol{1}_{\vv{r}^{\top}{\boldsymbol{X}} > \vv{r}^{\top}{\boldsymbol{X}_{\widehat{i}}} } =c_0 \mid \mathcal{X}_n ) =1. 
\end{equation}
Note that when \eqref{equivalencet.1} holds with $c_{0} = 1$, it holds that $\mathbb{P}( \boldsymbol{1}_{\vv{r}^{\top}{\boldsymbol{X}} > g} - \boldsymbol{1}_{\vv{r}^{\top}{\boldsymbol{X}} \le \vv{r}^{\top}{\boldsymbol{X}_{\widehat{i}}} } = 0 \mid \mathcal{X}_n ) =1$. Therefore, on $\mathcal{E}_{2}$, we have that
\begin{equation}
\begin{split}
    \label{lambda.5.b}    
    & \inf_{(\vv{w}^{'}, c^{'}) \in \widehat{W}_{p, s_0} } \mathbb{E}[\textnormal{Var}(m(\boldsymbol{X})| \boldsymbol{1}_{\boldsymbol{X} \in \boldsymbol{t}^{'}}, \boldsymbol{1}_{\vv{w}^{'\top}\boldsymbol{X}  > c^{'}} ) \boldsymbol{1}_{\boldsymbol{X} \in \boldsymbol{t}^{'}}] \\
      & = \inf_{(\vv{w}^{'}, c^{'}) \in W_{p, s_0} } \mathbb{E}[\textnormal{Var}( m(\boldsymbol{X})| \boldsymbol{1}_{\boldsymbol{X} \in \boldsymbol{t}^{'}}, \boldsymbol{1}_{\vv{w}^{'\top}\boldsymbol{X}  > c^{'}} ) \boldsymbol{1}_{\boldsymbol{X} \in \boldsymbol{t}^{'}}] .
      \end{split}
\end{equation}

Equation \eqref{lambda.5.b} bridges the global theory to the sample set $\widehat{W}_{p, s_0}$, which we further approximate using the candidate splits $\Lambda_q$. To facilitate this, we define the set of unique representative splits $\mathcal{K}^{\star}$ such that $\mathcal{K}^{\star} \subseteq \widehat{W}_{p, s_0}$ and $\widehat{W}_{p, s_0} = \bigcup_{(\vv{v}, a) \in \mathcal{K}^{\star}} \mathcal{Q}_{n}(\vv{v}, a)$. The relevant notations are summarized below:
\begin{align*}
\mathcal{Q}_{n}(\vv{r}, g) & = \{(\vv{u}, b) \in \widehat{W}_{p, s_0} : \sum_{i=1}^n |\boldsymbol{1}_{\vv{r}^{\top} \boldsymbol{X}_{i} > g} - \boldsymbol{1}_{\vv{u}^{\top} \boldsymbol{X}_{i} > b}| \in \{0, n\}\},\\    
\widehat{W}_{p, s_0} & = \{(\vv{r}, \vv{r}^{\top}\boldsymbol{X}_{i}) : i \in \{1, \dots, n\}, \vv{r} \in \mathcal{J}(s_0)\},\\
\Lambda_{q} & = \left\{ (\vv{w}_{q}, \vv{w}_{q}^{\top} \boldsymbol{X}_{i}) : i \in \{1, \dots, n\} \right\},
\end{align*}
where \(\vv{w}_{q}\) is sampled randomly from \(\mathcal{J}(s)\). Note that these notations follow those previously introduced and used in earlier sections.

We observe that on $\mathcal{E}_{2}$, two sample-equivalent splits $(\vv{u}, a)$ and $(\vv{r}, g)$ in \(\mathcal{Q}_{n}(\vv{v}, b)\) for some $\vv{v}\in \mathcal{J}(s_0)$ and $ b\in \mathbb{R}$ satisfy \(\sum_{i=1}^{n}|\boldsymbol{1}_{\vv{r}^{\top}\boldsymbol{X}_{i} > g} - \boldsymbol{1}_{\vv{u}^{\top}\boldsymbol{X}_{i} > a}| \in \{0, n\}\) if and only if they are equivalent in the sense that one of the following conditions holds: (i) \(\boldsymbol{1}_{\vv{r}^{\top}\boldsymbol{X} > g} = \boldsymbol{1}_{\vv{u}^{\top}\boldsymbol{X} > a}\) almost surely or (ii) \(\boldsymbol{1}_{\vv{r}^{\top}\boldsymbol{X} > g} = \boldsymbol{1}_{\vv{u}^{\top}\boldsymbol{X} \le a}\) almost surely. By this observation and the definition of $\mathcal{K}^{\star}$, we have that  on $\mathcal{E}_{2}$,
\begin{equation}
\begin{split}
    \label{lambda.5}
    & \inf_{(\vv{w}^{'}, c^{'}) \in \mathcal{K}^{\star} } \mathbb{E}[\textnormal{Var}( m(\boldsymbol{X})| \boldsymbol{1}_{\boldsymbol{X} \in \boldsymbol{t}^{'}}, \boldsymbol{1}_{\vv{w}^{'\top}\boldsymbol{X}  > c^{'}} ) \boldsymbol{1}_{\boldsymbol{X} \in \boldsymbol{t}^{'}}] \\
      & = \inf_{(\vv{w}^{'}, c^{'}) \in \widehat{W}_{p, s_0} } \mathbb{E}[\textnormal{Var}(m(\boldsymbol{X})| \boldsymbol{1}_{\boldsymbol{X} \in \boldsymbol{t}^{'}}, \boldsymbol{1}_{\vv{w}^{'\top}\boldsymbol{X}  > c^{'}} ) \boldsymbol{1}_{\boldsymbol{X} \in \boldsymbol{t}^{'}}] .
      \end{split}
\end{equation}

Let us define
$$\mathcal{E}_{1} = \cap_{(\vv{r}, g)\in \mathcal{K}^{\star}} \{ \mathcal{Q}_{n}(\vv{r}, g) \cap (\cup_{q=1}^B \Lambda_{q}) \not=\emptyset\}.$$
By similar arguments to those for \eqref{lambda.5}, it holds on $\mathcal{E}_{1}\cap \mathcal{E}_{2}$ that 
\begin{equation}
\begin{split}
    \label{lambda.6}
    & \inf_{(\vv{w}^{'}, c^{'}) \in \cup_{q=1}^B \Lambda_{q} } \mathbb{E}[\textnormal{Var}( m(\boldsymbol{X})| \boldsymbol{1}_{\boldsymbol{X} \in \boldsymbol{t}^{'}}, \boldsymbol{1}_{\vv{w}^{'\top}\boldsymbol{X}  > c^{'}} ) \boldsymbol{1}_{\boldsymbol{X} \in \boldsymbol{t}^{'}}] \\
      & \le \inf_{(\vv{w}^{'}, c^{'}) \in \mathcal{K}^{\star} } \mathbb{E}[\textnormal{Var}(m(\boldsymbol{X})| \boldsymbol{1}_{\boldsymbol{X} \in \boldsymbol{t}^{'}}, \boldsymbol{1}_{\vv{w}^{'\top}\boldsymbol{X}  > c^{'}} ) \boldsymbol{1}_{\boldsymbol{X} \in \boldsymbol{t}^{'}}] .
    \end{split}
\end{equation}

 Next, based on the definition of $\mathcal{B}$ provided before Corollary~\ref{corollary1}, it holds on $\mathcal{B}$ that
\begin{equation}
\begin{split}   \label{lambda.20}
      & \mathbb{E}[\textnormal{Var}(m(\boldsymbol{X})| \boldsymbol{1}_{\boldsymbol{X} \in \boldsymbol{t}^{'}}, \boldsymbol{1}_{\vv{w}^{\top}\boldsymbol{X}  > c} ) \boldsymbol{1}_{\boldsymbol{X} \in \boldsymbol{t}^{'}}] \\
      & \le \inf_{(\vv{w}^{'}, c^{'}) \in \cup_{q=1}^B \Lambda_{q} } \mathbb{E}[\textnormal{Var}( m(\boldsymbol{X})| \boldsymbol{1}_{\boldsymbol{X} \in \boldsymbol{t}^{'}}, \boldsymbol{1}_{\vv{w}^{'\top}\boldsymbol{X}  > c^{'}} ) \boldsymbol{1}_{\boldsymbol{X} \in \boldsymbol{t}^{'}}]  + 2\rho_{n}.
      \end{split}
 \end{equation}
By combining \eqref{lambda.5.b}--\eqref{lambda.20} and setting 
$$\mathcal{G} = \mathcal{E}_{1}\cap \mathcal{E}_{2},$$ 
the proof of the first part of \eqref{discrete.1} is complete.

We now proceed to establish the probability upper bound on $\mathbb{P}(\mathcal{G}^c)$. In light of that $1\le p\le n^{K_{0}}$ and $s_0$ is a constant, it holds that for all large $n$,
\begin{equation}
    \label{mathcal.2}
    \begin{split}        
    \mathbb{P}(\mathcal{E}_{2}) & \ge 1 - \binom{p}{s_0} \times 2^{s_0} \times (1-2^{-s_0} \times D_{\min})^n\\
    & \ge 1 - p^{s_0} \times 2^{s_0} \times \exp{(-2^{-s_0} D_{\min} n)}\\
    & \ge 1 - \frac{1}{2n},
    \end{split}
\end{equation}
where $D_{\min}$ is defined by (a) of Condition~\ref{regularity.1}, and the second inequality follows from that $1 + x \le e^{x}$ for every $x\in \mathbb{R}$. Here $(1-2^{-s_0} \times D_{\min})^n$ represent the probability upper bound on the event $\{(X_{ij}, j\in J) \not= a \textnormal{ for all } i\}$ for some $a\in \{0, 1\}^{s_0}$ and some $J\subseteq \{1, \dots, p\}$ with $\texttt{\#}J = s_0$, while $2^{s_0} = \texttt{\#}\{0, 1\}^{s_0}$. The last inequality holds for all large $n$ because $D_{\min} > 0$ and $s_0$ are assumed to be constant.

Next, we deal with establishing a probability lower bound for $\mathbb{P}(\mathcal{E}_{1})$.  We need the following technical Lemma~\ref{volume2}, whose proof is provided in Section~\ref{proof.lemma.volume2}.
\begin{lemma}\label{volume2}
	There are \( L_{s_0} \) distinct ways to separate the vertices of an \( s_0 \)-dimensional hypercube using a hyperplane with a normal vector from the \( s_0 \)-dimensional unit sphere \( U_{s_0} = \{\vv{w} \in \mathbb{R}^{s_0} : \norm{\vv{w}}_{2} = 1\} \) and a bias from the real line. For each $l\in \{1, \dots, L_{s_0}\}$, the $l$th separation  partitions the vertices \( \{0, 1\}^{s_0} \) into two sets, denoted by \( V_l \) and \( \{0, 1\}^{s_0} \backslash V_l \). It holds that  
	\( L_{s_0} \leq 2^{(s_0+1)^2} \), and  that for every $l\in \{1,\dots, L_{s_0}\}$,
	{\small \[
		\textnormal{Volume of } \{ \vv{u} \in U_{s_0} : \{ \vv{v} \in \{0,1\}^{s_0} : \vv{u}^{\top} \vv{v} - b > 0 \} = V_l \text{ for some } b \in \mathbb{R} \}  
		\geq c_{s_0} \times (\textnormal{Volume of } U_{s_0})
		\]  }%
	for some constant \( c_{s_0} > 0 \) that depends on \( s_0 \).
	
\end{lemma}

By \eqref{D.2} and an application of Lemma~\ref{volume2}, there exists some constant $C_{s_0}$ depending on $s_0$ such that on $\mathcal{E}_{2}$,
\begin{equation}
	\begin{split}\label{lambda.3.b}
		& \min_{(\vv{r}, g) \in \mathcal{K}^{\star}} \mathbb{P}(\mathcal{Q}_{n}(\vv{r}, g) \cap \Lambda_{1} \neq \emptyset \mid \mathcal{X}_{n}) \ge C_{s_0} \times \left[ s\times \binom{p}{s_0} \right]^{-1}.
		\end{split}
\end{equation}  
Here, the result of \eqref{lambda.3.b} results from the following three steps of  sampling some $\vv{u}$ from $\mathcal{J}(s_0)$.  
(1) Sample an integer from $\{1,\dots, s\}$ that determines the number of nonzero elements of \( \vv{u} \), which should be \( s_0 \). According to Condition~\ref{sampling.1} and \eqref{D.2}, the probability of achieving this is \( s^{-1} \).  
(2) Sample a set of \( s_0 \) coordinates, each corresponding to a nonzero element of the split weight vector \( \vv{u} \). This set must satisfy \( \{j : |r_{j}| > 0\} = \{ j : |u_{j}| > 0\} \). The probability of achieving this is \( [\binom{p}{s_0}]^{-1} \).  
(3) Let \( \vv{x}_{+} \) denote the subvector of nonzero elements of \( \vv{x} \). The pairs \( (\vv{u}_{+}, b) \) and \( (\vv{r}_{+}, g) \) must separate the vertices of an \( s_0 \)-dimensional hypercube identically for some \( b \in \mathbb{R} \). According to Lemma~\ref{volume2}, the probability of achieving this is \( c_{s_0} > 0 \).  With these facts and some simple calculations, we obtain \eqref{lambda.3.b}.

By \eqref{lambda.3.b} and that the normal vectors in $\Lambda_{q}$'s are i.i.d., it holds on $\mathcal{E}_2$ that
$$\min_{(\vv{r}, g) \in \mathcal{K}^{\star}} \mathbb{P}(\mathcal{Q}_{n}(\vv{r}, g) \cap (\cup_{q=1}^B \Lambda_{q}) =\emptyset\mid \mathcal{X}_{n}) \le(1 - \frac{C_{s_0}}{ s\times \binom{p}{s_0}})^{B}, $$
which implies that for all large $n$, on $\mathcal{E}_2$,
\begin{equation}
	\label{lambda.4}
	\begin{split}
		\mathbb{P}( \mathcal{E}_{1}^c \mid \mathcal{X}_{n}) &\le \texttt{\#}\mathcal{K}^{\star} \times \left(1- \frac{C_{s_0}}{ s\times  \binom{p}{s_0}}\right)^{B} \\        
		& \le  \binom{p}{s_0} \times 2^{(s_0+1)^2} \times \left(1- \frac{C_{s_0}}{ s\times  \binom{p}{s_0}}\right)^{B} \\
		& \le  p^{s_0} \times 2^{(s_0+1)^2} \times  \exp{\left(\frac{-B \times C_{s_0}}{ s\times  \binom{p}{s_0}} \right)} \\
		& \le  \frac{1}{2n},
	\end{split}
\end{equation}
where the second inequality is an application of Lemma~\ref{volume2} and the fact that we are considering $s_0$-sparse $p$-dimensional unit-length vectors, the third inequality follows from that $1+ x \le e^x$ for each $x\in \mathbb{R}$, and the fourth inequality holds because $1\le p\le n^{K_{0}}$, $s$ is a constant, and that $B\ge C_{3}s(s_0\log{n})^2 \binom{p}{s_0}$
for some large $C_{3}>0$ depending on  $s_0$.

Using \eqref{mathcal.2}, \eqref{lambda.4}, and \eqref{lambda.5.b}--\eqref{lambda.20} with $\mathcal{G} = \mathcal{E}_1\cap \mathcal{E}_2$, we establish the desired result in \eqref{discrete.1}, thereby completing the proof.

%%%
%%%%%%%%
%%%%

%%
%%
%%
\renewcommand{\thesubsection}{C.\arabic{subsection}}
\setcounter{equation}{0}
\renewcommand\theequation{C.\arabic{equation}}
\section{Proofs of Propositions, Corollaries, and Examples} \label{SecC}

\subsection{Proof of Proposition~\ref{prop1}}\label{proof.prop1}

    We begin by proving the first assertion of Proposition~\ref{prop1}, where $b_1 \ge 1$ and $b$ are arbitrary integers such that $b \ge b_1$. We will prove the desired result by induction; recall that \eqref{tspursuit.4} breaks ties randomly.

    1.) Suppose for some $q\in \{1, \dots, h\}$ and some $(\boldsymbol{t}^{'}, (\vv{w}, c), \boldsymbol{t}) \in \widehat{N}_{q}^{(b_{1}+1)}$, it holds that $\sum_{i=1}^{n} |\boldsymbol{1}_{\boldsymbol{X}_{i} \in \boldsymbol{t}^{'}} - \boldsymbol{1}_{\boldsymbol{X}_{i} \in \boldsymbol{s}^{'}}| = 0$ for some $(\boldsymbol{s}^{'}, (\vv{u}, a), \boldsymbol{s}) \in \widehat{N}_{q}^{(b_{1})}$. In what follows, we prove that $\sum_{i=1}^{n} |\boldsymbol{1}_{\boldsymbol{X}_{i} \in \boldsymbol{t}} - \boldsymbol{1}_{\boldsymbol{X}_{i} \in \boldsymbol{s}}| \in \{ 0, \sum_{i=1}^{n} \boldsymbol{1}_{\boldsymbol{X}_{i} \in \boldsymbol{t}^{'}} \}$ on $\Theta_{b_1, h}$ in this scenario, which implies that there exists some  $(\boldsymbol{s}^{'}, (\vv{u}, a), \boldsymbol{s}_{2}) \in \widehat{N}_{q}^{(b_{1})}$ such that $\sum_{i=1} |\boldsymbol{1}_{\boldsymbol{X}_{i} \in \boldsymbol{t}} - \boldsymbol{1}_{\boldsymbol{X}_{i} \in \boldsymbol{s}_{2}}| = 0$ on $\Theta_{b_1, h}$.

    Note that on $\Theta_{b_1, h}$, the split $(\vv{u}, a)$ is sample optimal among \(\bigcup_{q=1}^{B} \Lambda_{q}\) conditional on the subsample $\{\boldsymbol{X}_{i} \in \boldsymbol{t}^{'}\}$. Other splits from the available random set of splits \( W_{b_1 + 1} \) at the $(b_{1}+1)$th iteration (see Step (i) of Section~\ref{sec2.2.1}) are either inferior to $(\vv{u}, a)$ in terms of reducing the sample $\mathbb{L}^2$ loss or sample-equivalent to them, in the  sense of Condition~\ref{regular.tree}. If these splits are inferior than the optimal ones, they will not replace the  optimal split $(\vv{u}, a)$ from the previous $b_{1}$th iteration, and therefore $(\vv{w}, c) = (\vv{u}, a)$ and $\sum_{i=1}^{n} |\boldsymbol{1}_{\boldsymbol{X}_{i} \in \boldsymbol{t}} - \boldsymbol{1}_{\boldsymbol{X}_{i} \in \boldsymbol{s}}| = 0$. If they are sample-equivalent to $(\vv{u}, a)$ and that $(\vv{u}, a)$ is replaced with $(\vv{w}, c)$ at the $(b_{1} + 1)$th iteration, then it still holds that $\sum_{i=1}^{n} |\boldsymbol{1}_{\boldsymbol{X}_{i} \in \boldsymbol{t}} - \boldsymbol{1}_{\boldsymbol{X}_{i} \in \boldsymbol{s}}| \in \{ 0, \sum_{i=1}^{n} \boldsymbol{1}_{\boldsymbol{X}_{i} \in \boldsymbol{t}^{'}} \}$ due to Condition~\ref{regular.tree}. We conclude that $\sum_{i=1}^{n} |\boldsymbol{1}_{\boldsymbol{X}_{i} \in \boldsymbol{t}} - \boldsymbol{1}_{\boldsymbol{X}_{i} \in \boldsymbol{s}}| \in \{ 0, \sum_{i=1}^{n} \boldsymbol{1}_{\boldsymbol{X}_{i} \in \boldsymbol{t}^{'}} \}$ on $\Theta_{b_1, h}$ in this scenario.
    
2.) For $q = 1$, it holds that $\boldsymbol{t}^{'} = [0, 1]^p$ for each $(\boldsymbol{t}^{'}, (\vv{w}, c), \boldsymbol{t}) \in \widehat{N}_{q}^{(b_{1}+1)}$. Therefore, we have that for each $(\boldsymbol{t}^{'}, (\vv{w}, c), \boldsymbol{t}) \in \widehat{N}_{1}^{(b_{1}+1)}$, it holds that $\sum_{i=1}^{n} |\boldsymbol{1}_{\boldsymbol{X}_{i} \in \boldsymbol{t}^{'}} - \boldsymbol{1}_{\boldsymbol{X}_{i} \in \boldsymbol{s}^{'}}| = 0$ for some $(\boldsymbol{s}^{'}, (\vv{u}, a), \boldsymbol{s}) \in \widehat{N}_{1}^{(b_{1})}$. Then, by the arguments in 1), we conclude that $\widehat{N}_{1}^{(b_{1}+1)}$ and $\widehat{N}_{1}^{(b_{1})}$ are sample-equivalent on $\Theta_{b_1, h}$.

    3.) Next, suppose that for some \( q \in \{1, \dots, h\} \), the following holds: for each \( (\boldsymbol{t}^{'}, (\vv{w}, c), \boldsymbol{t}) \in \widehat{N}_{q}^{(b_{1}+1)} \), there exists \( (\boldsymbol{s}^{'}, (\vv{u}, a), \boldsymbol{s}) \in \widehat{N}_{q}^{(b_{1})} \) such that  
\[
\sum_{i=1}^{n} |\boldsymbol{1}_{\boldsymbol{X}_{i} \in \boldsymbol{t}^{'}} - \boldsymbol{1}_{\boldsymbol{X}_{i} \in \boldsymbol{s}^{'}}| = 0.
\]
Then, by the arguments in 1), it follows that on \( \Theta_{b_1, h} \), for each \( (\boldsymbol{t}^{'}, (\vv{w}, c), \boldsymbol{t}) \in \widehat{N}_{q}^{(b_{1}+1)} \), we have  
\[
\sum_{i=1}^{n} |\boldsymbol{1}_{\boldsymbol{X}_{i} \in \boldsymbol{t}^{'}} - \boldsymbol{1}_{\boldsymbol{X}_{i} \in \boldsymbol{s}^{'}}| = \sum_{i=1}^{n} |\boldsymbol{1}_{\boldsymbol{X}_{i} \in \boldsymbol{t}} - \boldsymbol{1}_{\boldsymbol{X}_{i} \in \boldsymbol{s}}| = 0
\]
for some \( (\boldsymbol{s}^{'}, (\vv{u}, a), \boldsymbol{s}) \in \widehat{N}_{q}^{(b_{1})} \). This implies that \( \widehat{N}_{q}^{(b_{1}+1)} \) and \( \widehat{N}_{q}^{(b_{1})} \) are sample-equivalent.

    4.) Furthermore, if, for some \( q \in \{1, \dots, h - 1\} \), the following holds: for each \( (\boldsymbol{t}^{'}, (\vv{w}, c), \boldsymbol{t}) \in \widehat{N}_{q}^{(b_{1}+1)} \), there exists \( (\boldsymbol{s}^{'}, (\vv{u}, a), \boldsymbol{s}) \in \widehat{N}_{q}^{(b_{1})} \) such that  
\[
\sum_{i=1}^{n} |\boldsymbol{1}_{\boldsymbol{X}_{i} \in \boldsymbol{t}} - \boldsymbol{1}_{\boldsymbol{X}_{i} \in \boldsymbol{s}}| = 0.
\]
Then, it follows that for each \( (\boldsymbol{t}^{'}, (\vv{w}, c), \boldsymbol{t}) \in \widehat{N}_{q+1}^{(b_{1}+1)} \), there exists \( (\boldsymbol{s}^{'}, (\vv{u}, a), \boldsymbol{s}) \in \widehat{N}_{q+1}^{(b_{1})} \) such that  
\[
\sum_{i=1}^{n} |\boldsymbol{1}_{\boldsymbol{X}_{i} \in \boldsymbol{t}^{'}} - \boldsymbol{1}_{\boldsymbol{X}_{i} \in \boldsymbol{s}^{'}}| = 0.
\]
    
    We begin with 2.) to demonstrate that \(\widehat{N}_{1}^{(b_{1}+1)}\) and \(\widehat{N}_{1}^{(b_{1})}\) are sample-equivalent on \( \Theta_{b_1, h} \) with $h\ge 1$. Then, by applying point 4.) with $q=1$ and point 3.) with $q=2$, we show that \(\widehat{N}_{2}^{(b_{1}+1)}\) and \(\widehat{N}_{2}^{(b_{1})}\) are also sample-equivalent on \( \Theta_{b_1, h} \). The remainder follows by induction to show that \(\widehat{N}_{q}^{(b_{1}+1)}\) and \(\widehat{N}_{q}^{(b_{1})}\) are sample-equivalent on \( \Theta_{b_1, h} \) for $q\in \{1, \dots, h\}$. But this implies that \(\widehat{N}_{q}^{(b_{1}+1)}\) and \(\widehat{N}_{q}^{(0)}\) are sample-equivalent on \( \Theta_{b_1, h} \) for $q\in \{1, \dots, h\}$ due to the definition of $\Theta_{b_1, h}$. 
    By induction again, we have shown that \(\widehat{N}_{q}^{(b_{1}+k)}\) and \(\widehat{N}_{q}^{(0)}\) are sample-equivalent on \( \Theta_{b_1, h} \) for $q\in \{1, \dots, h\}$ for each $k \ge 0$. Thus, we have established the desired result for the first assertion of Proposition~\ref{prop1}.
    
    Next, let us prove the second assertion of Proposition~\ref{prop1}.   Suppose $\widehat{N}_{s}^{(b)}$ and $\widehat{N}_{s}^{(0)}$ are sample-equivalent for each $s\in \{1, \dots, q\}$ for some $q\in \{1, \dots, h-1\}$. Then, for each $(\boldsymbol{t}^{'}, (\vv{w}, c), \boldsymbol{t})\in \widehat{N}_{q}^{(b)}$, there is a corresponding $(\boldsymbol{s}^{'}, (\vv{u}, a), \boldsymbol{s})\in \widehat{N}_{q+1}^{(0)}$ with $\sum_{i=1} |\boldsymbol{1}_{\boldsymbol{X}_{i} \in \boldsymbol{t}} - \boldsymbol{1}_{\boldsymbol{X}_{i} \in \boldsymbol{s}^{'}}| = 0$. If $(\vv{u}, a)\in W_{b+k}$ at the $(b+k)$th iteration (See Step (i) in Section~\ref{sec2.2.1} for notation details) for some $k\ge 1$, then the sample optimal split $(\vv{u}, a)$ or its sample equivalences will be selected and maintained by $\widehat{N}_{q+1}^{(b+k + l)}$ for every $l\ge 0$, according to the arguments in 1.) above.

    Since $H$ is given and that each $W_{l}$ is a random subset of $\bigcup_{q=1}^{B} \Lambda_{q}$ at each iteration with $\texttt{\#}W_{l} = n \times \texttt{\#}S$ and $\texttt{\#}S \ge 1$, the probability that $\widehat{N}_{q+1}^{(b)}$ and $\widehat{N}_{q+1}^{(0)}$ are not sample-equivalent for some $q\in \{1, \dots, H-1\}$, which implies that the sample optimal splits are not captured within the candidate sets until the $b$th iteration, can be arbitrarily small with a sufficiently large $b$ in this scenario. Therefore, we conclude by induction the desired result of the second assertion of Proposition~\ref{prop1}. We have finished the proof of Proposition~\ref{prop1}.

    \subsection{Proof of Proposition~\ref{prop3}}\label{proof.prop3}

    Despite the infinite possible splits in $\widehat{W}_{p, s} = \{(\vv{w}, \vv{w}^{\top}\boldsymbol{X}_{i}): i\in \{1, \dots, n\}, \vv{w}\in \mathcal{J}(s)\}$ where $\mathcal{J}(s) = \{\vv{w}\in\mathbb{R}^p: \norm{\vv{w}}_{2} = 1, \norm{\vv{w}}_{0} \le s \}$, only a finite number of splits must be evaluated for a given training sample $\mathcal{X}_{n} = \{\boldsymbol{X}_{i}, Y_{i}\}_{i=1}^{n}$. This is because the distinct ways of partitioning a sample into two subsamples are finite.

Therefore, we let $\mathcal{K}^{\star} = \{(\vv{w}_{q}^{\star}, c_{q}^{\star})\}_{q=1}^{L^{\star}} \subseteq \widehat{W}_{p, s}$ be a finite set of splits for some sample-dependent constant $L^{\star}>0$ (train sample is given and fixed here). This set is constructed such that every possible split $(\vv{w}, c) \in \widehat{W}_{p, s}$ is sample-equivalent to some element in $\mathcal{K}^{\star}$. Specifically, the set of all splits equivalent to some $(\vv{v}, a)$ is defined as
$$\mathcal{Q}_{n}(\vv{v}, a) = \left\{(\vv{u}, r) \in \widehat{W}_{p, s} : \sum_{i=1}^n |\boldsymbol{1}_{\vv{v}^{\top} \boldsymbol{X}_{i} > a} - \boldsymbol{1}_{\vv{u}^{\top} \boldsymbol{X}_{i} > r}| \in \{ 0, n\}\right\}.$$
It follows that $\widehat{W}_{p, s} = \cup_{(\vv{v}, a) \in \mathcal{K}^{\star}} \mathcal{Q}_{n}(\vv{v}, a)$. Notably, the construction of the representative set $\mathcal{K}^{\star}$ is not unique.

To establish the first assertion of Proposition~\ref{prop3}, we consider the event where the $B$ weight vectors sampled from $\mathcal{J}(s)$ successfully cover the entire representative set $\mathcal{K}^{\star}$. Specifically, we define this event $E_B$ by the condition that the full set of partitions is recovered:
\begin{equation}\label{e.approching.one.2}
E_{B} = \left\{\widehat{W}_{p, s} = \bigcup_{(\vv{v}, a) \in \bigcup_{q=1}^{B} \Lambda_q} \mathcal{Q}_{n}(\vv{v}, a)\right\}.
\end{equation}
When $E_B$ occurs, the union $\bigcup_{q=1}^{B} \Lambda_q$ is sufficiently comprehensive to recover the full set of possible partitions. Consequently, the one-shot tree constructed from the candidate split set $\mathcal{W} = \bigcup_{q=1}^{B} \Lambda_q$ is exactly equivalent to its corresponding ideal tree, which utilizes the full set $\mathcal{W} = \widehat{W}_{p, s}$. Therefore, the proof for the one-shot tree reduces to showing that the probability of event $E_B$ approaches one as $B \to \infty$, as demonstrated in \eqref{e.approching.one} below.

On the other hand, since the tree depth $H$ is a finite constant, we can without loss of generality define the full set of available oblique splits for the Breiman oblique tree as $\bigcup_{q=1}^{B(2^{H}-1)} \Lambda_q$. Here, the subset $\bigcup_{q=B(l-1) + 1}^{Bl} \Lambda_q$ denotes the candidate splits assigned to the $l$-th node, where $l \in \{1, \dots, 2^{H}-1\}$. Now, consider the intersection of events $\bigcap_{l=1}^{2^{H}-1} E_{l, B}$, where
$$E_{l, B} = \left\{\widehat{W}_{p, s} = \bigcup_{(\vv{v}, a) \in \bigcup_{q=B(l-1) + 1}^{Bl} \Lambda_q} \mathcal{Q}_{n}(\vv{v}, a)\right\}.$$
When this intersection of events occurs, the Breiman oblique tree becomes exactly equivalent to the ideal tree.

Because $H$ is finite and $B$ tends to infinity, the constant factor $2^{H}$ does not affect the limit. Therefore, to prove the first part of Proposition~\ref{prop3} regarding both the Breiman oblique tree and our one-shot tree, it suffices to show that
\begin{equation}\label{e.approching.one}
\lim_{B\rightarrow \infty}\mathbb{P}(E_{B} \mid \mathcal{X}_n) = 1.
\end{equation}

To show \eqref{e.approching.one}, given that $\texttt{\#}\mathcal{K}^{\star}$ is finite and the sets $\Lambda_q$ are independently and identically distributed, it is sufficient that
\begin{equation}\label{prop3.2}
\min_{(\vv{w}, c) \in \mathcal{K}^{\star} }\mathbb{P}(\mathcal{Q}_{n}(\vv{w}, c) \cap \Lambda_1 \neq \emptyset \mid \mathcal{X}_n) \ge q_0,
\end{equation}
for some generic constant $q_0>0$ that depends on $p$, $n$, and $s$, which are given and fixed in this scenario. In what follows, we establish \eqref{prop3.2}.

For each $(\vv{w}, c) \in \mathcal{K}^{\star}$, define the margin $d_{\vv{w}, c} = \min_{i: \vv{w}^{\top}\boldsymbol{X}_i \not= c} |\vv{w}^{\top}\boldsymbol{X}_i - c|$ if $\texttt{\#}\{\vv{w}^{\top}\boldsymbol{X}_i \not= c\} > 0$, and otherwise $d_{\vv{w}, c} = 1$. Since $\mathcal{X}_n$ is given and $\mathcal{K}^{\star}$ is finite, there exists a constant $\delta > 0$ such that $\min_{(\vv{w}, c) \in \mathcal{K}^{\star}} d_{\vv{w}, c} > \delta$. Therefore, for any $i\in \{1, \dots, n\}$, if $\vv{w}^{\top} \boldsymbol{X}_i > c$, it follows that
\begin{equation}
\label{prop3.3}
\vv{w}^{\top} \boldsymbol{X}_i > c + \delta.
\end{equation}

Now, let $\vv{u} \in \mathcal{J}(s)$ such that $\norm{\vv{u}- \vv{w}}_{2} \le \frac{\delta}{2\sqrt{p}}$. For any $\vv{x} \in [0, 1]^p$, the Cauchy-Schwarz inequality  ensures 
\begin{equation}
    \label{prop3.4}
    |(\vv{w} - \vv{u})^{\top} \vv{x} | \le \frac{\delta}{2}.
\end{equation}

Without loss of generality, let $c = \vv{w}^{\top} \boldsymbol{X}_{i_0}$ for some $i_0 \in \{1, \dots, n\}$.  If $\vv{w}^{\top} \boldsymbol{X}_i > \vv{w}^{\top} \boldsymbol{X}_{i_0}$, it follows from the assumption that $\boldsymbol{X}_i \in [0, 1]^p$, and \eqref{prop3.3}--\eqref{prop3.4} that:
$$\vv{u}^{\top} \boldsymbol{X}_i \ge \vv{w}^{\top} \boldsymbol{X}_i - \frac{\delta}{2} > \vv{w}^{\top} \boldsymbol{X}_{i_0} + \frac{\delta}{2} \ge  \vv{u}^{\top} \boldsymbol{X}_{i_0},$$
where the first the third inequalities results from \eqref{prop3.4}, while the second inequality holds due to \eqref{prop3.3}. A similar logic applies if $\vv{w}^{\top} \boldsymbol{X}_i < \vv{w}^{\top} \boldsymbol{X}_{i_0}$. Thus, the split induced by $\vv{u}$ is sample-equivalent to $(\vv{w}, c)$ such that $(\vv{u}, \vv{u}^{\top}\boldsymbol{X}_{i_0}) \in \mathcal{Q}_{n}(\vv{w}, c)$. 

Based on the definition of $\Lambda_{q}$ in \eqref{lambda.sampling} and the argument above, we conclude that \eqref{prop3.2} holds. This follows because randomly sampling $\vv{v} \in \mathcal{J}(s)$ such that $\vv{v} \in \{\vv{u} \mid \vv{u} \in \mathcal{J}(s), \norm{\vv{u}- \vv{w}}_{2} \le \frac{\delta}{2\sqrt{p}}\}$ has a strictly positive probability depending on $p ,s, \delta$. This result, combined with the previously established argument, completes the proof for the first part of Proposition~\ref{prop3}.

We now proceed to the second part of Proposition~\ref{prop3}. Under Condition~\ref{regular.tree}, the result follows directly from Proposition~\ref{prop1} and the first part of Proposition~\ref{prop3}. As a remark, under Condition~\ref{regular.tree}, the equivalence between two ideal trees can be defined more strictly such that they must partition the training samples in an identical manner at every node.

We have completed the proof Porposition~\ref{prop3}.

\subsection{Proof of Corollary~\ref{corollary1}}\label{proof.corollary1}

Let us outline the proof strategy. We treat discrete features separately to derive a sharper lower bound for $\texttt{\#}S \times b$. For the continuous case, the key step is to upper bound $\mathbb{P}(\mathcal{B}^c)$ by relating the progressive tree to the one-shot tree. As shown in \eqref{E_W_6} below, we decompose this probability into terms representing the sample-equivalence between the two tree types and the error bounds of the one-shot tree. We then establish individual upper bounds for each term on the right-hand side of \eqref{E_W_6}. The strategy for the discrete case follows a similar logic.

We begin with proving the first case with continuous features. Recall the progressive tree and the one-shot tree defined in Section~\ref{sec2.2.1}, which is reiterated as follows. Progressive tree: After $b$ iterations with $\texttt{\#}S \ge 1$ in Section~\ref{sec2.2.1}, we obtain $\widehat{N}_{1}^{(b)}, \dots, \widehat{N}_{H}^{(b)}$. One-shot tree: Running \eqref{tspursuit.4} with $\mathcal{W} = \bigcup_{q=1}^{B} \Lambda_{q}$, where we obtain $\widehat{N}_{1}^{(0)}, \dots, \widehat{N}_{H}^{(0)}$. As in the proof for Proposition~\ref{prop1}, for each iteration $b\ge 1$ and $h\in \{1, \dots, H\}$, we define the event
$$\Theta_{b, h} = \{ \widehat{N}_{q}^{(0)} \textnormal{ and } \widehat{N}_{q}^{(b)} \textnormal{ are sample-equivalent  for } q\in \{1,\dots, h\} \}. $$

In addition, let us analyze the one-shot tree based on Theorem~\ref{theorem1}. By this theorem, there exists an event denoted by $\mathcal{A}_n$ such that on $\mathcal{A}_n$, it holds that for all large $n$, each $h \in \{1, \dots, H\}$, and every $(\boldsymbol{t}^{'}, (\vv{w}, c), \boldsymbol{t})\in \widehat{N}_{h}^{(0)}$,
    \begin{equation}
        \begin{split}\label{E_W.2}
            (\vv{w}, c) \in & E(\rho_{n}, \boldsymbol{t}^{'}, \cup_{q=1}^{B} \Lambda_{q}),
        \end{split}
    \end{equation}
    where $\rho_{n} = C_{0}  \times n^{-\frac{1}{2}} s \sqrt{H(1\vee\log{s})} \log{n} \times \left(M_{\epsilon} + \sup_{\vv{x} \in [0, 1]^p} |m(\vv{x})|\right)^2$ for sufficiently large $C_{0} >0$  with probability at least $1 - 10\exp{ (- (\log{n})^2 / 2)}$. Then, we have that 
    \begin{equation}
        \label{E_W_6}
        \mathcal{A}_n \cap \Theta_{b, H}\subseteq \mathcal{B},
    \end{equation}
 where that event $\mathcal{B}$ depends on both $n$ and $b$, though this dependence is omitted for brevity.

In light of \eqref{E_W_6}, we deduce that 
\begin{equation}
    \begin{split} \label{B.upper.bound.1}
     \mathbb{P}(\mathcal{B}^c ) & \le \mathbb{P}( (\mathcal{A}_n \cap \Theta_{b, H})^c ) \\
     & =\mathbb{E}[\mathbb{P}( (\mathcal{A}_n \cap \Theta_{b, H})^c \mid \mathcal{X}_n  ) ]\\
    & \le \mathbb{E}[\mathbb{P}(\Theta_{b, H}^c \mid \mathcal{X}_n)]  + \mathbb{P}(\mathcal{A}_n^c).    
    \end{split}
\end{equation}
In the following, we derive an upper bound for the first term on the right-hand side of \eqref{B.upper.bound.1}.

Conditional on the event $\Theta_{b, h-1}$, we allocate $\lceil b \times 2^{-H} \rceil \times 2^{h-1}$ search iterations to identify the optimal splits in $\widehat{N}_{h}^{(0)}$. This allocation ensures that $\widehat{N}_{h}^{(0)}$ and $\widehat{N}_{h}^{(b)}$ are sample-equivalent with high probability. We employ this many iterations at the $h$th layer because there are $2^{h-1}$ nodes to process, with each split receiving a budget of $\lceil b \times 2^{-H} \rceil$ iterations.

Specifically, we define event $F_h$ as the successful identification of all equivalent splits at layer $h$ within the iteration range from $((2^{h-1} - 1) \times \lceil b \times 2^{-H} \rceil + 1)$ to $((2^h - 1) \times \lceil b \times 2^{-H} \rceil)$. Based on the memory transfer property from Proposition~\ref{prop1} and the tree structure in Condition~\ref{regular.tree}, we obtain the following sequence of inclusions:
\begin{equation}
\begin{split}\label{corollary1.2}
    F_{1}\cap F_{2} & \subseteq \Theta_{\lceil b \times 2^{-H} \rceil, 1} \cap F_{2} \\
    & \subseteq \Theta_{\lceil b \times 2^{-H} \rceil \times (2^{0} + 2^{1}), 1} \cap F_{2} \\
    & \subseteq \Theta_{\lceil b \times 2^{-H} \rceil \times (2^{0} + 2^{1}), 2},
\end{split}
\end{equation}
where the first subset operation follows from the definition of $F_{1}$ and $\Theta_{\lceil b \times 2^{-H} \rceil, 1}$ and the memory transfer property of Proposition~\ref{prop1}, the second subset operation holds because of the memory transfer property of Proposition~\ref{prop1}, and the third subset operation results from the definition of $F_{2}$ and $\Theta_{\lceil b \times 2^{-H} \rceil \times (2^{0} + 2^{1}), 2}$ and the memory transfer property of Proposition~\ref{prop1}.

A recursive application of \eqref{corollary1.2} concludes that for each $h\in \{1, \dots, H\}$,
\begin{equation}
	\begin{split}\label{corollary.1.as}
    \cap_{q=1}^h F_{q} & \subseteq \Theta_{\lceil b \times 2^{-H} \rceil \times (2^{h} - 1 ) , h} \\
    & \subseteq \Theta_{b, h}.
	\end{split}
\end{equation}
Define 
$$m_{h} = \lceil b \times 2^{-H} \rceil \times (2^{h} - 1 ).$$ 
The result of \eqref{corollary.1.as} in combination with the law of total expectation implies that almost surely
\begin{equation}
    \begin{split}\label{corollary.1}
   \mathbb{P}(\Theta_{b, H} \mid \mathcal{X}_{n} ) & \ge \mathbb{P}( \cap_{q=1}^H F_{q} \mid \mathcal{X}_{n}) \\
   & = \mathbb{E} \left[    \prod_{q=1}^{H}\boldsymbol{1}_{F_{q}} \ \bigg| \ \mathcal{X}_{n}\right] \\
   & =  \mathbb{E} \left[ \mathbb{E}( \boldsymbol{1}_{F_{H}} |   \boldsymbol{1}_{F_{1}}, \dots, \boldsymbol{1}_{F_{H-1}}, \mathcal{X}_{n} ) \prod_{q=1}^{H-1}\boldsymbol{1}_{F_{q}} \ \bigg| \ \mathcal{X}_{n}\right] \\   
   & =  \mathbb{E} \left[ \mathbb{E}( \boldsymbol{1}_{F_{H}} |   \boldsymbol{1}_{\Theta_{m_{H-1}, H-1}}, \mathcal{X}_{n} )   \prod_{q=1}^{H-1}\boldsymbol{1}_{F_{q}} \ \bigg| \ \mathcal{X}_{n}\right]\\
   & =  \mathbb{E} \left[ \prod_{q=1}^H \mathbb{P}( F_{q} |   \boldsymbol{1}_{\Theta_{m_{q-1}, q-1}}, \mathcal{X}_{n} )   \ \bigg| \ \mathcal{X}_{n}\right],
    \end{split}
\end{equation}
where the third equality follows because $\mathbb{E}( \boldsymbol{1}_{F_{H}} |   \boldsymbol{1}_{F_{1}}, \dots, \boldsymbol{1}_{F_{H-1}}, \mathcal{X}_{n} ) = \mathbb{E}( \boldsymbol{1}_{F_{H}} |   \boldsymbol{1}_{\Theta_{m_{H-1}, H-1}}, \mathcal{X}_{n} )$ on $\cap_{q=1}^{H-1} F_{q}$ due to \eqref{corollary.1.as} and that the random set of available splits in each iteration is independent. The forth equality holds due to a recursive application of the previous equalities. Here, we define $\Theta_{b, 0} = \Omega$ with $\mathbb{P}(\Omega) = 1$, and recall that $\mathcal{X}_n = \{(Y_{i}, \boldsymbol{X}_i)\}_{i=1}^n$.

%%%%%
%%%

%%%%%

To establish a lower bound for each $\mathbb{P}(F_{h} |  \boldsymbol{1}_{\Theta_{m_{h-1}, h-1}}, \mathcal{X}_{n})$ on $\Theta_{m_{h-1}, h-1}$ on the RHS of \eqref{corollary.1}, we have the following observation. Define
\begin{equation}
    \label{uh}
    u_{h} = \min_{ (\boldsymbol{t}^{'}, (\vv{w}, c), \boldsymbol{t}) \in \widehat{N}_{h}^{(0)}  } \sum_{q=1}^{B} \texttt{\#} (  \Lambda_{q} \cap \mathcal{Q}_{n}(\vv{w}, c) ),
\end{equation}
where $\mathcal{Q}_{n}(\vv{w}, c) = \{(\vv{u}, r) \in \widehat{W}_{p, s} : \sum_{i=1}^n |\boldsymbol{1}_{\vv{w}^{\top} \boldsymbol{X}_{i} > c} - \boldsymbol{1}_{\vv{u}^{\top} \boldsymbol{X}_{i} > r}| \in \{ 0, n\}\}$. Now, conditional on the event $\Theta_{m_{h-1}, h-1}$, if the random available split set at some iteration includes $(\vv{w}, c)$ for some $(\boldsymbol{t}^{'}, (\vv{w}, c), \boldsymbol{t}) \in \widehat{N}_{h}^{(0)}$ or its sample equivalences in 
$\{ (\vv{u}, a):\sum_{i=1}^{n} \boldsymbol{1}_{\boldsymbol{X}_{i} \in \boldsymbol{t}^{'}}| \boldsymbol{1}_{\vv{w}^{\top} \boldsymbol{X}_{i} > c} - \boldsymbol{1}_{\vv{u}^{\top} \boldsymbol{X}_{i} > a}| \in \{0, \sum_{i=1}^{n} \boldsymbol{1}_{\boldsymbol{X}_{i} \in \boldsymbol{t}^{'}}\} \}$, then following similar reasoning as for the memory transfer property and Condition~\ref{regular.tree}, we ensure that this sample-optimal split or one of its sample-equivalent counterparts is always preserved in the progressive tree during subsequent iterations. Thus, to establish a lower bound for $\mathbb{P}(F_{h} |  \boldsymbol{1}_{\Theta_{m_{h-1}, h-1}}, \mathcal{X}_{n})$ on $\Theta_{m_{h-1}, h-1}$, it suffices to compute the probability of including each $(\boldsymbol{t}^{'}, (\vv{w}, c), \boldsymbol{t}) \in \widehat{N}_{h}^{(0)}$, or their respective sample-equivalent counterparts, within $\lceil b \times 2^{-H} \rceil \times 2^{h-1}$ iterations.

To this end, by the definition of $u_{h}$ and given the sample $\mathcal{X}_{n}$, we derive that the probability of not including (in the random available splits of size $\texttt{\#}S$) some split $ (\vv{w}, c)$ with $(\boldsymbol{t}^{'}, (\vv{w}, c), \boldsymbol{t}) \in \widehat{N}_{h}^{(0)}$ or its sample equivalences during $ \lceil b \times 2^{-H } \rceil$ iterations is at most
\begin{equation}
	\begin{split}
    \label{corollary.4}
    \left(1 - B^{-1} \right)^{\texttt{\#}S \times  \lceil b \times 2^{-H }\rceil }  & \le   \left(1 -  \frac{\sum_{q=1}^{B} \texttt{\#} (  \Lambda_{q} \cap \mathcal{Q}_{n}(\vv{w}, c) ) } {\sum_{q=1}^{B} \texttt{\#}  \Lambda_{q}} \right)^{\texttt{\#}S \times  \lceil b \times 2^{-H }\rceil }  \\
    & \le \left(1 -  u_{h} \times \left(\sum_{q=1}^{B} \texttt{\#}  \Lambda_{q}\right)^{-1} \right)^{\texttt{\#}S \times  \lceil b \times 2^{-H }\rceil } ,
    \end{split}
\end{equation}
regardless of whether other splits among $\widehat{N}_{h}^{(0)}$ have been included or not. Here, the upper bound on the left-hand side of \eqref{corollary.4} holds because each $\Lambda_{q}$ is constructed by randomly picking up a split from the candidate set $\mathcal{W} = \cup_{q=1}^{B} \Lambda_{q}$ of the one-shot tree; consequently, the probability of selecting the specific split $(\vv{w}, c)$ is at least $B^{-1}$.

Note that there are at most $2^{h-1}$ splits to identify at the $h$th layer of the tree over a total of $\lceil b \times 2^{-H} \rceil \times 2^{h-1}$ iterations. Now, let us enumerate the unique splits in $\widehat{N}_{h}^{(0)}$ as $\{(\vv{w}_{k}, c_{k})\}_{k=1}^{k^{\star}}$ for some $k^{\star} \le 2^{h-1}$. Define an event $\mathcal{U}_{k}$ such that on $\mathcal{U}_{k}$, we select $(\vv{w}_{k}, c_{k})$ during $\lceil b \times 2^{-H} \rceil$ independent searching iterations for each $k\in \{1, \dots, k^{\star}\}$.

Given the upper bound in \eqref{corollary.4} and the construction of events $\mathcal{U}_{k}$, we deduce that on the event $\Theta_{m_{h-1}, h-1}$, the conditional probability of $F_h$ satisfies:
\begin{equation}
    \begin{split}\label{corollary.2}
    & \mathbb{P}(F_{h} |  \boldsymbol{1}_{\Theta_{m_{h-1}, h-1}}, \mathcal{X}_{n})\\   
    & \ge \mathbb{P}(\cap_{k=1}^{k^{\star}} \mathcal{U}_{k} |  \boldsymbol{1}_{\Theta_{m_{h-1}, h-1}}, \mathcal{X}_{n}) \\
  & \ge [1 - \left(1 -  u_{h} \times \left( \sum_{q=1}^{B} \texttt{\#}  \Lambda_{q}  \right)^{-1} \right)^{\texttt{\#}S \times  \lceil b \times 2^{-H }\rceil } ]^{2^{h-1}} \\
  &  \ge 1 - 2^{h-1}\times \left(1 -  u_{h} \times \left(\sum_{q=1}^{B} \texttt{\#}  \Lambda_{q}\right)^{-1} \right)^{\texttt{\#}S \times \lceil b \times 2^{-H }\rceil  }  \\
  & \ge  1 - 2^{h-1}\times \exp{\left( - u_{h} \times  \texttt{\#}S  \times \lceil b \times 2^{-H }\rceil \times  \left(\sum_{q=1}^{B} \texttt{\#}  \Lambda_{q}\right)^{-1}  \right)}\\
   & \eqqcolon \zeta_h,
    \end{split}
\end{equation}
where the second inequality follows because of \eqref{corollary.4} and the definition of $\mathcal{U}_{k}$, the third inequality is due to Bernoulli inequality, and the fourth inequality results from an application of $1+x \le \exp{(x)}$ for $x\in \mathbb{R}$.

By \eqref{corollary.1} and \eqref{corollary.2}, it holds that 
\begin{equation}
    \begin{split}\label{corollary.3}
        & \mathbb{P}(\Theta_{b, H} |\mathcal{X}_n) \\        
        & \ge \mathbb{E} \left[ \prod_{h=1}^H \zeta_{h} \ \middle| \ \mathcal{X}_n \right]\\
        & = \mathbb{E} \left[\prod_{h=1}^{H}\left[ 1 - 2^{h-1}\times \exp{( - u_{h} \times  \texttt{\#}S  \times \lceil b \times 2^{-H }\rceil\times  (\sum_{q=1}^{B} \texttt{\#}  \Lambda_{q})^{-1}  )}\right] \ \middle| \ \mathcal{X}_n \right] \\
        & \ge \mathbb{E} \left[1 - \sum_{h=1}^{H} 2^{h-1}\times \exp{( - u_{h} \times  \texttt{\#}S  \times \lceil b \times 2^{-H }\rceil \times  (\sum_{q=1}^{B} \texttt{\#}  \Lambda_{q})^{-1}  )} \ \middle| \ \mathcal{X}_n \right],
    \end{split}    
\end{equation}
where the first inequality follows from \eqref{corollary.1} and \eqref{corollary.2}. The final inequality holds because $\prod_{l=1}^{L} (1 - x_{l}) \ge 1 - \sum_{l=1}^{L} x_{l}$ for $x_l \in [0, 1]$. This simplification provides a clear lower bound for the probability that the progressive tree remains sample-equivalent to the one-shot tree.

By the definitions of $\Theta_{b, H}$ and $\mathcal{A}_n$, \eqref{E_W_6}, \eqref{corollary.3}, and the definition of $\mathcal{B}$, we deduce that 
\begin{equation}
    \label{E_W.5}
    \begin{split}        
     \mathbb{P}(\mathcal{B}^c ) & \le \mathbb{P}( (\mathcal{A}_n \cap \Theta_{b, H})^c ) \\    
    & \le \mathbb{E}[\mathbb{P}(\Theta_{b, H}^c | \mathcal{X}_n)]  + \mathbb{P}(\mathcal{A}_n^c)\\
    & \le \mathbb{E}\left[ \sum_{h=1}^{H} 2^{h-1}\times \exp{ \left( - u_{h} \times  \texttt{\#}S  \times \lceil b \times 2^{-H }\rceil \times  \left(\sum_{q=1}^{B} \texttt{\#}  \Lambda_{q}\right)^{-1}  \right)} \right] \\
    & \quad +  10\exp{ (- (\log{n})^2 / 2)}.
    \end{split}
\end{equation}

Lastly, by \eqref{E_W.5}, $u_{h}\ge 1$ almost surely, $\sum_{q=1}^{B} \texttt{\#}  \Lambda_{q} = n \times B$, $\sum_{h=1}^{H}2^{h-1} = 2^{H}-1$, and that $\texttt{\#}S  \times \lceil b \times 2^{-H } \rceil \ge \texttt{\#}S \times b \times 2^{-H } \ge n2^{H} B\log{(2^H B)}\times 2^{-H }  $ due to $\texttt{\#}S \times b \ge n2^{H} B\log{(2^H B)} $, it holds that for all large $n$,
$$\mathbb{P}(\mathcal{B}^c ) \le \frac{1}{2n} + B^{-1}.$$
We have completed the proof of the first assertion for the continuous features of Corollary~\ref{corollary1}.

We now move on to proving the second assertion for the discrete features of Corollary~\ref{corollary1}. Let $O^{\star} = \{(J, \vv{v}): \texttt{\#}J = s, J \subseteq \{1, \dots, p\}, \vv{v} \in \{0, 1\}^s\}$ denote the set of all possible discrete configurations. The size of this set is bounded as:
\begin{equation}
    \label{sample,equi.5}
    \texttt{\#}O^{\star} \le 2^s \times \binom{p}{s}.
\end{equation}

Define the $\mathcal{X}_n$-measurable event 
$$Z_{n} = \bigcap_{ (J,\vv{v})\in O^{\star}} \{ \texttt{\#}\{ i: (X_{ij}, j\in J)^{\top} = \vv{v} \} \ge n \times D_{\min}\times 2^{-s-1}\} .$$ 
By the definition of $u_{h}$'s in \eqref{uh}, it holds on $Z_{n}$ that 
\begin{equation}
    \label{sample.equi.3}
    \min_{1\le h\le H}u_{h} \ge n\times D_{\min} \times 2^{-s-1}.
\end{equation}
The result of \eqref{sample.equi.3} holds because  $\min_{1\le k\le n}\texttt{\#}\{i:(\vv{w}, \vv{w}^{\top}\boldsymbol{X}_i) = (\vv{w}, \vv{w}^{\top}\boldsymbol{X}_{k}), \norm{\vv{w}}_0 \le s, i\in \{1, \dots, n\}\}\ge n\times D_{\min} \times 2^{-s-1}$ on $Z_{n}$. Here, we recall that by construction of the one-shot tree, for each node and split $(\boldsymbol{t}^{'}, (\vv{w}, c), \boldsymbol{t}) \in \widehat{N}_{h}^{(0)}$, it holds that $(\vv{w}, c) \in \Lambda_{q}$ for some index $q \in \{1, \dots, B\}$.

In the following, we  establish a probability  upper bound on the complementary event of $Z_{n}$. For every $(J, \vv{v}) \in O^{\star}$, on the event 
$$\left\{\left| \sum_{i=1}^n (\boldsymbol{1}_{(X_{ij}, j\in J)^{\top} = \vv{v}} - \mathbb{P}((X_{j}, j\in J)^{\top} = \vv{v})) \right| < D_{\min}2^{-s-1} n \right\},$$ 
it holds that 
\begin{equation}
    \begin{split}\label{sample.equi.6}
      \texttt{\#}\{ i: (X_{ij}, j\in J)^{\top} = \vv{v} \} & \ge n \times \mathbb{P}((X_{j}, j\in J)^{\top} = \vv{v}) - D_{\min} \times 2^{-s-1} \times n\\
    & \ge n \times D_{\min} \times 2^{-s-1},
    \end{split}
\end{equation}
where the second inequality follows from (a) of Condition~\ref{regularity.1}. The result of \eqref{sample.equi.6} implies that 
\begin{equation}
    \begin{split}\label{sample.equi.4}
         \bigcap_{(J,\vv{v})\in O^{\star}} \left\{\left| \sum_{i=1}^n (\boldsymbol{1}_{(X_{ij}, j\in J)^{\top} = \vv{v}} - \mathbb{P}((X_{j}, j\in J)^{\top} = \vv{v})) \right| < D_{\min}2^{-s-1} n \right\} \subseteq Z_{n}.
    \end{split}
\end{equation}

Next, we deduce by Hoeffding's inequality that for every $\vv{v}\in \{0, 1\}^s$, every $J\subseteq \{1, \dots, p\}$ with $\texttt{\#}J = s$, each $t>0$,
\begin{equation}
    \begin{split}\label{sample.equi.2}
        \mathbb{P}\left(\left| \sum_{i=1}^n \big(\boldsymbol{1}_{(X_{ij}, j\in J)^{\top} = \vv{v}} - \mathbb{P}((X_{j}, j\in J)^{\top} = \vv{v}) \big) \right| \ge t \right) \le 2\exp{\left(-\frac{2t^2}{n}\right)}.
    \end{split}
\end{equation}

By \eqref{sample.equi.2} with $t = D_{\min} \times 2^{-s-1} \times n$, \eqref{sample.equi.4}, \eqref{sample,equi.5}, we derive that for all large $n$ with $1\le p\le n^{K_{0}}$ and $s < \frac{\log_{e}{n}}{2}$,
\begin{equation}
    \begin{split}\label{sample.equi.1}
        \mathbb{P} ( Z_{n}^c ) & \le  \binom{p}{s}\times 2^s \times 2\exp{\left(-2D_{\min}^2 2^{-2s-2} n\right)}\\
        & \le \frac{1}{2n},
    \end{split}
\end{equation}
where we use the fact that \( \log(2) < 0.7 \), which implies \( 2^{-2s} < n^{-0.7} \).

By \eqref{sample.equi.1}, \eqref{corollary.3}, \eqref{sample.equi.3}, $\sum_{q=1}^{B} \texttt{\#}  \Lambda_{q} = n \times B$, $\sum_{h=1}^{H}2^{h-1} = 2^{H}-1$, and that $\texttt{\#}S  \times \lceil b \times 2^{-H } \rceil \ge \texttt{\#}S \times b \times 2^{-H } \ge D_{\min}^{-1}2^{H+ s + 1} B\log{(2^H B)}\times 2^{-H }  $ due to $\texttt{\#}S \times b \ge D_{\min}^{-1}2^{H + s + 1} B\log{(2^H B)} $, it holds that for all large $n$,
\begin{equation}
    \label{E_W.5.b}
    \begin{split}        
      \mathbb{P}(\mathcal{B}^c ) & \le \mathbb{E}[\mathbb{P}(\Theta_{b, H}^c | \mathcal{X}_n)]  + \mathbb{P}(\mathcal{A}_n^c) \\
    & \le \mathbb{E}[\mathbb{P}(\Theta_{b, H}^c | \mathcal{X}_n) \boldsymbol{1}_{Z_{n}}] + \mathbb{P}(Z_{n}^c)  + \mathbb{P}(\mathcal{A}_n^c) \\
    & \le \mathbb{E}\left\{ \sum_{h=1}^{H} 2^{h-1}\times \left[\exp{\left( - u_{h} \times  \texttt{\#}S  \times \lceil b \times 2^{-H }\rceil \times  \left(\sum_{q=1}^{B} \texttt{\#}  \Lambda_{q}\right)^{-1}  \right)} \right] \times \boldsymbol{1}_{Z_{n}} \right\} \\
    & \quad + \mathbb{P}(Z_{n}^c)+  10\exp{ (- (\log{n})^2 / 2)}\\
    & \le B^{-1} + n^{-1}.
    \end{split}
\end{equation}

This completes the proof for the second assertion of Corollary~\ref{corollary1} regarding discrete features, and consequently establishes the overall proof of Corollary~\ref{corollary1}.

\subsection{Proof that Example~\ref{example.linear} satisfies Condition~\ref{sid}}\label{proof.example.linear}
Let an arbitrary node $\boldsymbol{t}$ satisfying the requirements of Condition~\ref{sid} be given.
 Then, in light of the model assumptions, the split $(\vv{\gamma}, \gamma_0) \in W_{p, s_0}$ is such that
$$\mathbb{E}[\textnormal{Var}( m(\boldsymbol{X}) | \boldsymbol{1}_{\boldsymbol{X}\in \boldsymbol{t}}, \boldsymbol{1}_{\vv{\gamma}^{\top}\boldsymbol{X} > \gamma_{0}} ) \boldsymbol{1}_{\boldsymbol{X}\in \boldsymbol{t}}] = 0 , $$
which concludes the desired result.

\subsection{Proof that Example~\ref{example.linear2} satisfies Condition~\ref{sid}}\label{proof.example.linear2}

    Let $h(\vv{x})$ denote the probability density function of the distribution of $\boldsymbol{X}$ with 
    \begin{equation}\label{linear2.1}
    D_{\min }\le h(\vv{x})   \le D_{\max}
    \end{equation}
    for some constant $D_{\max}\ge D_{\min} > 0$ by our model assumption of Example~\ref{example.linear2}. Consider a node in the form $$\boldsymbol{t} = \cap_{i=1}^{l} A_{i}$$ 
    where $A_{i} \in \{\{ \vv{x}\in [0, 1]^p : \vv{w}_{i}^{\top}\vv{x} > c_{i}\}, \{ \vv{x}\in [0, 1]^p : \vv{w}_{i}^{\top}\vv{x} \le c_{i}\}\}$ and $(\vv{w}_{i}, c_{i}) \in W_{p, p}$. Define the area of the cross-section as the function of the bias $t$ as follows: 
    $$S(t) = \textnormal{Area of } \{\vv{x}\in [0, 1]^p : \vv{\gamma}^{\top} \vv{x} = t\} \cap \boldsymbol{t},$$
where we refer to \eqref{geolemma.1} for the result that the area function is concave such that $\frac{d^2 S(t)}{dt^2}\le 0$ for each $t \in \mathbb{R}$.

% the change rate $\frac{d S(t)}{dt}\le C_0$ for some constant $C_0>0$ that does not depend on $\boldsymbol{t}$ and $\vv{\gamma}$.

%%%%

    The upper bound and lower bound of the probability density function of the distribution $\mathbb{P}(\vv{\gamma}^{\top}\boldsymbol{X} \le t  , \boldsymbol{X}\in \boldsymbol{t})$, which is denoted by $h_{\vv{\gamma}}(t)$,  can be derived as follows.
    $$h_{\vv{\gamma}}(t) = \int_{\vv{v}\in \{ \vv{x}\in \boldsymbol{t}: \vv{\gamma}^{\top} \vv{x} = t\}}  h(\vv{v})d \vv{v}  .$$
    
By \eqref{linear2.1} and the definitions of $S(t)$ and $h_{\vv{\gamma}}(t)$, it holds that
\begin{equation}
    \begin{split}\label{linear2.3}
        h_{\vv{\gamma}}(t) & \le D_{\max} \times S(t),\\
        h_{\vv{\gamma}}(t) & \ge D_{\min} \times S(t) .
    \end{split}
\end{equation}

To derive the desired reuslt, we first have to establish the lower bound of $\inf_{t\in [t_6, t_7]} h_{\vv{\gamma}}(t) $ in terms of $\sup_{t\in\mathbb{R}} S(t)$ as follows. Denote 
    \begin{equation*}
        \begin{split}
            t_0 & = \inf \{t\in\mathbb{R}: S(t)>0\} , \\
            t_8 & = \sup \{t\in\mathbb{R}: S(t)>0\},\\
            t_l & = t_0 + l\times \frac{t_8 - t_0}{8},\\
            \delta & = t_8 - t_0.
        \end{split}
    \end{equation*}
In the following, we examine two cases: Case 1 considers \( \frac{dS(t)}{dt} \le 0 \) at $t = t_7$, while Case 2 considers \( \frac{dS(t)}{dt} > 0 \) at $t = t_7$.

Let us begin with Case 1, where $\frac{ dS(t) }{dt} \le 0$ at $t = t_7$. The case assumption $\frac{ dS(t_7) }{dt} \le 0$ and that  \( S(t) \) is concave on \( [t_0, t_8] \) deduce that
\begin{equation}
    \label{linear2.6}
    \argmax_{t\in\mathbb{R}} S(t) \le t_{7}.
\end{equation}
Additionally, due to concavity, it holds that 
\begin{equation}
	\label{linear2.4}
	\inf_{t\in [t_6, t_7]} S(t) = \min \{S(t_6), S(t_7)\}.
\end{equation}

Next, we establish a lower bound for $S(t_7)$. We consider a straight-line function $u(t)$ such that \( u(t_8) = S(t_8) \le   S(t_7)= u(t_7) \). Since  \( S(t) \) is concave on \( [t_0, t_8] \), it holds that \( u(t) \ge S(t) \) for all \( t \le t_7 \). To see why, suppose, for the sake of contradiction,   that \( u(t_9) < S(t_9) \) for some \( t_9 < t_7 \). Then, there exists a \( \lambda \in (0, 1) \) such that  $(1-\lambda)t_{9} + \lambda t_8 = t_7$, and hence
$$	u((1 - \lambda) t_9 + \lambda t_8 )  = u(t_7) = S(t_7)= S((1 - \lambda) t_9 + \lambda t_8),$$
which, in combination with the condition that $u(t_9) < S(t_9)$, implies that
\begin{equation*}
	\begin{split}
	(1 - \lambda) S(t_9) + \lambda S(t_8) & >  (1 - \lambda) u(t_9) + \lambda u(t_8) \\
	& = u((1 - \lambda) t_9 + \lambda t_8 ) \\
	& \ge S((1 - \lambda) t_9 + \lambda t_8).
	\end{split}
\end{equation*}
However, such a result contradicts the concavity of \( S(t) \), and hence we conclude that \( u(t) \ge S(t) \) for all \( t \le t_7 \). 

This in combination with that $t_8 - t_7 = \frac{t_8 - t_0}{8}$ and \eqref{linear2.6} concludes that
\begin{equation*}
S(t_8) + 8\times (S(t_7) - S(t_8))  = u(t_0) = \sup_{t\in [t_0, t_7]} u(t)   \ge  \sup_{t\in\mathbb{R}} S(t),
\end{equation*}
leading to that 
\begin{equation}
	\label{linear2.5}
S(t_7) \ge \frac{1}{8} \sup_{t\in\mathbb{R}} S(t).
\end{equation}

Let us proceed to establish a lower bound for $S(t_6)$. In light of concavity of $S(\cdot)$, if $S(t_7) \ge S(t_6)$, we have that
\begin{equation}
    S(t_6)  \ge S(t_0) + \left(1-\frac{t_7 - t_6}{t_7 - t_0}\right) \times (S(t_7) - S(t_0))\ge  \frac{6}{7} \times S(t_7) ,
\end{equation}
and otherwise,
\begin{equation}
	\label{linear2.7}
	S(t_6) >  S(t_7) .
\end{equation}

By \eqref{linear2.5}--\eqref{linear2.7},
\begin{equation}
    \label{linear2.8}
    \inf_{t\in [t_6, t_7]} S(t) \ge \frac{3}{28} \times \sup_{t\in\mathbb{R}} S(t),
\end{equation}
which, in combination with \eqref{linear2.3}, implies that 
\begin{equation}
    \begin{split}\label{linear2.11}
        \inf_{t\in [t_6, t_7]} h_{\vv{\gamma}}(t) & \ge D_{\min} \frac{3}{28} \times \sup_{t\in\mathbb{R}} S(t) .        
    \end{split}
\end{equation}
For Case 1 with $\frac{ dS(t) }{dt} \le 0$ at $t = t_7$, we have obtained the results of \eqref{linear2.11}.

Now, let us turn to consider Case 2 with $\frac{ dS(t) }{dt} > 0$ at $t = t_7$. In this case, it holds that 
$$\inf_{t\in [t_6, t_7]} S(t) \ge \sup_{t\in [t_0, t_6]} S(t)$$
 due to concavity and the case assumption.  Additionally, due to the concavity of $S(\cdot)$, the case assumption that $\frac{ dS(t) }{dt} > 0$ at $t = t_7$, and the definition of $t_{i}$'s, it holds that
$S(t_0) + (S(t_1) - S(t_0)) \times 8 \ge \sup_{t\in\mathbb{R}} S(t),$
 leading to that 
 $$S(t_1) \ge \frac{1}{8} \times \sup_{t\in\mathbb{R}} S(t).$$ 
 Therefore, 
\begin{equation*} 
    \inf_{t\in [t_6, t_7]} S(t) \ge \sup_{t\in [t_0, t_6]} S(t)\ge S(t_1) \ge \frac{1}{8} \times \sup_{t\in\mathbb{R}} S(t),
\end{equation*}
which, in combination with \eqref{linear2.3}, concludes for Case 2 that
\begin{equation}
    \label{linear2.9}
    \inf_{t\in [t_6, t_7]} h_{\vv{\gamma}}(t) \ge D_{\min}\frac{1}{8} \times \sup_{t\in\mathbb{R}} S(t).
\end{equation}

%%%
%%%%
With \eqref{linear2.11} and \eqref{linear2.9}, we now derive the mean shift from $\mathbb{E}(m(\boldsymbol{X})  | \boldsymbol{X} \in \boldsymbol{t})$ to $\mathbb{E}(m(\boldsymbol{X}) | \{\boldsymbol{X} \in \boldsymbol{t}\}\cap \{\vv{\gamma}^{\top} \boldsymbol{X} > s_4\})$ after the split $(\vv{w}, c) = (\vv{r},  \mathbb{E}(m(\boldsymbol{X}) | \boldsymbol{X} \in \boldsymbol{t}))$. Let $s_4 = \mathbb{E}(m(\boldsymbol{X}) | \boldsymbol{X} \in \boldsymbol{t})$. In what follows, we consider the situation where $t_8 - s_4 \ge s_4 - t_0$, which implies that $t_4 \ge s_4$. Then, 
\begin{equation}
    \label{linear.2.12}
    \begin{split}
        &\mathbb{E}(m(\boldsymbol{X}) | \{\boldsymbol{X} \in \boldsymbol{t}\}\cap \{\vv{\gamma}^{\top} \boldsymbol{X} > s_4\}) \\
        & = \frac{\mathbb{E}(m(\boldsymbol{X}) \boldsymbol{1}_{\boldsymbol{X} \in \boldsymbol{t}} \boldsymbol{1}_{\vv{\gamma}^{\top} \boldsymbol{X} > s_4} )}{ \mathbb{P}(\boldsymbol{X} \in \boldsymbol{t}, \vv{\gamma}^{\top} \boldsymbol{X} > s_4)}\\
        & =   \int_{t= s_4}^{t_6} t \frac{ h_{\vv{\gamma}}(t)}{\mathbb{P}(\boldsymbol{X} \in \boldsymbol{t}, \vv{\gamma}^{\top} \boldsymbol{X} > s_4)}  dt + \int_{t=t_7}^{t_8} t \frac{ h_{\vv{\gamma}}(t)}{\mathbb{P}(\boldsymbol{X} \in \boldsymbol{t}, \vv{\gamma}^{\top} \boldsymbol{X} > s_4)}  dt \\
        & \quad + \int_{t=t_6}^{t_7} t \frac{ h_{\vv{\gamma}}(t)}{\mathbb{P}(\boldsymbol{X} \in \boldsymbol{t}, \vv{\gamma}^{\top} \boldsymbol{X} > s_4)}  dt \\
        & \ge \mathbb{E}(m(\boldsymbol{X})  | \boldsymbol{X} \in \boldsymbol{t}) \times (1- \int_{t=t_6}^{t_7}  \frac{ h_{\vv{\gamma}}(t)}{\mathbb{P}(\boldsymbol{X} \in \boldsymbol{t}, \vv{\gamma}^{\top} \boldsymbol{X} > s_4)}  dt ) \\
        & \quad + (\mathbb{E}(m(\boldsymbol{X}) | \boldsymbol{X} \in \boldsymbol{t}) + \frac{\delta}{4} )   \times \int_{t=t_6}^{t_7}  \frac{ h_{\vv{\gamma}}(t)}{\mathbb{P}(\boldsymbol{X} \in \boldsymbol{t}, \vv{\gamma}^{\top} \boldsymbol{X} > s_4)}  dt    \\
        & \ge \sup_{x_0 \le \int_{t=t_6}^{t_7}  \frac{ h_{\vv{\gamma}}(t)}{\mathbb{P}(\boldsymbol{X} \in \boldsymbol{t}, \vv{\gamma}^{\top} \boldsymbol{X} > s_4)}  dt } \bigg[ \mathbb{E}(m(\boldsymbol{X})  | \boldsymbol{X} \in \boldsymbol{t}) (1-x_0) + (\mathbb{E}(m(\boldsymbol{X}) | \boldsymbol{X} \in \boldsymbol{t}) + \frac{\delta}{4} )    x_0 \bigg]   \\
        & \ge  \bigg[ \mathbb{E}(m(\boldsymbol{X})  | \boldsymbol{X} \in \boldsymbol{t}) \times \left(1- \frac{3 D_{\min}}{224  D_{\max}}\right)  + (\mathbb{E}(m(\boldsymbol{X}) | \boldsymbol{X} \in \boldsymbol{t}) + \frac{\delta}{4} ) \times \frac{3 D_{\min}}{224  D_{\max}} \bigg] \\
        & = \mathbb{E}(m(\boldsymbol{X})  | \boldsymbol{X} \in \boldsymbol{t}) + \frac{\delta}{4} \times \frac{3 D_{\min}}{224  D_{\max}}.
    \end{split}
\end{equation}
The second equality results from $m(\boldsymbol{X}) = \vv{\gamma}^{\top} \boldsymbol{X}$. The first inequality holds because \( s_4 \le t_6 \le t_7 \le t_8 \) and \( t_6 - s_4 \ge t_6 - t_4 \ge \frac{\delta}{4} \). The second inequality follows from the condition \( \delta > 0 \) and simple calculations. In the third inequality, we let \( x_0 \) be the lower bound of $\int_{t=t_6}^{t_7}  \frac{ h_{\vv{\gamma}}(t)}{\mathbb{P}(\boldsymbol{X} \in \boldsymbol{t}, \vv{\gamma}^{\top} \boldsymbol{X} > s_4)}  dt $, which is established based on the following results. By \eqref{linear2.3}, \eqref{linear2.11}, and \eqref{linear2.9}, it holds that
\begin{equation}
    \begin{split}\label{linear2.13}
      \int_{t=t_6}^{t_7} h_{\vv{\gamma}}(t)  dt 
     &  \ge \frac{3 D_{\min}}{28 } \times \sup_{t\in\mathbb{R}} S(t)\times  (t_7 - t_6) \\
    & = \frac{3 D_{\min}}{28 } \times \frac{\delta}{8} \times \sup_{t\in\mathbb{R}} S(t), \\
       \mathbb{P}(\boldsymbol{X} \in \boldsymbol{t}, \vv{\gamma}^{\top} \boldsymbol{X} > s_4) & \le  \mathbb{P}( \boldsymbol{X} \in \boldsymbol{t})  \\
       & = \int_{t=t_0}^{t_8} h_{\vv{\gamma}}(t)  dt \\
        & \le D_{\max}\sup_{t\in\mathbb{R}} S(t) \times \delta.
    \end{split}
\end{equation}

%%%
%%%

Now, by the model assumption and the definition of $\delta$, the conditional variance is upper bounded as follows.
\begin{equation}
    \begin{split}
        \label{linear2.10}
        \mathbb{E} [ \textnormal{Var} ( m(\boldsymbol{X}) | \boldsymbol{1}_{\boldsymbol{X} \in \boldsymbol{t}} ) \boldsymbol{1}_{\boldsymbol{X} \in \boldsymbol{t}} ] & \le \mathbb{E} \{  [ m(\boldsymbol{X}) - \mathbb{E}(m(\boldsymbol{X}) | \boldsymbol{X} \in \boldsymbol{t} )  ]^2 \boldsymbol{1}_{\boldsymbol{X} \in \boldsymbol{t}}  \}\\
        & \le \delta^2 \times \mathbb{P}(\boldsymbol{X} \in \boldsymbol{t}) .
        % & \le \delta^2  \times \int_{t \in [t_0, t_8]} h_{\vv{\gamma}} (t)  dt\\
        % & \le \delta^2  \times \delta \times D_{\max}.
    \end{split}
\end{equation}

By \eqref{linear.2.12}--\eqref{linear2.10}, with the split $(\vv{w}, c) = (\vv{\gamma}, \mathbb{E}(m(\boldsymbol{X}) | \boldsymbol{X} \in \boldsymbol{t})) $, we deduce that
\begin{equation}
    \begin{split}\label{linear2.15}
    &\inf_{(\vv{w}^{'}, c^{'}) \in W_{p, s_0}}  \mathbb{E} [ \textnormal{Var} ( m(\boldsymbol{X}) \mid \boldsymbol{1}_{\boldsymbol{X} \in \boldsymbol{t}} , \boldsymbol{1}_{\vv{w}^{'\top}\boldsymbol{X}>c^{'}} ) \boldsymbol{1}_{\boldsymbol{X} \in \boldsymbol{t}} ] \\
         & \le \mathbb{E} [ \textnormal{Var} ( m(\boldsymbol{X}) \mid \boldsymbol{1}_{\boldsymbol{X} \in \boldsymbol{t}} , \boldsymbol{1}_{\vv{w}^{\top}\boldsymbol{X}>c} ) \boldsymbol{1}_{\boldsymbol{X} \in \boldsymbol{t}} ] \\
         & \le \mathbb{E} [ \textnormal{Var} ( m(\boldsymbol{X}) | \boldsymbol{1}_{\boldsymbol{X} \in \boldsymbol{t}} ) \boldsymbol{1}_{\boldsymbol{X} \in \boldsymbol{t}} ]   \\
         & \quad - \mathbb{P}( \boldsymbol{X} \in \boldsymbol{t} , \vv{w}^{\top}\boldsymbol{X}>c ) \times \{\mathbb{E} [m(\boldsymbol{X}) | \boldsymbol{X} \in \boldsymbol{t}] - \mathbb{E} [m(\boldsymbol{X}) | \{\boldsymbol{X} \in \boldsymbol{t} \}\cap \{ \vv{w}^{\top}\boldsymbol{X}>c\}] \}^2\\
         & \le  \mathbb{E} [ \textnormal{Var} ( m(\boldsymbol{X}) | \boldsymbol{1}_{\boldsymbol{X} \in \boldsymbol{t}} ) \boldsymbol{1}_{\boldsymbol{X} \in \boldsymbol{t}} ]  - \left[  \mathbb{P}( \boldsymbol{X} \in \boldsymbol{t} , \vv{w}^{\top}\boldsymbol{X}>c )  \times \frac{9 \delta^2 D_{\min}^2}{802816  D_{\max}^2}\right]\\ %%%%%
         & \le  \mathbb{E} [ \textnormal{Var} ( m(\boldsymbol{X}) | \boldsymbol{1}_{\boldsymbol{X} \in \boldsymbol{t}} ) \boldsymbol{1}_{\boldsymbol{X} \in \boldsymbol{t}} ]  \times \left[ 1 -  \frac{\mathbb{P}( \boldsymbol{X} \in \boldsymbol{t} , \vv{w}^{\top}\boldsymbol{X}>c ) }{\mathbb{P}( \boldsymbol{X} \in \boldsymbol{t}  ) } \times \frac{9 D_{\min}^2}{802816  D_{\max}^2}\right]\\
         & \le  \mathbb{E} [ \textnormal{Var} ( m(\boldsymbol{X}) | \boldsymbol{1}_{\boldsymbol{X} \in \boldsymbol{t}} ) \boldsymbol{1}_{\boldsymbol{X} \in \boldsymbol{t}} ]  \times \left[ 1 -  \frac{3 D_{\min}}{224  D_{\max}} \times \frac{9 D_{\min}^2}{802816  D_{\max}^2}\right],
    \end{split}
\end{equation}
where the second inequality follows from the variance decomposition formula, and the fourth inequality results from \eqref{linear2.10}, and the fifth inequality holds because $\frac{\mathbb{P}( \boldsymbol{X} \in \boldsymbol{t} , \vv{w}^{\top}\boldsymbol{X}>c ) }{\mathbb{P}( \boldsymbol{X} \in \boldsymbol{t}  ) } \ge  \frac{3 D_{\min}}{224  D_{\max}}$ due to the definition of $(\vv{w}, c)$ and \eqref{linear2.13}.

With \eqref{linear2.15} and  $\alpha_{0} = [ 1 -  \frac{3 D_{\min}}{224  D_{\max}} \times \frac{9 D_{\min}^2}{802816  D_{\max}^2} ]$, we conclude the desired results for the scenario with $t_8 - s_4 \ge s_4 - t_0$. The other case with $t_8 - s_4 < s_4 - t_0$ can be dealt with similarly, and we omit the detail.

We have completed the proof for Example~\ref{example.linear2}.
%%%
%%%%
%%%

\subsection{Proof that Example~\ref{example.xor} satisfies Condition~\ref{sid}}\label{proof.example.xor}

Let an arbitrary node $\boldsymbol{t}$ satisfying the requirements of Condition~\ref{sid} be given. Without loss of generality, let $\beta = 1$, and assume 
\begin{equation}
    \label{xor.1}
    \begin{split}    
    A_{1} & \coloneqq \mathbb{P}((X_{1}, \dots, X_{s_0}) \in \mathcal{V}_{1} \mid \boldsymbol{X}\in \boldsymbol{t}) \\
    & \ge \mathbb{P}((X_{1}, \dots, X_{s_0}) \in \mathcal{V}_{2} \mid \boldsymbol{X}\in \boldsymbol{t}) \\
    & \eqqcolon A_{2},
    \end{split}
\end{equation}
with $A_{1} + A_{2} = 1$, and that $\vv{v}_0 = \argmax_{\vv{u} \in \mathcal{V}_{2}}\mathbb{P}((X_{1}, \dots , X_{s_0}) = \vv{u} \mid \boldsymbol{X}\in \boldsymbol{t})$, which implies that for every $\vv{u} \in \mathcal{V}_2$,
\begin{equation}    
    \begin{split}\label{xor.3}
    \mathbb{P}((X_{1}, \dots , X_{s_0}) =\vv{v}_0 \mid \boldsymbol{X}\in \boldsymbol{t})&  \ge \frac{A_{2}}{\texttt{\#}\mathcal{V}_2} = \frac{A_{2}}{2^{s_0 - 1}} ,\\
     \mathbb{P}((X_{1}, \dots , X_{s_0}) =\vv{v}_0 \mid \boldsymbol{X}\in \boldsymbol{t}) & \ge\mathbb{P}((X_{1}, \dots , X_{s_0}) = \vv{u} \mid \boldsymbol{X}\in \boldsymbol{t}).
    \end{split}
\end{equation}

Next, we select $(\vv{w}, c)\in W_{p, s_0}$ such that $\boldsymbol{1}_{\vv{w}^{\top} \boldsymbol{X} > c} = \boldsymbol{1}_{(X_{1}, \dots , X_{s_0}) = \vv{v}_0}$ almost surely. This selection is possible because we are working with binary features. Hence,
\begin{equation}
    \label{xor.9}
    \mathbb{E}\{ [ m(\boldsymbol{X}) - \mathbb{E}(m(\boldsymbol{X}) \mid \boldsymbol{X}\in \boldsymbol{t}, \vv{w}^{\top} \boldsymbol{X} > c   ) ]^2 \times \boldsymbol{1}_{\boldsymbol{X} \in \boldsymbol{t}}\boldsymbol{1}_{\vv{w}^{\top} \boldsymbol{X} > c} \} = 0.  
\end{equation}

Define $\boldsymbol{t}_{1}$ and $\boldsymbol{t}_{2}$ such that
$\boldsymbol{1}_{\boldsymbol{X}\in \boldsymbol{t}_{1}} = \boldsymbol{1}_{\vv{w}^{\top} \boldsymbol{X} > c}\times \boldsymbol{1}_{\boldsymbol{X}\in \boldsymbol{t}}$ and $\boldsymbol{1}_{\boldsymbol{X}\in \boldsymbol{t}_{2}} = \boldsymbol{1}_{\vv{w}^{\top} \boldsymbol{X} \le c} \times \boldsymbol{1}_{\boldsymbol{X}\in \boldsymbol{t}}$. With these definitions, we deduce that 
\begin{equation}
    \begin{split}\label{xor.8}
        & \mathbb{E} [ \textnormal{Var} ( m(\boldsymbol{X}) \mid \boldsymbol{1}_{\boldsymbol{X} \in \boldsymbol{t}} ) \boldsymbol{1}_{\boldsymbol{X} \in \boldsymbol{t}} ] \\
        & = \mathbb{E}\{ [ m(\boldsymbol{X}) - \mathbb{E}(m(\boldsymbol{X}) \mid \boldsymbol{X}\in \boldsymbol{t}) ]^2 \times \boldsymbol{1}_{\boldsymbol{X} \in \boldsymbol{t}}\}\\
        & = \mathbb{E}\{ [ m(\boldsymbol{X}) - \mathbb{E}(m(\boldsymbol{X}) \mid \boldsymbol{X}\in \boldsymbol{t}) ]^2 \times \boldsymbol{1}_{\boldsymbol{X} \in \boldsymbol{t}_{1}}\} \\
        & \quad+ \mathbb{E}\{ [ m(\boldsymbol{X}) - \mathbb{E}(m(\boldsymbol{X}) \mid \boldsymbol{X}\in \boldsymbol{t}) ]^2 \times \boldsymbol{1}_{\boldsymbol{X} \in \boldsymbol{t}_{2}}\}.
    \end{split}
\end{equation}
In addition, by \eqref{xor.9} and simple calculations,
\begin{equation}
    \begin{split}\label{xor.4}
        & \mathbb{E} [ \textnormal{Var} ( m(\boldsymbol{X}) \mid \boldsymbol{1}_{\boldsymbol{X} \in \boldsymbol{t}}, \boldsymbol{1}_{\vv{w}^{\top} \boldsymbol{X} > c} ) \boldsymbol{1}_{\boldsymbol{X} \in \boldsymbol{t}} ] \\
        & = \mathbb{E}\{ [ m(\boldsymbol{X}) - \mathbb{E}(m(\boldsymbol{X}) \mid \boldsymbol{X}\in \boldsymbol{t}, \vv{w}^{\top} \boldsymbol{X} > c  )  ]^2 \times \boldsymbol{1}_{\boldsymbol{X} \in \boldsymbol{t}}\boldsymbol{1}_{\vv{w}^{\top} \boldsymbol{X} > c} \} \\
        & \quad + \mathbb{E}\{ [ m(\boldsymbol{X}) - \mathbb{E}(m(\boldsymbol{X}) \mid \boldsymbol{X}\in \boldsymbol{t}, \vv{w}^{\top} \boldsymbol{X} \le c ) ]^2 \times \boldsymbol{1}_{\boldsymbol{X} \in \boldsymbol{t}}\boldsymbol{1}_{\vv{w}^{\top} \boldsymbol{X} \le c} \} \\
        & = \mathbb{E}\{ [ m(\boldsymbol{X}) - \mathbb{E}(m(\boldsymbol{X}) \mid \boldsymbol{X}\in \boldsymbol{t}_{2} )  ]^2 \times \boldsymbol{1}_{\boldsymbol{X} \in \boldsymbol{t}_{2}} \} \\
        & \le  \mathbb{E}\{ [ m(\boldsymbol{X}) - \mathbb{E}(m(\boldsymbol{X}) \mid \boldsymbol{X}\in \boldsymbol{t} )  ]^2 \times \boldsymbol{1}_{\boldsymbol{X} \in \boldsymbol{t}_{2}} \} ,
    \end{split}
\end{equation}
where the last inequality holds because 
$$\mathbb{E}(m(\boldsymbol{X}) \mid \boldsymbol{X}\in \boldsymbol{t}_2 ) = \argmin_{x\in \mathbb{R}} \mathbb{E}\{ [ m(\boldsymbol{X}) - x  ]^2 \boldsymbol{1}_{\boldsymbol{X} \in \boldsymbol{t}_{2}} \}.$$

By \eqref{xor.8}--\eqref{xor.4},
\begin{equation}
    \begin{split}\label{xor.10}
    & \mathbb{E} [ \textnormal{Var} ( m(\boldsymbol{X}) \mid \boldsymbol{1}_{\boldsymbol{X} \in \boldsymbol{t}} ) \boldsymbol{1}_{\boldsymbol{X} \in \boldsymbol{t}} ] - \inf_{(\vv{w}^{'}, c^{'}) \in W_{p, s_0}} \mathbb{E} [ \textnormal{Var} ( m(\boldsymbol{X}) \mid \boldsymbol{1}_{\boldsymbol{X} \in \boldsymbol{t}}, \boldsymbol{1}_{\vv{w}^{'\top} \boldsymbol{X} > c^{'}} ) \boldsymbol{1}_{\boldsymbol{X} \in \boldsymbol{t}} ] \\
    & \ge \mathbb{E} [ \textnormal{Var} ( m(\boldsymbol{X}) \mid \boldsymbol{1}_{\boldsymbol{X} \in \boldsymbol{t}} ) \boldsymbol{1}_{\boldsymbol{X} \in \boldsymbol{t}} ] -  \mathbb{E} [ \textnormal{Var} ( m(\boldsymbol{X}) \mid \boldsymbol{1}_{\boldsymbol{X} \in \boldsymbol{t}}, \boldsymbol{1}_{\vv{w}^{\top} \boldsymbol{X} > c} ) \boldsymbol{1}_{\boldsymbol{X} \in \boldsymbol{t}} ]\\
            &\ge \mathbb{E}\{ [ m(\boldsymbol{X}) - \mathbb{E}(m(\boldsymbol{X}) \mid \boldsymbol{X}\in \boldsymbol{t}) ]^2 \times \boldsymbol{1}_{\boldsymbol{X} \in \boldsymbol{t}_{1}}\},
    \end{split}
\end{equation}
in which $\norm{\vv{w}}_{0}=s_0$ by the choice of $(\vv{w}, c)$ as in \eqref{xor.9}.

In the following, we will demonstrate that the RHS of \eqref{xor.10} is proportional to the total variance on the node $\boldsymbol{t}$, \(\mathbb{E} [ \textnormal{Var}(m(\boldsymbol{X}) | \boldsymbol{1}_{\boldsymbol{X} \in \boldsymbol{t}}) \boldsymbol{1}_{\boldsymbol{X} \in \boldsymbol{t}} ]\). In light of \eqref{xor.1} and the model assumptions on $m(\boldsymbol{X})$, 
\begin{equation}
    \label{xor.2}
    \mathbb{E}(m(\boldsymbol{X}) \mid \boldsymbol{X}\in \boldsymbol{t})  = A_{1} - A_{2} \ge 0.
\end{equation}

We additionally define $\boldsymbol{1}_{\boldsymbol{X} \in \boldsymbol{t}_{3}} = \boldsymbol{1}_{(X_{1}, \dots, X_{s_0}) \in \mathcal{V}_1}$. Now, by the definition of $\boldsymbol{t}_{1}$ to $\boldsymbol{t}_{3}$, \eqref{xor.3}, \eqref{xor.2}, and the model assumptions on $m(\boldsymbol{X})$, it holds that 
\begin{equation}
    \begin{split}\label{xor.6}
        & \mathbb{E}\{ [ m(\boldsymbol{X}) -\mathbb{E}(m(\boldsymbol{X}) \mid \boldsymbol{X}\in \boldsymbol{t}) ]^2 \times \boldsymbol{1}_{\boldsymbol{X} \in \boldsymbol{t}_{2}} \boldsymbol{1}_{\boldsymbol{X} \in \boldsymbol{t}_{3}} \} \\
         &\quad = \mathbb{E}\{ [ m(\boldsymbol{X}) -\mathbb{E}(m(\boldsymbol{X}) \mid \boldsymbol{X}\in \boldsymbol{t}) ]^2  \boldsymbol{1}_{\boldsymbol{X} \in \boldsymbol{t}_{3}} \boldsymbol{1}_{\boldsymbol{X} \in \boldsymbol{t}} \}\\
         & \quad \le (1-A_{1}+A_{2})^2 \times A_{1} \times \mathbb{P}(\boldsymbol{X}\in \boldsymbol{t}),\\   
         & \mathbb{E}\{ [ m(\boldsymbol{X}) -\mathbb{E}(m(\boldsymbol{X}) \mid \boldsymbol{X}\in \boldsymbol{t}) ]^2 \times \boldsymbol{1}_{\boldsymbol{X} \in \boldsymbol{t}_{1}}\} \\
         &\quad \ge (-1-A_{1}+A_{2})^2 \times \frac{A_{2}}{2^{s_0 - 1}} \times \mathbb{P}(\boldsymbol{X}\in \boldsymbol{t}).
    \end{split}
\end{equation}

By \eqref{xor.6} and that $A_{1} + A_{2} = 1$ and $A_{1} \ge \frac{1}{2} \ge A_{2}$ due to \eqref{xor.1}, it holds that 
\begin{equation}
    \begin{split}\label{xor.7}
    & \frac{\mathbb{E}\{ [ m(\boldsymbol{X}) -\mathbb{E}(m(\boldsymbol{X}) \mid \boldsymbol{X}\in \boldsymbol{t}) ]^2 \times \boldsymbol{1}_{\boldsymbol{X} \in \boldsymbol{t}_{1}}\} }{\mathbb{E}\{ [ m(\boldsymbol{X}) -\mathbb{E}(m(\boldsymbol{X}) \mid \boldsymbol{X}\in \boldsymbol{t}) ]^2 \times \boldsymbol{1}_{\boldsymbol{X} \in \boldsymbol{t}_{2}} \boldsymbol{1}_{\boldsymbol{X} \in \boldsymbol{t}_{3}} \}} \\
    & \ge \frac{(-1-A_{1}+A_{2})^2 \times \frac{A_{2}}{2^{s_0- 1}} \times \mathbb{P}(\boldsymbol{X}\in \boldsymbol{t})}{(1-A_{1}+A_{2})^2 \times A_{1} \times \mathbb{P}(\boldsymbol{X}\in \boldsymbol{t})} \\
    & = \frac{(-2A_{1})^2 (1-A_{1})}{ 2^{s_0 - 1} \times (2-2A_{1})^2 A_{1}} \\
    & = \frac{A_{1}}{2^{s_0 - 1}(1-A_{1})}\\
    & \ge \frac{1}{2^{s_0 - 1}}
    \end{split}
\end{equation}

Additionally, by the definitions of $\boldsymbol{t}_{1}$ to $\boldsymbol{t}_{3}$ and $\mathcal{V}_2$, \eqref{xor.3}, and the model assumptions,
{\small \begin{equation}
    \begin{split}\label{xor.5}
          &\mathbb{E}\{ [ m(\boldsymbol{X}) -\mathbb{E}(m(\boldsymbol{X}) \mid \boldsymbol{X}\in \boldsymbol{t}) ]^2 \times \boldsymbol{1}_{\boldsymbol{X} \in \boldsymbol{t}_{2}} \boldsymbol{1}_{\boldsymbol{X} \not\in \boldsymbol{t}_{3}} \} \\
        &  = \mathbb{E}\{ [ m(\boldsymbol{X}) -\mathbb{E}(m(\boldsymbol{X}) \mid \boldsymbol{X}\in \boldsymbol{t}) ]^2 \times (\boldsymbol{1}_{\boldsymbol{X}\in \boldsymbol{t}} \boldsymbol{1}_{(x_{1}, \dots, X_{s_0})\in \mathcal{V}_2} - \boldsymbol{1}_{\boldsymbol{X} \in \boldsymbol{t}_{1}} ) \} \\
        &  = \mathbb{E}\{ [ - 1 -\mathbb{E}(m(\boldsymbol{X}) \mid \boldsymbol{X}\in \boldsymbol{t}) ]^2\} \times \Big(\sum_{\vv{v} \in \mathcal{V}_2\backslash\{\vv{v}_0\}} \mathbb{P}( \{\boldsymbol{X}\in \boldsymbol{t}\} \cap \{(X_{1}, \dots , X_{s_0}) = \vv{v} \}) \Big) \\
        &  \le \mathbb{E}\{ [ - 1 -\mathbb{E}(m(\boldsymbol{X})  \boldsymbol{X}\in \boldsymbol{t}) ]^2\} \times \Big(\mathbb{P}( \{\boldsymbol{X}\in \boldsymbol{t}\} \cap \{(X_{1}, \dots , X_{s_0}) = \vv{v}_0 \}) \Big) \times ( \texttt{\#}\mathcal{V}_2 - 1) \\
        & =  \mathbb{E}\{ [ m(\boldsymbol{X}) -\mathbb{E}(m(\boldsymbol{X}) \mid \boldsymbol{X}\in \boldsymbol{t}) ]^2 \times \boldsymbol{1}_{\boldsymbol{X} \in \boldsymbol{t}_{1}}  \} \times (2^{s_0 - 1} - 1).
    \end{split}
\end{equation}}%

We deduce that
\begin{equation}
    \begin{split}\label{xor.11}
        & \frac{ \mathbb{E}\{ [ m(\boldsymbol{X}) - \mathbb{E}(m(\boldsymbol{X}) \mid \boldsymbol{X}\in \boldsymbol{t}) ]^2 \times \boldsymbol{1}_{\boldsymbol{X} \in \boldsymbol{t}_{1}}\} }{ \mathbb{E} [ \textnormal{Var} ( m(\boldsymbol{X}) \mid \boldsymbol{1}_{\boldsymbol{X} \in \boldsymbol{t}} ) \boldsymbol{1}_{\boldsymbol{X} \in \boldsymbol{t}} ] } \\
        & = \big[\mathbb{E}\{ [ m(\boldsymbol{X}) - \mathbb{E}(m(\boldsymbol{X}) \mid \boldsymbol{X}\in \boldsymbol{t}) ]^2 \times \boldsymbol{1}_{\boldsymbol{X} \in \boldsymbol{t}_{1}}\} \big]\\
        & \quad \times \big[ \mathbb{E}\{ [ m(\boldsymbol{X}) -\mathbb{E}(m(\boldsymbol{X}) \mid \boldsymbol{X}\in \boldsymbol{t}) ]^2 \times \boldsymbol{1}_{\boldsymbol{X} \in \boldsymbol{t}_{1}}  \} \\
        & \quad\quad + \mathbb{E}\{ [ m(\boldsymbol{X}) -\mathbb{E}(m(\boldsymbol{X}) \mid \boldsymbol{X}\in \boldsymbol{t}) ]^2 \times \boldsymbol{1}_{\boldsymbol{X} \in \boldsymbol{t}_{2}} \boldsymbol{1}_{\boldsymbol{X} \not\in \boldsymbol{t}_{3}} \} \\
        &\quad\quad + \mathbb{E}\{ [ m(\boldsymbol{X}) -\mathbb{E}(m(\boldsymbol{X}) \mid \boldsymbol{X}\in \boldsymbol{t}) ]^2 \times \boldsymbol{1}_{\boldsymbol{X} \in \boldsymbol{t}_{2}} \boldsymbol{1}_{\boldsymbol{X} \in \boldsymbol{t}_{3}} \}\big]^{-1}\\
        & \ge \big[\mathbb{E}\{ [ m(\boldsymbol{X}) - \mathbb{E}(m(\boldsymbol{X}) \mid \boldsymbol{X}\in \boldsymbol{t}) ]^2 \times \boldsymbol{1}_{\boldsymbol{X} \in \boldsymbol{t}_{1}}\} \big]\\
        & \quad \times \big[ [1 + 2^{s_0- 1}   + 2^{s_0-1} - 1] \times \mathbb{E}\{ [ m(\boldsymbol{X}) -\mathbb{E}(m(\boldsymbol{X}) \mid \boldsymbol{X}\in \boldsymbol{t}) ]^2 \times \boldsymbol{1}_{\boldsymbol{X} \in \boldsymbol{t}_{1}}  \} \big]^{-1}\\
        & \ge  \frac{1}{2^{s_0}},
    \end{split}
\end{equation}
in which the first equality follows from \eqref{xor.8}, and the first inequality follows from \eqref{xor.7}--\eqref{xor.5}.

The results of \eqref{xor.10} and \eqref{xor.11} conclude that 
\begin{equation}
\begin{split}\label{xor.12}
    & \mathbb{E} [ \textnormal{Var} ( m(\boldsymbol{X}) \mid \boldsymbol{1}_{\boldsymbol{X} \in \boldsymbol{t}} ) \boldsymbol{1}_{\boldsymbol{X} \in \boldsymbol{t}} ] \times (1- 2^{-s_0}) \\
    & \ge \mathbb{E} [ \textnormal{Var} ( m(\boldsymbol{X}) \mid \boldsymbol{1}_{\boldsymbol{X} \in \boldsymbol{t}} ) \boldsymbol{1}_{\boldsymbol{X} \in \boldsymbol{t}} ] - \mathbb{E}\{ [ m(\boldsymbol{X}) - \mathbb{E}(m(\boldsymbol{X}) \mid \boldsymbol{X}\in \boldsymbol{t}) ]^2 \times \boldsymbol{1}_{\boldsymbol{X} \in \boldsymbol{t}_{1}}\} \\
    & \ge \inf_{(\vv{w}^{'}, c^{'}) \in W_{p, s_0}} \mathbb{E} [ \textnormal{Var} ( m(\boldsymbol{X}) \mid \boldsymbol{1}_{\boldsymbol{X} \in \boldsymbol{t}}, \boldsymbol{1}_{\vv{w}^{'\top} \boldsymbol{X} > c^{'}} ) \boldsymbol{1}_{\boldsymbol{X} \in \boldsymbol{t}} ].
    \end{split}
\end{equation}

The result of \eqref{xor.12} applies to the scenario as in  \eqref{xor.1} and \eqref{xor.3}, but can be generalized to general cases. Therefore, we conclude the desired result that Example~\ref{example.xor} satisfies SID$(1 - \frac{1}{2^{s_0}}, s_0)$.% with $\alpha_{0} = 1 - \frac{1}{2^{s_0}}$ and $s\ge 2$.

%%%
%%%
%%%

%%%
%%%
%%%

\subsection{Proof that Example~\ref{example.mono} satisfies Condition~\ref{sid}}\label{proof.example.mono}

The assumptions on \(m(\cdot)\) imply that the sequence \(m(\vv{x}_{1}), \dots, m(\vv{x}_{L})\) is convex. Consequently, the model assumptions in Example~\ref{example.convex} are satisfied with \(K_{1} \leq 2\). Furthermore, the distributional assumptions on \(\boldsymbol{X}\) and other regularity conditions outlined in Example~\ref{example.convex} are also fulfilled. Thus, we can apply the results from Example~\ref{example.convex} to establish the desired conclusion for Example~\ref{example.mono}.

\subsection{Proof that Example~\ref{example.convex} satisfies Condition~\ref{sid}}\label{proof.example.convex}

Let us begin with introducing required terms for the proof. Under the assumptions of discrete \( m(\boldsymbol{X}) \) and the valley-shaped model, we refer to each region in $[0, 1]^p$ between two hyperplane decisions as a `platform.` Consider a decision region with a form given by
\begin{equation*}
    % \label{deision.region.1}
    \boldsymbol{t} = \cap_{i=1}^{l} A_{i},
\end{equation*}
 for some integer $l>0$, where $A_{i} \in \{\{ \vv{x}\in[0, 1]^p: \vv{w}_{i}^{\top}\vv{x} > c_{i}\}, \{ \vv{x}\in[0, 1]^p:\vv{w}_{i}^{\top}\vv{x} \le c_{i}\}\}$ and $(\vv{w}_{i}, c_{i}) \in W_{p, p}$.

Without loss of generality, assume 
\begin{equation}
    \label{without.1}
    m(\boldsymbol{X}) \textnormal{ takes discrete values in } \{r_{0}, \dots, r_{L_{0}}\}
\end{equation} 
with $r_{L_{0}} > \dots > r_{0}=0$ for some integer $L_{0}>0$ and 
\begin{equation}
    \label{without.2}
    \underline{r} = \min_{1\le j\le L_{0}} r_{j} - r_{j-1}  > 0.
\end{equation}
The lowest valley platform of the response is called `zero region.` Let $H_{1}, \dots, H_{K_{1}}$ with $H_{k} = \{\vv{x}\in [0, 1]^p: \vv{v}_{1}^{(k)\top} \vv{x} = a_{1}^{(k)}\}$ denote the $K_{1}$ hyperplanes surrounding the zero region, and we denote $\mathcal{C}_{k} = \{\vv{x}\in [0, 1]^p: \vv{v}_{1}^{(k)\top} \vv{x} \ge a_{1}^{(k)}\}$ as the corner region of $[0, 1]^{p}$ according to $H_{k}$ for each $k\in \{1, \dots, K_{1}\}$. The corner regions $\mathcal{C}_{1}, \dots, \mathcal{C}_{K_{1}}$, and the zero region, which is denoted by $\mathcal{C}_{0} = G_{0}^{(1)}$, form a partition of $[0, 1]^p$. For the node $\boldsymbol{t}$ and $k\in \{1, \dots, K_{1}\}$, let $h_{k}\subseteq \boldsymbol{t}$ denote the second highest platform in each corner region conditional on $\boldsymbol{t}$, and that $h_{k} = \emptyset$ if there is no more than one platform has intersection with $\boldsymbol{t}$ in the $k$th corner.

Next, we deal with important distributional inequalities as follows. To simplify the notation, we use $\{(\vv{\eta}_{l}, \theta_{l})\}_{l=1}^{L_{3}}$ with $L_{3} = \sum_{k=1}^{K_{1}} n_{k}$ to denote the set of $\{(\vv{v}_{l}^{(k)}, a_{l}^{(k)}): l\in \{1, \dots,  n^{k}\}, k\in\{1, \dots, K_{1}\}\}$. Define
\begin{equation*}
    \begin{split}
         q_{l} \coloneqq \inf \left(\bigcup_{k\not = l} \{|t|:\{\vv{x} \in [0, 1]^p :  \vv{\eta}_{l}^{\top}\vv{x} - \theta_{l} - t = 0\}\cap \{\vv{x} \in [0, 1]^p :  \vv{\eta}_{k}^{\top}\vv{x} - \theta_{k}  = 0\} \not=\emptyset \} \right),
    \end{split}
\end{equation*}
and  $c_{0} = \min_{1\le l \le L_3} q_{l}$. Then, it holds that $c_{0}>0$ due to the model assumption that there is no intersection between every two hyperplanes in $[0, 1]^p$. Define 
\begin{equation}
    \label{case1.7}
    \underline{\rho} =  \frac{c_{0}^2}{ 16s + 8 c_{0}\sqrt{s} + c_{0}^2} > 0.
\end{equation}

For any node \( \boldsymbol{t} \) that intersects with at least three platforms—\( \mathcal{A} \), \( \mathcal{B} \), and \( \mathcal{C} \)—where \( (\mathcal{A}, \mathcal{B}) \) and \( (\mathcal{C}, \mathcal{B}) \) are each separated by a different hyperplane, we conclude from Lemma~\ref{lower.volume.ratio} in Section~\ref{proof.lower.volume.ratio} that
\begin{equation}
    \label{case1.7.c}
    \begin{split}        
    \mathbb{P}(\boldsymbol{X} \in \boldsymbol{t} \cap \mathcal{B}) & \ge \left(\textnormal{Volume of }\boldsymbol{t} \cap \mathcal{B} \right) \times D_{\min}\\
    & \ge \left(\textnormal{Volume of }\boldsymbol{t}  \right) \times \underline{\rho} \times D_{\min}  \\
    & \ge \mathbb{P}(\boldsymbol{X} \in \boldsymbol{t} ) \times D_{\max}^{-1} \times \underline{\rho} \times D_{\min},
    \end{split}
\end{equation}
where the inequalities also rely on the distributional assumption that $(\textnormal{Volume of } A) \times D_{\min} \le \mathbb{P}(\boldsymbol{X} \in A) \le (\textnormal{Volume of } A) \times D_{\max}$  for each measurable $A\subseteq [0, 1]^p$ for some $D_{\max} \ge D_{\min}  >0$.

Particularly, the result applies to situations where \(\texttt{\#}\{k > 0 : \mathbb{P}(\boldsymbol{X} \in \boldsymbol{t} \cap \mathcal{C}_k) > 0\} > 1\), as \(\boldsymbol{t}\) spans two corner regions and the zero platform.

%%%%
%%%%

Our goal is to show that Example~\ref{example.convex} satisfies Condition~\ref{sid}. By the variance decomposition formula, we have that for every split $(\vv{w}, c)$,
\begin{equation}
    \begin{split}\label{decom.1}
         & \textnormal{Var} ( m(\boldsymbol{X}) \mid \{\boldsymbol{X}\in \boldsymbol{t}\} \cap \{\vv{w}^{\top}\boldsymbol{X} > c \}) \times \mathbb{P}(\vv{w}^{\top}\boldsymbol{X} > c \mid \boldsymbol{X}\in \boldsymbol{t} ) \\
         & \quad + \textnormal{Var} ( m(\boldsymbol{X}) \mid \{\boldsymbol{X}\in \boldsymbol{t}\} \cap \{\vv{w}^{\top}\boldsymbol{X} \le c \}) \times  \mathbb{P}(\vv{w}^{\top}\boldsymbol{X} \le c \mid \boldsymbol{X}\in \boldsymbol{t} ) \\
         & = \textnormal{Var} ( m(\boldsymbol{X}) \mid \boldsymbol{X}\in \boldsymbol{t} )  \\
         &\quad  - [\mathbb{E}( m(\boldsymbol{X}) \mid \{\boldsymbol{X}\in \boldsymbol{t}\} \cap \{\vv{w}^{\top}\boldsymbol{X} > c \}) -  \mathbb{E}( m(\boldsymbol{X}) \mid \boldsymbol{X}\in \boldsymbol{t} ) ]^2  \mathbb{P}(\vv{w}^{\top}\boldsymbol{X} > c \mid \boldsymbol{X}\in \boldsymbol{t} ) \\
         & \quad - [\mathbb{E}( m(\boldsymbol{X}) \mid \{\boldsymbol{X}\in \boldsymbol{t}\} \cap \{\vv{w}^{\top}\boldsymbol{X} \le c \}) -  \mathbb{E}( m(\boldsymbol{X}) \mid \boldsymbol{X}\in \boldsymbol{t} ) ]^2  \mathbb{P}(\vv{w}^{\top}\boldsymbol{X} \le c \mid \boldsymbol{X}\in \boldsymbol{t} ),
    \end{split}
\end{equation}
where the above formula can be interpreted as:  
\[
\textnormal{Remaining Bias after the split} = \textnormal{Var}( m(\boldsymbol{X}) \mid \boldsymbol{X} \in \boldsymbol{t}) - \textnormal{Bias Reduction due to the split},
\]
and $\textnormal{Var}( m(\boldsymbol{X}) \mid \boldsymbol{X} \in \boldsymbol{t})$ is the total variance (conditional on $\{\boldsymbol{X} \in \boldsymbol{t}\}$) before the split. To verify Condition~\ref{sid}, let us further derive that
\begin{equation}
    \begin{split}\label{decom.2}
        & \frac{\mathbb{E} [ \textnormal{Var} ( m(\boldsymbol{X}) | \boldsymbol{1}_{\boldsymbol{X} \in \boldsymbol{t}} , \boldsymbol{1}_{\vv{w}^{\top}\boldsymbol{X}>c} ) \boldsymbol{1}_{\boldsymbol{X} \in \boldsymbol{t}} ] }{\mathbb{P}(\boldsymbol{X}\in \boldsymbol{t})}\\            
       & =  \textnormal{Var} ( m(\boldsymbol{X}) \mid \{\boldsymbol{X}\in \boldsymbol{t}\} \cap \{\vv{w}^{\top}\boldsymbol{X} > c \}) \times \mathbb{P}(\vv{w}^{\top}\boldsymbol{X} > c \mid \boldsymbol{X}\in \boldsymbol{t} ) \\
        & \quad + \textnormal{Var} ( m(\boldsymbol{X}) \mid \{\boldsymbol{X}\in \boldsymbol{t}\} \cap \{\vv{w}^{\top}\boldsymbol{X} \le c \}) \times  \mathbb{P}(\vv{w}^{\top}\boldsymbol{X} \le c \mid \boldsymbol{X}\in \boldsymbol{t} ),
%        & \quad = [\mathbb{E}( m(\boldsymbol{X}) | \{\boldsymbol{X}\in \boldsymbol{t}\} \cap \{\vv{w}^{\top}\boldsymbol{X} > c \}) -  \mathbb{E}( m(\boldsymbol{X}) | \boldsymbol{X}\in \boldsymbol{t} ) ]^2  \mathbb{P}(\vv{w}^{\top}\boldsymbol{X} > c | \boldsymbol{X}\in \boldsymbol{t} ) \\
%         & \quad\quad + [\mathbb{E}( m(\boldsymbol{X}) | \{\boldsymbol{X}\in \boldsymbol{t}\} \cap \{\vv{w}^{\top}\boldsymbol{X} \le c \}) -  \mathbb{E}( m(\boldsymbol{X}) | \boldsymbol{X}\in \boldsymbol{t} ) ]^2  \mathbb{P}(\vv{w}^{\top}\boldsymbol{X} \le c | \boldsymbol{X}\in \boldsymbol{t} ),
    \end{split}
\end{equation}
and that 
\begin{equation*}
    \begin{split}
         \frac{\mathbb{E} [ \textnormal{Var} ( m(\boldsymbol{X}) | \boldsymbol{1}_{\boldsymbol{X} \in \boldsymbol{t}} ) \boldsymbol{1}_{\boldsymbol{X} \in \boldsymbol{t}} ] }{\mathbb{P}(\boldsymbol{X}\in \boldsymbol{t})}  = \textnormal{Var} ( m(\boldsymbol{X}) | \boldsymbol{X} \in \boldsymbol{t} ) .
    \end{split}
\end{equation*}

Building on these observations, we will demonstrate that for each node \(\boldsymbol{t}\), there exists a constant \(\alpha_{0} \in [0, 1)\) and a split \((\vv{w}, c)\in W_{p, s_0}\) such that the bias reduction achieved by the split is proportional to the total variance conditional on \(\boldsymbol{t}\) for each of the following four cases. Specifically, we will show that  
\[
\mathbb{E} [ \textnormal{Var}(m(\boldsymbol{X}) \mid \boldsymbol{1}_{\boldsymbol{X} \in \boldsymbol{t}}, \boldsymbol{1}_{\vv{w}^{\top} \boldsymbol{X} > c}) \boldsymbol{1}_{\boldsymbol{X} \in \boldsymbol{t}} ] \leq \alpha_{0} \times \mathbb{E} [ \textnormal{Var}(m(\boldsymbol{X}) \mid \boldsymbol{1}_{\boldsymbol{X} \in \boldsymbol{t}}) \boldsymbol{1}_{\boldsymbol{X} \in \boldsymbol{t}} ],
\]
thereby establishing that Condition~\ref{sid} is satisfied.

\bigskip
\begin{center}
\begin{varwidth}{\textwidth}
\begin{enumerate}
   \item[Case 1 : ] $\texttt{\#}\{k >0 : h_{k} \not= \emptyset\} = 0$ and $\texttt{\#}\{k>0: \mathbb{P} (\boldsymbol{X} \in \boldsymbol{t} \cap \mathcal{C}_{k}) >0\} >1$.
    \item[Case 2 : ] $\texttt{\#}\{k >0: h_{k} \not= \emptyset\} > 0$ and $\texttt{\#}\{k>0: \mathbb{P} (\boldsymbol{X} \in \boldsymbol{t} \cap \mathcal{C}_{k}) >0\} > 1$.
    \item[Case 3 : ] $\texttt{\#}\{k>0: \mathbb{P} (\boldsymbol{X} \in \boldsymbol{t} \cap \mathcal{C}_{k}) >0\} = 1$.
    \item[Case 4 : ] $\texttt{\#}\{k>0: \mathbb{P} (\boldsymbol{X} \in \boldsymbol{t} \cap \mathcal{C}_{k}) >0\} = 0$. 
\end{enumerate}
\end{varwidth}
\end{center}
\bigskip

\textbf{Analysis for Case 1} :

Let $\underline{\rho}_{2} =  \frac{1}{2} \times \min \{ \underline{r}\underline{\rho} D_{\max}^{-1}D_{\min}, r_{L_0}\}$, in which $\underline{r}$ and $\underline{\rho}$ are respectively given by \eqref{without.2} and \eqref{case1.7}. Consider the situation where there exists some $k^{\star}\in \{1, \dots, K_{1}\}$ such that 
\begin{equation}\label{case1.2}
    |\mathbb{E}( m(\boldsymbol{X})|\boldsymbol{X} \in \boldsymbol{t} \cap \mathcal{C}_{k^{\star}}) - \mathbb{E}( m(\boldsymbol{X})|\boldsymbol{X} \in \boldsymbol{t} )| < \underline{\rho}_{2}.
\end{equation}
% then we conclude the inequality as follows:
% \begin{equation}
% \begin{split}\label{case1.1}
%     &  \sum_{k \in Q} [\mathbb{E}( m(\boldsymbol{X})|\boldsymbol{X} \in \boldsymbol{t} \cap \mathcal{C}_{k}) - \mathbb{E}( m(\boldsymbol{X})|\boldsymbol{X} \in \boldsymbol{t} ) ]\times \mathbb{P} (\boldsymbol{X} \in \boldsymbol{t} \cap \mathcal{C}_{k})\\
%     & \quad + [0 - \mathbb{E}( m(\boldsymbol{X})|\boldsymbol{X} \in \boldsymbol{t} ) ]\times \mathbb{P} (\boldsymbol{X} \in \boldsymbol{t} \cap \mathcal{C}_{0}) \\
%     & \ge   \sum_{k=0}^{K} [\mathbb{E}( m(\boldsymbol{X})|\boldsymbol{X} \in \boldsymbol{t} \cap \mathcal{C}_{k}) - \mathbb{E}( m(\boldsymbol{X})|\boldsymbol{X} \in \boldsymbol{t} ) ]\times \mathbb{P} (\boldsymbol{X} \in \boldsymbol{t} \cap \mathcal{C}_{k})\\
%     & = 0,
%     \end{split}
% \end{equation}
% where 

Denote 
\begin{equation*}
	\begin{split}
%		Q & = \{q>0: \mathbb{E}( m(\boldsymbol{X})|\boldsymbol{X} \in \boldsymbol{t} \cap \mathcal{C}_{q}) > \mathbb{E}( m(\boldsymbol{X})|\boldsymbol{X} \in \boldsymbol{t}  ) - \underline{\rho}_{2}\},\\
		W & = \{q>0: \mathbb{E}( m(\boldsymbol{X})|\boldsymbol{X} \in \boldsymbol{t} \cap \mathcal{C}_{q}) \le \mathbb{E}( m(\boldsymbol{X})|\boldsymbol{X} \in \boldsymbol{t}  ) + \underline{\rho}_{2}\},
	\end{split}
\end{equation*}
and 
$$e_{k}=\mathbb{E}( m(\boldsymbol{X})|\boldsymbol{X} \in \boldsymbol{t} \cap \mathcal{C}_{k}) - \mathbb{E}( m(\boldsymbol{X})|\boldsymbol{X} \in \boldsymbol{t} ).$$

By the assumption of \eqref{case1.2}, there is some $k^{\star}$ such that $\mathcal{C}_{k^{\star}}$ is above $\mathcal{C}_0$ such that $e_{k^{\star}} \le \underline{\rho}_2$. Then, by the definitions of $\underline{\rho}_2$ and $\underline{r}$, it holds that $e_{0}\le e_{k^{\star}} - \underline{r}  \le \underline{\rho}_2 - \underline{r} < 0$. With this result and the definition of $r_{L_0}$, we conclude that
{\small \begin{equation}
	\begin{split}\label{case1.10.bb}
		 0&=  [\sum_{k \in W^c} \mathbb{P} (\boldsymbol{X} \in \boldsymbol{t} \cap \mathcal{C}_{k}) \times e_{k}] + [\sum_{k \in W} \mathbb{P} (\boldsymbol{X} \in \boldsymbol{t} \cap \mathcal{C}_{k}) \times e_{k}] + [ \mathbb{P} (\boldsymbol{X} \in \boldsymbol{t} \cap \mathcal{C}_{0}) \times e_{0}] \\
		& \le  [\sum_{k \in W^c} \mathbb{P} (\boldsymbol{X} \in \boldsymbol{t} \cap \mathcal{C}_{k}) \times r_{L_0}]  + [\sum_{k \in W} \mathbb{P} (\boldsymbol{X} \in \boldsymbol{t} \cap \mathcal{C}_{k}) \times \underline{\rho}_2] + \mathbb{P} (\boldsymbol{X} \in \boldsymbol{t} \cap \mathcal{C}_{0}) (\underline{\rho}_2 - \underline{r})\\
		& = \mathbb{P}(\boldsymbol{X}\in \boldsymbol{t})\underline{\rho}_2   + [\sum_{k \in W^c} \mathbb{P} (\boldsymbol{X} \in \boldsymbol{t} \cap \mathcal{C}_{k}) \times (r_{L_0} - \underline{\rho}_2 )] - \mathbb{P} (\boldsymbol{X} \in \boldsymbol{t} \cap \mathcal{C}_{0})  \times \underline{r},
	\end{split}
\end{equation}}%
which, in combination with \eqref{case1.7.c} and the definition of $\underline{\rho}_2$, implies that 
\begin{equation}
	\begin{split}\label{case1.30}
		\sum_{k \in W^c} \mathbb{P} (\boldsymbol{X} \in \boldsymbol{t} \cap \mathcal{C}_{k}) & \ge \frac{ \mathbb{P} (\boldsymbol{X} \in \boldsymbol{t} \cap \mathcal{C}_{0})  \underline{r}  - \mathbb{P}(\boldsymbol{X}\in \boldsymbol{t})\underline{\rho}_2 }{r_{L_0} - \underline{\rho}_2 } \\
		& \ge \frac{ \mathbb{P} (\boldsymbol{X} \in \boldsymbol{t}) \underline{\rho}D_{\max}^{-1}D_{\min} \underline{r}  - \mathbb{P}(\boldsymbol{X}\in \boldsymbol{t})\underline{\rho}_2 }{r_{L_0} - \underline{\rho}_2 }\\
		& = \mathbb{P} (\boldsymbol{X} \in \boldsymbol{t})\times \frac{  \underline{\rho}D_{\max}^{-1}D_{\min} \underline{r} -  \underline{\rho}_2 }{r_{L_0} - \underline{\rho}_2 }\\
		& \ge \mathbb{P} (\boldsymbol{X} \in \boldsymbol{t})\times \frac{  \underline{\rho}D_{\max}^{-1}D_{\min} \underline{r}  }{2 r_{L_0} }.
	\end{split}
\end{equation}

The result of \eqref{case1.30} and the definition of $W$ imply  there exists some $k_2^{\star}\in W^c$ such that
\begin{equation}
	\label{case1.40}
	\begin{split}
		e_{k_2^{\star}} & > \underline{\rho}_{2} ,\\
	\mathbb{P} (\boldsymbol{X} \in \boldsymbol{t} \cap \mathcal{C}_{k_{2}^{\star}}) & \ge \mathbb{P} (\boldsymbol{X} \in \boldsymbol{t})\times \frac{  \underline{\rho}D_{\max}^{-1}D_{\min}  \underline{r} }{ 2r_{L_0} K_{1}}.
		\end{split}
\end{equation}

In this case, we chose the split for the hyperplance $H_{k_{2}^{\star}}$ such that $\vv{w} = \vv{v}_{1}^{(k_{2}^{\star})}$ and $c = a_{1}^{(k_{2}^{\star}) }$, which implies that $(\vv{w}, c)\in W_{p, s_0}$ due to the assumption that \(\texttt{\#}\{j : |v_{l, j}^{(k)}| > 0, l \in \{1, \dots, n_{k}\}, k \in \{1, \dots, K_{1}\}\} \leq s_{0}\).  By \eqref{case1.40}, the bias reduction as in \eqref{decom.1} is lower bounded  by 
\begin{equation}
\begin{split}\label{case1.4}
    & [\mathbb{E}( m(\boldsymbol{X})|\boldsymbol{X} \in \boldsymbol{t} \cap \mathcal{C}_{k_{2}^{\star}}) - \mathbb{E}( m(\boldsymbol{X})|\boldsymbol{X} \in \boldsymbol{t} )]^2 \times \mathbb{P} (\boldsymbol{X} \in \boldsymbol{t} \cap \mathcal{C}_{k_{2}^{\star}} | \boldsymbol{X}\in\boldsymbol{t})\\
    & \ge \underline{\rho}_{2}^2 \times  \frac{  \underline{\rho}D_{\max}^{-1}D_{\min}  \underline{r} }{ 2r_{L_0} K_{1}} .
    \end{split}
\end{equation}

On the other hand, if there is no such $k^{\star}$ for \eqref{case1.2}, then let $k_{3}^{\star} = \argmax_{1\le k \le K_{1}} \mathbb{P} (\boldsymbol{X} \in \boldsymbol{t}\cap \mathcal{C}_{k})$. It holds that 
\begin{equation*}
    \begin{split}
        \mathbb{P} (\boldsymbol{X} \in \boldsymbol{t}\cap \mathcal{C}_{k_{3}^{\star}}) \ge K_{1}^{-1} \sum_{k=1}^{K_{1}}\mathbb{P} (\boldsymbol{X} \in \boldsymbol{t}\cap \mathcal{C}_{k}),
    \end{split}
\end{equation*}
and that
$$|\mathbb{E}( m(\boldsymbol{X})|\boldsymbol{X} \in \boldsymbol{t} \cap \mathcal{C}_{k}) - \mathbb{E}( m(\boldsymbol{X})|\boldsymbol{X} \in \boldsymbol{t} )| > \underline{\rho}_{2}.$$
In this case, we chose the split $(\vv{w}, c)\in W_{p, s_0}$ on $H_{k_{3}^{\star}}$, which implies that the bias reduction as in \eqref{decom.1}  is lower bounded  by 
\begin{equation}
\begin{split}\label{case1.5}
    & [\mathbb{E}( m(\boldsymbol{X})|\boldsymbol{X} \in \boldsymbol{t} \cap \mathcal{C}_{k_{3}^{\star}}) - \mathbb{E}( m(\boldsymbol{X})|\boldsymbol{X} \in \boldsymbol{t} )]^2 \times \mathbb{P} (\boldsymbol{X} \in \boldsymbol{t} \cap \mathcal{C}_{k_{3}^{\star}} | \boldsymbol{X}\in\boldsymbol{t})\\
    & \ge  \frac{\underline{\rho}_{2}^2 }{K_{1} \mathbb{P}(\boldsymbol{X} \in \boldsymbol{t})} \sum_{k=1}^{K_{1}}\mathbb{P} (\boldsymbol{X} \in \boldsymbol{t}\cap \mathcal{C}_{k}).
    \end{split}
\end{equation}

The results of \eqref{case1.4} and \eqref{case1.5} establish lower bounds of the bias reduction in this scenario. Let us derive an upper bound for the total variance as follows.
\begin{equation}
\begin{split}
    \label{case1.6}
    & \textnormal{Var}( m(\boldsymbol{X})| \boldsymbol{X} \in \boldsymbol{t}) \\
    & = \frac{1}{\mathbb{P}(\boldsymbol{X} \in \boldsymbol{t})  }\sum_{k=0}^{K_{1}} \mathbb{E}\{[m(\boldsymbol{X})- \mathbb{E}( m(\boldsymbol{X})|\boldsymbol{X}\in \boldsymbol{t} ) ]^2\times \boldsymbol{1}_{\boldsymbol{X}\in \boldsymbol{t} \cap \mathcal{C}_{k} } \} \\
    & = \sum_{k=0}^{K_{1}}[ \mathbb{E}( m(\boldsymbol{X})|\boldsymbol{X}\in \boldsymbol{t} \cap \mathcal{C}_{k}) - \mathbb{E}( m(\boldsymbol{X})|\boldsymbol{X}\in \boldsymbol{t} ) ]^2\times \mathbb{P}(\boldsymbol{X}\in \boldsymbol{t} \cap \mathcal{C}_{k}  | \boldsymbol{X}\in \boldsymbol{t}) \\
    & \le \left[0- \sum_{k=1}^{K_{1}} \mathbb{E}( m(\boldsymbol{X})|\boldsymbol{X}\in \boldsymbol{t} \cap \mathcal{C}_{k} ) \mathbb{P}(\boldsymbol{X}\in \boldsymbol{t} \cap \mathcal{C}_{k} )\right]^2 \times \mathbb{P}(\boldsymbol{X}\in \boldsymbol{t} \cap \mathcal{C}_{0} | \boldsymbol{X}\in \boldsymbol{t}) \\
    & \quad + \frac{(2\max_{ 1\le k\le K_{1}} \mathbb{E}( m(\boldsymbol{X})| \boldsymbol{X} \in \boldsymbol{t} \cap \mathcal{C}_{k}))^2}{\mathbb{P}(\boldsymbol{X}\in \boldsymbol{t} ) }  \times \left(\sum_{k=1}^{K_{1}}\mathbb{P}(\boldsymbol{X}\in \boldsymbol{t} \cap \mathcal{C}_{k})\right)\\
    & \le \frac{(2\max_{ k\ge 1} \mathbb{E}( m(\boldsymbol{X})| \boldsymbol{X} \in \boldsymbol{t} \cap \mathcal{C}_{k}))^2}{\mathbb{P}(\boldsymbol{X}\in \boldsymbol{t} ) }   \left[ \left[\sum_{k=1}^{K_{1}} \mathbb{P}(\boldsymbol{X} \in \boldsymbol{t} \cap \mathcal{C}_{k} ) \right]^2 + \sum_{k=1}^{K_{1}}\mathbb{P}(\boldsymbol{X}\in \boldsymbol{t} \cap \mathcal{C}_{k})  \right]\\
    &\le \frac{8r_{L_{0}}^2}{\mathbb{P}(\boldsymbol{X}\in \boldsymbol{t} ) }  \times \left[ \sum_{k=1}^{K_{1}}\mathbb{P}(\boldsymbol{X}\in \boldsymbol{t} \cap \mathcal{C}_{k})  \right],
    \end{split}
\end{equation}
where the second equality follows from the fact that \( m(\boldsymbol{X}) = \mathbb{E}( m(\boldsymbol{X})|\boldsymbol{X} \in \boldsymbol{t} \cap \mathcal{C}_{k}) \) on \( \{\boldsymbol{X} \in \boldsymbol{t} \cap \mathcal{C}_{k}\} \) in this scenario. The first inequality holds due to the assumption of \eqref{without.1}. The second inequality follows from \( \mathbb{P}(\boldsymbol{X} \in \boldsymbol{t} \cap \mathcal{C}_{0} | \boldsymbol{X} \in \boldsymbol{t}) \leq 1 \). The last inequality holds because \( \sum_{k=0}^{K_{1}} \mathbb{P}(\boldsymbol{X} \in \boldsymbol{t} \cap \mathcal{C}_{k}) \leq 1 \) and \( |m(\boldsymbol{X})| \leq r_{L_{0}} \) almost surely, due to the definition of \( r_{L_{0}} \) and our model assumptions.

By \eqref{decom.1} and \eqref{decom.2}, \eqref{case1.4}--\eqref{case1.6}, and \eqref{case1.7.c}, Condition~\ref{sid} holds with some constant $0\le \alpha_{0} < 1$ depending on the parameters $(r_{L_{0}}, \underline{\rho}_{2}, \underline{\rho}, K_{1}, \underline{r}, D_{\min}, D_{\max})$.

\textbf{Analysis for Case 2} :

Let $\mathcal{C}_{k}^{(2)}\subseteq \boldsymbol{t}$ denote the union of the highest and second highest platforms conditional on $\boldsymbol{t}$ in the $k$th corner, with $\mathcal{C}_{k}^{(2)} =\emptyset$ if $h_{k} = \emptyset$. Let $\mathcal{C}_{k}^{(3)} = (\mathcal{C}_{k}\cap \boldsymbol{t})\backslash \mathcal{C}_{k}^{(2)}$ if $h_{k} \not= \emptyset$, while $\mathcal{C}_{k}^{(3)} =\emptyset$ if $h_{k} = \emptyset$. To establish the upper bound of the total variance, note that by our model assumptions and \eqref{without.1}, 
\begin{equation}
    \label{case2.1}
    \textnormal{Var}( m(\boldsymbol{X})\mid \boldsymbol{X} \in \boldsymbol{t}) \le r_{L_{0}}^2.
\end{equation}

Additionally, by the assumption of this scenario (Case 2) that $\{q: h_{q}\not=\emptyset, 1\le q\le K_{1}\} \not = \emptyset$,  \eqref{without.1}, and \eqref{case1.7}, it holds that for each $k\in \{q: h_{q}\not=\emptyset, 1\le q\le K_{1}\}$,
\begin{equation}
	\begin{split}\label{case2.20}
		\mathbb{E}( m(\boldsymbol{X})\mid \boldsymbol{X} \in \boldsymbol{t}) & = \frac{\mathbb{E}( m(\boldsymbol{X}) \boldsymbol{1}_{\boldsymbol{X} \in \boldsymbol{t}} )   }{\mathbb{P}(\boldsymbol{X}\in \boldsymbol{t})} \\
		& \ge \frac{\mathbb{P}(\boldsymbol{X}\in \boldsymbol{t}\cap \mathcal{C}_{k}^{(2)} ) \times \mathbb{E}( m(\boldsymbol{X}) \boldsymbol{1}_{\boldsymbol{X} \in \boldsymbol{t}\cap \mathcal{C}_{k}^{(2)}} )   }{\mathbb{P}(\boldsymbol{X}\in \boldsymbol{t}\cap \mathcal{C}_{k}^{(2)}) \times \mathbb{P}(\boldsymbol{X}\in \boldsymbol{t})} \\
		& \ge r_{1}\times \underline{\rho} D_{\max}^{-1} D_{\min}.
	\end{split}
\end{equation}
Here, we apply similar arguments for \eqref{case1.7.c} to deduce that \(\mathbb{P}(\boldsymbol{X} \in \mathcal{C}_{k}^{(2)} \cap \boldsymbol{t}) \ge D_{\max}^{-1} D_{\min}\underline{\rho} \mathbb{P}(\boldsymbol{X} \in \boldsymbol{t})\) for each \(k \in \{q: h_{q}\not=\emptyset, 1\le q\le K_{1}\}\), based on the definition of \(\mathcal{C}_{k}^{(2)}\). Notably, in this case, \(\boldsymbol{t}\) intersects \(\mathcal{C}_0\) and the second highest and the highest platforms in the \(k\)th corner region for each \(k \in \{q: h_{q}\not=\emptyset, 1\le q\le K_{1}\}\).

Let $\underline{\delta}_{0} = \frac{\underline{\rho}D_{\min}}{3 \times K_{1} D_{\max}}$ be given with $\underline{\rho}>0$ defined by \eqref{case1.7}. Let us derive that
\begin{equation}
    \begin{split}\label{case2.4}
         &[\mathbb{E}( m(\boldsymbol{X})| \boldsymbol{X}\in \boldsymbol{t}) ]\times  \mathbb{P}(\boldsymbol{X}\in\boldsymbol{t}) \\
         & = \left[\sum_{k\in \mathcal{K}_{0}} [\mathbb{E} ( m(\boldsymbol{X})| \boldsymbol{X} \in \mathcal{C}_{k}^{(2)}\cap \boldsymbol{t}) ] \times \mathbb{P}(\boldsymbol{X} \in \mathcal{C}_{k}^{(2)}\cap \boldsymbol{t})\right] \\
        &\quad + \left[\sum_{k\in \mathcal{K}_{0}} [\mathbb{E} ( m(\boldsymbol{X})| \boldsymbol{X} \in \mathcal{C}_{k}^{(3)}\cap \boldsymbol{t} )] \times \mathbb{P}(\boldsymbol{X} \in \mathcal{C}_{k}^{(3)}\cap \boldsymbol{t})\right] \\
        & \quad + [\mathbb{E} ( m(\boldsymbol{X})| \boldsymbol{X} \in \mathcal{C}_{0} \cap \boldsymbol{t}) ] \times  \mathbb{P}(\boldsymbol{X} \in \mathcal{C}_{0} \cap \boldsymbol{t}) \\
        & \quad + \left[\sum_{k\in \mathcal{K}_{1}}[ \mathbb{E} ( m(\boldsymbol{X})| \boldsymbol{X} \in \mathcal{C}_{k} \cap \boldsymbol{t}) ]\times \mathbb{P}(\boldsymbol{X} \in \mathcal{C}_{k} \cap \boldsymbol{t})\right],
    \end{split}
\end{equation}
where $\mathcal{K}_{0} \coloneqq \{q: h_{q}\not=\emptyset, 1\le q\le K_{1}\}$ and $\mathcal{K}_{1} \coloneqq \{q: h_{q}=\emptyset, 1\le q\le K_{1}\}$.

If for each $k \in \mathcal{K}_{0}$,
\begin{equation}
    \label{case2.2}
    |\mathbb{E} ( m(\boldsymbol{X})\mid \boldsymbol{X} \in \mathcal{C}_{k}^{(2)}\cap \boldsymbol{t}) - \mathbb{E}( m(\boldsymbol{X})\mid \boldsymbol{X}\in\boldsymbol{t})| \le \mathbb{E}( m(\boldsymbol{X})| \boldsymbol{X}\in\boldsymbol{t}) \times\underline{\delta}_{0},
\end{equation} 
then by \eqref{case2.4},
\begin{equation}
    \begin{split}\label{deision.region.3}
    & \sum_{k\in \mathcal{K}_{1}} \left[\mathbb{E} ( m(\boldsymbol{X})| \boldsymbol{X} \in \mathcal{C}_{k} \cap \boldsymbol{t})  - \mathbb{E} ( m(\boldsymbol{X})| \boldsymbol{X} \in \boldsymbol{t}) 
 \right]\times \mathbb{P}(\boldsymbol{X} \in \mathcal{C}_{k} \cap \boldsymbol{t}) \\
         & \ge \left[\mathbb{E} ( m(\boldsymbol{X})| \boldsymbol{X} \in \boldsymbol{t})  - 0 \right] \times \mathbb{P}(\boldsymbol{X} \in \mathcal{C}_{0} \cap \boldsymbol{t}) \\
         & \quad -\left[\sum_{k\in \mathcal{K}_{0}} \mathbb{E} ( m(\boldsymbol{X})| \boldsymbol{X} \in \boldsymbol{t}) \times \underline{\delta}_{0}\times \mathbb{P}(\boldsymbol{X} \in \mathcal{C}_{k}^{(2)}\cap \boldsymbol{t}) \right] \\
        &\quad - \left[\sum_{k\in \mathcal{K}_{0}}  \mathbb{E} ( m(\boldsymbol{X})| \boldsymbol{X} \in \boldsymbol{t}) \times \underline{\delta}_{0} \times \mathbb{P}(\boldsymbol{X} \in \mathcal{C}_{k}^{(3)}\cap \boldsymbol{t})\right] \\
        & \ge [ \mathbb{E} ( m(\boldsymbol{X})| \boldsymbol{X} \in \boldsymbol{t}) ] \times  \frac{D_{\min} \underline{\rho}}{D_{\max}} \mathbb{P}(\boldsymbol{X} \in \boldsymbol{t} ) - 2K_{1} \mathbb{E} ( m(\boldsymbol{X})| \boldsymbol{X} \in \boldsymbol{t}) \times \underline{\delta}_{0}\times \mathbb{P}(\boldsymbol{X} \in \boldsymbol{t} )\\
        & \ge [ \mathbb{E} ( m(\boldsymbol{X})| \boldsymbol{X} \in \boldsymbol{t}) ] \times \frac{\underline{\rho} D_{\min}}{3D_{\max}}\times \mathbb{P}(\boldsymbol{X} \in \boldsymbol{t} ),
    \end{split}
\end{equation}
where the first inequality follows from that $\mathbb{E} ( m(\boldsymbol{X})| \boldsymbol{X} \in \boldsymbol{t}) \times (\underline{\delta}_{0} + 1) \ge \mathbb{E} ( m(\boldsymbol{X})| \boldsymbol{X} \in \mathcal{C}_{k}^{(2)}\cap \boldsymbol{t}) \ge \mathbb{E} ( m(\boldsymbol{X})| \boldsymbol{X} \in \mathcal{C}_{k}^{(3)}\cap \boldsymbol{t})$ for each $k\in \mathcal{K}_0$ due to \eqref{case2.2} and the definitions of $\mathcal{C}_{k}^{(2)}$'s and $\mathcal{C}_{k}^{(3)}$'s, and that $\mathbb{E} ( m(\boldsymbol{X})| \boldsymbol{X} \in \mathcal{C}_{0} \cap \boldsymbol{t}) = 0$. The second inequality results from \eqref{case1.7}, that $\mathbb{P}(\boldsymbol{X} \in \boldsymbol{t} ) \ge \max\{ \mathbb{P}(\boldsymbol{X} \in \mathcal{C}_{k}^{(2)}\cap \boldsymbol{t}),  \mathbb{P}(\boldsymbol{X} \in \mathcal{C}_{k}^{(3)}\cap \boldsymbol{t})\}$ and $m(\boldsymbol{X}) \ge 0$ almost surely here, and simple calculations. The third inequality is due to the choice of $\underline{\delta}_{0} = \frac{\underline{\rho}D_{\min}}{3 \times K_{1} D_{\max}}$.

Since the RHS of  \eqref{deision.region.3} is positive, we deduce that
\begin{equation}\label{case2.5}
    \mathcal{K}_{1} \not= \emptyset \textnormal{ when \eqref{case2.2} holds and } \texttt{\#}\{ k : h_{k} \not= \emptyset\} > 1.
\end{equation}

In light of \eqref{case2.5}, let 
$$k^{\star} = \argmax_{k\in \mathcal{K}_{1}} \left[\mathbb{E} ( m(\boldsymbol{X})| \boldsymbol{X} \in \mathcal{C}_{k} \cap \boldsymbol{t})  - \mathbb{E} ( m(\boldsymbol{X})| \boldsymbol{X} \in \boldsymbol{t}) 
 \right] \times \mathbb{P}(\boldsymbol{X} \in \mathcal{C}_{k} \cap \boldsymbol{t}) ,$$ 
which, combined with \eqref{deision.region.3}, implies that
 \begin{equation}
    \begin{split}\label{deision.region.2}
    &  \left[\mathbb{E} ( m(\boldsymbol{X})| \boldsymbol{X} \in \mathcal{C}_{k^{\star}} \cap \boldsymbol{t})  - \mathbb{E} ( m(\boldsymbol{X})| \boldsymbol{X} \in \boldsymbol{t}) 
 \right]\times \mathbb{P}(\boldsymbol{X} \in \mathcal{C}_{k^{\star}} \cap \boldsymbol{t}) \\
         & \ge K_{1}^{-1}[ \mathbb{E} ( m(\boldsymbol{X})| \boldsymbol{X} \in \boldsymbol{t}) ] \times \frac{\underline{\rho} D_{\min}}{3D_{\max}}\times \mathbb{P}(\boldsymbol{X} \in \boldsymbol{t} ).
    \end{split}
\end{equation}

Therefore, the bias reduction (see \eqref{decom.1}) due to the split on the hyperplane separating $\mathcal{C}_{k^{\star}}$ and the zero region, which is some split from $W_{p, s_0}$ due to the assumption that \(\texttt{\#}\{j : |v_{l, j}^{(k)}| > 0, l \in \{1, \dots, n_{k}\}, k \in \{1, \dots, K_{1}\}\} \leq s_{0}\), is at least
\begin{equation}
    \begin{split}\label{case2.7}
   &     \sum_{k\in \mathcal{K}_{1}} \left\{\left[\mathbb{E} ( m(\boldsymbol{X})| \boldsymbol{X} \in \mathcal{C}_{k} \cap \boldsymbol{t})  - \mathbb{E} ( m(\boldsymbol{X})| \boldsymbol{X} \in \boldsymbol{t}) 
 \right] \right\}^2 \times \mathbb{P}(\boldsymbol{X} \in \mathcal{C}_{k} \cap \boldsymbol{t} | \boldsymbol{X} \in \boldsymbol{t}) \\
 & \ge \frac{\left\{\left[\mathbb{E} ( m(\boldsymbol{X})| \boldsymbol{X} \in \mathcal{C}_{k^{\star}} \cap \boldsymbol{t})  - \mathbb{E} ( m(\boldsymbol{X})| \boldsymbol{X} \in \boldsymbol{t}) 
 \right] \times  \mathbb{P}(\boldsymbol{X} \in \mathcal{C}_{k^{\star}} \cap \boldsymbol{t}) \right\}^2} {\mathbb{P}(\boldsymbol{X} \in \mathcal{C}_{k^{\star}} \cap \boldsymbol{t}) \mathbb{P}(\boldsymbol{X} \in \boldsymbol{t})} \\
 & \ge  \frac{D_{\min}^2 \underline{\rho}^2\times [ \mathbb{E} ( m(\boldsymbol{X})| \boldsymbol{X} \in \boldsymbol{t}) ]^2\times \mathbb{P}(\boldsymbol{X} \in \boldsymbol{t} )}{9 D_{\max}^2 \times K_{1}^2\times \mathbb{P}(\boldsymbol{X} \in \mathcal{C}_{k^{\star}} \cap \boldsymbol{t})}\\
 & \ge  \frac{\underline{\rho}^2 D_{\min}^2 \times [ \mathbb{E} ( m(\boldsymbol{X})| \boldsymbol{X} \in \boldsymbol{t}) ]^2}{9\times K_{1}^2 D_{\max}^2},
    \end{split}
\end{equation}
where the second inequality is due to \eqref{deision.region.2}.

The results of \eqref{case2.7} and \eqref{case2.20} lead to the conclusion that
\begin{equation}\label{case2.8}
    \textnormal{RHS of \eqref{case2.7}} \ge \frac{\underline{\rho}^4\times  r_{1}^2 D_{\max}^{-4} D_{\min}^4 }{9\times K_{1}^2}.
\end{equation}

On the other hand, if \eqref{case2.2} does not hold, meaning that there exists some \(k_{3}^\star \in \mathcal{K}_{0}\), then
$$|\mathbb{E} ( m(\boldsymbol{X})\mid \boldsymbol{X} \in \mathcal{C}_{k_{3}^{\star}}^{(2)}\cap \boldsymbol{t}) - \mathbb{E}( m(\boldsymbol{X})\mid \boldsymbol{X}\in\boldsymbol{t})| > \mathbb{E}( m(\boldsymbol{X})| \boldsymbol{X}\in\boldsymbol{t})\times \underline{\delta}_{0},$$ 
then the bias reduction (see \eqref{decom.1}) due to the split on the hyperplane separating $\mathcal{C}_{k_{3}^{\star}}^{(2)}$ and $\mathcal{C}_{k_{3}^{\star}}^{(3)}$ (or separating $\mathcal{C}_{k_{3}^{\star}}^{(2)}$ and $\mathcal{C}_{0}$ if $\mathcal{C}_{k_{3}^{\star}}^{(3)} = \emptyset$), which is some split from $W_{p, s_0}$ due to the assumption that \(\texttt{\#}\{j : |v_{l, j}^{(k)}| > 0, l \in \{1, \dots, n_{k}\}, k \in \{1, \dots, K_{1}\}\} \leq s_{0}\), is at least
\begin{equation}
    \begin{split}
        \label{case2.9}   
  [ \mathbb{E}( m(\boldsymbol{X})| \boldsymbol{X}\in\boldsymbol{t}) ]^2\times
        \underline{\delta}_{0}^2 \times \mathbb{P}(\boldsymbol{X} \in \mathcal{C}_{k_{3}^{\star}}^{(2)}\cap \boldsymbol{t} \mid \boldsymbol{X} \in  \boldsymbol{t}) \ge \underline{\delta}_{0}^2 \times r_1^2 \times \underline{\rho}^3 D_{\min}^3 D_{\max}^{-3},
    \end{split}
\end{equation}
where the inequality follows from \eqref{case2.20} and arguments similar to those for \eqref{case1.7.c}, with \( \underline{\rho} \) given by \eqref{case1.7} and the definition of \( k_{3}^{\star} \in \mathcal{K}_{0} \).

By \eqref{case2.1}, \eqref{case2.8}, and \eqref{case2.9}, we conclude desired result that Condition~\ref{sid} holds with some $\alpha_{0} \in [0, 1)$  depending on the parameters $(r_{L_{0}}, r_1, \underline{\rho}, K_{1}, \underline{r}, D_{\min}, D_{\max})$. in this case.

\textbf{Analysis for Cases 3 and 4} :

Case 4 corresponds to $\textnormal{Var}(m(\boldsymbol{X}) \mid \boldsymbol{X} \in \boldsymbol{t}) = 0$, and hence Condition~\ref{sid} holds. Meanwhile, the analysis for Case 3 can be done in a similar way to the other two cases, and hence we omit the detail.

We have completed the proof of the desired result that Example~\ref{example.convex} satisfies Condition~\ref{sid}.

%%%
%%%
%%

\renewcommand{\thesubsection}{D.\arabic{subsection}}
\setcounter{equation}{0}
\renewcommand\theequation{D.\arabic{equation}}
\section{Auxiliary Lemmas}\label{SecD}
\subsection{Statement and proof of Lemma~\ref{lemma.3}}\label{proof.lemma3}

We introduce some technical notation. Let \( u_{h}: \{1, \dots, 2^{h}\} \rightarrow \{0, 1\}^h \) be given as a function that converts a decimal number to its binary representation in \( H \) digits. In addition, $(b_{l, 1}, \dots, b_{l, h})^{\top} = u_{h}(l)$. For each integer $ h \ge 1$, each $l\in \{1, \dots, 2^{h} \}$,  and each set of splits $\{(\vv{w}_{k}, c_{k})\}_{k=1}^h$, define
$$\mathcal{P}_{l}(\boldsymbol{X}, T) = \prod_{k=1}^{\texttt{\#} T} \left[(1-b_{l, k})\boldsymbol{1}_{\vv{w}_{k}^{\top}\boldsymbol{X}\le c_{k}} + b_{l, k}\left(1- \boldsymbol{1}_{\vv{w}_{k}^{\top}\boldsymbol{X}\le c_{k}}\right) \right], $$ 
where $T = \{\boldsymbol{1}_{\vv{w}_{k}^{\top}\boldsymbol{X}> c_{k}}\}_{k=1}^{h}$. The notation \(\mathcal{P}_{l}(\boldsymbol{X}, T)\) allows for a systematic representation of \(\boldsymbol{1}_{\boldsymbol{X} \in \boldsymbol{t}}\), where \(\boldsymbol{t}\) is a leaf node formed by  splits in \(T\).
 
We give a remark for Lemma~\ref{lemma.3}. In this context, the number of features \(p\), the feature vector \(\boldsymbol{X}\), the regression function \(m(\boldsymbol{X})\), and the model error \(\varepsilon\) are assumed to satisfy the conditions stated in Lemma~\ref{lemma.3}. It is important to note that when \(p\), \(\boldsymbol{X}\), \(m(\boldsymbol{X})\), and \(\varepsilon\) are given and fixed, the derived concentration inequalities apply uniformly over \(n\), \(s\), \(H\), and the associated split sets. Additionally, $D_{\max}$ is not needed for the case with discrete features.

\begin{lemma}
    \label{lemma.3}

Assume that $\mathbb{P}(\boldsymbol{X} \in \{0, 1\}^p) = 1$ or that $\mathbb{P}(\boldsymbol{X} \in [0, 1]^p) = 1$ and $\mathbb{P}(\boldsymbol{X} \in A) \le (\textnormal{Volume of } A) \times D_{\max}$ for every measurable $A\subseteq [0, 1]^p$ for some constant $D_{\max}\ge 1$. Assume $Y = m(\boldsymbol{X}) + \varepsilon$ with $\sup_{\vv{x}\in [0, 1]^p}|m(\vv{x})| < \infty$ and $|\varepsilon| \le M_{\epsilon}$ almost surely. Let a sample of size $n$ be given such that $(\boldsymbol{X}_{1}, Y_{1}), \dots , (\boldsymbol{X}_{n}, Y_{n}), (\boldsymbol{X}, Y)$ are i.i.d. There exists some sufficiently large constant $N \ge 1$ and an event $E_{n}$ such that the following results hold. On $E_{n}$, for all $n\ge N$, each $1\le p \le n^{K_{0}}$ for some constant $K_{0}$, each $s\ge 1$, each $H\ge h\ge 1$, each $l \in\{1, \dots, 2^{h}\}$, and every $T = \{\boldsymbol{1}_{\vv{w}_{l}^{\top}\boldsymbol{X} > c_{l}}: (\vv{w}_{l}, c_{l})\in W_{p, s}, 1\le l \le h\}$ with $\sqrt{2}n^{-\frac{1}{2}} s \sqrt{H(1\vee\log{s})} \log{n}\le 1$, it holds that
 \begin{equation*}
    \begin{split}\label{theorem1.20.b}
         \left|\left[n^{-1}\sum_{i=1}^{n}  \mathcal{P}_{l}(\boldsymbol{X}_{i}, T)\right]  -  \mathbb{E}(\mathcal{P}_{l}(\boldsymbol{X}, T)  )\right| & \le 3\sqrt{2} n^{-\frac{1}{2}} s \sqrt{H(1\vee\log{s})} \log{n},\\       
        \left| \left[ n^{-1} \sum_{i=1}^{n}  m(\boldsymbol{X}_{i})  \mathcal{P}_{l}(\boldsymbol{X}_{i}, T) \right] - \mathbb{E}(m(\boldsymbol{X})  \mathcal{P}_{l}(\boldsymbol{X}, T)) \right| & \le 12M_{01}n^{-\frac{1}{2}} s \sqrt{H(1\vee\log{s})} \log{n},\\
        \left| n^{-1} \sum_{i=1}^{n}  \varepsilon_{i}  \mathcal{P}_{l}(\boldsymbol{X}_{i}, T)  \right| & \le 12 M_{01} n^{-\frac{1}{2}} s \sqrt{H(1\vee\log{s})} \log{n}, \\
        \left| \left[ n^{-1} \sum_{i=1}^{n}  [m(\boldsymbol{X}_{i})]^2  \mathcal{P}_{l}(\boldsymbol{X}_{i}, T) \right] - \mathbb{E}( [m(\boldsymbol{X})]^2  \mathcal{P}_{l}(\boldsymbol{X}, T)) \right| & \le 12M_{01}n^{-\frac{1}{2}} s \sqrt{H(1\vee\log{s})} \log{n},\\
        \left| \left[ n^{-1} \sum_{i=1}^{n}  \varepsilon_{i}^2  \mathcal{P}_{l}(\boldsymbol{X}_{i}, T) \right] - \mathbb{E}( \varepsilon^2  \mathcal{P}_{l}(\boldsymbol{X}, T)) \right| & \le 12M_{01} n^{-\frac{1}{2}} s \sqrt{H(1\vee\log{s})} \log{n},\\
        \left| \left[ n^{-1} \sum_{i=1}^{n}  \varepsilon_{i}m(\boldsymbol{X}_{i})  \mathcal{P}_{l}(\boldsymbol{X}_{i}, T) \right] - \mathbb{E}( \varepsilon m(\boldsymbol{X})  \mathcal{P}_{l}(\boldsymbol{X}, T)) \right| & \le 12M_{01} n^{-\frac{1}{2}} s \sqrt{H(1\vee\log{s})} \log{n},
    \end{split}
\end{equation*}
where $M_{01} = \left(M_{\epsilon} + \sup_{\vv{x}\in[0, 1]^p} |m(\vv{x})|\right)$. Additionally, 
$$\mathbb{P}(E_{n}^c) \le 10\exp{\left(-\frac{1}{2} s^2 (\log{n})^2 (1\vee\log{s}) \right)}.$$

\end{lemma}

\textit{Proof of Lemma~\ref{lemma.3}: } Let us begin by addressing the case where $\mathbb{P}(\boldsymbol{X} \in [0, 1]^p) = 1$ and $\mathbb{P}(\boldsymbol{X} \in A) \leq (\text{Volume of } A) \times D_{\max}$ for some constant $D_{\max} \geq 1$, applicable to every measurable set $A \subseteq [0, 1]^p$.

First, we note that if there is any split $(\vv{w}, c)$ with $\norm{\vv{w}}_{2} = 1$ and $ c\in [-\sqrt{s}, \sqrt{s}]^c$, then the corresponding split is trivial in that $\mathbb{P}(\vv{w}^{\top} \boldsymbol{X} > c) =\mathbb{P}(\vv{w}^{\top} \boldsymbol{X} \le -c)= 0$  because of $\mathbb{P}(\boldsymbol{X} \in [0, 1]^p) = 1$ and an application of the Cauchy-Schwarz inequality. In these cases, the desired deviation upper bounds hold naturally. Therefore, we only need to consider biases within $[-\sqrt{s}, \sqrt{s}]$.

For the general case, to derive the desired upper bounds for all splits \( (\vv{w}, c) \) with \( \vv{w} \in \mathcal{J}(s) \) and \( c \in [-\sqrt{s}, \sqrt{s}] \), we utilize the concept of covering splits. Specifically, we construct concentration inequalities for a selected subset of splits whose size depends on \( (p, s, n) \). This subset is designed to cover the entire available split space such that every available split is sufficiently close to at least one split in the subset, allowing the deviation to be effectively controlled. The detail is given as follows.

Let  
\begin{equation}
    \label{dn}
    d_{n} = \frac{1}{2 n s^{s+1}}
\end{equation}
denote the edge length of each small \(s\)-dimensional hypercube. To completely fill an \(s\)-dimensional hypercube centered at the origin with edge length 2, at most \((\frac{2}{d_{n}} + 1)^s\) small hypercubes are required. Label these small hypercubes as \(A_{1}, \dots, A_{L}\), where \(L\) is a positive integer. Now, let $\vv{w}_{l} \in \{\vv{w} \in \mathbb{R}^s, \norm{\vv{w}}_{2} =1 \} \cap A_{l}$ be given for each $l\in \{1, \dots, L\}$. Simple calculation shows that 
\begin{equation}\label{theorem1.26}
    \{\vv{w} \in \mathbb{R}^s, \norm{\vv{w}}_{2} =1 \} \subseteq \cup_{l=1}^{L} \{\vv{w} :  \norm{\vv{w} - \vv{w}_{l}}_{2} \le d_{n}\sqrt{s} \}.
\end{equation}
Similar arguments apply to show that to cover $\mathcal{J}(s) = \{\vv{w} \in \mathbb{R}^p, \norm{\vv{w}}_{2} =1, \norm{\vv{w}}_{0} \le s \}$, it is required at most $\binom{p}{s} \times \left(\frac{2}{d_{n}} + 1\right)^s$ balls, each is of radius $d_{n}\sqrt{s}$  and is centered at some distinct unit weight $\vv{w}$ with $\norm{\vv{w}}_{0}\le s$. Denote this set of distinct unit weights as $\overline{\mathcal{W}}$, with $\texttt{\#}\overline{\mathcal{W}} \le \binom{p}{s} \times \left(\frac{2}{d_{n}} + 1\right)^s$. On the other hand, by the Cauchy-Schwarz inequality, for each $\norm{\vv{w}}_{2} = 1$ and $\norm{\vv{w}}_{0} \le s$, on $\{\boldsymbol{X} \in [0, 1]^p\}$, it holds that
$|\vv{w}^{\top}\boldsymbol{X}| \le \norm{\vv{w}}_{2} \sqrt{s} = \sqrt{s}$. Hence, $\mathbb{P}(\vv{w}^{\top}\boldsymbol{X} > c) = \mathbb{P}(\vv{w}^{\top}\boldsymbol{X} \le -c) = 0$ for every $c\ge \sqrt{s}$, which implies we only need to consider biases within $[-\sqrt{s}, \sqrt{s}]$. As a result, let $\mathcal{C} = \{-\sqrt{s} + t \times d_{n}\}_{t=1}^{\ceil{2\sqrt{s} / d_{n}}}$.
Moreover, let 
\begin{equation}
    \label{theorem1.10}
    \mathcal{W}(p, s, d_{n})  = \{(\vv{w}, c): \vv{w} \in \overline{\mathcal{W}}, c\in \mathcal{C}\}
\end{equation}
represent the set of all covering splits to be considered, where \(p\) is the feature dimensionality, \(s\) is the sparsity parameter indicating the number of non-zero elements in the weights, and $d_{n}$ denotes the length of each edge of the small $s$-dimensional hypercube. Then 
\begin{equation}
    \label{theorem1.9}
    \texttt{\#}\mathcal{W}(p, s, d_{n}) = \texttt{\#}\overline{\mathcal{W}} \times \texttt{\#} \mathcal{C} \le \binom{p}{s} \times \left(\frac{2}{d_{n}} + 1\right)^s \times \left(1+\frac{2\sqrt{s} }{d_{n}}\right).
\end{equation}

An essential property of \(\mathcal{W}(p, s, d_{n})\) is that for every split $(\vv{w}, c)$ with \(\vv{w} \in \mathcal{J}(s) \) and \(c \in [-\sqrt{s}, \sqrt{s}]\), there exists \((\vv{v}, a) \in \mathcal{W}(p, s, d_{n})\) satisfying \(\|\vv{w} - \vv{v}\|_{0} \leq s\) such that  
\begin{equation}
    \label{theorem1.2.e}
    |(\vv{w} - \vv{v})^{\top} \boldsymbol{X} - (c - a)| \leq \|\vv{w} - \vv{v}\|_{2} \sqrt{s} + |c - a| \leq d_{n}(s + 1),
\end{equation}
where the last inequality holds due to the Cauchy-Schwarz inequality, \eqref{theorem1.26}, the condition \(\|\vv{w} - \vv{v}\|_{0} \leq s\), and the assumption \(\mathbb{P}(\boldsymbol{X} \in [0, 1]^p) = 1\).

In what follows, we complete the proof of Lemma~\ref{lemma.3} in three steps.

\textbf{STEP 1: Definition of some required events and corresponding concentration inequalities}

We need the Bernstein's inequality~\citep{vershynin2018high} as follows.
\begin{lemma}[Bernstein's inequality]\label{berstein.lemma}
For a sequence of i.i.d. random variables \( Z_i \) with zero mean, \( \mathrm{Var}(Z_i) \le Q \) and \( |Z_i| \leq M \) for some constants $Q\ge 0 $ and $M>0$, it holds that for each $t> 0$,
\[
\mathbb{P}\left( \left|\sum_{i=1}^n Z_i \right| \geq t \right) \leq 2\exp\left( -\frac{t^2}{2n Q + \frac{2}{3} M t} \right).
\]
\end{lemma}

We begin by establishing concentration inequalities concerning the covering splits. Define $t_{11} = s^2(\log{n})^2(1\vee\log{s})$. By Lemma~\ref{lemma.4} in Section~\ref{proof.lemma4}, Lemma~\ref{berstein.lemma}, and the definition of $d_{n}$ in \eqref{dn}, we have that for all large $n$,  each $1\le p \le n^{K_{0}}$, each $s\ge 1$, and each $(\vv{w}, c) \in \mathcal{W}(p, s, d_{n})$,
\begin{equation}
    \begin{split}\label{theorem1.5}
         \mathbb{P}\left(W_{11}(\vv{w}, c) \right) & \le 2\exp{\left( -\frac{t_{11}^2 }{ 2n D_{\max}s^{ s}  d_{n}(s+1) + \frac{2}{3} t_{11} } \right)}\\
        & \le 2\exp{\left( -t_{11} \right)},
    \end{split}
\end{equation}
where 
$$W_{11}(\vv{w}, c) = \left\{\left|\left[n^{-1}\sum_{i=1}^{n} \boldsymbol{1}_{|\vv{w}^{\top}\boldsymbol{X}_{i} - c| \le d_{n}(s+1)} \right] - \mathbb{P}(|\vv{w}^{\top}\boldsymbol{X} - c| \le d_{n}(s+1)) \right| \ge  \frac{ t_{11}}{n}\right\} .$$

Let 
\begin{equation}
    \begin{split}
        \label{theorem1.22}
        W_{21} = \cap_{(\vv{w}, c) \in \mathcal{W}(p, s, d_{n})} (W_{11}(\vv{w}, c))^c,
    \end{split}
\end{equation}
where $W_{11}(\vv{w}, c)$ is define by \eqref{theorem1.5} and $A^c$ denote the complementary event of $A$. Then, by \eqref{theorem1.9}, \eqref{theorem1.5}, and $t_{11} = s^2(\log{n})^2(1\vee\log{s})$, it holds for each $1\le p \le n^{K_{0}}$, each $s\ge 1$, and all large $n$ that
\begin{equation}
    \begin{split}
        \label{theorem1.21}
        \mathbb{P}(W_{21}^c) & \le \sum_{(\vv{w}, c) \in \mathcal{W}(p, s, d_{n})} \mathbb{P}(W_{11}(\vv{w}, c) )\\
        & \le \binom{p}{s} \times \left(\frac{2}{d_{n}} + 1\right)^s \times (1+\frac{2\sqrt{s} }{d_{n}})\times 2\exp{\left( -t_{11} \right)}\\
        & \le p^s \times 3^{s+2} \times n^{s+1} \times s^{(s+1)^2 + s+2} \times \exp{\left( -t_{11} \right)}\\
        & \le \exp{\left(-\frac{1}{2} s^2 (\log{n})^2 (1\vee\log{s})\right)}.
    \end{split}
\end{equation}

Next, we address the concentration inequalities concerning one tree leaf. Define 
\begin{equation}
    \label{theorem1.24}
    \mathcal{Q} = \{\{\boldsymbol{1}_{\vv{w}_{l}^{\top}\boldsymbol{X} > c_{l}}\}_{l=1}^{q}: (\vv{w}_{l}, c_{l}) \in \mathcal{W}(p, s, d_{n}) , l\in \{1, \dots,  q\},  q\in \{1, \dots, H \}\}.
\end{equation}

Let us proceed to define related events as follows. For each $t_{12} >0, \dots, t_{17} >0$,
\begin{equation}
    \begin{split}\label{theorem1.23}
    W_{12} & = \cup_{T \in \mathcal{Q}} \cup_{ l=1}^{ 2^{\texttt{\#}T}  } \left\{ \left| \left[n^{-1} \sum_{i=1}^{n}    \mathcal{P}_{l}(\boldsymbol{X}_{i}, T) \right] - \mathbb{E}[\mathcal{P}_{l}(\boldsymbol{X}, T)  ] \right| \ge t_{12} \right\},\\
         W_{13}  &= \cup_{T \in \mathcal{Q}}\cup_{l=1}^{2^{\texttt{\#}T}}\left\{ \left| \left[n^{-1} \sum_{i=1}^{n}  m(\boldsymbol{X}_{i})  \mathcal{P}_{l}(\boldsymbol{X}_{i}, T) \right] - \mathbb{E}[m(\boldsymbol{X})  \mathcal{P}_{l}(\boldsymbol{X}, T)  ] \right| \ge t_{13} \right\},\\
         W_{14}  &= \cup_{T \in \mathcal{Q}}\cup_{l=1}^{2^{\texttt{\#}T}}\left\{ \left| \left[n^{-1} \sum_{i=1}^{n}  \varepsilon_{i}  \mathcal{P}_{l}(\boldsymbol{X}_{i}, T) \right] - \mathbb{E}[\varepsilon  \mathcal{P}_{l}(\boldsymbol{X}, T)  ] \right| \ge t_{14} \right\},\\
         W_{15}  &= \cup_{T \in \mathcal{Q}}\cup_{l=1}^{2^{\texttt{\#}T}}\left\{ \left| \left[n^{-1} \sum_{i=1}^{n}  \varepsilon_{i} m(\boldsymbol{X}_{i}) \mathcal{P}_{l}(\boldsymbol{X}_{i}, T) \right] - \mathbb{E}[\varepsilon m(\boldsymbol{X}) \mathcal{P}_{l}(\boldsymbol{X}, T)  ] \right| \ge t_{15} \right\},\\
         W_{16}  &= \cup_{T \in \mathcal{Q}}\cup_{l=1}^{2^{\texttt{\#}T}}\left\{ \left| \left[n^{-1} \sum_{i=1}^{n}  \varepsilon_{i}^2  \mathcal{P}_{l}(\boldsymbol{X}_{i}, T) \right] - \mathbb{E}[\varepsilon^2  \mathcal{P}_{l}(\boldsymbol{X}, T)  ] \right| \ge t_{16} \right\},\\
         W_{17}  &= \cup_{T \in \mathcal{Q}}\cup_{l=1}^{2^{\texttt{\#}T}}\left\{ \left| \left[n^{-1} \sum_{i=1}^{n}  [ m(\boldsymbol{X}_{i})]^2 \mathcal{P}_{l}(\boldsymbol{X}_{i}, T) \right] - \mathbb{E}[[m(\boldsymbol{X})]^2  \mathcal{P}_{l}(\boldsymbol{X}, T)  ] \right| \ge t_{17} \right\},
    \end{split}
\end{equation}
where $\mathcal{Q}$ is defined by \eqref{theorem1.24}. Note that $\mathbb{E}[\varepsilon  \mathcal{P}_{l}(\boldsymbol{X}, T)  ] = \mathbb{E}[\varepsilon m(\boldsymbol{X}) \mathcal{P}_{l}(\boldsymbol{X}, T)  ] = 0$. 

By \eqref{theorem1.9}, $d_{n}\le 1$, $s\ge 1$, and an application of Hoeffding's inequality with 
\begin{equation}
    \begin{split}\label{theorem1.23.b}
%    t_{11} & = s^2(\log{n})^2(1\vee\log{s}),\\
t_{12} &= \sqrt{2}n^{-\frac{1}{2}} s \sqrt{H(1\vee\log{s})} (\log{n}) , \\
t_{13} & = \sqrt{2\sup_{\vv{x}\in[0, 1]^p} |m(\vv{x})|}n^{-\frac{1}{2}} s \sqrt{H(1\vee\log{s})} \log{n}, \\ t_{14} & = \sqrt{2M_{\epsilon}}n^{-\frac{1}{2}} s \sqrt{H(1\vee\log{s})} \log{n},\\
t_{15} & = \sqrt{2M_{\epsilon} \sup_{\vv{x}\in[0, 1]^p} |m(\vv{x})| }n^{-\frac{1}{2}} s \sqrt{H(1\vee\log{s})} \log{n},\\
t_{16} & = \sqrt{2M_{\epsilon}^2}n^{-\frac{1}{2}} s \sqrt{H(1\vee\log{s})} \log{n},\\
t_{17} & = \sqrt{2\sup_{\vv{x}\in[0, 1]^p} |m(\vv{x})|^2}n^{-\frac{1}{2}} s \sqrt{H(1\vee\log{s})} \log{n},
    \end{split}
\end{equation}
 it holds that for each $s\ge 1$, $H\ge 1$, each $1\le p\le n^{K_{0}}$, and all large $n$,
\begin{equation}
    \begin{split}\label{theorem1.21.b}
    \mathbb{P}(W_{12}) & \le \left[  \texttt{\#}\mathcal{W}(p, s, d_{n}) + 1\right]^{H}\times 2^{H+1}\exp{\left( - \frac{n t_{12}^2}{2} \right)}\\
        & \le \left[\binom{p}{s} \times \left(\frac{2}{d_{n}} + 1\right)^s \times (1+\frac{2\sqrt{s} }{d_{n}}) + 1\right]^{H}\times 2^{H+1}\exp{\left( - \frac{n t_{12}^2}{2} \right)} \\
        & \le \left[p^s \times \left(\frac{3}{d_{n}}\right)^s \times \left(\frac{3}{d_{n}}\right) \times \sqrt{s} \times 3\right]^{H}\times 2^{H+1}\exp{\left( - \frac{ n t_{12}^2}{2} \right)}\\
        & \le p^{sH} \times 3^{(2+s)H} \times s^{\frac{H}{2}} \times d_{n}^{-(1+s)H} \times 2^{H+1}\exp{\left( - \frac{n t_{12}^2}{2} \right)}\\
        & \le \exp{\left(-\frac{1}{2} s^2 (\log{n})^2 H (1\vee\log{s})\right)},
    \end{split}
\end{equation}
where the first inequality is due to an application of Hoeffding's inequality. Here, \(  \texttt{\#}\mathcal{W}(p, s, d_n) + 1 \) represents the total count of all possible splits, including trivial splits $(\vv{w}, c)$ with $\mathbb{P}(\vv{w}^{\top}\boldsymbol{X} > c) \in \{0, 1\}$. The split set $\texttt{\#} T = h< H$ can be seen as a split set of $H$ splits but $H-h$ of them are trivial splits. Similarly,
\begin{equation}
    \begin{split}\label{theorem1.25}
        \mathbb{P}(W_{13}) & \le  \left[ \texttt{\#}\mathcal{W}(p, s, d_{n}) + 1\right]^{H} \times 2^{H + 1}\times \exp{\left( - \frac{n t_{13}^2}{2 \sup_{\vv{x}\in[0, 1]^p} |m(\vv{x})|  } \right)}\\
        & \le \exp{(-\frac{1}{2} s^2 (\log{n})^2 H (1\vee\log{s}))} ,\\
        \mathbb{P}(W_{14}) & \le  \left[ \texttt{\#}\mathcal{W}(p, s, d_{n}) + 1\right]^{H} \times 2^{H + 1}\times \exp{\left( - \frac{n t_{14}^2}{2 M_{\epsilon} } \right)}\\
        & \le \exp{(-\frac{1}{2} s^2 (\log{n})^2 H (1\vee\log{s}))},\\
        \mathbb{P}(W_{15}) & \le  \left[ \texttt{\#}\mathcal{W}(p, s, d_{n}) + 1\right]^{H} \times 2^{H + 1}\times \exp{\left( - \frac{n t_{15}^2}{2 M_{\epsilon} \sup_{\vv{x}\in[0, 1]^p} |m(\vv{x})| } \right)}\\
        & \le \exp{(-\frac{1}{2} s^2 (\log{n})^2 H (1\vee\log{s}))},\\
        \mathbb{P}(W_{16}) & \le  \left[ \texttt{\#}\mathcal{W}(p, s, d_{n}) + 1\right]^{H} \times 2^{H + 1}\times \exp{\left( - \frac{n t_{16}^2}{2 M_{\epsilon}^2 } \right)}\\
        & \le \exp{(-\frac{1}{2} s^2 (\log{n})^2 H (1\vee\log{s}))},\\
        \mathbb{P}(W_{17}) & \le  \left[ \texttt{\#}\mathcal{W}(p, s, d_{n}) + 1\right]^{H} \times 2^{H + 1}\times \exp{\left( - \frac{n t_{17}^2}{2 \sup_{\vv{x}\in[0, 1]^p} |m(\vv{x})|^2 } \right)}\\
        & \le \exp{(-\frac{1}{2} s^2 (\log{n})^2 H (1\vee\log{s}))}.
    \end{split}
\end{equation}

\textbf{STEP 2: Application of derived inequalities for deviation bounds}

Consider a set of splits  $\{(\vv{w}_{1t}, c_{1t})\}_{t=1}^{t_{0}} $ with \(c_{1t} \in [-\sqrt{s}, \sqrt{s}]\), and let $\{(\vv{w}_{0t}, c_{0t})\}_{t=1}^{t_{0}} \in \mathcal{Q}$ be given such that 
\begin{equation}
    \label{theorem1.2.d}
|(\vv{w}_{1t} - \vv{w}_{0t})^{\top}\boldsymbol{X} - (c_{1t} - c_{0t})| \le d_{n}(s+1)
\end{equation}
holds for each $(\vv{w}_{1t}, c_{1t}, \vv{w}_{0t}, c_{0t})$, which can be done due to \eqref{theorem1.2.e} and the definitions of $\mathcal{Q}$ and $\mathcal{W}(p, s, d_n)$. Note that  $t_{0} \le H$. By \eqref{theorem1.2.d}, \eqref{theorem1.4} of Lemma~\ref{lemma.4}, and the definition of constant $D_{\max}>0$ from the assumption of Lemma~\ref{lemma.3}, it holds that 
\begin{equation}
    \begin{split}\label{theorem1.6}
        & \mathbb{E}(|\boldsymbol{1}_{\vv{w}_{1t}^{\top}\boldsymbol{X} > c_{1t}} - \boldsymbol{1}_{\vv{w}_{0t}^{\top}\boldsymbol{X} > c_{0t}}|) \\
        & \le \mathbb{P}( |\vv{w}_{0t}^{\top}\boldsymbol{X} - c_{0t}| \le d_{n}(s+1) ) \\
        & \le s^{ s} \times d_{n}(s+1) \times D_{\max},
    \end{split}
\end{equation}
where the first inequality holds beacsue
\begin{equation}
\label{theorem1.6.e}
\{|\vv{w}_{0t}^{\top}\boldsymbol{X} - c_{0t}| > d_{n}(s+1)\} \subseteq \{ \boldsymbol{1}_{\vv{w}_{1t}^{\top}\boldsymbol{X} > c_{1t}} = \boldsymbol{1}_{\vv{w}_{0t}^{\top}\boldsymbol{X} > c_{0t}} \}
\end{equation}  
given \eqref{theorem1.2.d}.

For  $l\in \{0, \dots t_{0}\}$, define 
$$T(l) = \{\boldsymbol{1}_{ \vv{w}_{1t}^{\top}\boldsymbol{X} > c_{1t}  }\}_{t=1}^{t_{0}-l}\cup \{\boldsymbol{1}_{ \vv{w}_{0t}^{\top}\boldsymbol{X} > c_{0t}  }\}_{t=t_{0}-l+1}^{t_{0}}.$$

By \eqref{theorem1.6} and the definition of $W_{21}$ in \eqref{theorem1.22}, it holds that on the event $W_{21}$, for all large $n$  and each $k \in \{1, \dots, 2^{t_{0}}\}$,
\begin{equation}
    \begin{split}\label{theorem1.12}
    & \left| \left[\sum_{i=1}^{n}    \mathcal{P}_{k}(\boldsymbol{X}_{i}, T(0))  \right]- \sum_{i=1}^{n} \mathcal{P}_{k}(\boldsymbol{X}_{i}, T(t_{0}))  \right| \\
        & \le \sum_{i=1}^{n} | \mathcal{P}_{k}(\boldsymbol{X}_{i}, T(0))  -  \mathcal{P}_{k}(\boldsymbol{X}_{i}, T(t_{0})) | \\
        &\le \sum_{t=1}^{t_{0}} \sum_{i=1}^{n} | \mathcal{P}_{k}(\boldsymbol{X}_{i}, T(t-1))  -  \mathcal{P}_{k}(\boldsymbol{X}_{i}, T(t)) | \\         
        &\le \sum_{t=1}^{t_{0}} \sum_{i=1}^{n} |\boldsymbol{1}_{ \vv{w}_{1t}^{\top}\boldsymbol{X}_{i} > c_{1t}  } - \boldsymbol{1}_{ \vv{w}_{0t}^{\top}\boldsymbol{X}_{i} > c_{0t}  } |  \\
        & \le \sum_{t=1}^{t_{0}} \sum_{i=1}^{n} \boldsymbol{1}_{|\vv{w}_{0t}^{\top}\boldsymbol{X} - c_{0t}| \le d_{n}(s+1)} \\
        & \le t_{0} (t_{11} + ns^{ s} d_{n}(s+1) D_{\max}),
    \end{split}
\end{equation}
where the fifth inequality here follows from the second inequality of \eqref{theorem1.6} and the definition of $W_{21}$, and the fourth inequality follows from \eqref{theorem1.6.e}.

Next, we deduce that on $W_{12}$ defined in \eqref{theorem1.23}, for all large $n$ and each $k \in \{1, \dots, 2^{t_{0}}\}$,
\begin{equation}
    \begin{split}\label{theorem1.12.b}
        \left|\left[n^{-1}\sum_{i=1}^{n}\mathcal{P}_{k}(\boldsymbol{X}_{i}, T(t_{0})) \right] - \mathbb{E}(  \mathcal{P}_{k}(\boldsymbol{X}, T(t_{0}))) \right| \le t_{12}.
    \end{split}
\end{equation}
In addition, by \eqref{theorem1.6}, it holds that for all large $n$ and each $k \in \{1, \dots, 2^{t_{0}}\}$,
\begin{equation}
    \begin{split}\label{theorem1.12.c}
    & \left| \mathbb{E} [ \mathcal{P}_{k}(\boldsymbol{X}, T(0))] - \mathbb{E}[  \mathcal{P}_{k}(\boldsymbol{X}, T(t_{0})) ] \right| \\        
        &\le \sum_{t=1}^{t_{0}}  |\mathbb{E}[ \mathcal{P}_{k}(\boldsymbol{X}, T(t-1)) ] - \mathbb{E}[ \mathcal{P}_{k}(\boldsymbol{X}, T(t))] | \\        
        &\le \sum_{t=1}^{t_{0}} \mathbb{E} | \boldsymbol{1}_{ \vv{w}_{1t}^{\top}\boldsymbol{X} > c_{1t}  } - \boldsymbol{1}_{ \vv{w}_{0t}^{\top}\boldsymbol{X} > c_{0t}  } |  \\
        & \le t_{0} (s^{ s} \times d_{n}(s+1) \times D_{\max}).
    \end{split}
\end{equation}

%%%
%%%

We now turn to deal with establishing the upper bound for the deviations involving $m(\boldsymbol{X})$ and model error $\varepsilon$. By \eqref{theorem1.6} and the definition of $W_{21}$ in \eqref{theorem1.22}, it holds that on the event $W_{21}$, for all large $n$, and each $k \in \{1, \dots, 2^{t_{0}}\}$,
\begin{equation}
    \begin{split}\label{theorem1.12.m}
    & \left| \left[\sum_{i=1}^{n} m(\boldsymbol{X}_{i})  \mathcal{P}_{k}(\boldsymbol{X}_{i}, T(0))  \right]- \sum_{i=1}^{n} m(\boldsymbol{X}_{i}) \mathcal{P}_{k}(\boldsymbol{X}_{i}, T(t_{0}))  \right| \\        
        &\le \sum_{t=1}^{t_{0}} \sum_{i=1}^{n} | m(\boldsymbol{X}_{i}) \mathcal{P}_{k}(\boldsymbol{X}_{i}, T(t-1))  - m(\boldsymbol{X}_{i})  \mathcal{P}_{k}(\boldsymbol{X}_{i}, T(t)) | \\        
        &\le \sup_{\vv{x}\in[0 , 1]^p}| m(\vv{x})|\times  \sum_{t=1}^{t_{0}} \sum_{i=1}^{n} |\boldsymbol{1}_{ \vv{w}_{1t}^{\top}\boldsymbol{X}_{i} > c_{1t}  } - \boldsymbol{1}_{ \vv{w}_{0t}^{\top}\boldsymbol{X}_{i} > c_{0t}  } | \\
        & \le \sup_{\vv{x}\in[0 , 1]^p}| m(\vv{x})| \times \sum_{t=1}^{t_{0}} \sum_{i=1}^{n} \boldsymbol{1}_{|\vv{w}_{0t}^{\top}\boldsymbol{X} - c_{0t}| \le d_{n}(s+1)} \\
        & \le t_{0}  (t_{11} + ns^{ s} d_{n}(s+1) D_{\max}) \left(\sup_{\vv{x}\in[0 , 1]^p}| m(\vv{x})| \right),
    \end{split}
\end{equation}
where the fourth inequality here follows from the second inequality of \eqref{theorem1.6} and the definition of $W_{21}$, and the third inequality follows from \eqref{theorem1.6.e}.
  Similarly,
\begin{equation}
    \begin{split}\label{theorem1.12.e}
    & \left| \left[\sum_{i=1}^{n} \varepsilon_{i}  \mathcal{P}_{k}(\boldsymbol{X}_{i}, T(0))  \right]- \sum_{i=1}^{n} \varepsilon_{i} \mathcal{P}_{k}(\boldsymbol{X}_{i}, T(t_{0}))  \right|  \le t_{0}  (t_{11} + ns^{ s} d_{n}(s+1) D_{\max}) M_{\epsilon},\\
    & \left| \left[\sum_{i=1}^{n} \varepsilon_{i} m(\boldsymbol{X}_{i}) \mathcal{P}_{k}(\boldsymbol{X}_{i}, T(0))  \right]- \sum_{i=1}^{n} \varepsilon_{i} m(\boldsymbol{X}_{i})\mathcal{P}_{k}(\boldsymbol{X}_{i}, T(t_{0}))  \right|  \\
    & \quad \le t_{0}  (t_{11} + ns^{ s} d_{n}(s+1) D_{\max}) M_{\epsilon} \left(\sup_{\vv{x}\in[0 , 1]^p}| m(\vv{x})| \right),\\
    & \left| \left[\sum_{i=1}^{n} \varepsilon_{i}^2  \mathcal{P}_{k}(\boldsymbol{X}_{i}, T(0))  \right]- \sum_{i=1}^{n} \varepsilon_{i}^2 \mathcal{P}_{k}(\boldsymbol{X}_{i}, T(t_{0}))  \right|  \le t_{0}  (t_{11} + ns^{ s} d_{n}(s+1) D_{\max}) M_{\epsilon}^2,\\
    & \left| \left[\sum_{i=1}^{n} [m(\boldsymbol{X}_{i})]^2  \mathcal{P}_{k}(\boldsymbol{X}_{i}, T(0))  \right]- \sum_{i=1}^{n} [m(\boldsymbol{X}_{i})]^2 \mathcal{P}_{k}(\boldsymbol{X}_{i}, T(t_{0}))  \right|  \\
    & \quad \le t_{0}  (t_{11} + ns^{ s} d_{n}(s+1) D_{\max}) \left(\sup_{\vv{x}\in[0 , 1]^p}| m(\vv{x})| \right)^2.
    \end{split}
\end{equation}

Next, we deduce that on $W_{13}^c\cap \dots \cap W_{17}^c$ respectively defined below, it holds that for each $k \in \{1, \dots, 2^{t_{0}}\}$,
\begin{equation}
    \begin{split}\label{theorem1.12.b.m}
        \left|\left[n^{-1}\sum_{i=1}^{n} m(\boldsymbol{X}_{i}) \mathcal{P}_{k}(\boldsymbol{X}_{i}, T(t_{0})) \right] - \mathbb{E}( m(\boldsymbol{X}) \mathcal{P}_{k}(\boldsymbol{X}, T(t_{0}))) \right| & \le t_{13},\\
        \left|\left[n^{-1}\sum_{i=1}^{n} \varepsilon_{i} \mathcal{P}_{k}(\boldsymbol{X}_{i}, T(t_{0})) \right] - \mathbb{E}( \varepsilon \mathcal{P}_{k}(\boldsymbol{X}, T(t_{0}))) \right| & \le t_{14},\\
        \left|\left[n^{-1}\sum_{i=1}^{n} \varepsilon_{i} m(\boldsymbol{X}_{i}) \mathcal{P}_{k}(\boldsymbol{X}_{i}, T(t_{0})) \right] - \mathbb{E}( \varepsilon m(\boldsymbol{X}) \mathcal{P}_{k}(\boldsymbol{X}, T(t_{0}))) \right| & \le t_{15},\\
        \left|\left[n^{-1}\sum_{i=1}^{n} \varepsilon_{i}^2 \mathcal{P}_{k}(\boldsymbol{X}_{i}, T(t_{0})) \right] - \mathbb{E}( \varepsilon^2 \mathcal{P}_{k}(\boldsymbol{X}, T(t_{0}))) \right| & \le t_{16},\\
        \left|\left[n^{-1}\sum_{i=1}^{n} [m(\boldsymbol{X}_{i})]^2 \mathcal{P}_{k}(\boldsymbol{X}_{i}, T(t_{0})) \right] - \mathbb{E}( [m(\boldsymbol{X})]^2 \mathcal{P}_{k}(\boldsymbol{X}, T(t_{0}))) \right| & \le t_{17}.
    \end{split}
\end{equation}

%%%

In addition, by \eqref{theorem1.2.d} and \eqref{theorem1.6}, it holds that for all large $n$, and each $k \in \{1, \dots, 2^{t_{0}}\}$,
\begin{equation}
    \begin{split}\label{theorem1.12.c.v2}
    & \left| \mathbb{E} [ m(\boldsymbol{X})  \mathcal{P}_{k}(\boldsymbol{X}, T(0))] - \mathbb{E}[m(\boldsymbol{X}) \mathcal{P}_{k}(\boldsymbol{X}, T(t_{0})) ] \right| \\        
        &\le \sum_{t=1}^{t_{0}}  |m(\boldsymbol{X})  \mathcal{P}_{k}(\boldsymbol{X}, T(t-1)) ] - \mathbb{E}[m(\boldsymbol{X})  \mathcal{P}_{k}(\boldsymbol{X}, T(t))] | \\      
        &\le (\sup_{\vv{x}\in[0 , 1]^p}| m(\vv{x})| ) \sum_{t=1}^{t_{0}} \mathbb{E} | \boldsymbol{1}_{ \vv{w}_{1t}^{\top}\boldsymbol{X} > c_{1t}  } - \boldsymbol{1}_{ \vv{w}_{0t}^{\top}\boldsymbol{X} > c_{0t}  } | \\
        & \le t_{0} (s^{ s} \times d_{n}(s+1) \times D_{\max}) (\sup_{\vv{x}\in[0 , 1]^p}| m(\vv{x})| ).
    \end{split}
\end{equation}
Similarly,
\begin{equation}
    \begin{split}\label{theorem1.12.c.2}
    & \left| \mathbb{E} [ [m(\boldsymbol{X})]^2  \mathcal{P}_{k}(\boldsymbol{X}, T(0))] - \mathbb{E}[ [m(\boldsymbol{X})]^2 \mathcal{P}_{k}(\boldsymbol{X}, T(t_{0})) ] \right| \\           
        & \quad\le t_{0} (s^{ s} \times d_{n}(s+1) \times D_{\max}) (\sup_{\vv{x}\in[0 , 1]^p}| m(\vv{x})|^2 ),\\
        & \left| \mathbb{E} [ \varepsilon^2  \mathcal{P}_{k}(\boldsymbol{X}, T(0))] - \mathbb{E}[ \varepsilon^2\mathcal{P}_{k}(\boldsymbol{X}, T(t_{0})) ] \right| \\           
        &\quad \le t_{0} (s^{ s} \times d_{n}(s+1) \times D_{\max}) M_{\epsilon}^2 .
    \end{split}
\end{equation}
%%%
%%%
%%%

\textbf{STEP 3: Completion of the proof}

We are now prepared to finalize the establishment of the upper bound for the deviations. Let $T$ be given as in Lemma~\ref{lemma.3} in Section~\ref{proof.lemma3}, and recall that each of these sets of splits contains a split that is denoted by $(\vv{w}, c)$ with $\vv{w} \in \mathcal{J}(s)$ and $ c\in \mathbb{R}$. Additionally, recall from the discussion at the beginning of the proof that we only need to consider splits with $c \in [-\sqrt{s}, \sqrt{s}]$.

Now, let $T^{\dagger} \in \mathcal{Q}$, defined in \eqref{theorem1.24}, be given such that $\texttt{\#}T^{\dagger} = h$  (recall that $\texttt{\#}T = h \in \{1, \dots, H\}$) and that \eqref{theorem1.2.d} holds for each $(\vv{w}, c)\in T$ and the corresponding $(\vv{u}, b)\in T^{\dagger}$.

By \eqref{theorem1.23.b}, \eqref{theorem1.12}--\eqref{theorem1.12.c}, it holds that on $W_{21}\cap W_{12}^c$, for each $l\in \{1, \dots, 2^h\}$, each  $1\le p \le n^{K_{0}}$, and all large $n$,
\begin{equation}
    \begin{split}\label{theorem1.20}
        & \left| \left[ n^{-1} \sum_{i=1}^{n}    \mathcal{P}_{l}(\boldsymbol{X}_{i}, T) \right] - \mathbb{E}(  \mathcal{P}_{l}(\boldsymbol{X}, T)) \right| \\
        & = \bigg| \left[ n^{-1} \sum_{i=1}^{n}    \mathcal{P}_{l}(\boldsymbol{X}_{i}, T) \right] - \left[ n^{-1} \sum_{i=1}^{n}    \mathcal{P}_{l}(\boldsymbol{X}_{i}, T^{\dagger}) \right] \\
        & \quad + \left[ n^{-1} \sum_{i=1}^{n}  \mathcal{P}_{l}(\boldsymbol{X}_{i}, T^{\dagger}) \right] - \mathbb{E}(  \mathcal{P}_{l}(\boldsymbol{X}, T^{\dagger}))  \\
        & \quad + \mathbb{E}( \mathcal{P}_{l}(\boldsymbol{X}, T^{\dagger}))  - \mathbb{E}(  \mathcal{P}_{l}(\boldsymbol{X}, T)) \bigg|\\
        & \le 2Ht_{11}n^{-1} + t_{12} + 2H n^{-1}\log{n}\\
        & \le 3\sqrt{2}n^{-\frac{1}{2}} s \sqrt{H(1\vee\log{s})} \log{n},
    \end{split}
\end{equation}
where we simplify the expression for the last inequality by the assumption that $\sqrt{2}n^{-\frac{1}{2}} s \sqrt{H(1\vee\log{s})} \log{n}\le 1$ and that $x\le \sqrt{x}$ when $0\le x\le 1$.

Additionally, with \eqref{theorem1.23.b}, it holds that on $W_{21}\cap W_{13}^{c}$ (defined respectively by \eqref{theorem1.22} and \eqref{theorem1.23}), for each $l\in \{1, \dots, 2^h\}$ and all large $n$,
\begin{equation}
    \begin{split}\label{theorem1.20.c}
        &\left| \left[ n^{-1} \sum_{i=1}^{n}  m(\boldsymbol{X}_{i}) \times \mathcal{P}_{l}(\boldsymbol{X}_{i}, T) \right] - \mathbb{E}(m(\boldsymbol{X}) \times \mathcal{P}_{l}(\boldsymbol{X}, T)) \right| \\
        & = \bigg| \left[ n^{-1} \sum_{i=1}^{n}  m(\boldsymbol{X}_{i})  \mathcal{P}_{l}(\boldsymbol{X}_{i}, T) \right] - \left[ n^{-1} \sum_{i=1}^{n}  m(\boldsymbol{X}_{i})  \mathcal{P}_{l}(\boldsymbol{X}_{i}, T^{\dagger}) \right]\\
        & \quad + \left[ n^{-1} \sum_{i=1}^{n}  m(\boldsymbol{X}_{i})  \mathcal{P}_{l}(\boldsymbol{X}_{i}, T^{\dagger}) \right] - \mathbb{E}(m(\boldsymbol{X})  \mathcal{P}_{l}(\boldsymbol{X}, T^{\dagger})) \\
        &\quad + \mathbb{E}(m(\boldsymbol{X})  \mathcal{P}_{l}(\boldsymbol{X}, T^{\dagger}))  - \mathbb{E}(m(\boldsymbol{X}) \mathcal{P}_{l}(\boldsymbol{X}, T)) \bigg|\\
        & \le 2Ht_{11}n^{-1} + t_{13} + 2H n^{-1}\log{n}\\
        & \le (\sqrt{2\sup_{\vv{x}\in[0, 1]^p} |m(\vv{x})| } \vee 1) \times 3\sqrt{2}n^{-\frac{1}{2}} s \sqrt{H(1\vee\log{s})} \log{n},
    \end{split}
\end{equation}
in which the first inequality is due to \eqref{theorem1.12.m}, \eqref{theorem1.12.b.m}, and \eqref{theorem1.12.c.v2}, and we simplify the expression for the last inequality by the assumption that $\sqrt{2}n^{-\frac{1}{2}} s \sqrt{H(1\vee\log{s})} \log{n}\le 1$  and that $x\le \sqrt{x}$ when $0\le x\le 1$. 

Similarly, in light of \eqref{theorem1.12.e}--\eqref{theorem1.12.c.2}, we deduce that on $W_{21}\cap W_{14}^{c}\cap W_{15}^{c}\cap W_{16}^{c}\cap W_{17}^{c}$, for each $l\in \{1, \dots, 2^h\}$ and all large $n$,
{\small \begin{equation}
    \begin{split}\label{theorem1.20.d}
         &\left| n^{-1} \sum_{i=1}^{n}  \varepsilon_{i}  \mathcal{P}_{l}(\boldsymbol{X}_{i}, T)  \right|  \le  \sqrt{2M_{\epsilon}}\times 2\sqrt{2}n^{-\frac{1}{2}} s \sqrt{H(1\vee\log{s})} \log{n},\\
         &\left| n^{-1} \sum_{i=1}^{n}  \varepsilon_{i}m(\boldsymbol{X}_{i})  \mathcal{P}_{l}(\boldsymbol{X}_{i}, T)  \right|  \le (\sqrt{2M_{\epsilon} \sup_{\vv{x}\in[0, 1]^p} |m(\vv{x})|}\vee 1)  2\sqrt{2}n^{-\frac{1}{2}} s \sqrt{H(1\vee\log{s})} \log{n} ,\\
         &\left| \left[n^{-1} \sum_{i=1}^{n}  \varepsilon_{i}^2  \mathcal{P}_{l}(\boldsymbol{X}_{i}, T)\right]  - \mathbb{E}[\varepsilon^2  \mathcal{P}_{l}(\boldsymbol{X}, T) ] \right|  \\
         &\quad \le \sqrt{2M_{\epsilon}^2} \times 3\sqrt{2}n^{-\frac{1}{2}} s \sqrt{H(1\vee\log{s})} \log{n}  ,\\
         &\left| \left[n^{-1} \sum_{i=1}^{n}  [m(\boldsymbol{X}_{i})]^2  \mathcal{P}_{l}(\boldsymbol{X}_{i}, T)  \right] -\mathbb{E}[[m(\boldsymbol{X})]^2  \mathcal{P}_{l}(\boldsymbol{X}, T)] \right| \\
         &\quad \le \left(\sqrt{2 \sup_{\vv{x}\in[0, 1]^p} |m(\vv{x})|^2} \vee 1\right) \times 3\sqrt{2}n^{-\frac{1}{2}} s \sqrt{H(1\vee\log{s})} \log{n} .
    \end{split}
\end{equation}}%

Since $M_{\epsilon}\ge 1$, simple calculations show that
$$\max\left\{\sqrt{M_{\epsilon}} + \sqrt{\sup_{\vv{x}\in[0, 1]^p} |m(\vv{x})|}, \sqrt{M_{\epsilon}\times\sup_{\vv{x}\in[0, 1]^p} |m(\vv{x})|} \right\} \le 2\left(M_{\epsilon} + \sup_{\vv{x}\in[0, 1]^p} |m(\vv{x})|\right),$$
which implies that 
\begin{equation}
    \label{theorem1.20.e}
    \begin{split}        
    & \textnormal{ Each upper bound in  \eqref{theorem1.20.c}--\eqref{theorem1.20.d} } \\
    & \le 2\sqrt{2}\left(M_{\epsilon} + \sup_{\vv{x}\in[0, 1]^p} |m(\vv{x})|\right) \times 3\sqrt{2}n^{-\frac{1}{2}} s \sqrt{H(1\vee\log{s})} \log{n}\\
    & = 12\left(M_{\epsilon} + \sup_{\vv{x}\in[0, 1]^p} |m(\vv{x})|\right) \times n^{-\frac{1}{2}} s \sqrt{H(1\vee\log{s})} \log{n}.
    \end{split}
\end{equation}

Let $E_{n} = W_{21}\cap W_{12}^c\cap \dots \cap W_{17}^c$. Then the upper bounds of the deviation of Lemma~\ref{lemma.3} are established on $E_{n}$ by \eqref{theorem1.20} and \eqref{theorem1.20.e} with $\sqrt{2}n^{-\frac{1}{2}} s \sqrt{H(1\vee\log{s})} \log{n}\le 1$. Finally, using \eqref{theorem1.21}, \eqref{theorem1.21.b}, \eqref{theorem1.25}, and the above specified values of \( t_{11}, \dots, t_{17}\) as in \eqref{theorem1.23.b}, we establish when $1\le p\le n^{K_{0}}$ that for all large $n$, each $s\ge 1$, and each $H\ge 1$, it holds that
\begin{equation*}
    \mathbb{P}(E_{n}^c) \le 10\exp{\left(-\frac{1}{2} s^2 (\log{n})^2 H (1\vee\log{s})\right)}.
\end{equation*}
This completes the proof of Lemma~\ref{lemma.3} for the case with continuous features.

We now shift our focus to addressing discrete features. For discrete cases, we do not have to consider covering splits as in \eqref{theorem1.10}. In contrast, the total number of sample equivalent decision rules can be counted as follows.

According to Theorem 2 of~\citep{chow1961characterization}, the number of distinct partitions of the vertices of an \(s\)-dimensional hypercube induced by a hyperplane (let us call it $\kappa$) is determined by: (1) the number of vertices classified as positive, and (2) the \(s\)-dimensional vector obtained by summing the positive vertices. Here, a positive vertex is defined as a vertex \(\vv{v}\) for which \(\vv{w}^\top \vv{v} - c > 0\) with respect to the hyperplane \(\{\vv{x} \in \mathbb{R}^s : \vv{w}^\top \vv{x} - c = 0\}\). In summary, $\kappa$ is determined by \(s+1\) integer values, each ranging from \(0\) to \(2^s\). As a result, there are at most 
\begin{equation}
    \label{chow.1}
    (2^s + 1)^{s+1} \le 2^{(s+1)^2}
\end{equation} 
distinct ways to separate the vertices. Note that the exponential order of ${s^2}$ is similar to the one in \eqref{theorem1.9}.

With \eqref{chow.1}, the number of distinct sets of sample-equivalent splits in $W_{p, s}$, where two splits $(\vv{w}, c)$ and $(\vv{u}, a)$ are sample-equivalent splits if $\sum_{i=1}^{n} |\boldsymbol{1}_{\vv{w}^{\top}\boldsymbol{X}_{i} > c} - \boldsymbol{1}_{\vv{u}^{\top}\boldsymbol{X}_{i} > a} | = 0$, is upper bounded by 
$$ \binom{p}{s} \times 2^{(s+1)^2}.$$
Therefore, the number of distinct leaf nodes formed by at most $H$ splits in the discrete case is upper bounded by 
\begin{equation}
    \label{chow.2}
    \left[\binom{p}{s} \times 2^{(s+1)^2} + 1\right]^{H} \times 2^H, 
\end{equation}
in which the `plus one` here count the trivial splits as in \eqref{theorem1.21.b}.

In conclusion, to address the discrete case, we define events analogous to \eqref{theorem1.23}, accounting for all distinct leaf nodes formed by at most \(H\) splits. By applying the upper bound \eqref{chow.2} and following similar reasoning as in the continuous case, while using the same choice of \(t_{12}, \dots, t_{17}\) as in \eqref{theorem1.23.b}, we arrive at the same result for the discrete case as for the continuous case. Note that \(t_{11}\), Lemma~\ref{lemma.4}, and \(D_{\max}\) are not required in the discrete case. In fact, much of STEP 1 through STEP 3 can be significantly simplified for the discrete case. For brevity, we omit the details.
    
    We have completed the proof of Lemma~\ref{lemma.3}.

\subsection{Statement and proof of Lemma~\ref{lemma.4}}\label{proof.lemma4}
The notation used in Lemma~\ref{lemma.4} aligns with the context established in \eqref{theorem1.2.e}.
\begin{lemma}
    \label{lemma.4}
    Let $(\vv{w}_{0}, c_{0}) \in \mathcal{W}(p, s, d_{n})$ be given. Then,
    \begin{equation*}
    \begin{split}         
         \textnormal{Var}(\boldsymbol{1}_{|\vv{w}_{0}^{\top}\boldsymbol{X} - c_{0}| \le d_{n}(s+1)} ) &\le s^{ s}  d_{n}(s+1) \times D_{\max},
    \end{split}
\end{equation*}
where constant $D_{\max}>0$ is defined by Condition~\ref{regularity.1}.
\end{lemma}

\textit{Proof of Lemma~\ref{lemma.4}: } Let $(\vv{w}_{0}, c_{0}) \in \mathcal{W}(p, s, d_{n})$, which implies $\norm{\vv{w}_{0}}_{0} \le s$ due to the definition of $\mathcal{W}(p, s, d_{n})$ be given. By $\norm{\vv{w}_{0}}_{0}\le s$ due to the definition of $\vv{w}_{0}$ and the arguments similar to those for \eqref{exponential.prob.7}, we have that 
\begin{equation}\label{theorem1.4}
    \textnormal{ Volume of } \{\vv{x} \in [0, 1]^{p}: |\vv{w}_{0}^{\top} \vv{x} - c_{0}| \le d_{n}(s+1) \} \le s^{ s} \times d_{n}(s+1).
\end{equation}

Then,
\begin{equation}
    \begin{split}\label{theorem1.3}
        & \textnormal{Var}(\boldsymbol{1}_{|\vv{w}_{0}^{\top}\boldsymbol{X} - c_{0}| \le d_{n}(s+1)} ) \\
        & = \mathbb{P}(|\vv{w}_{0}^{\top}\boldsymbol{X} - c_{0}| \le d_{n}(s+1)) \times [1 - \mathbb{P}(|\vv{w}_{0}^{\top}\boldsymbol{X} - c_{0}| \le d_{n}(s+1))]\\
        & \le \mathbb{P}( |\vv{w}_{0}^{\top}\boldsymbol{X} - c_{0}| \le d_{n}(s+1) ) \\
        & \le s^{ s}  d_{n}(s+1) \times D_{\max},
    \end{split}
\end{equation}
where the last inequality follows from the definition of constant $D_{\max}>0$ from Condition~\ref{regularity.1}. 

With \eqref{theorem1.3}, we have completed the proof of Lemma~\ref{lemma.4}.
    
\subsection{Statement and proof of Lemma~\ref{lower.volume.ratio}}\label{proof.lower.volume.ratio}

\begin{lemma}
    \label{lower.volume.ratio}
    Consider any two hyperplanes $H_{1} = \{\vv{x} \in \mathbb{R}^p: \vv{v}_{1}^{\top}\vv{x} - a_{1} = 0 \}$ and $H_{2} = \{\vv{x} \in \mathbb{R}^p: \vv{v}_{2}^{\top}\vv{x} - a_{2} = 0 \}$ such that (i) $\norm{\vv{v}_{l}}_{2} = 1$, (ii) $H_{1}\cap H_{2} \cap [0, 1]^p = \emptyset$, (iii) $\texttt{\#} ( \{j: |v_{1j}| > 0\}\cup\{j: |v_{2j}| > 0\}) \le s$, and (iv) they separate $[0, 1]^p$ into a partition denoted by $(C_{1},C_{2}, C_{3})$, with $C_{2}$ denoting the disagreement region between $H_{1}$ and $H_{2}$ such that 
    {\small$$  C_{2} = \{\vv{x} \in [0, 1]^p: \vv{v}_{1}^{\top}\vv{x} - a_{1} > 0 , \vv{v}_{2}^{\top}\vv{x} - a_{2} \le 0 \}\cup \{\vv{x} \in [0, 1]^p: \vv{v}_{1}^{\top}\vv{x} - a_{1} \le 0 , \vv{v}_{2}^{\top}\vv{x} - a_{2} > 0 \}  .$$    }%
    Additionally, let a decision region $\boldsymbol{t} = \cap_{i=1}^{l} A_{i}$ for some integer $l>0$, in which $A_{i} \in \{\{ \vv{w}_{i}^{\top}\vv{x} > c_{i}\}, \{ \vv{w}_{i}^{\top}\vv{x} \le c_{i}\}\}$ and $(\vv{w}_{i}, c_{i}) \in W_{p, p}$, be given such that $\boldsymbol{t}\cap C_{l} \not = \emptyset$ for $l\in \{1, 2, 3\}$. Then the ratio of the volume of $C_{2} \cap \boldsymbol{t}$ to the volume of $\boldsymbol{t}$ is lower bounded by  $ \underline{\rho} =  \frac{c_{0}^2}{ 16s + 8 c_{0}\sqrt{s} + c_{0}^2} > 0$, where $c_{0} = \min\{q_{1}, q_{2}\} > 0$ and that
    \begin{equation*}
    \begin{split}
         q_{l} \coloneqq \inf \{|t|:\{\vv{x} \in [0, 1]^p : \vv{v}_{l}^{\top} \vv{x} - a_{l} - t = 0\}\cap \{\vv{x} \in [0, 1]^p : \vv{v}_{3 - l}^{\top} \vv{x} - a_{3 - l} = 0\} \not=\emptyset \}.
    \end{split}
\end{equation*}

\end{lemma}

\textit{Proof of Lemma~\ref{lower.volume.ratio}: } Before presenting the formal proof, we note that \( c_0 > 0 \) follows from assumption (ii) of Lemma~\ref{lower.volume.ratio}. Under the same assumption, the partition consists of exactly three subsets, \( C_1, C_2, \) and \( C_3 \). Recall that two hyperplanes can partition the space into at most four subsets. Additionally, it holds either that
    $$  C_{2} = \{\vv{x} \in [0, 1]^p: \vv{v}_{1}^{\top}\vv{x} - a_{1} > 0 , \vv{v}_{2}^{\top}\vv{x} - a_{2} \le 0 \},$$
or 
    $$  C_{2} =  \{\vv{x} \in [0, 1]^p: \vv{v}_{1}^{\top}\vv{x} - a_{1} \le 0 , \vv{v}_{2}^{\top}\vv{x} - a_{2} > 0 \}  .$$

Now, let us begin the formal proof of Lemma~\ref{lower.volume.ratio}.  Let 
$$S(\boldsymbol{t}, \vv{w},  t) = \textnormal{Area of } \{ \vv{x}\in \boldsymbol{t}: \vv{w}^{\top} \vv{x} = t\}.$$ 
Here, area refers to the $(p-1)$-dimensional Lebesgue measure. In light of the definition of $\boldsymbol{t}$ in Lemma~\ref{lower.volume.ratio}, it holds that for every $\vv{w}\in \{\vv{x}\in \mathbb{R}^p: \norm{\vv{x}}_{2} =1 \}$, 
\begin{equation}
    \label{geolemma.1}
    S(\boldsymbol{t}, \vv{w},  t) \textnormal{ is concave with respect to } t.
\end{equation}

To see the intuition of \eqref{geolemma.1}, let $U(\boldsymbol{t}, \vv{w},  t) = \{ \vv{x}\in \boldsymbol{t}: \vv{w}^{\top} \vv{x} = t\}$ be the cross-section between $\boldsymbol{t}$ and the hyperplane at $t$. Note that the boundaries of $U(\boldsymbol{t}, \vv{w},  t)$ are all decision functions and the boundaries of $[0, 1]^p$ due to the definition of $\boldsymbol{t}$ in Lemma~\ref{lower.volume.ratio}. Intuitively, as $t$ moving forward, the area changing rate of $U(\boldsymbol{t}, \vv{w},  t)$ cannot increase since $\boldsymbol{t}$ is the intersection of several hyper decision planes and the boundaries of the unit hypercube. On the other hand, it is easy to see that the changing rate may decrease (e.g., if $\boldsymbol{t}$ is inside the unit hypercube). Also, see page 197 of \citet{acosta2004optimal} for some details.

In addition, by Fubini's theorem, we have that 
\begin{equation}
    \label{cross-section.v}
    \textnormal{Vloume of } \boldsymbol{t} = \int_{t=-\infty}^{\infty} S(\boldsymbol{t}, \vv{w},  t) dt.
\end{equation}

In light of \eqref{geolemma.1}, it holds for every $t \in [ t_{0}, t_{1}]$ that
\begin{equation}
    \label{cross-section.v.10}
    \begin{split}
S(\boldsymbol{t}, \vv{w},  t) & \ge \frac{b-a}{t_{1} -t_{0}} \times (t - t_{0}) + a    ,
    \end{split}
\end{equation}
 for $t\in [t_{0}, t_{1}]$, in which $a =  S(\boldsymbol{t}, \vv{w},  t_{0})$ and $b = S(\boldsymbol{t}, \vv{w},  t_{1})$.  The result of \eqref{cross-section.v.10} in combination with \eqref{cross-section.v} concludes that
\begin{equation}
\begin{split}
    \label{cross-section.v.2}
    & \textnormal{Volume of } \boldsymbol{t} \cap \{\vv{x}: t_{0}\le \vv{w}^{\top}\vv{x}  \le t_{1}\} \\
    & \ge \int_{t=t_{0}}^{t_{1}} \left(\frac{b-a}{t_{1} -t_{0}} \times (t - t_{0}) + a\right) dt \\
    & = \frac{ a + b }{ 2} \times (t_{1} - t_{0}).
    \end{split}
\end{equation}

In addition, in light of \eqref{geolemma.1}, the changing rate of $S(\boldsymbol{t}, \vv{w},  t)$ is no larger than $\frac{|b-a|}{t_{1}-t_{0}}\le \frac{b+a}{t_{1}-t_{0}}$ for every $t \ge t_{1}$, it holds for each $t_{2} > t_{1}$ and  every $t \in [t_{1}, t_{2}]$ that
\begin{equation}
	\label{cross-section.v.5.b}
	S(\boldsymbol{t}, \vv{w},  t) \le \frac{b+a}{t_{1} -t_{0}} \times (t - t_{1}) + b.
\end{equation}
With \eqref{cross-section.v.5.b}, it holds for every $t_{2} > t_{1}$ that
\begin{equation}
\begin{split}
    \label{cross-section.v.5}
    & \textnormal{Volume of } \boldsymbol{t} \cap \{\vv{x}: t_{1}\le \vv{w}^{\top}\vv{x}  \le t_{2}\} \\
    & \le \int_{t=t_{1}}^{t_{2}} \left(\frac{b+a}{t_{1} -t_{0}} \times (t - t_{1}) + b\right) dt \\
    & \le (a+b)\int_{t=t_{1}}^{t_{2}} \left(\frac{1}{t_{1} -t_{0}} \times (t - t_{1}) + 1\right) dt \\
    & \le (a+b) (t_{2} - t_{1})\left(\frac{t_{2} - t_{1}}{t_{1} -t_{0}}  + 1\right).
    \end{split}
\end{equation}

Meanwhile, by Cauchy-Schwarz inequality and the definition of $\vv{v}_{1}$ and $\vv{v}_{2}$ (assumption (iii) of Lemma~\ref{lower.volume.ratio}), it holds that 
\begin{equation}\label{cross-section.v.4}
    \sup_{\vv{x} \in [0, 1]^p, l\in \{1, 2\}}|\vv{v}_{l}^{\top} \vv{x} |\le  \sqrt{s}.
\end{equation}

Without loss of generality, assume that the partition $\{C_{1}, C_{2}, C_{3}\}$ of $[0, 1]^p$ is such that
\begin{equation}
    \label{cross-section.v.7}
    \begin{split}        
    C_{1} = \{\vv{x}\in [0, 1]^p: \vv{v}_{1}^{\top}\vv{x} \le a_{1}\} \cap\{\vv{x}\in [0, 1]^p: \vv{v}_{2}^{\top}\vv{x} \le a_{2}\} ,\\
    C_{2} = \{\vv{x}\in [0, 1]^p: \vv{v}_{1}^{\top}\vv{x} > a_{1}\} \cap\{\vv{x}\in [0, 1]^p: \vv{v}_{2}^{\top}\vv{x} \le a_{2}\} ,\\
    C_{3} = \{\vv{x}\in [0, 1]^p: \vv{v}_{1}^{\top}\vv{x} > a_{1}\} \cap\{\vv{x}\in [0, 1]^p: \vv{v}_{2}^{\top}\vv{x} > a_{2}\} ,
    \end{split}
\end{equation}
and that 
\begin{equation}
    \label{cross-section.v.8}
    H_{3} \cap H_{2}\cap [0, 1]^{p} = \emptyset,
\end{equation}
where $H_{3} = \{ \vv{x}\in \mathbb{R}^p: \vv{v}_{1}^{\top} \vv{x} -a_{1} - \frac{c_{0}}{2} = 0\}$ in light of the definition of $c_{0}$ from Lemma~\ref{lower.volume.ratio}. With \eqref{cross-section.v.7}--\eqref{cross-section.v.8} and the fact that $C_{l}$'s are a partition of $[0, 1]^p$, we deduce that 
\begin{equation}\label{cross-section.v.6}
\begin{split}
    C_{3}&\subseteq \{\vv{x}\in [0, 1]^p: \vv{v}_{1}^{\top}\vv{x} > a_{1} + \frac{c_{0}}{2}\},\\
     \{\vv{x}\in [0, 1]^p: a_{1} < \vv{v}_{1}^{\top}\vv{x} \le a_{1} + \frac{c_{0}}{2}\} &\subseteq C_{2}.
     \end{split}
\end{equation}

Next, (Area of $\boldsymbol{t} \cap H_{1}$) $= S(\boldsymbol{t},\vv{v}_{1}, a_{1})$ and (Area of $ \boldsymbol{t} \cap H_{3}$) $= S(\boldsymbol{t},\vv{v}_{1}, a_{1} + \frac{c_{0}}{2}) $ are respectively denoted by  $a$ and $b$. By the assumption that $\boldsymbol{t}\cap C_{l} \not = \emptyset$ for $l\in \{1, 2, 3\}$,
\begin{equation}
    \label{cross-section.v.9}
    a + b >0.
\end{equation}

By \eqref{cross-section.v.2} and \eqref{cross-section.v.6}, 
\begin{equation}
	\label{cross-section.v.20}
	\textnormal{Volume of } C_{2}\cap \boldsymbol{t}\ge \int_{t=a_{1}}^{a_{1} + \frac{c_{0}}{2}} S(\boldsymbol{t}, \vv{v}_{1}, t) dt \ge \frac{c_{0}}{4} (a + b).
\end{equation}
 Furthermore, we have that
 \begin{equation}
 	\label{cross-section.v.21}
 	\textnormal{Volume of } C_{3}\cap \boldsymbol{t}\le \int_{t=a_{1} + \frac{c_{0}}{2}}^{a_{1} + \frac{c_{0}}{2}  +\sqrt{s}} S(\boldsymbol{t}, \vv{v}_{1}, t) dt \le \left(\frac{2s}{c_{0}} + \sqrt{s}\right) (a + b),
 \end{equation}
where the first inequality holds because of \eqref{cross-section.v.4} and \eqref{cross-section.v.6}, and the second inequality is due to \eqref{cross-section.v.5} with $(t_{0}, t_{1}, t_{2}) = (a_{1}, a_{1} + \frac{c_{0}}{2}, a_{1} + \frac{c_{0}}{2} + \sqrt{s})$. On the other hand, by similar arguments as in \eqref{cross-section.v.10}--\eqref{cross-section.v.5} but with $(t_{0}^{'}, t_{1}^{'}, t_{2}^{'}) = (a_{1} + \frac{c_{0}}{2}, a_{1} , a_{1}- \sqrt{s})$, we concludes that the volume of $C_{1}\cap \boldsymbol{t}$ is no larger than $\left(\frac{2s}{c_{0}} + \sqrt{s}\right) (a + b)$.

Therefore, the ratio of the volume of $C_{2} \cap \boldsymbol{t}$ to the volume of $\boldsymbol{t}$ is lower bounded by 
\begin{equation}
	\begin{split}
		\frac{\textnormal{Volume of } C_{2} \cap \boldsymbol{t} }{\textnormal{Volume of } \boldsymbol{t}} &=  \frac{\textnormal{Volume of } C_{2} \cap \boldsymbol{t} }{\textnormal{Volume of } C_{1} \cap \boldsymbol{t} + \textnormal{Volume of } C_{2} \cap \boldsymbol{t} + \textnormal{Volume of } C_{3} \cap \boldsymbol{t}} \\
		& \ge \frac{\textnormal{Volume of } C_{2} \cap \boldsymbol{t} }{2\left(\frac{2s}{c_{0}} + \sqrt{s}\right) (a + b)+ \textnormal{Volume of } C_{2} \cap \boldsymbol{t}} \\
				& \ge \inf_{x\ge \frac{c_{0}}{4} (a + b)}\frac{x }{2\left(\frac{2s}{c_{0}} + \sqrt{s}\right) (a + b)+ x} \\
		& \ge \frac{\frac{c_{0}}{4} (a + b)}{ 2\left(\frac{2s}{c_{0}} + \sqrt{s}\right) (a + b) + \frac{c_{0}}{4} (a + b)} \\
		& =  \frac{c_{0}^2}{ 16s + 8 c_{0}\sqrt{s} + c_{0}^2} \\
		&  = \underline{\rho},
	\end{split}
\end{equation}
where the first inequality results from \eqref{cross-section.v.21}, the second inequality follows from \eqref{cross-section.v.20}, and the second equality relies on \eqref{cross-section.v.9}. This result completes the proof of Lemma~\ref{lower.volume.ratio} in the scenario of \eqref{cross-section.v.7}.

Lastly, we note that the above arguments apply to the general case beyond \eqref{cross-section.v.7}, thereby completing the proof of Lemma~\ref{lower.volume.ratio}.

\subsection{Proof of Lemma~\ref{volume2}}\label{proof.lemma.volume2}

    The first part of Lemma~\ref{volume2} is given by \eqref{chow.1}. In the following, we prove the second assertion in four steps.

Step 1: Let $\mathcal{Q}^{\star}$ be the set of pairs $(\vv{w}, c)$ defining the $L_{s_0}$ distinct ways to split the vertices of the $s_0$-dimensional hypercube, $\{0, 1\}^{s_0}$. For any $(\vv{w}, c)$, we define the gap $d_{\vv{w}, c}$ as the minimum distance between the threshold $c$ and the values $\vv{w}^{\top}\vv{v}$ for all vertices $\vv{v}$ not lying on the hyperplane (i.e., $\vv{w}^{\top}\vv{v} \neq c$).

Since there are only $2^{s_0}$ vertices and a finite number of unique partitions $L_{s_0}$, there must exist a positive constant $\delta$, depending only on $s_0$, such that:
\begin{equation}\label{mathcal.3}
\min_{(\vv{w}, c) \in \mathcal{Q}^{\star}, d_{\vv{w}, c} > 0} d_{\vv{w}, c} > \delta. \tag{*}
\end{equation}

Step 2: For a chosen split $(\vv{w}, c)$, this $\delta$ ensures a "buffer zone" around the threshold. Specifically, for every vertex $\vv{v}$, if $\vv{w}^{\top}\vv{v} > c$, then $\vv{w}^{\top}\vv{v} > c + \delta$. Additionally, if $\vv{w}^{\top}\vv{v} < c$, then $\vv{w}^{\top}\vv{v} < c - \delta$.

Step 3: Consider a new normal vector $\vv{u}$ on the unit sphere such that $\norm{\vv{u} - \vv{w}}_2 < \frac{\delta}{3\sqrt{s_0}}$. For any vertex $\vv{v}$, the change in the dot product is bounded by:
$$|\vv{u}^{\top}\vv{v} - \vv{w}^{\top}\vv{v}| \leq \norm{\vv{u} - \vv{w}}_2 \times \norm{\vv{v}}_2 \le \frac{\delta}{3\sqrt{s_0}} \times \sqrt{s_0} = \frac{\delta}{3}.$$

We pick a vertex $\vv{v}_1$ that satisfied $\vv{w}^{\top}\vv{v}_1 = c$, and let $\bar{c}\in\mathbb{R}$ be such that $\bar{c} = \vv{u}^{\top}\vv{v}_1$. By the bound above, $|\bar{c} - c| \le \frac{\delta}{3}$. Now, we check the classification for all other vertices.
For any vertex where $\vv{w}^{\top}\vv{v} > c$:
$$\vv{u}^{\top}\vv{v} \ge \vv{w}^{\top}\vv{v} - \frac{\delta}{3} > (c + \delta) - \frac{\delta}{3} = c + \frac{2\delta}{3}.$$
Since $\bar{c} < c + \frac{\delta}{3}$, it follows that $\vv{u}^{\top}\vv{v} > \bar{c}$.

For any vertex where $\vv{w}^{\top}\vv{v} < c$:
$$\vv{u}^{\top}\vv{v} \le \vv{w}^{\top}\vv{v} + \frac{\delta}{3} < (c - \delta) + \frac{\delta}{3} = c - \frac{2\delta}{3}.$$
Since $\bar{c} > c - \frac{\delta}{3}$, it follows that $\vv{u}^{\top}\vv{v} < \bar{c}$.
Thus, the pair $(\vv{u}, \bar{c}) = (\vv{u}, \vv{u}^{\top}\vv{v}_1)$ induces the same partition as $(\vv{w}, c)$.

Step 4: Since any vector $\vv{u}$ within a distance of $\frac{\delta}{3\sqrt{s_0}}$ from $\vv{w}$ results in the same vertex partition, the set of all such valid normal vectors contains a spherical cap on the unit sphere $U_{s_0}$. The volume of a spherical cap is a strictly positive and continuous function of its radius and the dimension of the sphere. Because the radius of this cap is determined solely by $\delta$ (which is a constant for a fixed $s_0$) and the dimension $s_0$ itself, the volume of this region is at least a constant fraction $c_{s_0} > 0$ of the total volume of $U_{s_0}$. This establishes the second assertion of Lemma~\ref{volume2} and completes the proof.

%%
%%%
%%%

\end{document}